\documentclass[pdflatex,sn-nature]{sn-jnl}

\usepackage{graphicx}%
\usepackage{multirow}%
\usepackage{tcolorbox}
\usepackage{amsmath,amssymb,amsfonts}%
\usepackage{amsthm}%
\usepackage{mathrsfs}%
\usepackage[title]{appendix}%
\usepackage{xcolor}%
\usepackage{textcomp}%
\usepackage{manyfoot}%
\usepackage{booktabs}%
\usepackage{algorithm}%
\usepackage{algorithmicx}%
\usepackage{algpseudocode}%
\usepackage{listings}%
\usepackage{dirtytalk}
\usepackage{booktabs}
\usepackage{tabularx}
\usepackage{tikz}
\usepackage{bbm}
\definecolor{ForestGreen}{HTML}{009B55}
\definecolor{BrickRed}{HTML}{B6321C}
\definecolor{BurntOrange}{HTML}{F7921D}
\definecolor{MidnightBluee}{HTML}{006795}
\definecolor{Hpurple}{HTML}{69005f}
\definecolor{Hred}{HTML}{ff506e}
\usetikzlibrary{positioning}
\usepackage{array}
\newcolumntype{L}[1]{>{\raggedright\let\newline\\\arraybackslash\hspace{0pt}}m{#1}}
\newcolumntype{C}[1]{>{\centering\let\newline\\\arraybackslash\hspace{0pt}}m{#1}}
\newcolumntype{R}[1]{>{\raggedleft\let\newline\\\arraybackslash\hspace{0pt}}m{#1}}
\usepackage{amssymb}
\usepackage{pifont}
\newcommand{\cmark}{\ding{51}}%
\newcommand{\xmark}{\ding{55}}%
\usepackage{siunitx}

\begin{document}

\title[Article Title]{Centaur: a foundation model of human cognition}


\author*[1]{\fnm{Marcel} \sur{Binz}}\email{marcel.binz@helmholtz-munich.de}
\author[1]{\fnm{Elif} \sur{Akata}}
\author[2]{\fnm{Matthias} \sur{Bethge}}
\author[3,5]{\fnm{Franziska} \sur{Brändle}}
\author[4]{\fnm{Fred} \sur{Callaway}}
\author[1]{\fnm{Julian} \sur{Coda-Forno}}
\author[2,5]{\fnm{Peter} \sur{Dayan}}
\author[1]{\fnm{Can} \sur{Demircan}}
\author[6]{\fnm{Maria K.} \sur{Eckstein}}
\author[5]{\fnm{Noémi} \sur{Éltető}}
\author[7]{\fnm{Thomas L.} \sur{Griffiths}}
\author[1,13]{\fnm{Susanne} \sur{Haridi}}
\author[1,2,5]{\fnm{Akshay K.} \sur{Jagadish}}
\author[8]{\fnm{Li} \sur{Ji-An}}
\author[1]{\fnm{Alexander} \sur{Kipnis}}
\author[7]{\fnm{Sreejan} \sur{Kumar}}
\author[2,5]{\fnm{Tobias} \sur{Ludwig}}
\author[1]{\fnm{Marvin} \sur{Mathony}}
\author[4]{\fnm{Marcelo} \sur{Mattar}}
\author[1]{\fnm{Alireza} \sur{Modirshanechi}}
\author[2,5,13]{\fnm{Surabhi S.} \sur{Nath}}
\author[9]{\fnm{Joshua C.} \sur{Peterson}}
\author[1]{\fnm{Milena} \sur{Rmus}}
\author[7]{\fnm{Evan M.} \sur{Russek}}
\author[5]{\fnm{Tankred} \sur{Saanum}}
\author[5]{\fnm{Johannes A.} \sur{Schubert}}
\author[1]{\fnm{Luca M.} \sur{Schulze Buschoff}}
\author[14]{\fnm{Nishad} \sur{Singhi}}
\author[2,5]{\fnm{Xin} \sur{Sui}}
\author[1]{\fnm{Mirko} \sur{Thalmann}}
\author[1]{\fnm{Fabian} \sur{Theis}}
\author[5]{\fnm{Vuong} \sur{Truong}}
\author[2,15]{\fnm{Vishaal} \sur{Udandarao}}
\author[1]{\fnm{Konstantinos} \sur{Voudouris}}
\author[10]{\fnm{Robert} \sur{Wilson}}
\author[1]{\fnm{Kristin} \sur{Witte}}
\author[1]{\fnm{Shuchen} \sur{Wu}}
\author[11,12]{\fnm{Dirk U.} \sur{Wulff}}
\author[10]{\fnm{Huadong} \sur{Xiong}}
\author[1]{\fnm{Eric} \sur{Schulz}}

\affil[1]{\orgdiv{Helmholtz Munich}}
\affil[2]{\orgdiv{University of Tuebingen}}
\affil[3]{\orgdiv{University of Oxford}}
\affil[4]{\orgdiv{New York University}}
\affil[5]{\orgdiv{Max Planck Institute for Biological Cybernetics}}
\affil[6]{\orgdiv{Google DeepMind}}
\affil[7]{\orgdiv{Princeton University}}
\affil[8]{\orgdiv{University of California San Diego}}
\affil[9]{\orgdiv{Boston University}}
\affil[10]{\orgdiv{Georgia Institute of Technology}}
\affil[11]{\orgdiv{University of Basel}}
\affil[12]{\orgdiv{Max Planck Institute for Human Development}}
\affil[13]{\orgdiv{Max Planck School of Cognition}}
\affil[14]{\orgdiv{TU Darmstadt}}
\affil[15]{\orgdiv{University of Cambridge}}


\abstract{Establishing a unified theory of cognition has been a major goal of psychology \cite{anderson1983ao, Newell1990-NEWUTO}. While there have been previous attempts to instantiate such theories by building computational models \cite{anderson1983ao, Newell1990-NEWUTO}, we currently do not have one model that captures the human mind in its entirety. A first step in this direction is to create a model that can predict human behavior in a wide range of settings. Here we introduce Centaur, a computational model that can predict and simulate human behavior in any experiment expressible in natural language. We derived Centaur by finetuning a state-of-the-art language model on a novel, large-scale data set called Psych-101. Psych-101 reaches an unprecedented scale, covering trial-by-trial data from over 60,000 participants performing over 10,000,000 choices in 160 experiments. Centaur not only captures the behavior of held-out participants better than existing cognitive models, but also generalizes to new cover stories, structural task modifications, and entirely new domains. Furthermore, we find that the model’s internal representations become more aligned with human neural activity after finetuning.  Taken together, our results demonstrate that it is possible to discover computational models that capture human behavior across a wide range of domains. We believe that such models provide tremendous potential for guiding the development of cognitive theories and present a case study to demonstrate this.
}


\keywords{cognitive science, cognitive modeling, unified theory of cognition, large language models}

\maketitle

\section*{Introduction}\label{sec1}

The human mind is remarkably general \cite{lake2017building, lake2023human, wu2024unifying}. Not only do we routinely make mundane decisions, like choosing a breakfast cereal or selecting an outfit, but we also tackle complex challenges, such as figuring out how to cure cancer or explore outer space. We learn new skills from only a few demonstrations \cite{lake2015human}, reason causally \cite{goddu2024development}, and fuel our actions through curiosity \cite{chu2020play}. Whether we are climbing mountains, playing video games, or creating captivating art, our versatility defines what it means to be human.

In contrast to this, most contemporary computational models -- whether in machine learning or the cognitive sciences -- are domain-specific. They are designed to excel at one particular problem and that problem alone. Take, for instance, AlphaGo -- a computer system created by Google DeepMind to master the game of Go  \cite{silver2017mastering}. Even though the system can play this particular game at an impressive level, it can do not much beyond that. A similar pattern can be observed in the cognitive sciences. Prospect theory, one of the most influential accounts of human cognition, for instance, offers valuable insights into how people make choices \cite{kahneman2013prospect}, but it tells us nothing about how we learn, plan, or explore.

\begin{figure}

    \definecolor{fire1}{rgb}{0.974638, 0.797692, 0.206332}
    \definecolor{fire2}{rgb}{0.954506, 0.468744, 0.0998740}
    \definecolor{fire3}{rgb}{0.769556, 0.236077, 0.307485}
    \definecolor{fire4}{rgb}{0.497257, 0.119379, 0.424488}
    \definecolor{fire5}{rgb}{0.211095, 0.0370300, 0.378563}
    \definecolor{fire6}{rgb}{0.00146200, 0.000466, 0.0138660}
    \definecolor{Nblue}{HTML}{404c62}
    \definecolor{Nblue1}{HTML}{002752}
    \definecolor{Nblue2}{HTML}{004586}
    \definecolor{Nblue3}{HTML}{c23637}
    \definecolor{Nblue4}{HTML}{d95d5b}
    \definecolor{Nblue5}{HTML}{e9999c}
    \definecolor{Nblue6}{HTML}{f9c9c7}
    \definecolor{Nlightblue}{HTML}{e2e2ea}
    \definecolor{Ngrey}{HTML}{8e99ab}
    
    \centering
    \scalebox{0.71}{\begin{tikzpicture}[font={\fontfamily{phv}\selectfont\footnotesize\sffamily}]
    
    \draw[rounded corners, draw=Nlightblue!80, fill=Nlightblue!80] (-1.15, -7.35) rectangle ++(18.6,6.9);

    \draw[rounded corners, draw=Nlightblue!80, fill=Nlightblue!80] (-1.15, -11.7) rectangle ++(18.6,3.8);

    \node[anchor=west] (A) at (-1.125, -0.75) {{\color{Nblue!100}\textbf{\small\includegraphics[scale=0.135]{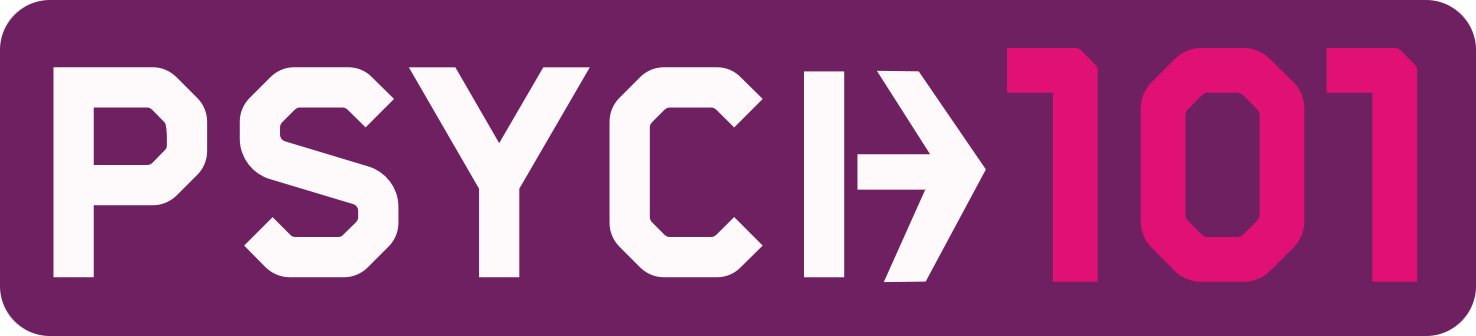}}}};

    \node[anchor=west] (A) at (0.625, -0.775) {{\color{Nblue!100}\textbf{\small 160 psychological experiments, 60,092 individual participants, 10,681,650 human choices, 253,597,411 text tokens}}};

    \node[anchor=west] (A) at (-1.125, -0.25) {\small\textbf{a}};

    \node[anchor=west] (A) at (-1.125, -7.65) {\small\textbf{b}};

    \node[anchor=west] (A) at (-1.125, -8.225) {{\color{Nblue!100}\textbf{\small\textsc{Centaur}: a foundation model of human cognition}}};

    \node[rounded corners, minimum width=6cm, text width=5.75cm] (A) at (3.4, -1.35) {\small{\color{Nblue!100}\text{Multi-armed bandits\phantom{y}}}};
    \node[rounded corners, minimum width=6cm, text width=5.75cm] (A) at (9.8, -1.35) {\small{\color{Nblue!100}\text{Decision-making}}};
    \node[rounded corners, minimum width=6cm,  text width=5.75cm] (A) at (16.5, -1.35) {\small{\color{Nblue!100}\text{Memory}}};

    \node[rounded corners, minimum width=6cm, text width=5.75cm] (A) at (3.5, -4.45) {\small{\color{Nblue!100}\text{Supervised learning}}};
    \node[rounded corners, minimum width=6cm, text width=5.75cm] (A) at (9.2, -4.45) {\small{\color{Nblue!100}\text{Markov decision processes}}};
    \node[rounded corners, minimum width=6cm,  text width=5.75cm] (A) at (16.2, -4.45) {\small{\color{Nblue!100}\text{Miscellaneous\phantom{y}}}};

    \node[fill=Nblue!100, rounded corners, anchor=west, draw=Nblue!100, minimum width=6cm, minimum height=2.5cm, align=left, text width=5.75cm] (A) at (-1, -2.85) {\color{white}In this task, you have to repeatedly choose between two slot machines labeled B and C. When you select one of the machines, you will win or lose points. Your goal is to choose the slot machines that will give you the most points. \\
    You press $<<$C$>>$ and get -8 points.  \\
    You press $<<$B$>>$ and get 0 points. \\
    You press $<<$B$>>$ and get 1 points.};
    \node[fill=Nblue!100, rounded corners, anchor=west, draw=Nblue!100, minimum width=6cm, minimum height=2.5cm, align=left, text width=5.75cm] (A) at (5.15, -2.85) {\color{white}You will choose from two monetary lotteries by pressing N or U. Your choice will trigger a random draw from the chosen lottery that will be added to your bonus. \\
    Lottery N offers 4.0 points with 80.0\% or 0.0 points with 20.0\%. \\
    Lottery U offers 3.0 points with 100.0\%. \\
    You press $<<$U$>>$.};
    \node[fill=Nblue!100, rounded corners, anchor=west, draw=Nblue!100, minimum width=6cm, minimum height=2.5cm, align=left, text width=5.75cm] (A) at (11.3, -2.85) {\color{white}You will view a stream of letters on the screen, one letter at a time. You have to remember the last two letters you saw since the beginning of the block. If the letter you see matches the letter two trials ago, press E, otherwise press K.\\
    You see the letter V and press $<<$K$>>$. \\
    You see the letter X and press $<<$K$>>$. \\
    You see the letter V and press $<<$E$>>$. 
    };

    \node[fill=Nblue!100, rounded corners, anchor=west, draw=Nblue!100, minimum width=6cm, minimum height=2.5cm, align=left, text width=5.75cm] (A) at (-1, -5.95) {\color{white}In each trial, you will see between one and three tarot cards. Your task is to decide if the combination of cards presented predicts rainy weather (by pressing P) or fine weather (by pressing L). \\
    You are seeing the following: card 3, card 4. You press $<<$L$>>$. You are wrong, the weather is rainy. \\  You are seeing the following: card 1, card 4. You press $<<$P$>>$. You are right, the weather is rainy.};
    \node[fill=Nblue!100, rounded corners, anchor=west, draw=Nblue!100, minimum width=6cm, minimum height=2.5cm, align=left, text width=5.75cm] (A) at (5.15, -5.95) {\color{white}You will be taking one of the spaceships F or V to one of the planets M or S. When you arrive at each planet, you will ask one of the aliens for space treasure. \\
    You are presented with spaceships V and F. \\You press $<<$V$>>$. You end up on planet M and see aliens G and W. You press $<<$G$>>$.\\ You find 1 pieces of space treasure.};
    \node[fill=Nblue!100, rounded corners, anchor=west, draw=Nblue!100, minimum width=6cm, minimum height=2.5cm, align=left, text width=5.75cm] (A) at (11.3, -5.95) {\color{white}You will be presented with triplets of objects, which will be assigned to the keys E, Z, and B. In each trial, please indicate which object you think is the odd one out by pressing the corresponding key. \\ 
    E: tablet, Z: fox, and B: vent. You press $<<$Z$>>$. \\
    E: ivy, Z: coop, and B: drink. You press $<<$B$>>$. \\
    E: kite, Z: flan, and B: jar. You press $<<$E$>>$. \\
    E: wand, Z: flag, and B: globe. You press $<<$Z$>>$. };

    \draw[thick, rounded corners, draw=Nblue!100, fill=Nblue!100] (1.8, -10.65) rectangle node[text width=2.32cm, align=left]{{\color{white}In this task, you have to repeatedly choose between two slot machines labeled B and C. $[\ldots]$  \\
    You press $<<$}} ++(2.5,2);

    \draw[thick, rounded corners, draw=gray!50, draw=Ngrey!80, fill=Ngrey!80] (5.05, -10.65) rectangle node[text width=1.3cm, align=center, rotate=90]{{\color{white}Token embedding}} ++(1,2);

    \draw[thick, rounded corners, rounded corners, draw=Nblue!100, fill=Nblue!100] (16.085, -10.65) rectangle  node[text width=2.32cm, align=center, rotate=90]{{\color{white}\tiny OUTPUT} \\[0.cm]{\color{white}C}} ++(1,2);

    \draw[anchor=west, draw=black, thick, dashed, rounded corners] (6.8, -10.65) rectangle ++(3.5,2);

    \draw[anchor=west, draw=black, thick, dashed, rounded corners] (11.7, -10.65) rectangle ++(3.5,2);

    \draw[rounded corners, draw=Hred!80, fill=Hred!80] (6.975, -10.525) rectangle node[text width=1.05cm, align=center]{{\color{white}Self-attention}} ++(1.5,1.75);

     \draw[rounded corners, draw=Hred!80, fill=Hred!80] (8.625, -10.525) rectangle node[text width=1.4cm, align=center]{{\color{white}Feedforward network}} ++(1.5,1.75);

     \draw[rounded corners, draw=Hred!80, fill=Hred!80] (11.875, -10.525) rectangle node[text width=1.05cm, align=center]{{\color{white}Self-attention}} ++(1.5,1.75);

     \draw[rounded corners, draw=Hred!80, fill=Hred!80] (13.525, -10.525) rectangle node[text width=1.4cm, align=center]{{\color{white}Feedforward network}} ++(1.5,1.75);

    \draw[rounded corners, draw=Hpurple!80, fill=Hpurple!80] (6.975, -11.525) rectangle node[text width=1.05cm, align=center]{{\color{white}Low-rank adapter}} ++(1.5,0.75);

     \draw[rounded corners, draw=Hpurple!80, fill=Hpurple!80] (8.625, -11.525) rectangle node[text width=1.05cm, align=center]{{\color{white}Low-rank adapter}} ++(1.5,0.75);

     \draw[rounded corners, draw=Hpurple!80, fill=Hpurple!80] (11.875, -11.525) rectangle node[text width=1.05cm, align=center]{{\color{white}Low-rank adapter}} ++(1.5,0.75);

     \draw[rounded corners, draw=Hpurple!80, fill=Hpurple!80] (13.525, -11.525) rectangle node[text width=1.05cm, align=center]{{\color{white}Low-rank adapter}} ++(1.5,0.75);

    \node[anchor=west] (A) at (4.3, -9.65) {\includegraphics[scale=0.008]{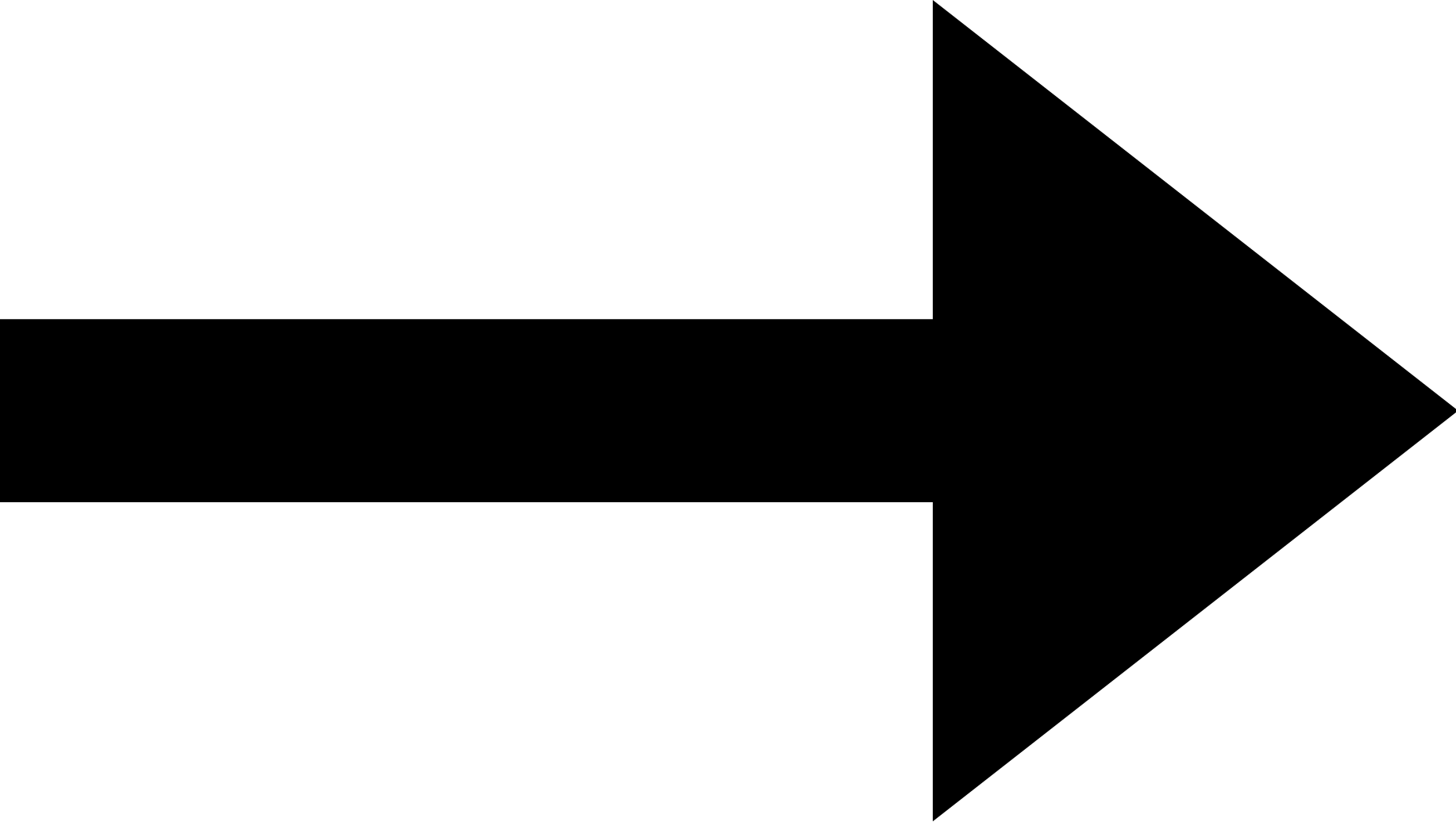} };
    \node[anchor=west] (A) at (6.05, -9.65) {\includegraphics[scale=0.008]{figures/arrowshort.png} };
    \node[anchor=west] (A) at (15.3, -9.65) {\includegraphics[scale=0.008]{figures/arrowshort.png} };
    \node[anchor=west] (A) at (10.55, -9.65) {\Large$\mathbf{\ldots}$};

    \node[anchor=west] (cent) at (-0.85, -9.65) {\includegraphics[scale=0.093]{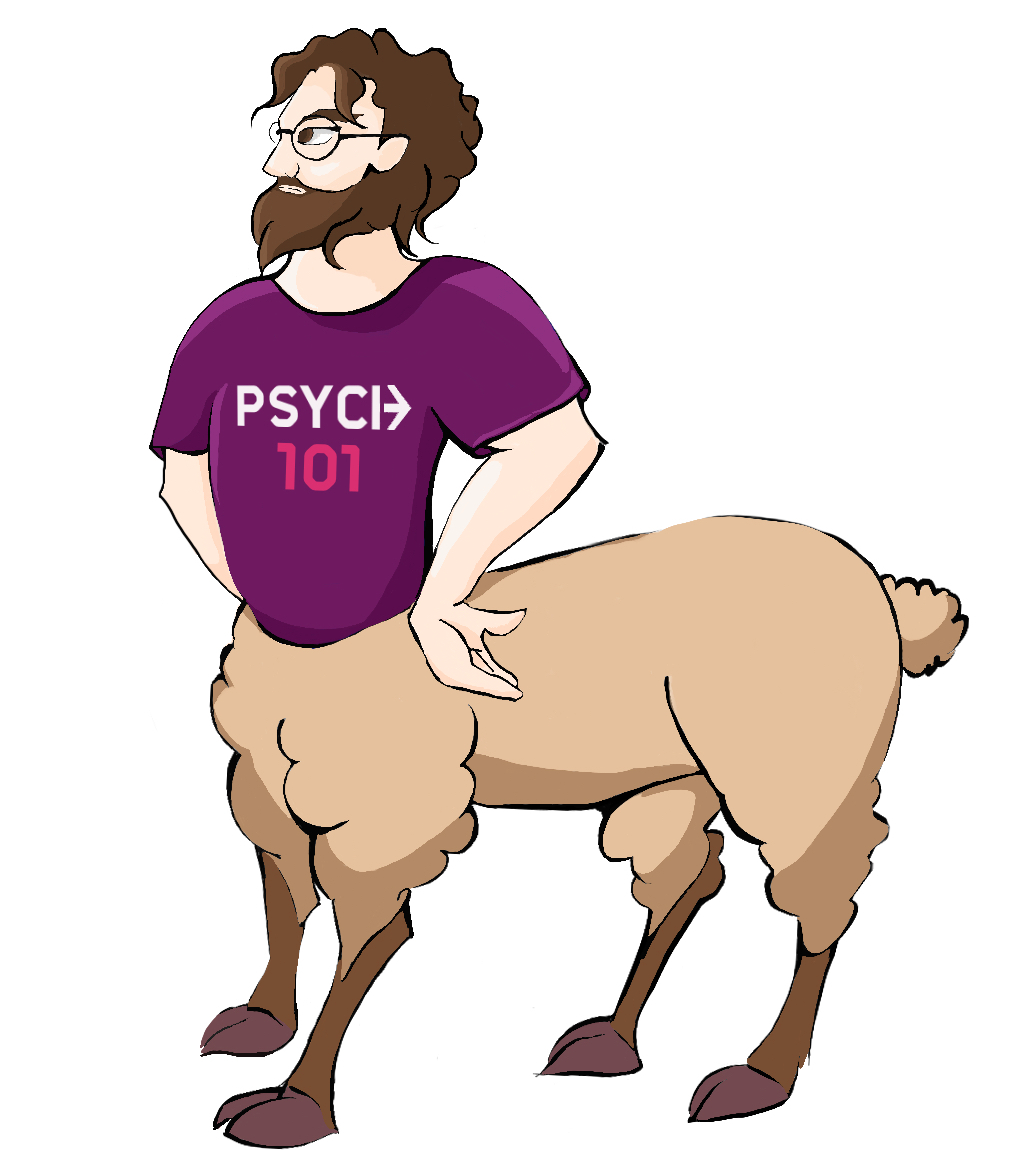}};

    \end{tikzpicture}}
    \caption{Psych-101 and Centaur overview. \textbf{a}, Psych-101 comprises of trial-by-trial data from 160 psychological experiments and 60,092 participants, making 10,681,650 choices in total. It contains domains such as multi-armed bandits, decision-making, memory, supervised learning, Markov decision processes, and others (shown examples are stylized and abbreviated for readability). \textbf{b}, Centaur is a foundation of model human cognition that is obtained by adding low-rank adapters to a state-of-the-art language model and finetuning it on Psych-101.} 
    \label{fig:fig1}
\end{figure}

If we want to understand the human mind in its entirety, we must move from domain-specific theories to an integrated one. The importance of such a unified approach has already been recognized by the pioneers of our field. For example, in 1990, Newell stated that \say{unified theories of cognition are the only way to bring [our] wonderful, increasing fund of knowledge under intellectual control} \cite{Newell1990-NEWUTO}. How can we make meaningful progress toward such theories? 

An important step towards a unified theory of cognition is to build a computational model that can predict and simulate human behavior in any domain \cite{Newell1990-NEWUTO,  riveland2024natural}. The present paper takes up this challenge and introduces Centaur -- the first foundation model of human cognition \cite{bommasani2021opportunities}. Centaur was designed in a data-driven manner by finetuning a state-of-the-art large language model \cite{dubey2024llama} on a large corpus of human behavior. For this purpose, we curated a novel, large-scale data set called Psych-101, covering trial-by-trial data from 160 psychological experiments. We transcribed each of these experiments into natural language, which provides a common format for expressing vastly different experimental paradigms \cite{binz2023using, binz2024turning}. The resulting data set reaches an unprecedented scale, containing over 10,000,000 human choices and including many canonical studies from domains such as multi-armed bandits, decision-making, memory, supervised learning, Markov decision processes, and others (see Figure \ref{fig:fig1}a for an overview and examples).

We subject Centaur to a series of rigorous tests and demonstrate that it captures human behavior at several levels of generalization. First, we show that Centaur predicts behavior of held-out participants (i.e., participants that are not part of the training data) better than existing cognitive models in almost every single experiment. We then demonstrate that its ability to capture human behavior also generalizes to held-out experiments. In this context, we find that Centaur accurately predicts human behavior under modified cover stories, problem structures, and even in entirely novel domains. Finally, we show that Centaur's internal representations become more human-aligned, even though it was never explicitly trained to capture human neural activity. 

Taken together, our results demonstrate that it is possible to discover computational models that capture human behavior across a wide range of domains. We believe that such a predictive model offers many direct opportunities to obtain a better understanding of the human mind \cite{hofman2021integrating, rocca2021putting} and present a case study that demonstrates this potential.

\section*{Results}\label{sec2}

\subsection*{Model overview}

We built Centaur on top of the open-source language model Llama 3.1 70B -- a state-of-the-art model pre-trained by Meta AI \cite{dubey2024llama} (hereafter, we refer to this model simply as Llama). Having a large language model as the backbone allowed us to rely on the vast amounts of knowledge that is present in these models \cite{petroni-etal-2019-language}. The training process involved finetuning on Psych-101 using a parameter-efficient finetuning technique known as quantized low-rank adaptation (QLoRA) \cite{dettmers2024qlora}. QLoRA relies on a frozen four-bit quantized language model as a base model. While the parameters of the base model are left unchanged, it adds so-called low-rank adapters, which contain only a few additional, trainable parameters (typically represented in a half-precision floating-point format). In our case, we added low-rank adapters of rank $r = 8$ to all non-embedding layers (i.e., all linear layers of the self-attention mechanisms and the feedforward networks) as illustrated in Figure \ref{fig:fig1}b. With these settings, the newly added parameters amount to 0.15\% of the base model's parameters. We then trained the model for one epoch on the entire data set using a standard cross-entropy loss. We masked out the loss for all tokens that do not correspond to human responses, thereby ensuring that the model focuses on capturing human behavior and not on completing experimental instructions. The entire training process took approximately five days on an A100 80GB GPU. Further details on the finetuning procedure are provided in the Methods section.

\subsection*{Centaur predicts human behavior better than domain-specific cognitive models}\label{sec3}

\begin{figure}[!ht]
    \centering
    \scalebox{0.74}{\includegraphics[]{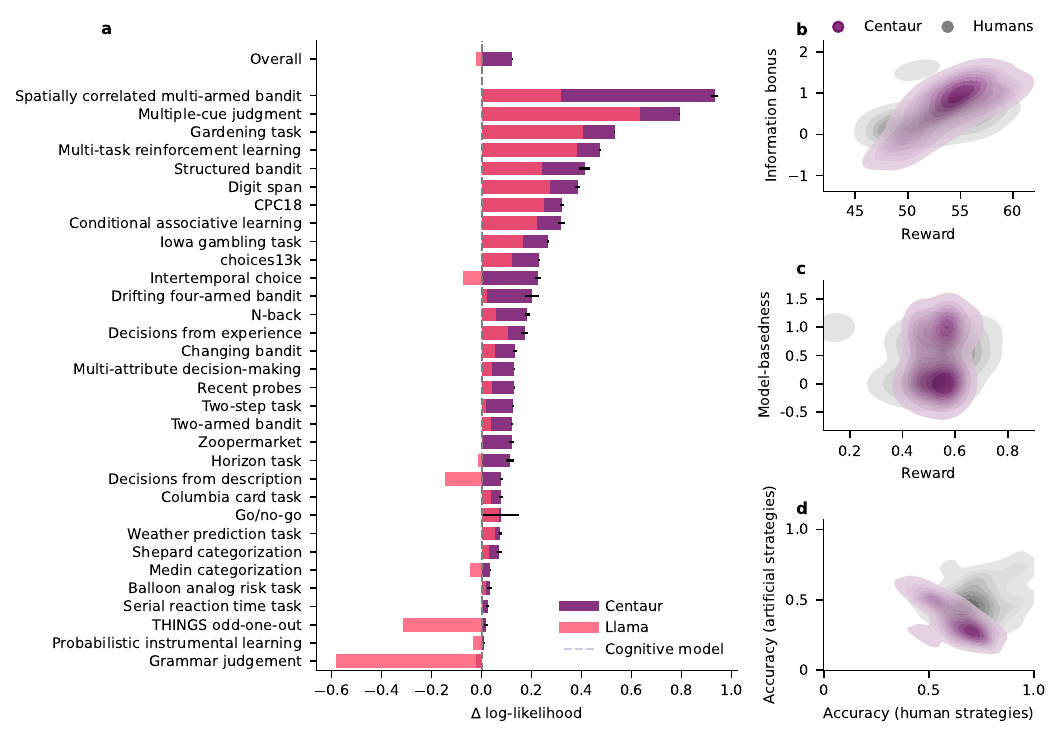}}
    \caption{Goodness-of-fit on Psych-101. \textbf{a}, Difference in log-likelihood of Centaur and Llama relative to a domain-specific cognitive model for each experiment. A value of zero corresponds to the goodness-of-fit of the domain-specific cognitive model while a value above zero indicates improved goodness-of-fit to human responses. Error bars correspond to the standard error of the mean, taken over responses. Centaur outperforms both Llama and a collection of domain-specific cognitive models in almost every experiment. Note that we only included experiments for which we have implemented a domain-specific cognitive model in this graphic and merged different studies using the same paradigm. A full table for all experiments can be found in the Supplementary Information. \textbf{b}, Model simulations on the horizon task. The plot visualizes probability densities over reward and an information bonus parameter for both people and simulated runs of Centaur. \textbf{c}, Model simulations on the two-step task. The plot visualizes probability densities over reward and a parameter indicating how model-based learning was for both people and simulated runs of Centaur.  \textbf{d}, Model simulations on a social prediction game. The plot visualizes probability densities over accuracies of predicting human strategies and strategies of an artificial agent with matched statistics for both people and simulated runs of Centaur.}
    \label{fig:fig2}
\end{figure}

We evaluated Centaur on different types of held-out data to demonstrate that it robustly captures human behavior. In our first analysis, we tested whether it can predict behavior of participants that were not part of the training data. For this, we split each transcribed experiment into two parts, and used 90\% of participants for training and retained 10\% for testing. We measured goodness-of-fit to human choices using negative log-likelihoods averaged across responses. Figure \ref{fig:fig2}a presents the result of this analysis, comparing Centaur against the base model without finetuning and collection of domain-specific models that represent the state-of-the-art in the cognitive science literature. While there was substantial variance in predictability between the experiments, finetuning always improved goodness-of-fit. The average difference in log-likelihoods across experiments after finetuning was $0.14$ (Centaur negative log-likelihood: $0.44$; Llama negative log-likelihood: $0.58$; $t = -144.22, p \leq 0.0001$).

Furthermore, we compared Centaur against the previously mentioned collection of domain-specific cognitive models. These models include, amongst others, the generalized context model \cite{nosofsky2011generalized}, a prospect theory model \cite{peterson2021using}, and various reinforcement learning models \cite{daw2011model, wilson2014humans}. Further technical details can be found in the Supplementary Information. We observed that Centaur outperforms domain-specific cognitive models in all but one experiment. The average difference in predicting human behavior to the domain-specific cognitive models was $0.13$ (cognitive models negative log-likelihood: $0.56$; $t = -178.7, p \leq 0.0001$).


The previous analyses have focused on predicting human responses conditioned on previously executed behavior. We may ask whether Centaur can also generate human-like behavior when simulated in an open-loop fashion (i.e., when feeding its own responses back into the model). This setting arguably provides a much stronger test for the model’s capabilities and is sometimes also referred to as model falsification \cite{palminteri2017importance}. To check whether Centaur survives this test, we ran open-loop simulations in three different experimental paradigms and inspected the distributions of statistics that resulted from these simulations. First, we simulated Centaur on the horizon task paradigm, a two-armed bandit task used to detect different types of exploration strategies \cite{wilson2014humans}. We found that Centaur achieved a performance comparable to human participants, as well that it engaged in a similar level of uncertainty-guided, directed exploration (see Figure \ref{fig:fig2}b), a pattern that is absent in many contemporary language models \cite{binz2023using}. We also observed that Centaur’s behavior is well-aligned with the human population, demonstrating that it does not merely model the behavior of the average participant but rather the distribution over trajectories produced by the entire population. For example, in the two-step task -- a well-known paradigm to tease apart model-free and model-based reinforcement learning \cite{daw2011model} -- Centaur, like human subjects, produced trajectories in which learning is purely model-free, purely model-based, and mixtures thereof (see Figure \ref{fig:fig2}c). Lastly, we verified that Centaur fails at predicting non-human behavior. For this, we considered a study that required participants to predict either human responses or responses of an artificial agent with matched statistics in four canonical economic games \cite{van2022latent}. Mirroring the results of the original human study, Centaur accurately predicted human responses ($64\%$ accuracy) but struggled to predict artificial responses ($35\%$ accuracy; see Figure \ref{fig:fig2}d). Taken together, these results demonstrate that Centaur exhibits human-like characteristics across various settings, confirming that it can generate meaningful open-loop behavior.

\subsection*{Probing increasingly complex generalization abilities}

Thus far, we have shown that Centaur generalizes to previously unseen participants performing experiments that were part of the training data. A true foundation model of human cognition, however, must also capture behavior in any arbitrary experiment, even if that experiment was not part of the training data. To probe whether Centaur has this ability, we exposed it to a series of increasingly complex out-of-distribution evaluations.

\begin{figure}
    
    \scalebox{0.7}{\begin{minipage}{\textwidth}%
    \includegraphics[]{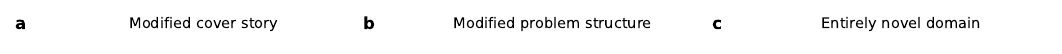} \\[-0.5cm]
    \begin{tikzpicture}     
        \node[] at (-3.1, 0) (b1) {$~$};
        \node[] at (0., 0) (b1) {\includegraphics[width=5.3cm]{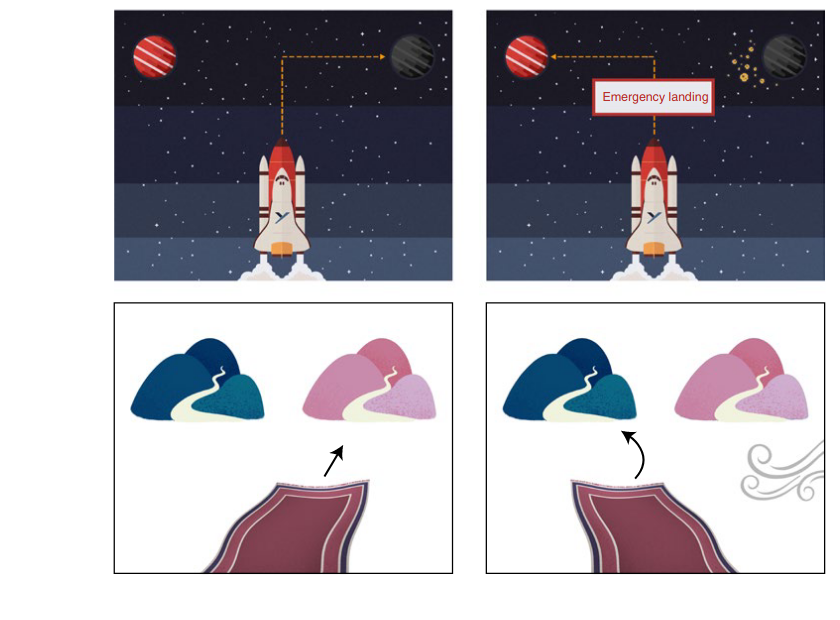} };
        \node[] at (11.9, 0.) (b1) {\includegraphics[width=4.9cm]{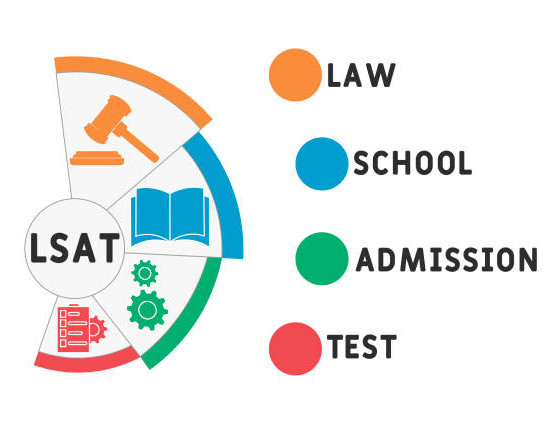}};
        \node[] at (6.1, 0.1) (b1) {\scalebox{0.715}{\begin{tikzpicture}     
        \footnotesize     

        \node[draw, minimum width=0.5cm,minimum height=0.5cm, BrickRed] at (7.25, 0) (b1) {};
        \node[draw, minimum width=0.5cm,minimum height=0.5cm, BrickRed] at (7.25, 0.5) (b2) {};
        \node[draw, minimum width=0.5cm,minimum height=0.5cm, BrickRed] at (7.25, 1.) (b3) {};
        \node[draw, minimum width=0.5cm,minimum height=0.5cm, BrickRed] at (7.25, 1.5) (b4) {};
        \node[draw, minimum width=0.5cm,minimum height=0.5cm, BrickRed] at (7.25, 2.0) (b5) {};
        \node[draw, minimum width=0.5cm,minimum height=0.5cm, BrickRed] at (7.25, 2.5) (b6) {};
        \node[draw, minimum width=0.5cm,minimum height=0.5cm, BrickRed, fill=gray!30] at (7.25, 3) (b7) {};
        \node[draw, minimum width=0.5cm,minimum height=0.5cm, BrickRed, fill=gray!30] at (7.25, 3.5) (b8) {};
        \node[draw, minimum width=0.5cm,minimum height=0.5cm, BrickRed, fill=gray!30] at (7.25, 4) (b9) {};
        \node[draw, minimum width=0.5cm,minimum height=0.5cm, BrickRed, text=black] at (7.25, 4.5) (b10) {51};
        
        \node[draw, minimum width=0.5cm,minimum height=0.5cm, ForestGreen] at (8.0,0) (b1) {};
        \node[draw, minimum width=0.5cm,minimum height=0.5cm, ForestGreen] at (8.0,0.5) (b2) {};
        \node[draw, minimum width=0.5cm,minimum height=0.5cm, ForestGreen] at (8.0,1.) (b3) {};
        \node[draw, minimum width=0.5cm,minimum height=0.5cm, ForestGreen] at (8.0,1.5) (b4) {};
        \node[draw, minimum width=0.5cm,minimum height=0.5cm, ForestGreen] at (8.0,2.0) (b5) {};
        \node[draw, minimum width=0.5cm,minimum height=0.5cm, ForestGreen] at (8.0,2.5) (b6) {};
        \node[draw, minimum width=0.5cm,minimum height=0.5cm, ForestGreen, text=black] at (8.0,3) (b7) {26};
        \node[draw, minimum width=0.5cm,minimum height=0.5cm, ForestGreen, text=black] at (8.0,3.5) (b8) {40};
        \node[draw, minimum width=0.5cm,minimum height=0.5cm, ForestGreen, text=black] at (8.0,4) (b9) {39};
        \node[draw, minimum width=0.5cm,minimum height=0.5cm, ForestGreen, fill=gray!30] at (8.0,4.5) (b10) {};

        \node[] at (9.125,2.25) (b10) {\includegraphics[scale=0.01]{figures/arrowshort.png}};

        \node[draw, minimum width=0.5cm,minimum height=0.5cm, BrickRed] at (10.25, 0) (b1) {};
        \node[draw, minimum width=0.5cm,minimum height=0.5cm, BrickRed] at (10.25, 0.5) (b2) {};
        \node[draw, minimum width=0.5cm,minimum height=0.5cm, BrickRed] at (10.25, 1.) (b3) {};
        \node[draw, minimum width=0.5cm,minimum height=0.5cm, BrickRed] at (10.25, 1.5) (b4) {};
        \node[draw, minimum width=0.5cm,minimum height=0.5cm, BrickRed] at (10.25, 2.0) (b5) {};
        \node[draw, minimum width=0.5cm,minimum height=0.5cm, BrickRed] at (10.25, 2.5) (b6) {};
        \node[draw, minimum width=0.5cm,minimum height=0.5cm, BrickRed, fill=gray!30] at (10.25, 3) (b7) {};
        \node[draw, minimum width=0.5cm,minimum height=0.5cm, BrickRed, fill=gray!30] at (10.25, 3.5) (b8) {};
        \node[draw, minimum width=0.5cm,minimum height=0.5cm, BrickRed, fill=gray!30] at (10.25, 4) (b9) {};
        \node[draw, minimum width=0.5cm,minimum height=0.5cm, BrickRed, text=black] at (10.25, 4.5) (b10) {51};
        
        \node[draw, minimum width=0.5cm,minimum height=0.5cm, ForestGreen] at (11,0) (b1) {};
        \node[draw, minimum width=0.5cm,minimum height=0.5cm, ForestGreen] at (11,0.5) (b2) {};
        \node[draw, minimum width=0.5cm,minimum height=0.5cm, ForestGreen] at (11,1.) (b3) {};
        \node[draw, minimum width=0.5cm,minimum height=0.5cm, ForestGreen] at (11,1.5) (b4) {};
        \node[draw, minimum width=0.5cm,minimum height=0.5cm, ForestGreen] at (11,2.0) (b5) {};
        \node[draw, minimum width=0.5cm,minimum height=0.5cm, ForestGreen] at (11,2.5) (b6) {};
        \node[draw, minimum width=0.5cm,minimum height=0.5cm, ForestGreen, text=black] at (11,3) (b7) {26};
        \node[draw, minimum width=0.5cm,minimum height=0.5cm, ForestGreen, text=black] at (11,3.5) (b8) {40};
        \node[draw, minimum width=0.5cm,minimum height=0.5cm, ForestGreen, text=black] at (11,4) (b9) {39};
        \node[draw, minimum width=0.5cm,minimum height=0.5cm, ForestGreen, fill=gray!30] at (11,4.5) (b10) {};

        \node[draw, minimum width=0.5cm,minimum height=0.5cm, MidnightBluee] at (11.75,0) (b1) {};
        \node[draw, minimum width=0.5cm,minimum height=0.5cm, MidnightBluee] at (11.75,0.5) (b2) {};
        \node[draw, minimum width=0.5cm,minimum height=0.5cm, MidnightBluee] at (11.75,1.) (b3) {};
        \node[draw, minimum width=0.5cm,minimum height=0.5cm, MidnightBluee] at (11.75,1.5) (b4) {};
        \node[draw, minimum width=0.5cm,minimum height=0.5cm, MidnightBluee] at (11.75,2.0) (b5) {};
        \node[draw, minimum width=0.5cm,minimum height=0.5cm, MidnightBluee] at (11.75,2.5) (b6) {};
        \node[draw, minimum width=0.5cm,minimum height=0.5cm, MidnightBluee, text=black, fill=gray!30] at (11.75,3) (b7) {};
        \node[draw, minimum width=0.5cm,minimum height=0.5cm, MidnightBluee, text=black, fill=gray!30] at (11.75,3.5) (b8) {};
        \node[draw, minimum width=0.5cm,minimum height=0.5cm, MidnightBluee, text=black, fill=gray!30] at (11.75,4) (b9) {};
        \node[draw, minimum width=0.5cm,minimum height=0.5cm, MidnightBluee, fill=gray!30] at (11.75,4.5) (b10) {};
        
\end{tikzpicture}} };
    \end{tikzpicture}
    \includegraphics[]{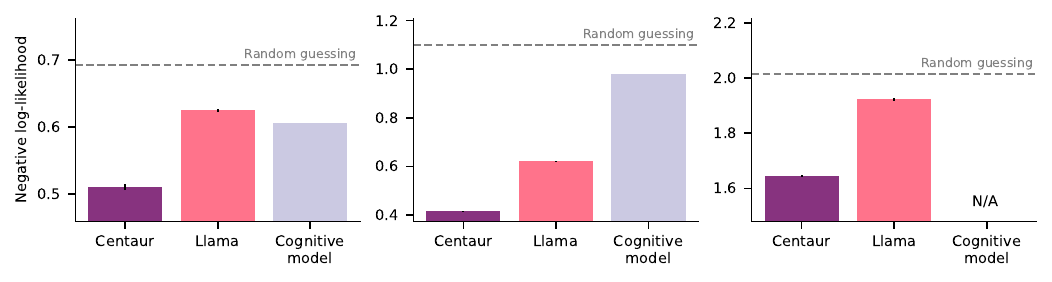}  
    \end{minipage}}
    \caption{Evaluation in different held-out settings. \textbf{a}, Negative log-likelihoods for the two-step task with a modified cover story \cite{feher2020humans}. \textbf{b}, Negative log-likelihoods for a three-armed bandit experiment \cite{dubois2022value}. \textbf{c}, Negative log-likelihoods for an experiment probing logical reasoning \cite{jansen2021rational} with items based on the Law School Admission Test (LSAT). Centaur outperforms both Llama and domain-specific cognitive models when faced with modified cover stories, problem structures, and entirely novel domains. Error bars correspond to the standard error of the mean, taken over responses.}
    \label{fig:fig3}
\end{figure}

First, we investigated whether Centaur is robust in the face of changes to the cover story. For this analysis, we relied on data collected by Feher da Silva and Hare \cite{feher2020humans}, who conducted a study using the aforementioned two-step task. In addition to the canonical cover story (spaceships traveling to foreign planets in search of treasures), their study introduced a novel cover story involving magical carpets. Importantly, Psych-101 includes experiments using the canonical spaceship cover story \cite{kool2016does, kool2017cost} but no experiments with the magical carpet cover story. Yet, we still found that Centaur captures human behavior in the magical carpet experiment of Feher da Silva and Hare (see Figure \ref{fig:fig3}a). Like in our previous analysis, we observed an improvement after finetuning, as well as a favorable goodness-of-fit when compared to a domain-specific cognitive model (Centaur negative log-likelihood: $0.51$; Llama negative log-likelihood: $0.63$; cognitive model negative log-likelihood: $0.61$; all pairwise comparisons significant with $p \leq 0.0001$). 

In a second out-of-distribution evaluation, we probed whether Centaur is robust to modifications in task structure. To test this, we exposed it to a paradigm known as Maggie's farm \cite{dubois2022value}. Maggie's farm extends the horizon task paradigm by adding a third choice option. Psych-101 encompasses several two-armed bandit experiments (including the horizon task) but not Maggie's farm or any other three-armed bandit experiments.\footnote{It does, however, contain multi-armed bandit experiments with more than three choice options.} Thus, this analysis provides a test of Centaur's robustness to structural task modifications. We found that Centaur captures human behavior on Maggie's farm as shown in Figure \ref{fig:fig3}b. We again observed a benefit of finetuning, as well as a favorable goodness-of-fit compared to a domain-specific cognitive model, which did not generalize well to this setting (Centaur negative log-likelihood: $0.42$; Llama negative log-likelihood: $0.62$; cognitive model negative log-likelihood: $0.98$; all pairwise comparisons significant with $p \leq 0.0001$). 

Finally, we investigated whether Centaur can capture human behavior even in entirely novel domains. In this context, we considered a study investigating logical reasoning \cite{jansen2021rational}. While Psych-101 includes probabilistic and causal reasoning problems, we purposefully excluded any studies involving logical reasoning. Like in the previous analyses, there was again a positive effect of finetuning (Centaur negative log-likelihood: $1.65$; Llama negative log-likelihood: $1.92$; $t = -50.39, p \leq 0.0001$; see Figure \ref{fig:fig3}c). Note that we did not compare to any domain-specific cognitive model in this setting, as it is unclear how to construct a model that would make any meaningful transfer from training data that does not include any related problems.

We consolidated these results by analyzing Centaur on six additional out-of-distribution experimental paradigms that were not part of the training data in any shape or form (including moral decision-making \cite{awad2018moral}, economic games \cite{akata2023playing},  naturalistic category and reward learning \cite{demircan2024evaluating}, behavioral propensities \cite{singh2022representing}, and a deep sequential decision task \cite{xu2021novelty}). Centaur robustly captured human behavior in all of these settings, while smaller and non-finetuned models did not do so consistently. We present the corresponding results in the Supplementary Information.

In addition to analyzing human choice data, we also examined whether Centaur can predict human response times. Hick's law \cite{hick1952rate} suggests that individual response times are a linear function of the response entropy. Therefore, we extracted nearly 4,000,000 response times for a subset of experiments in Psych-101 and fitted three linear mixed effects models each predicting response time based on the response entropy derived from a different model. We found that the response entropy derived from Centaur captured a larger proportion of the variance in response times (conditional R$^2$: 0.58) than those derived from Llama (conditional R$^2$: 0.4) and the cognitive models (conditional R$^2$: 0.38), thereby highlighting Centaur’s ability to predict measures beyond pure choice data.

To demonstrate that the model does not degrade on problems it was pretrained for, we furthermore verified it on a collection of benchmarks from the machine learning literature \cite{pmlr-v235-coda-forno24a, kipnis2024texttt}. We found that Centaur remains stable in performance-based benchmarks, even improving over the base model in some of them \cite{kipnis2024texttt}. Finally, in benchmarks that measure human alignment, we observed a shift towards human-like characteristics. Figure \ref{fig:fig4}a depicts this improved alignment on a low-dimensional embedding derived from ten behavioral metrics in CogBench, a benchmark to test the cognitive abilities of large language models \cite{pmlr-v235-coda-forno24a}. A more detailed description of these additional benchmarking results can be found in the Supplementary Information.

\subsection*{Internal representations become more aligned to human neural activity}

\begin{figure}
    \centering
    \scalebox{0.74}{\includegraphics[]{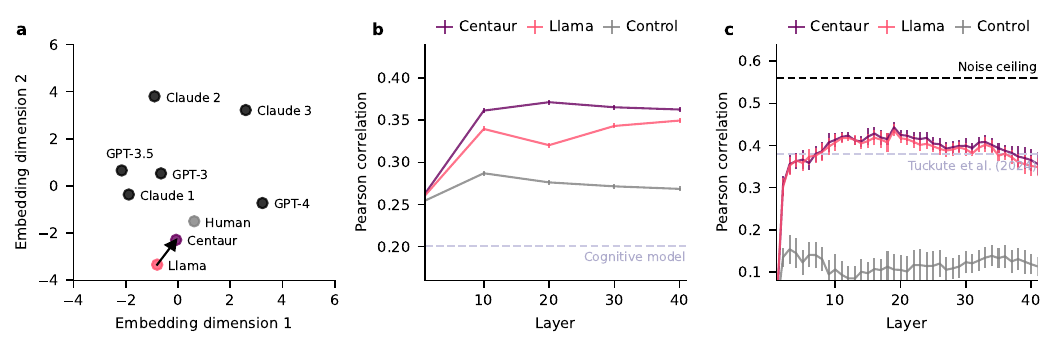}}    
    \caption{Human alignment. \textbf{a}, Multidimensional scaling embedding of the ten behavioral metrics in CogBench \cite{pmlr-v235-coda-forno24a} for different models. \textbf{b}, Pearson correlation coefficients indicating how well human neural activity in the two-step task \cite{feher2023rethinking} can be decoded using Centaur's internal representations extracted from different layers. \textbf{c}, Pearson correlation coefficients indicating how well human neural activity in a sentence-reading task \cite{tuckute2024driving} can be decoded using Centaur's internal representations extracted from different layers. Control refers to a model that uses representations extracted from a randomly-initialized transformer model with matched architecture.}
    \label{fig:fig4}
\end{figure}

Despite only being trained to match human behavior, we also wondered whether Centaur’s internal representations become more aligned with human neural activity. To check whether this is the case, we conducted two analyses in which we predicted human neural activity using the model's internal representations \cite{yamins2014performance, schrimpf2021neural}. We first conducted a whole-brain analysis, in which we predicted fMRI measurements of people performing the two-step task \cite{feher2023rethinking}. For this, we relied on data collected in a previous study \cite{feher2023rethinking}, involving 94 participants each making 300 choices. Participants were tested on either the magical carpet cover story (that we already used in one of our earlier generalization analyses) or an abstract cover story. Neither of these two cover stories were part of Centaur’s training data. We extracted recordings from models' residual stream before each choice and after feedback. We then aggregated human neural activity in each region and regressed the aggregated activity on Centaur's internal representations. This procedure was then repeated separately for each participant and region. Further details can be found in the Methods section. Figure \ref{fig:fig4}b shows the resulting Pearson correlation coefficients across layers for both Centaur and Llama. We found that Centaur's representations consistently outperform Llama's representations in predicting human neural activity (all $p \leq 0.001$), suggesting that finetuning a model on large-scale behavioral data aligned its internal representations to human neural activity. It is worth noting that this type of analysis was only possible due to the expressivity of Centaur's representations, and that using representations of a traditional cognitive model led to a substantially drop in performance (dashed line in Figure \ref{fig:fig4}b).

We expanded these results in a second analysis, for which we relied on a previously collected data set involving fMRI measurements of people reading simple, six-word sentences, such as \say{That is such a beautiful picture!}\cite{tuckute2024driving}. We focused our analysis on a subset of five participants who each passively read 1,000 sentences, spread across 20 experimental runs and two scanning sessions. The presented sentences were extracted from nine corpora and selected to maximize semantic diversity. We closely followed the protocol of the original study and predicted aggregated neural activity across participants in the language network. We repeated this procedure for representations extracted from different layers in both Centaur and Llama. Predictability peaked at around layer 20 as shown in Figure \ref{fig:fig4}c. This peak is consistent with the hypothesis that the intermediate layers of such models contain the most amount of information \cite{lad2024remarkable}.  We performed an inverse-weighted meta-analysis \cite{cochran1954combination} on the difference in correlations between Centaur and Llama which indicated that there was a significant benefit of finetuning when pooling across layers ($\beta = 0.007$, 95\% CI $[0.0002, 0.013]$, $p$ = $0.045$). While this effect was consistent across layers, it was not statistically significant for any individual layer. Yet, this analysis demonstrates that Centaur's alignment to human neural activity increases (although only marginally) even for stimuli far away from its training data, thereby highlighting the generalization abilities of the model. 

\subsection*{Model-guided scientific discovery using Psych-101 and Centaur} 

\begin{figure}
    \centering
    \scalebox{0.7}{
\begin{tikzpicture}[font={\fontfamily{phv}\selectfont\footnotesize\sffamily}]

    \node[] (A) at (2.7, 5) {\color{black}\begin{tabular}{lcc}
        & \includegraphics[height=1cm]{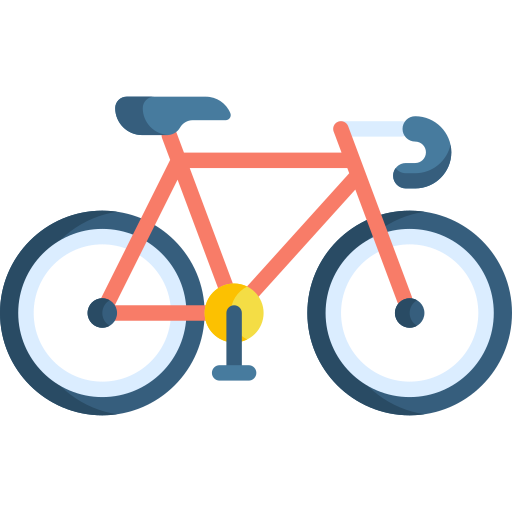}  & \includegraphics[height=1cm]{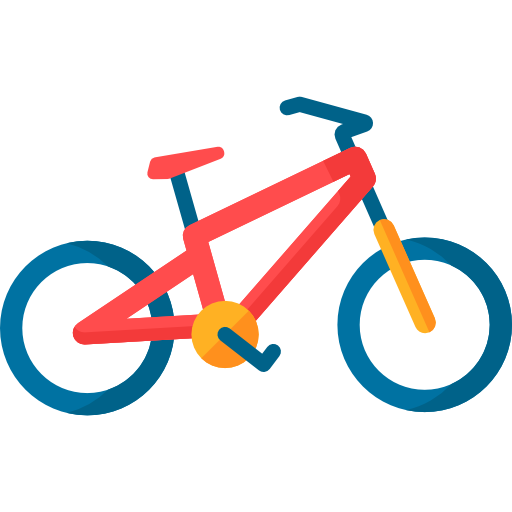} \\
        Expert 1 (90\%) & 1 & 0\\
        Expert 2 (80\%) & 0 & 1\\
        Expert 3 (70\%) & 0 & 1\\
        Expert 4 (60\%) & 1 & 1\\
    \end{tabular}};
    \node[] (A) at (2.7, 3.5) {\color{black}Which product is superior in quality?};

    \node[] (A) at (2.7, 6.3) {\color{black}\textbf{Original study}};

    \node[] (A) at (8.7, 6.3) {\color{black}\textbf{DeepSeek-R1 explanation}};

    \node[text width=4cm, align=justify] (A) at (8.7, 4.65) {\color{black}The participant employed a two-step decision-making strategy. First, they determined which product had the majority of positive ratings (1s) across all experts. If the products were tied in the number of positive ratings, the participant then considered the rating from the highest validity expert to break the tie.};

    \node[] (A) at (15.5, 6.3) {\color{black}\textbf{Scientific regret minimization}};

    \node[text width=5cm] (A) at (15.8, 4.75) {\color{black}\begin{tabular}{ccc}
        \toprule
        \textbf{Product A} & \textbf{Product B} & \textbf{Response} \\
        \midrule
        1 0 0 1 & 0 1 1 1 & A\\
        0 1 1 0 & 1 0 0 0 & B\\
        0 1 1 0 & 1 0 0 0 & B\\
        1 0 0 0 & 0 1 1 0 & A\\
        0 1 1 1 & 1 0 0 1 & B\\
        1 0 0 1 & 0 1 1 1 & A\\
        0 1 1 0 & 1 0 0 0 & B \\
        \bottomrule
\end{tabular}};
    
    \node[anchor=west] (A) at (0, 0) {\includegraphics[]{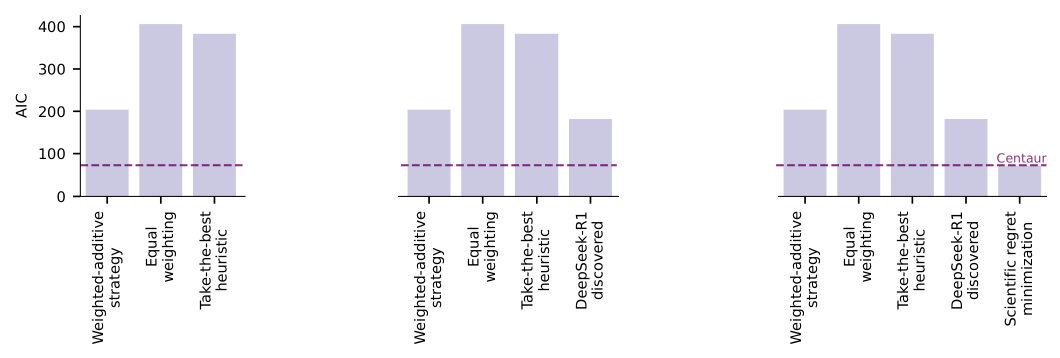}};

    \node[] (A) at (5.5, 0.) {\includegraphics[scale=0.015]{figures/arrowshort.png}};

    \node[] (A) at (11.8, 0.) {\includegraphics[scale=0.015]{figures/arrowshort.png}};

    \node[] (A) at (5.5, 0.8) {\includegraphics[height=0.45cm]{figures/psych101.png}};

    \node[] (A) at (11.8, 1.25) {\includegraphics[height=1.5cm]{figures/centaur_cropped.png}};
    
\end{tikzpicture}}
    \caption{Model-guided scientific discovery. We used Psych-101 and Centaur to guide the development of a cognitive model for a multi-attribute decision-making study. Each panel shows the Akaike information criterion (AIC) for the set of models considered at the given stage, starting with the models considered in the original study. We first asked DeepSeek-R1 to generate an explanation for human responses and formalized the resulting verbal strategy into a formal computational model. We then further refined this model through scientific regret minimization using Centaur as a reference model. Eight data points are visualized for which Centaur makes accurate predictions but the DeepSeek-R1-discovered model does not. We then used this information to design a domain-specific cognitive model that is as predictive as Centaur yet still interpretable.}
    \label{fig:deepseek}
\end{figure}

Psych-101 and Centaur both constitute valuable tools for scientific discovery. In the following section, we present an example for how each of them can be used to improve our understanding of human decision-making. The individual steps of this process are illustrated in Figure \ref{fig:deepseek}.

Psych-101 contains human behavioral data in a natural language format, which means that it can be readily processed and analyzed by language-based reasoning model, such as DeepSeek-R1 \cite{deepseekai2025deepseekr1incentivizingreasoningcapability}. To demonstrate this use case, we asked DeepSeek-R1 to generate an explanation of participants’ behavior in a multi-attribute decision-making experiment \cite{hilbig2014generalized}. The model produced several explanations, one of which caught our attention: \say{The participant employed a two-step decision-making strategy. First, they determined which product had the majority of positive ratings (1s) across all experts. If the products were tied in the number of positive ratings, the participant then considered the rating from the highest validity expert to break the tie.} This strategy combines two well-known heuristic decision-making strategies that -- best to our knowledge -- have not been considered in this particular combination before. We then took this verbal strategy, implemented it as a formal computational model, and found that it indeed explained human response behavior more accurately than the three strategies considered in the original study (a weighted-additive strategy, equal weighting, and the take-the-best heuristic). However, the DeepSeek-R1-discovered model (AIC: $181.7$) still fell short of the goodness-of-fit of Centaur (AIC: $72.5$), indicating that there is still room for improvement. We, therefore, turned towards a method known as scientific regret minimization, which uses a black-box predictive model as a reference to identify responses that are in principle predictable but are not captured by a given model \cite{agrawal2020scaling}. Typically, scientific regret minimization requires the collection of a large-scale experiment-specific data set to train this predictive model. Centaur, on the other hand, can be used out-of-the-box and without the need to collect any domain-specific data, thereby circumventing this step and broadening the scope of scientific regret minimization considerably (indeed, the multi-attribute decision-making data set under consideration contains less than a hundred participants, placing it far out of reach for traditional scientific regret minimization). When inspecting the responses that are well-predicted by Centaur but not the DeepSeek-R1-discovered model, we observed that they all involve choices in which participants choose the option with fewer positive ratings overall but which was rated positively by a higher validity expert. This pattern suggests that the switch between the two heuristics is likely not as strict as initially suggested by the DeepSeek-R1-discovered strategy. To capture this, we replaced the either-or rule with a weighted average of both heuristics. We found that the model that resulted from this process matched Centaur in terms of its goodness-of-fit (AIC: $71.7$) while still being interpretable. Notably, the final result of this model comparison stands in contrast to the one that was conducted with the original set of models and indicates that people rely on a combination of heuristics when making decisions as opposed to following a weighted-additive strategy \cite{binz2022heuristics}.

\section*{Discussion}\label{sec12}

The present paper introduced Centaur, the first foundation model of human cognition. We obtained Centaur by finetuning a state-of-the-art language model on Psych-101 -- a novel, large-scale data set of human behavior. This approach allowed us to leverage the vast knowledge embedded in large language models, while, at the same time, aligning them with human behavior \cite{binz2024turning}. Centaur successfully captures human behavior and passes a wide range of out-of-distribution checks. It generalizes not only to unseen participants but also to new cover stories, structural variations, and entirely novel domains. In addition to analyzing the model on a behavioral level, we also conducted a series of analyses on its internal representations, in which we found an increased alignment with human neural activity.

We furthermore conducted a case study demonstrating how both Psych-101 and Centaur can be used for guiding the development of predictive, yet interpretable, cognitive models. Looking beyond this example, Centaur finds many additional applications in the context of automated cognitive science \cite{musslick2024automatingpracticescience, musslick2024closed, rmus2025towards}. It may, for instance, be used for in-silico prototyping of experimental studies \cite{dillion2023can}. In this context, one could use the model to figure out which designs lead to the largest effect sizes, how to design a study to reduce the number of required participants, or to estimate the power of an effect. 

While the present paper takes initial steps in leveraging Centaur to gain deeper insights into human cognition, it also opens up exciting new avenues for future exploration. First, one could further probe Centaur's internal representations to understand how it represents knowledge and processes information. The resulting insights could, in turn, be used to generate new hypotheses about knowledge representation and information processing in humans that could be validated in future experimental studies. We believe that tools like sparse autoencoders \cite{huben2024sparse} and attention map visualization \cite{Chefer_2021_CVPR} provide promising avenues towards accomplishing this goal and hope to explore them in future studies.

In addition, it might also be possible to train models with different architectures from scratch using the data set that we have created in the process of this paper. Doing so would enable us to investigate the neural architecture of human cognition at a scale that could not have been done before. We might, for example, ask questions such as whether human information processing is better described by attention-based architectures \cite{vaswani2017attention} or architectures with a vector-based memory \cite{smithsimplified}, or how much we can improve by incorporating theories from the neuroscience literature \cite{zador2023catalyzing, doerig2023neuroconnectionist}. We expect an eventual outcome of such an approach to contain both domain-specific and domain-general modules, thereby allowing us to investigate the interplay between the two.

Even though Psych-101 is already the broadest and largest data set of human behavior out there, we view its development as an ongoing process and plan to further develop it. The focus in its current state is largely on learning and decision-making, but eventually, we intend to include additional domains such as psycholinguistics, social psychology, and economic games. Experiments with information about individual differences are another source of neglected data in the current iteration of Psych-101. Ideally, we want to include all types of relevant information about subjects (e.g., age, personality traits, or socioeconomic status) in the prompt, such that a model trained on this data can capture individual differences. Experiments from developmental psychology or computational psychiatry provide an ideal source for this purpose. Finally, while we have already included some cross-cultural and meta-studies \cite{ruggeri2022globalizability, wulff2018meta, frey2017risk, enkavi2019large}, the current iteration has still a strong bias toward a Western, educated, industrialized, rich, and democratic (WEIRD) population \cite{henrich2010most}. To address all of these shortcomings, we have created an open-source repository and invite everyone to contribute to the next iteration of Psych-101 in an open research collaboration (\url{https://github.com/marcelbinz/Psych-201}). The goal of this effort is to provide psychological data in a standardized format that facilitates benchmarking, thereby complementing existing efforts from the neuroscience community \cite{schrimpf2020integrative, schrimpf2018brain, poldrack2024past, markiewicz2021openneuro}. While the natural language format (together with quite a bit of reverse-engineering) used in this work allows us to express a vast range of experimental paradigms, it introduces a selection bias against experiments that cannot be expressed in natural language. The long-term objective should be to move towards a multi-modal data format instead \cite{schulze2025visual}.

\section*{Conclusion}

When the idea of a unified model of cognition was first proposed, researchers expressed concern that established areas of cognitive science might react negatively to such a model. In particular, they feared that the new approach might be seen as unfamiliar or incompatible with existing theories, just like an \say{intruder with improper pheromones} \cite{vere1992cognitive}. This could lead to an \say{attack of the killer bees}, where researchers in traditional fields would fiercely critique or reject the new model to defend their established approaches. To mitigate these concerns, the concept of a cognitive decathlon was proposed: a rigorous evaluation framework in which competing models of cognition are tested across ten experiments and judged based on their cumulative performance in them. In the current work, we applied Centaur to the equivalent of sixteen such cognitive decathlons, where it was tested against numerous established models and consistently won every competition. This outcome suggests that the data-driven discovery of domain-general models of cognition is a promising research direction. The next step for future research should be to translate this domain-general computational model into a unified theory of human cognition as envisioned by Newell \cite{Newell1990-NEWUTO}.

\newpage 

\section*{Methods}\label{sec11}

\subsection*{Data collection}

We constructed Psych-101 by transcribing data from 160 psychological experiments into natural language. Each prompt was designed to include the entire trial-by-trial history of a complete session from a single participant. The included experiments were selected using the following criteria: (a) publicly available data on a trial-by-trial level, (b) possibility of transcription into text without a significant loss of information, and (c) coverage of a broad spectrum of domains. The transcription of each experiment was done manually by the authors. We designed our natural language prompts using the following principles: (a) instructions follow the original study as closely as possible, (b) simplifications were made where appropriate, and (c) a maximum prompt length of roughly 32,768 tokens. Full information about all included experiments is provided in the Supplementary Information.

\subsection*{Finetuning procedure}

We used Llama 3.1 70B as the base model for our finetuning procedure. We relied on a parameter-efficient finetuning technique known as QLoRA \cite{dettmers2024qlora}. QLoRA adds so-called low-rank adapters to each layer of a four-bit quantized base model. The base model was kept fixed during finetuning and only the parameters of the low-rank adapters were adjusted. We added low-rank adapters to all linear layers of the self-attention mechanisms and the feedforward networks. Each low-rank adapter modifies the forward pass as follows:
\begin{align*}
    \mathbf{Y} = \mathbf{X}\mathbf{W} + \alpha \mathbf{X} \mathbf{L}_1 \mathbf{L}_2 \qquad  \qquad
    \mathbf{W} \in \mathbb{R}^{h \times o}; ~ \mathbf{L}_1 \in \mathbb{R}^{h \times r}; ~ \mathbf{L}_2 \in \mathbb{R}^{r \times o}
\end{align*}

where $\mathbf{X}\mathbf{W}$ is the (quantized) linear transformation of the base model and $\mathbf{X} \mathbf{L}_1 \mathbf{L}_2$ the low-rank adapter. The hyperparameter $\alpha$ controls the trade-off between the two. Low-rank adapter computations are performed in half-precision floating-point format. For further details on this technique, we refer the reader to \cite{dettmers2024qlora}. 

We finetuned the model for one epoch on the entire data set using a standard cross-entropy loss (we experimented with prolonged training but found that this led to overfitting). We only backpropagated the loss at human responses and masked out the loss for all other tokens. The effective batch size was set to 32, the learning rate to 0.00005, and the weight decay to 0.01. We used an 8-bit AdamW optimizer \cite{loshchilov2019decoupledweightdecayregularization} with a linearly increasing warm up over the first 100 gradient steps. The finetuning procedure was implemented using the unsloth library (\url{https://unsloth.ai/}). 

We have also trained a smaller version of Centaur -- called Minitaur -- that uses Llama 3.1 8B as base model following the same recipe. Minitaur captures human behavior close to its training distribution but generalizes less robustly than the larger model to out-of-distribution experiments. Nevertheless, we believe that Mintaur is useful for prototyping as it does not require access to any specific hardware (it runs, for instance, on the free GPU instances in Google Colab). More detailed results for this model are provided in the Supplementary Information.

\subsection*{Evaluation metric}

We use (negative) log-likelihoods averaged over responses as our evaluation metric. For experiments with multi-token responses, we summed log-likelihoods within a response and averaged across responses.

\subsection*{Domain-specific cognitive models}

We selected 14 cognitive and statistical models that together cover most of the experiments in Psych-101 as our baseline models. Further details regarding the included models and their specifications are provided in the Supplementary Information. 

For our main analysis, we were interested in predicting the behavior of held-out participants. Therefore, we fitted a joint set of parameters for all participants in the training data and evaluated how well a model with these parameters predicts responses of held-out participants. Mirroring the evaluation metric for the language-based models, we evaluated goodness-of-fit using (negative) log-likelihoods averaged over responses. 

For the out-of-distribution evaluations, we fitted model parameters using the most similar experiment in the training set, and then evaluated how well a model with the resulting parameters predicts human responses in the unseen setting. The most similar experiment for the magical carpet version of the two-step task was a two-step task experiment with the default spaceship cover story. The most similar experiment for Maggie's farm was the horizon task. We included no baseline model for the logical reasoning task, as none of the experiments in the training data were similar to it.

\subsection*{Neural alignment}

The neural alignment analysis on the two-step task was conducted using data collected in a previous study \cite{feher2023rethinking}. We used a regularized linear regression model to predict fMRI data from internal representations of Centaur and Llama (a separate model was used for each participant and region). We fitted each of these models on data from two scanning blocks and evaluated them on data from the third. The regularization strength was selected using a nested cross-validation procedure.  For each run, we split the beta maps into cortical and subcortical regions of interest (ROI) using the Schaefer 2018 atlas \cite{schaefer2018local}. We averaged the betas within each ROI, reducing the number of betas from the number of voxels to the number of ROIs. All cortical and subcortical ROIs from the atlas were evaluated. Reported Pearson correlation coefficients correspond to the average across all ROIs. \\
Internal representations were extracted from the models' residual stream and transformed using a principal component analysis. We set the number of retained components such that they explain 95\% of the variance. \\
The fMRI data were preprocessed using fMRIPrep 24.0 \cite{esteban2019fmriprep}. We used the default settings of fMRIPrep, and all the scans were aligned to the MNI152NLin2009cAsym atlas \cite{fonov2009unbiased}. To extract effect estimates for each subtrial of the task (e.g., second step of the fifth trial, feedback of the tenth trial), we built separate general linear models (GLMs). Each GLM included the subtrial of interest as a separate regressor, whose z-scored beta estimates were used for the alignment analysis. This part of the data was not modeled using other regressors. Additionally, we included different regressors capturing all the first steps, all the second steps, and all the feedback steps. Lastly, we used $6$ rotation and translation estimates as well as framewise displacement as noise regressors. The hemodynamic response was modeled using the spm \cite{penny2011statistical} model. A high pass filter of $0.01$ Hz and a Gaussian kernel with $6$mm full-width at half-maximum was applied. The GLMs were built using nilearn \cite{Nilearn}. \\

The neural alignment analysis on the sentence-reading task was conducted using publicly available code from the original study \cite{tuckute2024driving}. No other changes were made besides replacing \textsc{GPT2-XL} with Centaur and Llama. We refer the reader to \cite{tuckute2024driving} for further details.

\subsection*{Model-guided scientific discovery}

In our model-guided scientific discovery analysis, we focused on participants in the test set to avoid any potential contamination issues. We fitted parameters of all cognitive models individually for each participant using a maximum likelihood estimation. Models were compared against each other using the Akaike information criterion (AIC). The three models from the original study were implemented via the following equations:
\begin{align*}
    p(a = \text{A} | \mathbf{x}_\text{A}, \mathbf{x}_\text{B}, \text{WADD}) &\propto \exp \left(\beta \cdot \mathbf{w}_{\text{WADD}}^{\top}\mathbf{x}_\text{A} \right)  \\
    \mathbf{w}_{\text{WADD}} &= \left[0.9, 0.8, 0.7, 0.6 \right] \\
    p(a = \text{A} | \mathbf{x}_\text{A}, \mathbf{x}_\text{B}, \text{EW}) &\propto \exp \left(\beta \cdot \mathbf{w}_{\text{EW}}^{\top}\mathbf{x}_\text{A} \right)  \\
    \mathbf{w}_{\text{EW}} &= \left[1, 1, 1, 1 \right] \\
    p(a = \text{A} | \mathbf{x}_\text{A}, \mathbf{x}_\text{B}, \text{TTB}) &\propto \exp \left(\beta \cdot \mathbf{w}_{\text{TTB}}^{\top}\mathbf{x}_\text{A} \right)  \\
    \mathbf{w}_{\text{TTB}} &=  \left[1, 0.5, 0.25, 0.125 \right]
\end{align*}

\noindent where $\mathbf{x}_{\text{A}}$ and $\mathbf{x}_{\text{B}}$ are vectors containing four expert ratings (either 0 or 1) and $\beta$ is a free parameter controlling the noise level. 

We prompted DeepSeek-R1 (in the Distill-Llama-70B variant) to generate explanations of human decision-making; the corresponding prompt is provided in the Supplementary Information. We then formalized the explanation shown in Figure \ref{fig:deepseek} into the following computational model:
\begin{align*}
    p(a = \text{A} | \mathbf{x}_\text{A}, \mathbf{x}_\text{B}, \text{DeepSeek-R1}) &\propto 
    \begin{cases}
    \exp \left(\beta \cdot \mathbf{w}_{\text{TTB}}^{\top}\mathbf{x}_\text{A} \right) & \text{if } \sum_i \mathbf{x}_{\text{A}, i} = \sum_i \mathbf{x}_{\text{B}, i} \\
    \exp \left(\beta \cdot \mathbf{w}_{\text{EW}}^{\top}\mathbf{x}_\text{A} \right)              & \text{otherwise}
\end{cases} 
\end{align*}

For the scientific regret minimization pipeline, we computed the difference in log-likelihoods between Centaur and the DeepSeek-R1-discovered model. We visualized and inspected the ten data points with the highest difference. This process resulted in the following computational model:
\begin{align*}
    p(a = \text{A} | \mathbf{x}_\text{A}, \mathbf{x}_\text{B}, \text{SRM}) &\propto \exp \left(\beta \cdot  \left(\sigma \cdot \mathbf{w}_{\text{TTB}}^{\top}\mathbf{x}_\text{A} + \left(1-\sigma \right) \cdot \mathbf{w}_{\text{EW}}^{\top}\mathbf{x}_\text{A} \right) \right) 
\end{align*}
\noindent where $\sigma$ is a free parameter constrained between $0$ and $1$ controlling the trade-off between the two strategies.

\subsection*{Data availability}

Psych-101 is publicly available on the Huggingface platform: \url{https://huggingface.co/datasets/marcelbinz/Psych-101}. The test set is accessible under a CC-BY-ND-4.0 license via a gated repository: \url{https://huggingface.co/datasets/marcelbinz/Psych-101-test}. 

\subsection*{Code availability}
Centaur is available on the Huggingface platform: \url{https://huggingface.co/marcelbinz/Llama-3.1-Centaur-70B-adapter}. We provide additional code to reproduce our results under \url{https://github.com/marcelbinz/Llama-3.1-Centaur-70B}.
 
\backmatter

\bmhead{Acknowledgements}

Funding was from the Max Planck Society (PD), the Humboldt Foundation (PD), and the NOMIS Foundation (TLG). PD is a member of the Machine Learning Cluster of Excellence, EXC number 2064/1 – Project number 39072764 and of the Else Kr\"oner Medical Scientist Kolleg "ClinbrAIn: Artificial Intelligence for Clinical Brain
Research”. SK is supported by a Google PhD Fellowship.

\bmhead{Author contributions} $~$

\noindent\textbf{Project lead:} Marcel Binz

\noindent\textbf{Data curation:} Elif Akata, Franziska Brändle, Marcel Binz, Fred Callaway, Julian Coda-Forno, Can Demircan, Maria Eckstein, Noemi Elteto, Susanne Haridi, Akshay Jagadish, Li Ji-An, Alexander Kipnis, Sreejan Kumar, Tobias Ludwig, Surabhi Nath, Joshua Peterson, Evan Russek, Tankred Saanum, Johannes Schubert, Luca Schulze Buschoff, Nishad Singhi, Xin Sui, Mirko Thalmann, Vuong Truong, Kristin Witte, Shuchen Wu, Dirk Wulff, Huadong Xiong

\noindent\textbf{Data quality control:} Elif Akata, Marcel Binz, Julian Coda-Forno, Can Demircan, Susanne Haridi, Luca Schulze Buschoff

\noindent\textbf{Model training:} Marcel Binz, Vishaal Udandarao

\noindent\textbf{Model evaluation:} Marcel Binz, Julian Coda-Forno, Alexander Kipnis, Mirko Thalmann, Konstantinos Voudouris 

\noindent\textbf{Domain-specific models:} Marcel Binz, Julian Coda-Forno, Can Demircan, Akshay Jagadish, Marvin Mathony, Alireza Modirshanechi, Milena Rmus, Tobias Ludwig 

\noindent\textbf{Neural analyses:} Marcel Binz, Can Demircan, Sreejan Kumar, Marcelo Mattar, Evan Russek 

\noindent\textbf{First draft:} Marcel Binz, Eric Schulz

\noindent\textbf{Review and editing:} All authors

\noindent No researchers at Google DeepMind used Llama for this research. We thank Natalia Scharfenberg for her contributions to the data collection.

\newpage 

\section*{Supplementary Information}\label{secA1}

\subsection*{Psych-101}

\begin{figure}[H]
    \centering
    \scalebox{0.73}{\begin{tabular}{@{}cc}
         \includegraphics[]{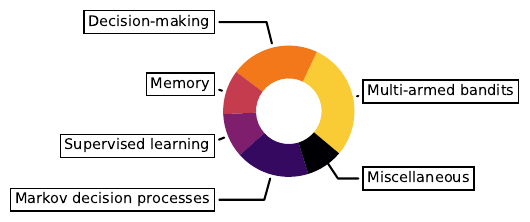} & \includegraphics[width=8.7cm]{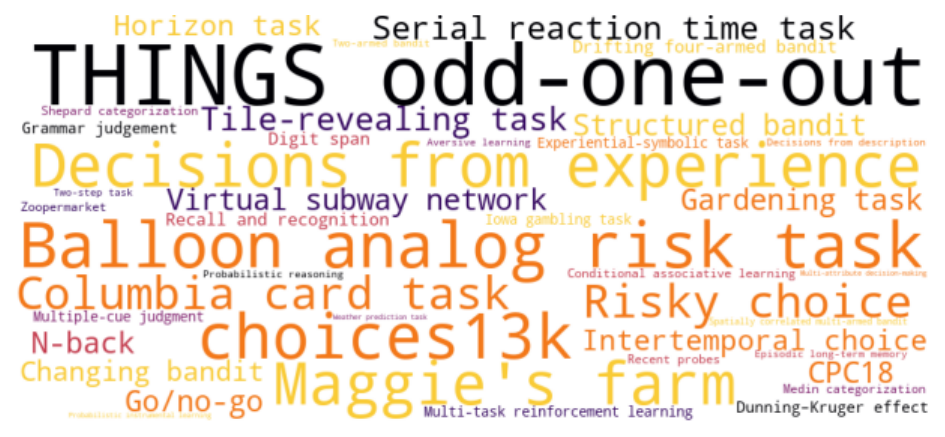}
    \end{tabular}}
    \caption{Visual summary of Psych-101. The left panel shows the proportion of included domains. The right panel shows a word cloud with the included experimental paradigms.}
    \label{fig:fig5}
\end{figure}

\subsubsection*{Data contamination analysis}

We performed a data contamination analysis using the LogProber method proposed in \cite{yax2024assessing}. LogProber fits the following two-parameter exponential model to the cumulative log-likelihood of each sequence being checked for contamination:
\begin{equation*}
f(x) = -A \left( 1-e^{-Bx}\right)
\end{equation*}

\begin{figure}[H]
    \centering
    \scalebox{0.74}{\includegraphics[]{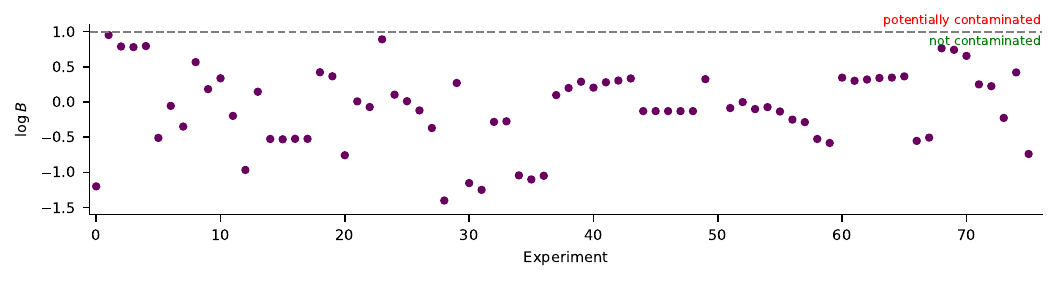}}
    \caption{Data contamination analysis using the LogProber method \cite{yax2024assessing}. LogProber fits an exponential model to the cumulative log-likelihood of each sequence being checked for contamination. High acceleration ($\log B$) suggests that a prompt is memorized from the pretraining data.}
    \label{fig:logprober}
\end{figure}

Prompts that are memorized from the pretraining data will show a high acceleration ($B$). Following the results presented in \cite{yax2024assessing}, we set a threshold for possible contamination to $\log B \geq 1$. We repeat this analysis for each experiment in Psych-101. For this, we used prompts up to the point of the first human choice. This analysis indicated no evidence of contamination as shown in Figure \ref{fig:logprober}.

\subsubsection*{Experiment embeddings}

\begin{figure}[H]
    \centering
    \scalebox{0.74}{\includegraphics[]{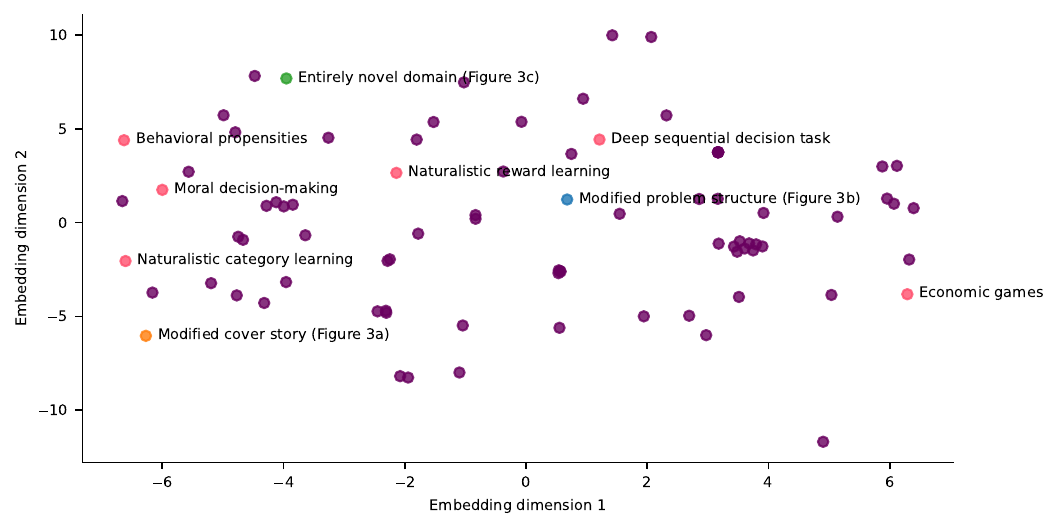}}
    \caption{Multidimensional scaling embedding of the experiments used in this paper. Purple dots correspond to experiments from Psych-101, whereas the colored dots correspond to the indicated evaluation experiment.}
    \label{fig:12}
\end{figure}

Figure \ref{fig:12} shows a two-dimensional embedding of the experiments used in this paper. To obtain this embedding, we took the corresponding natural language prompts up to the point of the first human choice, extracted a vector-based representation for them using ModernBERT \cite{modernbert}, and finally projected these representations onto two dimensions using multidimensional scaling. We observed that most of our out-of-distribution evaluation experiments are placed on the boundaries of the embedding space, confirming that they indeed probe the generalization abilities of Centaur.

\newpage

\subsection*{Full goodness-of-fit results}

\begin{table}[h]
\centering 
\begin{tabular}{@{}lccc@{}} 
\toprule  
\textbf{Experiment} & \textbf{Centaur} & \textbf{Llama} & \textbf{Cognitive model} \\ 
\midrule 
Shepard categorization & 0.5394 & 0.5818 & 0.6108 \\ 
Drifting four-armed bandit & 0.7029 & 0.8810 & 0.9043 \\ 
N-back & 0.3954 & 0.5209 & 0.5787 \\ 
Digit span & 0.5520 & 0.6618 & 0.9359 \\ 
Go/no-go & 0.0000 & 0.0062 & 0.0757 \\ 
Recent probes & 0.2572 & 0.3433 & 0.3868 \\ 
Horizon task & 0.4032 & 0.5237 & 0.3595 \\ 
Gardening task & 0.3783 & 0.5040 & 0.9105 \\ 
Columbia card task & 0.1867 & 0.2261 & 0.2629 \\ 
Balloon analog risk task & 0.0593 & 0.0753 & 0.0922 \\ 
Two-armed bandit & 0.2963 & 0.3829 & 0.4187 \\ 
Conditional associative learning & 0.5380 & 0.6373 & 0.8575 \\ 
THINGS odd-one-out & 0.8068 & 1.1386 & 0.8253 \\ 
Multi-attribute decision-making & 0.0619 & 0.1502 & 0.1922 \\ 
Two-step task & 0.4998 & 0.6075 & 0.6043 \\ 
Probabilistic instrumental learning & 0.4937 & 0.5382 & 0.5047 \\ 
Medin categorization & 0.4967 & 0.5772 & 0.5313 \\ 
Zoopermarket & 0.4850 & 0.6026 & 0.6047 \\ 
choices13k & 0.4274 & 0.5342 & 0.6563 \\ 
CPC18 & 0.3390 & 0.4118 & 0.6607 \\ 
Intertemporal choice & 0.4340 & 0.7336 & 0.6591 \\ 
Structured bandit & 0.6410 & 0.8114 & 1.0530 \\ 
Weather prediction task & 0.5514 & 0.5749 & 0.6267 \\ 
Iowa gambling task & 0.8890 & 0.9880 & 1.1555 \\ 
Virtual subway network & 1.1271 & 1.5347 & nan \\ 
Multi-task reinforcement learning & 0.5672 & 0.6604 & 1.0424 \\ 
Serial reaction time task & 0.1718 & 0.1900 & 0.1962 \\ 
Decisions from description & 0.5336 & 0.7569 & 0.6120 \\ 
Decisions from experience & 0.3686 & 0.4339 & 0.5404 \\ 
Changing bandit & 0.3025 & 0.3824 & 0.4378 \\ 
Multiple-cue judgment & 1.1236 & 1.2818 & 1.9157 \\ 
Recall and recognition & 1.0591 & 1.3759 & nan \\ 
Experiential-symbolic task & 0.4536 & 0.6983 & nan \\ 
Grammar judgement & 1.4355 & 1.9949 & 1.4127 \\ 
Risky choice & 0.4281 & 0.6475 & nan \\ 
Tile-revealing task & 1.8713 & 2.7380 & nan \\ 
Episodic long-term memory & 0.8684 & 1.1344 & nan \\ 
Aversive learning & 4.0733 & 5.1066 & nan \\ 
Spatially correlated multi-armed bandit & 1.8319 & 2.4479 & 2.7635 \\ 
Probabilistic reasoning & 2.3731 & 2.6406 & nan \\ 
\bottomrule \\ 
\end{tabular} 
\caption{Full negative log-likelihoods on held-out participants.}
\label{tab:tab2} 
\end{table}

\subsection*{Minitaur}

\begin{figure}[H]
    \centering
    \scalebox{0.74}{\includegraphics[]{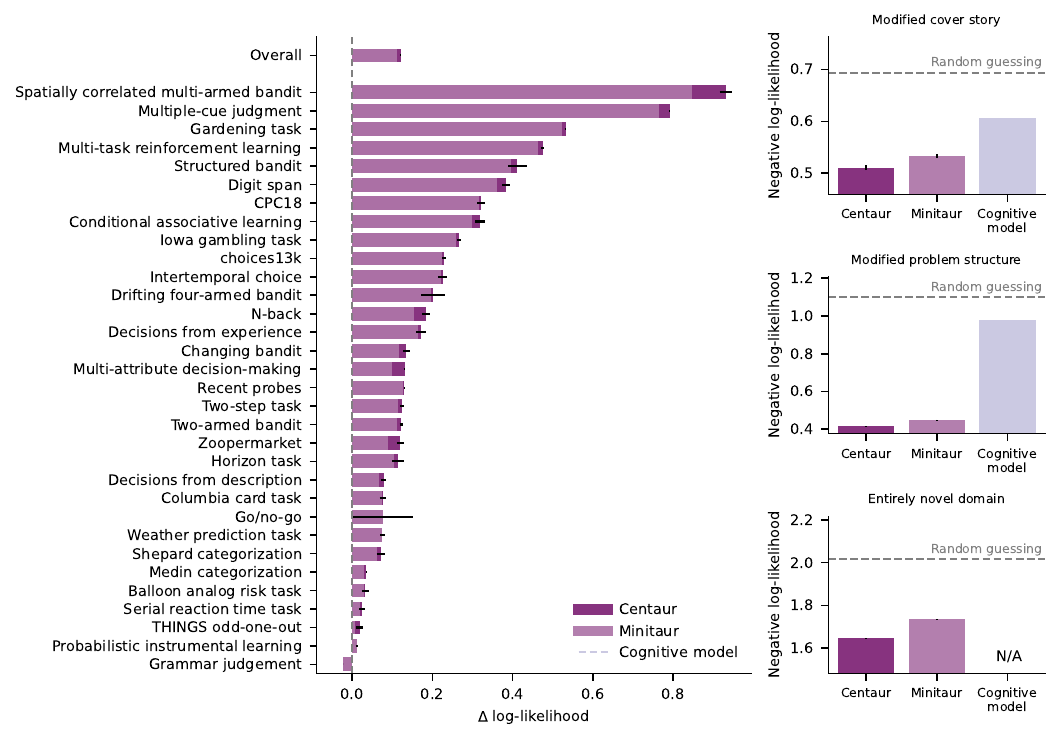}}
    \caption{Log-likelihood comparison between Centaur and Minitaur on the analyses from the main text. The left subplot contains results from Psych-101, while the right subplot shows contains results for different held-out settings. Error bars correspond to the standard error of the mean, taken over responses.}
    \label{fig:8b}
\end{figure}

Figure \ref{fig:8b} compares Centaur and to Minitaur on the analyses from the main text. 

\subsection*{Further out-of-distribution evaluations}

\begin{figure}[H]
    \centering
    \scalebox{0.74}{\includegraphics[]{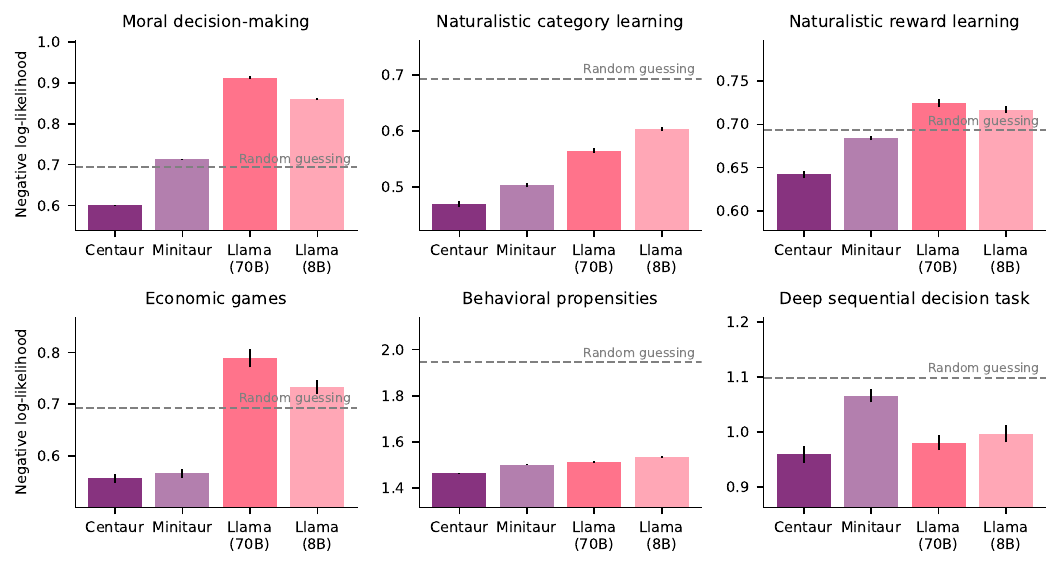}}
    \caption{Further out-of-distribution evaluations. Each subplot shows negative log-likelihoods for a different experiment. Error bars correspond to the standard error of the mean, taken over responses.}
    \label{fig:10}
\end{figure}

Figure \ref{fig:10} contains results for  six additional out-of-distribution experimental paradigms, including moral decision-making \cite{awad2018moral}, economic games \cite{akata2023playing},  naturalistic category and reward learning \cite{demircan2024evaluating}, behavioral propensities \cite{singh2022representing}, and a deep sequential decision task \cite{xu2021novelty}. None of these paradigms were included in Psych-101, hence they provide a stress test for a model's generalization capabilities. Centaur robustly captured human behavior in all of these settings, while smaller and non-finetuned models did not do so consistently. 

\subsection*{Further model  comparisons}

To rule out the hypothesis that finetuning on any data aligns a model with human behavior, we compared Centaur to various Llama variants finetuned for other purposes (i.e. non-cognitive tasks). These models include the following:
\begin{itemize}
\item Nemotron \cite{wang2024helpsteer2}: finetuned for instruction-following.
\item Hermes \cite{teknium2024hermes}: finetuned for various purposes, including agentic capabilities, roleplaying, reasoning, multi-turn conversation, and long context coherence.
\item Reflection: finetuned for reasoning.
\end{itemize}

\begin{figure}[H]
    \centering
    \scalebox{0.74}{\includegraphics[]{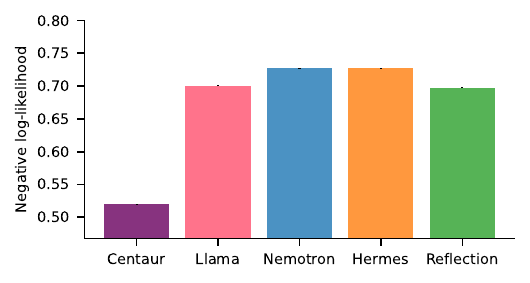}}
    \caption{Negative log-likelihoods of Centaur and alternative Llama variants on Psych-101. Error bars correspond to the standard error of the mean, taken over responses.}
    \label{fig:noncog}
\end{figure}

Figure \ref{fig:noncog} illustrates the results of this analysis. We observed that none of the Llama variants captures human behavior better than the base model, ruling out the hypothesis that finetuning generally leads to models that are better at predicting human behavior.

\subsection*{Noise ceiling analysis}

\begin{figure}[H]
    \centering
    \scalebox{0.74}{\includegraphics[]{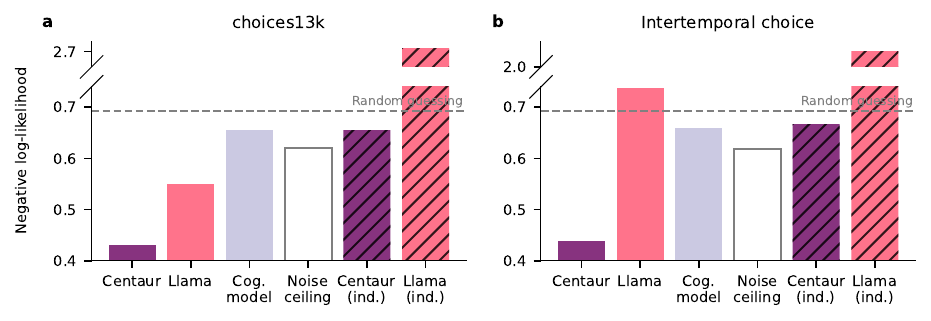}}
    \caption{Noise ceiling analysis. We use the suffix "ind." to indicate models that are prompted with to predict each response independently.}
    \label{fig:noise}
\end{figure}

We conducted a noise ceiling analysis to better understand the capabilities of Centaur. It is not straightforward to estimate the noise ceiling for experiments with sequential dependencies, which includes the majority of Psych-101. Hence, we focused on two experiments for which such an analysis is possible: the choices13k data set \cite{peterson2021using} and an intertemporal choice experiment \cite{ruggeri2022globalizability}. In both cases, we found that Centaur substantially exceeds the estimated noise ceiling. This is possible because Centaur can pick up on context-dependent patterns that are not captured by standard noise ceiling analyses. Therefore, we have performed an additional analysis testing how well Centaur can predict human responses if we prompt it to predict each response independently. We use the suffix “ind.” to indicate this way of prompting the model. Centaur still matches the performance of domain-specific cognitive models when context-independent prompts are used, amounting to roughly half of the estimated noise ceiling. Figure \ref{fig:noise} visualizes these results.

\subsection*{Benchmarks}

\subsubsection*{\texttt{metabench}}

\begin{figure}[H]
    \centering
    \scalebox{0.74}{\includegraphics[]{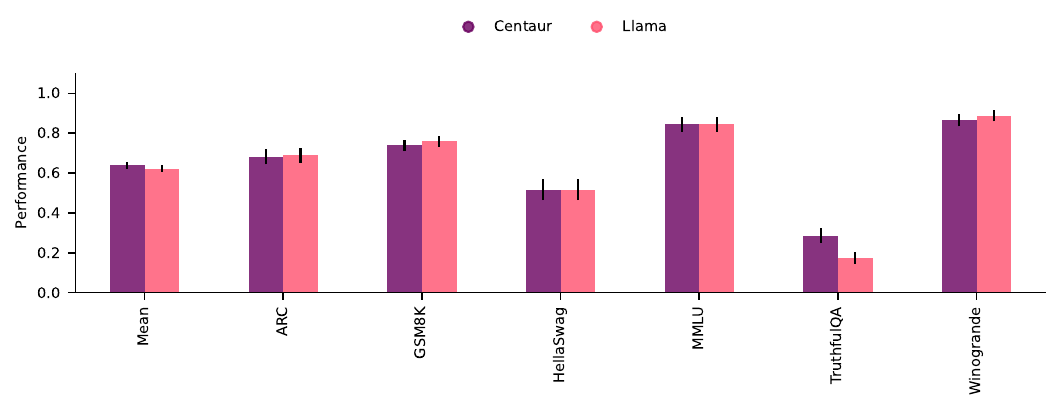}}
    \caption{\texttt{metabench} \cite{kipnis2024texttt} results. }
    \label{fig:metabench}
\end{figure}

Figure \ref{fig:metabench} shows additional results in \texttt{metabench}, a sparse benchmark containing several canonical benchmarks from the machine learning literature \cite{kipnis2024texttt}. We find that Centaur maintains the level of performance of Llama, indicating that finetuning on human behavior did not lead to deterioration in other tasks. Performance on TruthfulQA \cite{DBLP:journals/corr/abs-2109-07958} -- which measures how models mimic human falsehoods -- even improved significantly with finetuning. We refer the reader to \cite{kipnis2024texttt} for further details.

\subsubsection*{CogBench}

\begin{figure}[H]
    \centering
    \scalebox{0.74}{\includegraphics[]{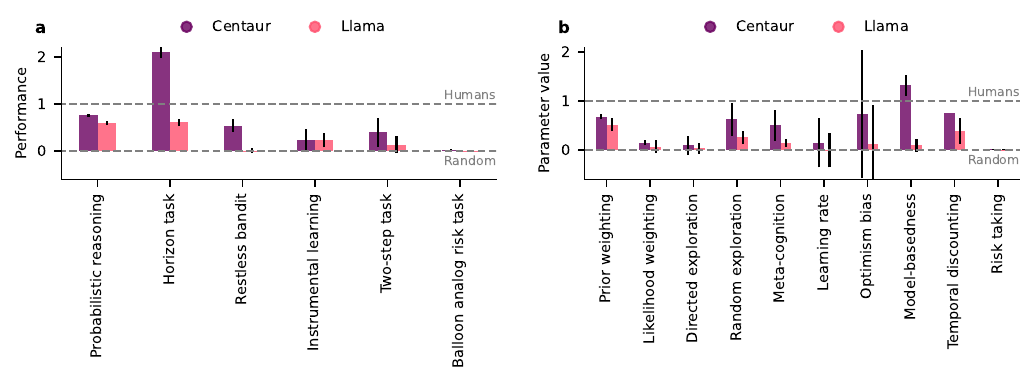}}
    \caption{CogBench \cite{pmlr-v235-coda-forno24a} results. \textbf{a}, Performance-based metrics. \textbf{b}, Behavioral metrics. All metrics are human-normalized: a value of zero corresponds to a random agent, while a value of one corresponds to the average human subject.}
    \label{fig:cogbench}
\end{figure}

Figure \ref{fig:cogbench} shows additional results in CogBench, a benchmark that includes ten behavioral metrics derived from seven cognitive psychology experiments \cite{pmlr-v235-coda-forno24a}. We find that -- relative to Llama -- Centaur's performance improves in all experiments (see Figure \ref{fig:cogbench}a). Furthermore, Centaur becomes more similar to human subjects in all ten behavioral metrics (see  Figure  \ref{fig:cogbench}b). We refer the reader to \cite{pmlr-v235-coda-forno24a} for further details.

\subsubsection*{Fine-grained neural analysis results}

\begin{figure}[H]
    \centering
    \scalebox{0.74}{\includegraphics[]{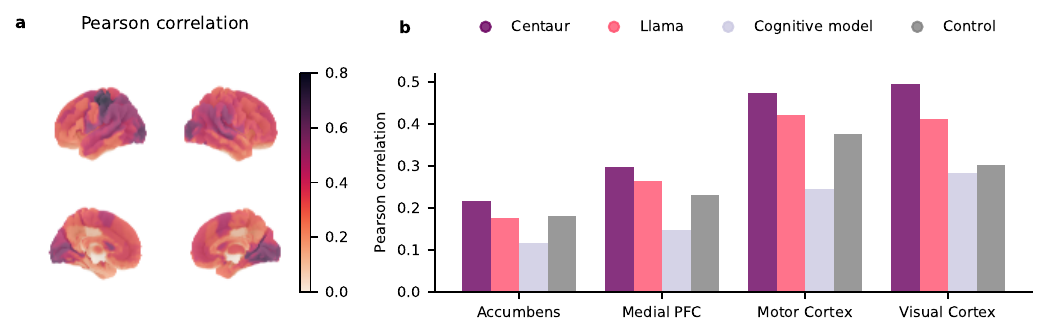}}
    \caption{Neural alignment results in the two-step task. \textbf{a}, Pearson correlation coefficients between the predicted activity from Centaur’s representations and the BOLD data shown on a surface brain. \textbf{b}, Predictive performance of Centaur’s representations against alternatives for specific ROIs. Cortical scores are averaged over the corresponding bilateral parcels in the Schaefer atlas. The accumbens is defined based on the Harvard-Oxford atlas. Pearson correlation coefficients are shown for layer 20 but exhibit a similar pattern across all layers.}
    \label{fig:furtherbrain}
\end{figure}

Figure \ref{fig:furtherbrain} shows fine-grained neural alignment results on the two-step task data. Centaur achieves the most accurate predictions in the left motor cortex (see Figure \ref{fig:furtherbrain}a). As participants performed the task with their right hand in the scanner, this effect may be explained by Centaur’s strong performance in predicting choices.

Figure \ref{fig:furtherbrain}b contains numerical results for ROIs that have been identified as behaviorally relevant in previous work. Centaur outperformed Llama and the cognitive model in predicting activity in accumbens, the ROI from the original study that showed a reward prediction error effect \cite{daw2011model, feher2023rethinking}. We found a similar pattern in the medial PFC, another region that showed an effect in the original article \cite{feher2023rethinking}, as well as in the sensory and motor cortices.

\subsection*{Model-guided scientific discovery}

We generated an explanation for the behavior of every participant in the multi-attribute decision-making study of \cite{hilbig2014generalized} using DeepSeek-R1 based on the following prompt template:

\begin{tcolorbox}[width=\textwidth,
                  boxsep=3pt,
                  left=3pt,
                  right=3pt,
                  top=3pt,
                  ]
In the following, you will see a transcript of a psychological experiment. Participant responses are highlighted within \say{$<<$} and \say{$>>$} characters. Your job will be to explain what strategies that participant used to solve the given task. Your explanation should be targeted at an expert cognitive scientist.

\#\#\# START OF TRANSCRIPT \#\#\#

[TRANSCRIPT]

\#\#\# END OF TRANSCRIPT \#\#\#

Please explain what strategies that participant used to solve the given task in less than two hundred words:

\end{tcolorbox}

\subsection*{Newell test}

Together with his call for unified theories of cognition \cite{Newell1990-NEWUTO, anderson2003newell}, Newell outlined a set of criteria that a unified computational model should fulfill. Centaur is the first model to satisfy the majority of these criteria (see Table \ref{tab:tab1}). Most importantly, it (1) behaves as an almost arbitrary function of the environment, (2) operates in real time, and (3) relies on vast amounts of knowledge about the world. We provide an extended discussion on Newell's criteria in the following. 

\begin{table}[]
    \centering
    \begin{tabular}{@{}L{9.7cm}R{3cm}@{}}
        \toprule
         \textbf{Criterion} & \textbf{Fulfilled by Centaur} \\
         \midrule 
         Behave as an (almost) arbitrary function of the environment & {\color{ForestGreen}\cmark} \\
         Operate in real time & {\color{ForestGreen}\cmark} \\
         Exhibit rational, that is, effective adaptive behavior & {\color{ForestGreen}\cmark} \\
         Use vast amounts of knowledge about the environment &  {\color{ForestGreen}\cmark} \\
         Behave robustly in the face of error, the unexpected, and the unknown &  {\color{ForestGreen}\cmark} \\
         Integrate diverse knowledge & {\color{ForestGreen}\cmark} \\
         Use (natural) language &  {\color{ForestGreen}\cmark} \\
         Exhibit self-awareness and a sense of self & {\color{gray}$\bullet$} \\  
         Learn from its environment & {\color{ForestGreen}\cmark} \\
         Acquire capabilities through development & {\color{BrickRed}\xmark} \\
         Arise through evolution &  {\color{BrickRed}\xmark} \\
         Be realizable within the brain & {\color{ForestGreen}\cmark} \\
         \bottomrule \\
    \end{tabular}
    \caption{Newell test for a theory of cognition.}
    \label{tab:tab1}
\end{table}

\subsubsection*{Behave as an (almost) arbitrary function of the environment}

This is the most important criterion according to Newell. Centaur fulfills it more than any previous model as shown by our extensive analysis. Yet, Centaur's scope is still limited to psychological experiments that can be expressed in natural language. It will be an important avenue for future research to transfer this ability to real-world applications.

\subsubsection*{Operate in real time}

Centaur can simulate human behavior in (almost) real-time. For example, running an open-loop simulation of a typical two-step task experiment takes around 30 minutes, while it takes around 20 minutes for the average human participant. We believe that inference time could be further optimized to fully close this gap.

\subsubsection*{Exhibit rational, that is, effective adaptive behavior}

Bayesian inference is the gold standard for rational and adaptive behavior \cite{anderson90}. Previous work has shown that systems that engage in in-context learning implement Bayesian inference implicitly \cite{binz2023meta}. In-context learning is at the heart of Centaur, thereby making it a rational and adaptive system.

\subsubsection*{Use vast amounts of knowledge about the environment}

Large language models are the biggest knowledge bases we have to date. As Centaur is built on top of a state-of-the-art language model, it fulfills this criterion by design.

\subsubsection*{Behave robustly in the face of error, the unexpected, and the unknown}

Our extensive out-of-distribution evaluations clearly demonstrate that Centaur has this ability.

\subsubsection*{Integrate diverse knowledge}

This was originally a criterion on symbols and abstractions \cite{Newell1990-NEWUTO}. At the basic level, Centaur is a system that processes language. Language is a symbolic system, meaning that Centaur fulfills this criterion.

\subsubsection*{Use (natural) language}

No further elaboration is required.

\subsubsection*{Exhibit self-awareness and a sense of self}

This is a tricky criterion as Centaur is trained on a population of individuals. Yet, to make good predictions about the future, Centaur needs to make inferences about the person who has produced a given trajectory, and in order to do that it requires a representation of that person \cite{andreas-2022-language}.

\subsubsection*{Learn from its environment}

Psych-101 contains many experiments that require learning from an environment. Centaur does well in those experiments, thereby clearly satisfying this criterion.

\subsubsection*{Acquire capabilities through development}

Centaur makes no claims about how human behavior might arise through development.

\subsubsection*{Arise through evolution}

Centaur makes no claims about how human behavior might arise through evolution.

\subsubsection*{Be realizable within the brain}

We have shown that Centaur's internal representations are robust predictors of human neural activity. Even though there is clearly a gap between the transformer architecture that Centaur is based on and the human brain, Centaur still represents the current state-of-the-art when looking at neural alignment to human subjects.

\subsection*{Modeling details}

In the following, we list the domain-specific cognitive and statistical models used in our comparison. Each model was implemented in PyTorch \cite{10.5555/3454287.3455008}. We optimized a joint set of parameters for all participants in the training data by maximizing log-likelihoods of their choices and then evaluated how well a model with these parameters predicts choices of held-out participants. The optimization procedure involved 1000 iterations over the entire training set, and relied on a gradient-based algorithm \cite{defazio2024roadscheduled} with an initial learning rate of 0.1. We use $\mathbbm{1}$ to denote the indicator function that takes a value of one if the argument is true and zero otherwise.

\subsubsection*{Generalized context model}

Reference: \cite{nosofsky2011generalized} 

\noindent This model was used for the following experiments:
\begin{itemize}
    \item Shepard categorization
    \item Medin categorization
    \item Weather prediction task
\end{itemize}

\noindent It uses the following log-likelihood:
\begin{equation*}
    p(c_t = i | x_t = \mathbf{x}_t) \propto \text{exp} \left( \beta \sum_{k=1}^{t-1} \text{exp} \left(-||\mathbf{x}_k - \mathbf{x}_t ||_2\right) \cdot \mathbbm{1}\left[y_k = i\right] \right)
\end{equation*}

\noindent where $\mathbf{x}_t$ are the features of the item observed at trial $t$ and $y_t$ is the corresponding class label. $\beta$ is a free parameter of the model.
 
\subsubsection*{Prospect theory model}

Reference: \cite{peterson2021using} 

\noindent This model was used for the following experiments:
\begin{itemize}
    \item CPC18
    \item choices13k
    \item Decisions from description
\end{itemize}

\noindent It uses the following log-likelihood:
\begin{align*}
    p(c_t = i | p_i = \mathbf{p}_i, x_i  = \mathbf{x}_i) &\propto \exp\left(\exp\left(\beta\right)\left(\pi\left(\mathbf{p}_i\right)^{\top} u\left(\mathbf{x}_i\right)\right) \right) \\
    \pi\left(\mathbf{p}_i\right) &= \text{sigmoid}\left(a\right) + \text{sigmoid}\left(b\right) \cdot\mathbf{p}_i\\
    u\left(\mathbf{x}_i\right) &= 
    \begin{cases} 
        \text{sigmoid}\left(c\right) \cdot \mathbf{x}_i^{\text{sigmoid}\left(d\right)} & \text{where } \mathbf{x}_i\geq 0\\
        -\text{sigmoid}\left(e\right) \cdot \left(-\text{sigmoid}\left(f\right)\mathbf{x}_i\right)^{\text{sigmoid}\left(g\right)}              & \text{where } \mathbf{x}_i < 0
    \end{cases}
\end{align*}

\noindent where $\mathbf{p}_i$ is the vector of probabilities and $\mathbf{x}_i$ is the vector of values for each possible outcome in option $i$. $\beta$, $a$, $b$, $c$, $d$, $e$, $f$, and $g$ are free parameters of the model.

\subsubsection*{Hyperbolic discounting model}

Reference: \cite{van2013towards} 

\noindent This model was used for the following experiments:
\begin{itemize}
    \item Intertemporal choice
\end{itemize}

\noindent It uses the following log-likelihood:
\begin{align*}
    p(c_t = i | x_i = x_i, \gamma_i  =  \gamma_i) &\propto \exp\left(\beta\left( x_i \cdot \frac{1}{1+\left(a \cdot \gamma_i \right)}\right)\right)
\end{align*}

\noindent where $x_i$ is the reward and $\gamma_i$ is the delay of delivery for option $i$. $\beta$ and $a$ are free parameters of the model.

\subsubsection*{Dual-systems model}

Reference: \cite{daw2011model} 

\noindent This model was used for the following experiments:
\begin{itemize}
    \item Two-step task
\end{itemize}

\noindent It uses the following log-likelihood:
\begin{align*}
p(c_t = i | s_t = s) &\propto 
\begin{cases} 
       \text{exp} \left( \beta \left( \text{sigmoid}\left( \tau \right) Q_{s, i}^{\text{MB}}  + \left(1 - \text{sigmoid}\left( \tau \right)\right)    Q_{s, i}^{\text{MF}} \right) \right) & \text{if } s = 0 \\
        \text{exp} \left( \beta Q_{s, i}^{\text{MF}}\right)         & \text{if } s > 0
\end{cases} 
\end{align*}

\noindent where $Q_{s, i}^{\text{MB}}$ and $Q_{s, i}^{\text{MF}}$ are model-based and model-free value estimates that are computed as described in \cite{daw2011model}. $\beta$ and $\tau$ are free parameters of the model. We also included a stickiness term for the first stage choices, which is omitted for brevity in the equations above.

\subsubsection*{Rescorla-Wagner model}

Reference: \cite{wilson2014humans} 

\noindent This model was used for the following experiments:
\begin{itemize}
    \item Drifting four-armed bandit
    \item Horizon task
    \item Two-armed bandit
    \item Probabilistic instrumental learning
    \item Iowa gambling task
    \item Changing bandit
    \item Decisions from experience
\end{itemize}

\noindent It uses the following log-likelihood:
\begin{align*}
    p(c_t = i) &\propto \exp\left(a V_{i,t} + b S_{i,t} + c I_{i,t} \right) \\
    V_{i,t} &=  
    \begin{cases} 
        V_{i, t-1} + \text{sigmoid}\left(\alpha^+\right) \left(r_{t-1} - V_{i, t-1}  \right) & \text{if } c_{t-1} = i \text{ and } r_{t-1} - V_{i, t-1} \geq 0 \\
        V_{i, t-1} + \text{sigmoid}\left(\alpha^-\right) \left(r_{t-1} - V_{i, t-1}  \right) & \text{if } c_{t-1} = i \text{ and } r_{t-1} - V_{i, t-1} < 0 \\
        V_{i, t-1}             & \text{otherwise}
    \end{cases} \\
    S_{i,t} &= \mathbbm{1}\left[c_{t-1} = i\right] \\
    I_{i,t} &= \sum_{k=1}^{t-1} \mathbbm{1}\left[c_k = i\right]\\
    V_{i,1} &= d \\
    S_{i,1} &= 0 \\
    I_{i,1} &= 0
\end{align*}

\noindent where $r_t$ is the reward obtained in trial $t$. $\alpha^+$, $\alpha^-$, $a$, $b$, $c$, and $d$ are free parameters of the model.

\subsubsection*{Rescorla-Wagner model with context}

Reference: \cite{sutton2018reinforcement} 

\noindent This model was used for the following experiments:
\begin{itemize}
    \item Conditional associative learning
\end{itemize}

\noindent It uses the following log-likelihood:
\begin{align*}
    p(c_t = i | s_t = s) &\propto \exp\left(\beta V_{s,i,t} \right) \\
    V_{s,i, t} &=  
    \begin{cases} 
        V_{s,i, t-1} + \text{sigmoid}\left(\alpha\right) \left(r_{t-1} - V_{s,i,t-1}  \right) & \text{if } c_{t-1} = i \text{ and } s_{t-1} = s \\
        V_{s,i,t-1}             & \text{otherwise}
    \end{cases} \\
    V_{s,i,1} &= d 
\end{align*}

\noindent where $r_t$ is the reward obtained in trial $t$. $\alpha$, $\beta$, and $d$ are free parameters of the model.

\subsubsection*{Linear regression model}

Reference: \cite{gershman2015unifying} 

\noindent This model was used for the following experiments:
\begin{itemize}
    \item Multiple-cue judgment
    \item Gardening task
\end{itemize}

\noindent It uses the following log-likelihood for multiple-cue judgment:
\begin{align*}
    p(c_t = i | x_t = \mathbf{x}_t) &\propto 
        \text{exp}\left( \beta \left( \mathbf{w}_{t}^{\top}\mathbf{x}_{t} - i\right)^2 + \gamma \right) 
\end{align*}

\noindent It uses the following log-likelihood for the gardening task:
\begin{align*}
    p(c_t = \text{accept} | x_t = \mathbf{x}_t) &\propto \text{exp}\left( \beta  \mathbf{w}_{t}^{\top}\mathbf{x}_{t} \right) \\
    p(c_t = \text{reject} | x_t = \mathbf{x}_t) &\propto \text{exp}\left( 0 \right)
\end{align*}

\noindent and the following learning rule for both tasks:
\begin{align*}
    \mathbf{w}_t &= \mathbf{w}_{t-1} + \alpha \left( r_{t-1} - \mathbf{w}_{t-1}^{\top}\mathbf{x}_{t-1}\right) \mathbf{x}_{t-1} \\
    \mathbf{w}_1 &= \mathbf{d}
\end{align*}

\noindent where $r_t$ is the reward obtained in trial $t$ and $\mathbf{x}_t$ are the observed features. $\alpha$, $\beta$, $\gamma$, and $\mathbf{d}$ are free parameters of the model.

\subsubsection*{Weighted-additive model}

Reference: \cite{czerlinski1999good} 

\noindent This model was used for the following experiments:
\begin{itemize}
    \item Multi-attribute decision-making
\end{itemize}

\noindent It uses the following log-likelihood:
\begin{align*}
    p(c_t = i | x_i = \mathbf{x}_i) &\propto \exp\left(\mathbf{w}^{\top} \mathbf{x}_i\right) 
\end{align*}

\noindent where $\mathbf{x}_i$ is the vector of features for option $i$. $\mathbf{w}$ are free parameters of the model.

\subsubsection*{Decision-updated reference point model}

Reference: \cite{pedroni2018prospect} 

\noindent This model was used for the following experiments:
\begin{itemize}
    \item Columbia card task
\end{itemize}

\noindent It uses the following log-likelihood:
\begin{align*}
    p(c_t = \text{sample} | x_{\text{win}}, x_{\text{loss}}, p_{\text{win}}, p_{\text{loss}}) &\propto \exp\left(  h \cdot \left( x_{\text{win}} \cdot p_{\text{win}} + x_{\text{loss}} \cdot p_{\text{loss}}\right) + i \right) \\
    p(c_t = \text{stop} | x_{\text{win}}, x_{\text{loss}}, p_{\text{win}}, p_{\text{loss}}) &\propto \text{exp} \left(j\right) \\
    \pi\left(p\right) &= \text{sigmoid}\left(a\right) + \text{sigmoid}\left(b\right) \cdot p\\
    u\left(v\right) &= 
    \begin{cases} 
        \text{sigmoid}\left(c\right) \cdot v^{\text{sigmoid}\left(d\right)} & \text{where }v\geq 0\\
        -\text{sigmoid}\left(e\right) \cdot \left(-\text{sigmoid}\left(f\right)v\right)^{\text{sigmoid}\left(g\right)}              & \text{where } v < 0
    \end{cases}
\end{align*}

\noindent where $x_{\text{win}}$ and $x_{\text{loss}}$ are the values that can be won or lost respectively, and $p_{\text{win}}$ and $p_{\text{loss}}$ are the corresponding probabilities. $a$, $b$, $c$, $d$, $e$, $f$, $g$, $h$, $i$, and $j$ are free parameters of the model.

\subsubsection*{Odd-one-out model}

Reference: \cite{hebart2020revealing}

\noindent This model was used for the following experiments:
\begin{itemize}
    \item THINGS odd-one-out
\end{itemize}

\noindent It uses the following log-likelihood:
\begin{align*}
    p(c_t = i | x_i, x_j, x_k) &\propto \exp\left(\mathbf{x}_j^{\top} \mathbf{x}_k\right) 
\end{align*}

\noindent where $x_i$, $x_j$, and $x_k$ are the observed objects with their  corresponding embeddings $\mathbf{x}_i$, $\mathbf{x}_j$, and $\mathbf{x}_k \in \mathbb{R}^{16}$ that are free parameters of the model.

\subsubsection*{Multi-task reinforcement learning model}

Reference: \cite{tomov2021multi} 

\noindent This model was used for the following experiments:
\begin{itemize}
    \item Multi-task reinforcement learning
    \item Zoopermarket
\end{itemize}

\subsubsection*{GP-UCB model}

Reference: \cite{wu2018generalization} 

\noindent This model was used for the following experiments:
\begin{itemize}
    \item Spatially correlated multi-armed bandit
    \item Structured bandit
\end{itemize}

\noindent It uses the following log-likelihood:
\begin{align*}
    p(c_t = i) &\propto \exp\left(\beta \left( \mathbf{m}_{i,t} + \text{exp} \left( \gamma \right) \mathbf{s}_{i,t} \right) \right)
\end{align*}

\noindent where $\mathbf{m}_{i,t}$ and $\mathbf{s}_{i,t}$ are obtained via Gaussian Process regression with a radial basis function kernel as described in \cite{wu2018generalization}. $\beta$ and $\gamma$ are free parameters of the model.

\subsubsection*{Rational model}

Reference: N/A 

\noindent This model was used for the following experiments:
\begin{itemize}
    \item Balloon analog risk task
    \item N-back
    \item Digit span
    \item Go/no-go
    \item Recent probes
    \item Serial reaction time task
\end{itemize}

\noindent It uses the following log-likelihood:
\begin{align*}
    p(c_t = i | o_t = j) &\propto \exp\left(\Theta_{j, i}\right)
\end{align*}

\noindent where $j$ is the optimal choice at trial $t$. $\Theta \in \mathbb{R}^{N_c \times N_c}$ are free parameters of the model.

\subsubsection*{Lookup table model}

Reference: N/A 

\noindent This model was used for the following experiments:
\begin{itemize}
    \item Grammar judgement
\end{itemize}

\noindent It uses the following log-likelihood:
\begin{align*}
    p(c_t = i) &\propto \exp\left(\Theta_{t,  i}\right)
\end{align*}

\noindent where $\Theta  \in \mathbb{R}^{T \times N_c}$ are free parameters of the model. 

\subsection*{Example prompts}

\noindent The following sections present the experiments contained in Psych-101 in detail. Example prompts are truncated to $4096$ characters but otherwise shown as is.

\subsubsection*{Shepard categorization}

Data source: \cite{badham2017deficits} \\ $~$ \\
Number of experiments: 1 $~$\\ 
Number of participants: 85 $~$\\ 
Number of choices: 29776 $~$\\ 
 $~$\\ 
\textbf{Example prompt:}
 $~$\\ 
You will be shown several examples of geometric objects. $~$\\ 
Your task is to learn a rule that allows you to tell whether an object belongs to the E or K category. $~$\\ 
For each presented object, you will be asked to make a category judgment by pressing the corresponding key and then you will receive feedback. $~$\\ 
You will encounter four different problems with different rules. $~$\\ 
 $~$\\ 
You encounter a new problem with a new rule determining which objects belong to each category: $~$\\ 
You see a big black square. You press $<<$K$>>$. The correct category is K. $~$\\ 
You see a small black triangle. You press $<<$K$>>$. The correct category is E. $~$\\ 
You see a big white triangle. You press $<<$E$>>$. The correct category is K. $~$\\ 
You see a small white triangle. You press $<<$E$>>$. The correct category is E. $~$\\ 
You see a small white square. You press $<<$E$>>$. The correct category is E. $~$\\ 
You see a small black square. You press $<<$K$>>$. The correct category is E. $~$\\ 
You see a big white square. You press $<<$E$>>$. The correct category is K. $~$\\ 
You see a big black triangle. You press $<<$E$>>$. The correct category is K. $~$\\ 
You see a big white square. You press $<<$E$>>$. The correct category is K. $~$\\ 
You see a big black square. You press $<<$E$>>$. The correct category is K. $~$\\ 
You see a small white triangle. You press $<<$K$>>$. The correct category is E. $~$\\ 
You see a small black triangle. You press $<<$K$>>$. The correct category is E. $~$\\ 
You see a big white triangle. You press $<<$E$>>$. The correct category is K. $~$\\ 
You see a small white square. You press $<<$K$>>$. The correct category is E. $~$\\ 
You see a small black square. You press $<<$K$>>$. The correct category is E. $~$\\ 
You see a big black triangle. You press $<<$E$>>$. The correct category is K. $~$\\ 
You see a small white square. You press $<<$E$>>$. The correct category is E. $~$\\ 
You see a small black square. You press $<<$E$>>$. The correct category is E. $~$\\ 
You see a big white triangle. You press $<<$K$>>$. The correct category is K. $~$\\ 
You see a small black triangle. You press $<<$K$>>$. The correct category is E. $~$\\ 
You see a small white triangle. You press $<<$E$>>$. The correct category is E. $~$\\ 
You see a big white square. You press $<<$K$>>$. The correct category is K. $~$\\ 
You see a big black triangle. You press $<<$E$>>$. The correct category is K. $~$\\ 
You see a big black square. You press $<<$E$>>$. The correct category is K. $~$\\ 
You see a small black triangle. You press $<<$E$>>$. The correct category is E. $~$\\ 
You see a small white square. You press $<<$K$>>$. The correct category is E. $~$\\ 
You see a small black square. You press $<<$E$>>$. The correct category is E. $~$\\ 
You see a big black square. You press $<<$K$>>$. The correct category is K. $~$\\ 
You see a big white square. You press $<<$K$>>$. The correct category is K. $~$\\ 
You see a big white triangle. You press $<<$K$>>$. The correct category is K. $~$\\ 
You see a small white triangle. You press $<<$E$>>$. The correct category is E. $~$\\ 
You see a big black triangle. You press $<<$K$>>$. The correct category is K. $~$\\ 
You see a small white square. You press $<<$E$>>$. The correct category is E. $~$\\ 
You see a big black triangle. You press $<<$K$>>$. The correct category is K. $~$\\ 
You see a big black triangle. You press $<<$K$>>$. The correct category is K. $~$\\ 
You see a small white triangle. You press $<<$E$>>$. The correct category is E. $~$\\ 
You see a big white triangle. You press $<<$K$>>$. The correct category is K. $~$\\ 
You see a small white triangle. You press $<<$E$>>$. The correct category is E. $~$\\ 
You see a small black triangle. You press $<<$E$>>$. The correct category is E. $~$\\ 
You see a small black square. You press $<<$E$>>$. The correct category is E. $~$\\ 
You see a small black triangle. You press $<<$E$>>$. The correct category is E. $~$\\ 
You see a big white square. You press $<<$K$>>$. The correct category is K. $~$\\ 
You see a big black square. You press $<<$K$>>$. The correct category is K. $~$\\ 
You see a big white square. You press $<<$K$>>$. The correct category is K. $~$\\ 
You see a big black square. You press $<<$K$>>$. The correct category is K. $~$\\ 
You see a small black square. You press $<<$E$>>$. The correct category is E. $~$\\ 
You see a big white triangle. You press $<<$E$>>$. The correct category is K. $~$\\ 
You see a small white square. You press $<<$E$>>$. The correct category is E. $~$\\ 
You see a big white square. You press $<<$K$>>$. The correct category is K. $~$\\ 
You see  

\subsubsection*{Drifting four-armed bandit}
Data source: \cite{Bahrami_Navajas_2022} \\ $~$ \\
Number of experiments: 1 $~$\\ 
Number of participants: 869 $~$\\ 
Number of choices: 125952 $~$\\ 
 $~$\\ 
\textbf{Example prompt:}
 $~$\\ 
You will be asked to repeatedly choose between four different options labeled L, G, O, and U. $~$\\ 
You select an option by pressing the corresponding key on your keyboard. $~$\\ 
Each time you select an option, you will get a different number of points. $~$\\ 
Your goal is to win as many points as possible. $~$\\ 
 $~$\\ 
You press $<<$L$>>$ and get 84.0 points. $~$\\ 
You press $<<$G$>>$ and get 90.0 points. $~$\\ 
You press $<<$O$>>$ and get 53.0 points. $~$\\ 
You press $<<$U$>>$ and get 24.0 points. $~$\\ 
You press $<<$G$>>$ and get 92.0 points. $~$\\ 
You press $<<$G$>>$ and get 78.0 points. $~$\\ 
You press $<<$L$>>$ and get 71.0 points. $~$\\ 
You press $<<$L$>>$ and get 75.0 points. $~$\\ 
You press $<<$G$>>$ and get 80.0 points. $~$\\ 
You press $<<$G$>>$ and get 80.0 points. $~$\\ 
You press $<<$G$>>$ and get 91.0 points. $~$\\ 
You press $<<$G$>>$ and get 90.0 points. $~$\\ 
You press $<<$U$>>$ and get 29.0 points. $~$\\ 
You press $<<$O$>>$ and get 45.0 points. $~$\\ 
You press $<<$G$>>$ and get 81.0 points. $~$\\ 
You press $<<$G$>>$ and get 75.0 points. $~$\\ 
You press $<<$G$>>$ and get 82.0 points. $~$\\ 
You press $<<$G$>>$ and get 82.0 points. $~$\\ 
You press $<<$G$>>$ and get 87.0 points. $~$\\ 
You press $<<$G$>>$ and get 85.0 points. $~$\\ 
You press $<<$G$>>$ and get 87.0 points. $~$\\ 
You press $<<$G$>>$ and get 87.0 points. $~$\\ 
You press $<<$G$>>$ and get 79.0 points. $~$\\ 
You press $<<$G$>>$ and get 75.0 points. $~$\\ 
You press $<<$L$>>$ and get 61.0 points. $~$\\ 
You press $<<$O$>>$ and get 40.0 points. $~$\\ 
You press $<<$U$>>$ and get 37.0 points. $~$\\ 
You press $<<$G$>>$ and get 72.0 points. $~$\\ 
You press $<<$G$>>$ and get 73.0 points. $~$\\ 
You press $<<$L$>>$ and get 66.0 points. $~$\\ 
You press $<<$G$>>$ and get 57.0 points. $~$\\ 
You press $<<$L$>>$ and get 64.0 points. $~$\\ 
You press $<<$L$>>$ and get 63.0 points. $~$\\ 
You press $<<$L$>>$ and get 61.0 points. $~$\\ 
You press $<<$O$>>$ and get 54.0 points. $~$\\ 
You press $<<$U$>>$ and get 30.0 points. $~$\\ 
You press $<<$L$>>$ and get 59.0 points. $~$\\ 
You press $<<$O$>>$ and get 56.0 points. $~$\\ 
You press $<<$O$>>$ and get 46.0 points. $~$\\ 
You press $<<$L$>>$ and get 59.0 points. $~$\\ 
You press $<<$G$>>$ and get 63.0 points. $~$\\ 
You press $<<$G$>>$ and get 63.0 points. $~$\\ 
You press $<<$G$>>$ and get 58.0 points. $~$\\ 
You press $<<$L$>>$ and get 53.0 points. $~$\\ 
You press $<<$O$>>$ and get 60.0 points. $~$\\ 
You press $<<$O$>>$ and get 59.0 points. $~$\\ 
You press $<<$L$>>$ and get 52.0 points. $~$\\ 
You press $<<$O$>>$ and get 54.0 points. $~$\\ 
You press $<<$U$>>$ and get 23.0 points. $~$\\ 
You press $<<$G$>>$ and get 52.0 points. $~$\\ 
You press $<<$O$>>$ and get 52.0 points. $~$\\ 
You press $<<$L$>>$ and get 67.0 points. $~$\\ 
You press $<<$L$>>$ and get 71.0 points. $~$\\ 
You press $<<$L$>>$ and get 71.0 points. $~$\\ 
You press $<<$L$>>$ and get 69.0 points. $~$\\ 
You press $<<$G$>>$ and get 46.0 points. $~$\\ 
You press $<<$O$>>$ and get 47.0 points. $~$\\ 
You press $<<$U$>>$ and get 19.0 points. $~$\\ 
You press $<<$L$>>$ and get 63.0 points. $~$\\ 
You press $<<$L$>>$ and get 58.0 points. $~$\\ 
You press $<<$L$>>$ and get 62.0 points. $~$\\ 
You press $<<$L$>>$ and get 53.0 points. $~$\\ 
You press $<<$L$>>$ and get 59.0 points. $~$\\ 
You press $<<$L$>>$ and get 65.0 points. $~$\\ 
You press $<<$G$>>$ and get 58.0 points. $~$\\ 
You press $<<$O$>>$ and get 54.0 points. $~$\\ 
You press $<<$L$>>$ and get 61.0 points. $~$\\ 
You press $<<$G$>>$ and get 66.0 points. $~$\\ 
You press $<<$O$>>$ and get 62.0 points. $~$\\ 
You press $<<$L$>>$ and get 61.0 points. $~$\\ 
You press $<<$G$>>$ and get 56.0 points. $~$\\ 
You press $<<$O$>>$ and get 58.0 points. $~$\\ 
You press $<<$L$>>$ and get 49.0 points. $~$\\ 
You press $<<$O$>>$ and get 50.0 points. $~$\\ 
You press $<<$G$>>$ and get 63.0 points. $~$\\ 
You press $<<$G$>>$ and get 68.0 points. $~$\\ 
You press $<<$G$>>$ and get 56.0 points. $~$\\ 
You press $<<$G$>>$ and get 59.0 points. $~$\\ 
You press $<<$G$>>$ and get 66.0 points. $~$\\ 
You press $<<$G$>>$ and get 56.0 points. $~$\\ 
You press $<<$G$>>$ and get 56.0 points. $~$\\ 
You press $<<$L$>>$ and get 48.0 points. $~$\\ 
You press $<<$G$>>$ and get 59.0 points. $~$\\ 
You press $<<$G$>>$ and get 55.0 points. $~$\\ 
You press $<<$O$>>$ and get 51.0 points. $~$\\ 
You press $<<$O$>>$ and get 58.0 points. $~$\\ 
You press $<<$O$>>$ and get 51.0 points. $~$\\ 
You press $<<$O$>>$ and get 62.0 points. $~$\\ 
You press $<<$O$>>$ and get 64.0 points. $~$\\ 
You press $<<$O$>>$ and get 60.0 points. $~$\\ 
You press $<<$O$>>$ and get 62.0 points. $~$\\ 
You press $<<$O$>>$ and get 64.0 points. $~$\\ 
You press $<<$O$>>$ and get 50.0 points. $~$\\ 
You press $<<$U$>>$ and get 40.0 points. $~$\\ 
You press $<<$O$>>$ and get 50.0 points. $~$\\ 
You press $<<$G$>>$ and get 49.0 points. $~$\\ 
You press $<<$L$>>$ and get 33.0 points. $~$\\ 
You press $<<$G$>>$ and get 55.0 points. $~$\\ 
You press $<<$O$>>$ and get 78.0 points. $~$\\ 
You press $<<$O$>>$ and get 79.0 points. $~$\\ 
You press $<<$O$>>$ and get 85.0 points. $~$\\ 
You press $<<$O$>>$ and get 82.0 points. $~$\\ 
You press $<<$O$>>$ and get 83.0 po 

\subsubsection*{Multiple-cue judgment}
Data source: \cite{collsioo2023numerical} \\ $~$ \\
Number of experiments: 3 $~$\\ 
Number of participants: 232 $~$\\ 
Number of choices: 52464 $~$\\ 
 $~$\\ 
\textbf{Example prompt:}
 $~$\\ 
Your task is to estimate the blood concentration of the hormone Caldionine based on information about the amount of two other hormones, Progladine and Amalydine, in multiple individuals' urine. $~$\\ 
Both Progladine and Amalydine can take five values (very little, a little, average, a lot, very much). $~$\\ 
Caldionine can take nine values (extremely low, very low, low, somewhat low, normal, somewhat high, high, very high, extremely high). $~$\\ 
Your goal is to estimate the concentration of Caldionine correctly. $~$\\ 
You will receive feedback about the actual concentration after making your estimate. $~$\\ 
This feedback will stop at some point. $~$\\ 
 $~$\\ 
Progladine: a lot. Amalydine: very much. You say that the Caldionine concentration is $<<$high$>>$. That is incorrect. The correct concentration of Caldionine is somewhat low. $~$\\ 
Progladine: average. Amalydine: a lot. You say that the Caldionine concentration is $<<$somewhat low$>>$. That is correct. The correct concentration of Caldionine is indeed somewhat low. $~$\\ 
Progladine: a lot. Amalydine: average. You say that the Caldionine concentration is $<<$normal$>>$. That is incorrect. The correct concentration of Caldionine is somewhat high. $~$\\ 
Progladine: average. Amalydine: a little. You say that the Caldionine concentration is $<<$low$>>$. That is incorrect. The correct concentration of Caldionine is somewhat high. $~$\\ 
Progladine: a little. Amalydine: average. You say that the Caldionine concentration is $<<$normal$>>$. That is incorrect. The correct concentration of Caldionine is somewhat low. $~$\\ 
Progladine: a lot. Amalydine: a little. You say that the Caldionine concentration is $<<$very high$>>$. That is incorrect. The correct concentration of Caldionine is high. $~$\\ 
Progladine: very little. Amalydine: very little. You say that the Caldionine concentration is $<<$normal$>>$. That is correct. The correct concentration of Caldionine is indeed normal. $~$\\ 
Progladine: very much. Amalydine: a little. You say that the Caldionine concentration is $<<$very high$>>$. That is correct. The correct concentration of Caldionine is indeed very high. $~$\\ 
Progladine: a lot. Amalydine: very little. You say that the Caldionine concentration is $<<$very high$>>$. That is correct. The correct concentration of Caldionine is indeed very high. $~$\\ 
Progladine: a little. Amalydine: a little. You say that the Caldionine concentration is $<<$normal$>>$. That is correct. The correct concentration of Caldionine is indeed normal. $~$\\ 
Progladine: a little. Amalydine: very little. You say that the Caldionine concentration is $<<$somewhat high$>>$. That is correct. The correct concentration of Caldionine is indeed somewhat high. $~$\\ 
Progladine: very little. Amalydine: a little. You say that the Caldionine concentration is $<<$somewhat low$>>$. That is correct. The correct concentration of Caldionine is indeed somewhat low. $~$\\ 
Progladine: very much. Amalydine: a lot. You say that the Caldionine concentration is $<<$very high$>>$. That is incorrect. The correct concentration of Caldionine is somewhat high. $~$\\ 
Progladine: a little. Amalydine: very much. You say that the Caldionine concentration is $<<$very low$>>$. That is correct. The correct concentration of Caldionine is indeed very low. $~$\\ 
Progladine: average. Amalydine: very much. You say that the Caldionine concentration is $<<$low$>>$. That is correct. The correct concentration of Caldionine is indeed low. $~$\\ 
Progladine: very much. Amalydine: average. You say that the Caldionine concentration is $<<$somewhat high$>>$. That is incorrect. The correct concentration of Caldionine is high. $~$\\ 
Progladine: average. Amalydine: very little. You say that the Caldionine concentration is $<<$high$>>$. That is correct. The correct concentration of Caldionine is indeed high. $~$\\ 
Progladine: a little. Amalydine: a lot. You say that the Caldionine concentration is $<<$normal$>>$. That is incorrect. The correct concentration of Caldionine is low. $~$\\ 
Progladine: very much. Amalydine: very much. You say that the Caldionine concentration is $<<$normal$>>$. That is correct. The correct concentration of Caldionine is indeed normal. $~$\\ 
Progladine: a lot. Amalydine: a lot. You say that the Caldionine concentration is $<<$normal$>>$. That is correct.  

\subsubsection*{Recall and recognition}
Data source: \cite{cox2018information} \\ $~$ \\
Number of experiments: 1 $~$\\ 
Number of participants: 424 $~$\\ 
Number of choices: 109201 $~$\\ 
 $~$\\ 
\textbf{Example prompt:}
 $~$\\ 
You study the following 20 word pairs: $~$\\ 
FILE, GERMAN $~$\\ 
STANDS, RISES $~$\\ 
OFFICER, FUEL $~$\\ 
CLASSROOM, JOURNEY $~$\\ 
TERRITORY, EDUCATIONAL $~$\\ 
TRANSPORTATION, AGREEMENT $~$\\ 
SIGNIFICANT, SPECIALIZED $~$\\ 
TUBE, ENEMY $~$\\ 
CLUB, RAPID $~$\\ 
LEGAL, CONCEPT $~$\\ 
TOWARDS, VICTORY $~$\\ 
RARELY, BAY $~$\\ 
ELECTION, ROYAL $~$\\ 
BELONG, SUPREME $~$\\ 
FRIGHTENED, PLAINS $~$\\ 
SLAVES, FILM $~$\\ 
STOMACH, WHISPERED $~$\\ 
NURSE, GOLDEN $~$\\ 
POLITICS, WINGS $~$\\ 
BOXES, TOM $~$\\ 
 $~$\\ 
You will now view a single pair of words. $~$\\ 
Your task is to indicate if the pair of words you see on the screen was studied as a pair on the list you just studied (by pressing D) or was not a pair (by pressing N). $~$\\ 
 $~$\\ 
You view the word pair SPECIALIZED, POLITICS. You press $<<$N$>>$. $~$\\ 
You view the word pair TOWARDS, VICTORY. You press $<<$D$>>$. $~$\\ 
You view the word pair SUPREME, BELONG. You press $<<$D$>>$. $~$\\ 
You view the word pair WINGS, OFFICER. You press $<<$D$>>$. $~$\\ 
You view the word pair CONCEPT, FRIGHTENED. You press $<<$N$>>$. $~$\\ 
You view the word pair GERMAN, NURSE. You press $<<$N$>>$. $~$\\ 
You view the word pair BAY, RARELY. You press $<<$D$>>$. $~$\\ 
You view the word pair FILE, GOLDEN. You press $<<$N$>>$. $~$\\ 
You view the word pair JOURNEY, LEGAL. You press $<<$N$>>$. $~$\\ 
You view the word pair CLASSROOM, ENEMY. You press $<<$N$>>$. $~$\\ 
You view the word pair ROYAL, ELECTION. You press $<<$D$>>$. $~$\\ 
You view the word pair BOXES, TOM. You press $<<$D$>>$. $~$\\ 
You view the word pair TERRITORY, EDUCATIONAL. You press $<<$D$>>$. $~$\\ 
You view the word pair WHISPERED, STOMACH. You press $<<$D$>>$. $~$\\ 
You view the word pair STANDS, RISES. You press $<<$D$>>$. $~$\\ 
You view the word pair AGREEMENT, TUBE. You press $<<$N$>>$. $~$\\ 
You view the word pair FUEL, TRANSPORTATION. You press $<<$N$>>$. $~$\\ 
You view the word pair CLUB, RAPID. You press $<<$D$>>$. $~$\\ 
You view the word pair SLAVES, FILM. You press $<<$D$>>$. $~$\\ 
You view the word pair PLAINS, SIGNIFICANT. You press $<<$N$>>$. $~$\\ 
 $~$\\ 
You study the following 20 word pairs: $~$\\ 
METALS, BEAUTY $~$\\ 
SHORE, COLONISTS $~$\\ 
HIGHEST, FASHION $~$\\ 
INFLUENCED, FLEW $~$\\ 
SHADOW, GRANDFATHER $~$\\ 
MOTOR, DISAPPEARED $~$\\ 
DETAIL, SHOULDERS $~$\\ 
REPORTS, MALE $~$\\ 
PAINT, COMMUNITIES $~$\\ 
STORM, RESULTING $~$\\ 
SELECTION, NIGHTS $~$\\ 
APPLICATION, FELLOW $~$\\ 
DESIGN, PRINCIPLE $~$\\ 
POSSIBILITY, AFRICAN $~$\\ 
REMOVE, TRAIL $~$\\ 
STAFF, JUSTICE $~$\\ 
SECTIONS, TRADITIONAL $~$\\ 
CLAY, MIXED $~$\\ 
TEA, AGRICULTURE $~$\\ 
PATIENT, LIMIT $~$\\ 
 $~$\\ 
You will now view a single word item. $~$\\ 
Your task is to type the other word in the pair. $~$\\ 
For example if you studied BRICK, BRACK and you now see BRICK, your response should be BRACK. $~$\\ 
The spelling does not matter; focus on providing as many responses as possible. $~$\\ 
If you cannot recall the word, answer DON'T REMEMBER. $~$\\ 
 $~$\\ 
You view the word STAFF. You answer $<<$JUSTICE$>>$. $~$\\ 
You view the word MOTOR. You answer $<<$DON'T REMEMBER$>>$. $~$\\ 
You view the word INFLUENCED. You answer $<<$DON'T REMEMBER$>>$. $~$\\ 
You view the word DETAIL. You answer $<<$SHOULDER$>>$. $~$\\ 
You view the word REMOVE. You answer $<<$DON'T REMEMBER$>>$. $~$\\ 
You view the word PATIENT. You answer $<<$LIMITED$>>$. $~$\\ 
You view the word PAINT. You answer $<<$COMMUNITY$>>$. $~$\\ 
You view the word STORM. You answer $<<$RESULTS$>>$. $~$\\ 
You view the word TEA. You answer $<<$ARIGCULTURE$>>$. $~$\\ 
You view the word SHADOW. You answer $<<$GRANDFATHER$>>$. $~$\\ 
You view the word APPLICATION. You answer $<<$FELLOW$>>$. $~$\\ 
You view the word METALS. You answer $<<$BEAUTY$>>$. $~$\\ 
You view the word POSSIBILITY. You answer $<<$AFRICANS$>>$. $~$\\ 
You view the word DESIGN. You answer $<<$PRINCPLE$>>$. $~$\\ 
You view the word SHORE. You answer $<<$DON'T REMEMBER$>>$. $~$\\ 
You view the word HIGHEST. You answer $<<$DON'T REMEMBER$>>$. $~$\\ 
You view the word CLAY. You answer $<<$MIXED$>>$. $~$\\ 
You view the word REPORTS. You answer $<<$DON'T REMEMBER$>>$. $~$\\ 
You view the word SECTIONS. You answer $<<$DON'T REMEMBER$>>$. $~$\\ 
You view the word SELECTION. You answer $<<$DON'T REMEMBER$>>$. $~$\\ 
 $~$\\ 
You study the following 20 word pairs: $~$\\ 
DEFENSE, EXPANSION $~$\\ 
PROGRESS, PROMISE $~$\\ 
EGG, ATTRACTIVE $~$\\ 
CRYING, RECOGNIZE $~$\\ 
PURE, AUNT $~$\\ 
OBTAINED, FEATURE $~$\\ 
TRAVELED, FLOWER $~$\\ 
INCREASINGLY, FEMALE $~$\\ 
POURED, VALUABLE $~$\\ 
TIGHT, HOLDS $~$\\ 
BIGGER, HONOR $~$\\ 
DRUG, SPENDING $~$\\ 
DOUBLE, OUTPUT $~$\\ 
ESTABLISH, CRITICAL $~$\\ 
DRIED, CHRISTMAS $~$\\ 
PROCESSING, NUMEROUS $~$\\ 
MEASURES, PARAGRAPH $~$\\ 
AFFAIRS, MOVEMENTS $~$\\ 
DAMAGE, FED $~$\\ 
WARS, CONTRACT $~$\\ 
 $~$\\ 
You will now view a single word item. $~$\\ 
Your task is to indicate if the item you view was on the list you just studied (by pressing D) or not on the list (by pressing N). $~$\\ 
 $~$\\ 
You view the word EXPANSION. You press $<<$D$>>$. 

\subsubsection*{N-back}
Data source: \cite{enkavi2019large} \\ $~$ \\
Number of experiments: 1 $~$\\ 
Number of participants: 470 $~$\\ 
Number of choices: 200821 $~$\\ 
 $~$\\ 
\textbf{Example prompt:}
 $~$\\ 
You will view a stream of letters on the screen, one letter at a time. $~$\\ 
At the beginning of a block, you are told a number N. $~$\\ 
You have to remember the last N letters you saw since the beginning of the block. $~$\\ 
If the letter you see matches the letter N trials ago, press W, otherwise press D. $~$\\ 
The case of the letter is irrelevant, so "t" matches "T" for example. $~$\\ 
If you make more than 5 mistakes in a block, N is decreased by 1. $~$\\ 
If you make fewer than 3 mistakes in a block, N is increased by 1. $~$\\ 
You will go through 20 blocks with 20+N trials each. $~$\\ 
     $~$\\ 
Block 0, N = 2: $~$\\ 
You see the letter d. $~$\\ 
You see the letter g. $~$\\ 
You see the letter D and press $<<$W$>>$. $~$\\ 
You see the letter v. $~$\\ 
You see the letter B. $~$\\ 
You see the letter V. $~$\\ 
You see the letter d. $~$\\ 
You see the letter g. $~$\\ 
You see the letter D and press $<<$W$>>$. $~$\\ 
You see the letter t. $~$\\ 
You see the letter G. $~$\\ 
You see the letter B and press $<<$D$>>$. $~$\\ 
You see the letter G. $~$\\ 
You see the letter g and press $<<$W$>>$. $~$\\ 
You see the letter g and press $<<$W$>>$. $~$\\ 
You see the letter v. $~$\\ 
You see the letter V. $~$\\ 
You see the letter B. $~$\\ 
You see the letter G. $~$\\ 
You see the letter G. $~$\\ 
You see the letter G and press $<<$W$>>$. $~$\\ 
You see the letter G. $~$\\ 
 $~$\\ 
Block 1, N = 1: $~$\\ 
You see the letter g. $~$\\ 
You see the letter D and press $<<$D$>>$. $~$\\ 
You see the letter D. $~$\\ 
You see the letter d and press $<<$W$>>$. $~$\\ 
You see the letter D. $~$\\ 
You see the letter d and press $<<$W$>>$. $~$\\ 
You see the letter v. $~$\\ 
You see the letter b and press $<<$D$>>$. $~$\\ 
You see the letter V. $~$\\ 
You see the letter t and press $<<$D$>>$. $~$\\ 
You see the letter d. $~$\\ 
You see the letter b and press $<<$D$>>$. $~$\\ 
You see the letter G. $~$\\ 
You see the letter t and press $<<$D$>>$. $~$\\ 
You see the letter D. $~$\\ 
You see the letter t and press $<<$D$>>$. $~$\\ 
You see the letter t. $~$\\ 
You see the letter b and press $<<$D$>>$. $~$\\ 
You see the letter B. $~$\\ 
You see the letter B and press $<<$W$>>$. $~$\\ 
You see the letter v. $~$\\ 
 $~$\\ 
Block 2, N = 1: $~$\\ 
You see the letter g. $~$\\ 
You see the letter g and press $<<$W$>>$. $~$\\ 
You see the letter d. $~$\\ 
You see the letter d and press $<<$W$>>$. $~$\\ 
You see the letter T. $~$\\ 
You see the letter v. $~$\\ 
You see the letter V. $~$\\ 
You see the letter D and press $<<$W$>>$. $~$\\ 
You see the letter g. $~$\\ 
You see the letter G and press $<<$W$>>$. $~$\\ 
You see the letter t. $~$\\ 
You see the letter D and press $<<$D$>>$. $~$\\ 
You see the letter t. $~$\\ 
You see the letter g and press $<<$D$>>$. $~$\\ 
You see the letter g. $~$\\ 
You see the letter D and press $<<$W$>>$. $~$\\ 
You see the letter d. $~$\\ 
You see the letter V. $~$\\ 
You see the letter B. $~$\\ 
You see the letter v and press $<<$D$>>$. $~$\\ 
You see the letter V. $~$\\ 
 $~$\\ 
Block 3, N = 1: $~$\\ 
You see the letter T. $~$\\ 
You see the letter d and press $<<$W$>>$. $~$\\ 
You see the letter t. $~$\\ 
You see the letter V and press $<<$D$>>$. $~$\\ 
You see the letter v. $~$\\ 
You see the letter B and press $<<$D$>>$. $~$\\ 
You see the letter b. $~$\\ 
You see the letter D and press $<<$D$>>$. $~$\\ 
You see the letter d. $~$\\ 
You see the letter d and press $<<$W$>>$. $~$\\ 
You see the letter G. $~$\\ 
You see the letter B and press $<<$D$>>$. $~$\\ 
You see the letter G. $~$\\ 
You see the letter d. $~$\\ 
You see the letter G. $~$\\ 
You see the letter v and press $<<$D$>>$. $~$\\ 
You see the letter B. $~$\\ 
You see the letter t and press $<<$D$>>$. $~$\\ 
You see the letter T. $~$\\ 
You see the letter T and press $<<$W$>>$. $~$\\ 
You see the letter T. $~$\\ 
 $~$\\ 
Block 4, N = 1: $~$\\ 
You see the letter d. $~$\\ 
You see the letter d and press $<<$W$>>$. $~$\\ 
You see the letter d. $~$\\ 
You see the letter V and press $<<$D$>>$. $~$\\ 
You see the letter V. $~$\\ 
You see the letter v and press $<<$D$>>$. $~$\\ 
You see the letter B. $~$\\ 
You see the letter B and press $<<$W$>>$. $~$\\ 
You see the letter g. $~$\\ 
You see the letter B. $~$\\ 
You see the letter t. $~$\\ 
You see the letter D. $~$\\ 
You see the letter d. $~$\\ 
You see the letter B. $~$\\ 
You see the letter B. $~$\\ 
You see the letter t. $~$\\ 
You see the letter d. $~$\\ 
You see the letter G. $~$\\ 
You see the letter d. $~$\\ 
You see the letter v. $~$\\ 
You see the letter t. $~$\\ 
 $~$\\ 
Block 5, N = 1: $~$\\ 
You see the letter v. $~$\\ 
You see the letter t and press $<<$D$>>$. $~$\\ 
You see the letter D. $~$\\ 
You see the letter V and press $<<$D$>>$. $~$\\ 
You see the letter v. $~$\\ 
You see the letter D and press $<<$D$>>$. $~$\\ 
You see the letter D. $~$\\ 
You see the letter B and press $<<$D$>>$. $~$\\ 
You see the letter g. $~$\\ 
You see the letter v and press $<<$D$>>$. $~$\\ 
You see the letter d. $~$\\ 
You see the letter g. $~$\\ 
You see the letter G. $~$\\ 
You see the letter D and press $<<$D$>>$. $~$\\ 
You see the letter b. $~$\\ 
You see the letter v. $~$\\ 
You see the letter V. $~$\\ 
You see the letter D and press $<<$D$>>$. $~$\\ 
You see the  

\subsubsection*{Digit span}
Data source: \cite{enkavi2019large} \\ $~$ \\
Number of experiments: 1 $~$\\ 
Number of participants: 472 $~$\\ 
Number of choices: 97012 $~$\\ 
 $~$\\ 
\textbf{Example prompt:}
 $~$\\ 
You will view a series of digits and are then asked to recall them in the order you have seen them by pressing the corresponding keys. $~$\\ 
After having recalled all digits, please press 'S' to indicate the end of your recalled sequence. $~$\\ 
 $~$\\ 
The digits are the following: [4, 8, 2] $~$\\ 
You press $<<$4$>>$. $~$\\ 
You press $<<$8$>>$. $~$\\ 
You press $<<$2$>>$. $~$\\ 
You press $<<$S$>>$.  $~$\\ 
 $~$\\ 
You will view a series of digits and are then asked to recall them in the order you have seen them by pressing the corresponding keys. $~$\\ 
After having recalled all digits, please press 'S' to indicate the end of your recalled sequence. $~$\\ 
 $~$\\ 
The digits are the following: [5, 2, 8, 5] $~$\\ 
You press $<<$5$>>$. $~$\\ 
You press $<<$2$>>$. $~$\\ 
You press $<<$8$>>$. $~$\\ 
You press $<<$5$>>$. $~$\\ 
You press $<<$S$>>$.  $~$\\ 
 $~$\\ 
You will view a series of digits and are then asked to recall them in the order you have seen them by pressing the corresponding keys. $~$\\ 
After having recalled all digits, please press 'S' to indicate the end of your recalled sequence. $~$\\ 
 $~$\\ 
The digits are the following: [9, 2, 8, 2, 5] $~$\\ 
You press $<<$9$>>$. $~$\\ 
You press $<<$2$>>$. $~$\\ 
You press $<<$8$>>$. $~$\\ 
You press $<<$2$>>$. $~$\\ 
You press $<<$5$>>$. $~$\\ 
You press $<<$S$>>$.  $~$\\ 
 $~$\\ 
You will view a series of digits and are then asked to recall them in the order you have seen them by pressing the corresponding keys. $~$\\ 
After having recalled all digits, please press 'S' to indicate the end of your recalled sequence. $~$\\ 
 $~$\\ 
The digits are the following: [8, 1, 4, 8, 4, 9] $~$\\ 
You press $<<$8$>>$. $~$\\ 
You press $<<$1$>>$. $~$\\ 
You press $<<$4$>>$. $~$\\ 
You press $<<$8$>>$. $~$\\ 
You press $<<$8$>>$. $~$\\ 
You press $<<$1$>>$. $~$\\ 
You press $<<$S$>>$.  $~$\\ 
 $~$\\ 
You will view a series of digits and are then asked to recall them in the order you have seen them by pressing the corresponding keys. $~$\\ 
After having recalled all digits, please press 'S' to indicate the end of your recalled sequence. $~$\\ 
 $~$\\ 
The digits are the following: [5, 1, 7, 1, 6, 3] $~$\\ 
You press $<<$7$>>$. $~$\\ 
You press $<<$1$>>$. $~$\\ 
You press $<<$5$>>$. $~$\\ 
You press $<<$1$>>$. $~$\\ 
You press $<<$6$>>$. $~$\\ 
You press $<<$3$>>$. $~$\\ 
You press $<<$S$>>$.  $~$\\ 
 $~$\\ 
You will view a series of digits and are then asked to recall them in the order you have seen them by pressing the corresponding keys. $~$\\ 
After having recalled all digits, please press 'S' to indicate the end of your recalled sequence. $~$\\ 
 $~$\\ 
The digits are the following: [5, 9, 5, 9, 1] $~$\\ 
You press $<<$5$>>$. $~$\\ 
You press $<<$9$>>$. $~$\\ 
You press $<<$5$>>$. $~$\\ 
You press $<<$9$>>$. $~$\\ 
You press $<<$1$>>$. $~$\\ 
You press $<<$S$>>$.  $~$\\ 
 $~$\\ 
You will view a series of digits and are then asked to recall them in the order you have seen them by pressing the corresponding keys. $~$\\ 
After having recalled all digits, please press 'S' to indicate the end of your recalled sequence. $~$\\ 
 $~$\\ 
The digits are the following: [7, 2, 5, 8, 2, 6] $~$\\ 
You press $<<$7$>>$. $~$\\ 
You press $<<$7$>>$. $~$\\ 
You press $<<$2$>>$. $~$\\ 
You press $<<$8$>>$. $~$\\ 
You press $<<$5$>>$. $~$\\ 
You press $<<$2$>>$. $~$\\ 
You press $<<$S$>>$.  $~$\\ 
 $~$\\ 
You will view a series of digits and are then asked to recall them in the order you have seen them by pressing the corresponding keys. $~$\\ 
After having recalled all digits, please press 'S' to indicate the end of your recalled sequence. $~$\\ 
 $~$\\ 
The digits are the following: [8, 5, 8, 3, 9, 4] $~$\\ 
You press $<<$8$>>$. $~$\\ 
You press $<<$5$>>$. $~$\\ 
You press $<<$8$>>$. $~$\\ 
You press $<<$3$>>$. $~$\\ 
You press $<<$9$>>$. $~$\\ 
You press $<<$4$>>$. $~$\\ 
You press $<<$S$>>$.  $~$\\ 
 $~$\\ 
You will view a series of digits and are then asked to recall them in the order you have seen them by pressing the corresponding keys. $~$\\ 
After having recalled all digits, please press 'S' to indicate the end of your recalled sequence. $~$\\ 
 $~$\\ 
The digits are the following: [9, 6, 9, 5, 1, 7, 1] $~$\\ 
You press $<<$9$>>$. $~$\\ 
You press $<<$6$>>$. $~$\\ 
You press $<<$9$>>$. $~$\\ 
You press $<<$5$>>$. $~$\\ 
You press $<<$7$>>$. $~$\\ 
You press $<<$1$>>$. $~$\\ 
You press $<<$7$>>$. $~$\\ 
You press $<<$S$>>$.  $~$\\ 
 $~$\\ 
You will view a series of digits and are then asked to recall them in the order you have seen them by pressing the corresponding keys. $~$\\ 
After having recalled all digits, please press 'S' to indicate the end of your recalled sequence. $~$\\ 
 $~$\\ 
The digits are the following: [7, 2, 5, 9, 1, 8] $~$\\ 
You press $<<$7$>>$. $~$\\ 
You press $<<$2$>>$. $~$\\ 
You press $<<$5$>>$. $~$\\ 
You press $<<$9$>>$. $~$\\ 
You press $<<$1$>>$. $~$\\ 
You press $<<$8$>>$. $~$\\ 
You press $<<$S$>>$.  $~$\\ 
 $~$\\ 
You will view a series of digits and are then asked to recall them in the order you have seen them by pressing the corresponding keys. $~$\\ 
After having recalled all digits, please 

\subsubsection*{Go/no-go}
Data source: \cite{enkavi2019large} \\ $~$ \\
Number of experiments: 1 $~$\\ 
Number of participants: 463 $~$\\ 
Number of choices: 150517 $~$\\ 
 $~$\\ 
\textbf{Example prompt:}
 $~$\\ 
In this task, you need to emit responses to certain stimuli and omit responses to others. $~$\\ 
You will see one of two colours, colour1 or colour2, on the screen in each trial. $~$\\ 
You need to press button X when you see colour1 and press nothing when you see colour2. $~$\\ 
You need to respond as quickly as possible. $~$\\ 
You will be doing 10 practice trials followed by 350 test trials. $~$\\ 
 $~$\\ 
You see colour1 and press nothing. $~$\\ 
You see colour2 and press $<<$X$>>$ in 753.0ms. $~$\\ 
You see colour2 and press $<<$X$>>$ in 381.0ms. $~$\\ 
You see colour2 and press nothing. $~$\\ 
You see colour1 and press $<<$X$>>$ in 473.0ms. $~$\\ 
You see colour1 and press $<<$X$>>$ in 713.0ms. $~$\\ 
You see colour2 and press nothing. $~$\\ 
You see colour1 and press $<<$X$>>$ in 364.0ms. $~$\\ 
You see colour2 and press nothing. $~$\\ 
You see colour1 and press $<<$X$>>$ in 378.0ms. $~$\\ 
You see colour1 and press $<<$X$>>$ in 794.0ms. $~$\\ 
You see colour1 and press $<<$X$>>$ in 436.0ms. $~$\\ 
You see colour1 and press $<<$X$>>$ in 427.0ms. $~$\\ 
You see colour1 and press $<<$X$>>$ in 337.0ms. $~$\\ 
You see colour1 and press $<<$X$>>$ in 269.0ms. $~$\\ 
You see colour1 and press $<<$X$>>$ in 312.0ms. $~$\\ 
You see colour1 and press $<<$X$>>$ in 273.0ms. $~$\\ 
You see colour2 and press nothing. $~$\\ 
You see colour1 and press $<<$X$>>$ in 288.0ms. $~$\\ 
You see colour1 and press $<<$X$>>$ in 276.0ms. $~$\\ 
You see colour1 and press $<<$X$>>$ in 314.0ms. $~$\\ 
You see colour1 and press $<<$X$>>$ in 309.0ms. $~$\\ 
You see colour1 and press $<<$X$>>$ in 320.0ms. $~$\\ 
You see colour1 and press $<<$X$>>$ in 342.0ms. $~$\\ 
You see colour1 and press $<<$X$>>$ in 301.0ms. $~$\\ 
You see colour1 and press $<<$X$>>$ in 289.0ms. $~$\\ 
You see colour2 and press nothing. $~$\\ 
You see colour1 and press $<<$X$>>$ in 360.0ms. $~$\\ 
You see colour2 and press $<<$X$>>$ in 424.0ms. $~$\\ 
You see colour2 and press nothing. $~$\\ 
You see colour1 and press $<<$X$>>$ in 525.0ms. $~$\\ 
You see colour1 and press $<<$X$>>$ in 306.0ms. $~$\\ 
You see colour1 and press $<<$X$>>$ in 387.0ms. $~$\\ 
You see colour1 and press $<<$X$>>$ in 292.0ms. $~$\\ 
You see colour1 and press $<<$X$>>$ in 317.0ms. $~$\\ 
You see colour1 and press $<<$X$>>$ in 270.0ms. $~$\\ 
You see colour1 and press $<<$X$>>$ in 278.0ms. $~$\\ 
You see colour2 and press nothing. $~$\\ 
You see colour1 and press $<<$X$>>$ in 277.0ms. $~$\\ 
You see colour1 and press $<<$X$>>$ in 311.0ms. $~$\\ 
You see colour1 and press $<<$X$>>$ in 338.0ms. $~$\\ 
You see colour1 and press $<<$X$>>$ in 323.0ms. $~$\\ 
You see colour1 and press $<<$X$>>$ in 304.0ms. $~$\\ 
You see colour1 and press $<<$X$>>$ in 323.0ms. $~$\\ 
You see colour1 and press $<<$X$>>$ in 354.0ms. $~$\\ 
You see colour1 and press $<<$X$>>$ in 292.0ms. $~$\\ 
You see colour1 and press $<<$X$>>$ in 302.0ms. $~$\\ 
You see colour1 and press $<<$X$>>$ in 309.0ms. $~$\\ 
You see colour2 and press nothing. $~$\\ 
You see colour1 and press $<<$X$>>$ in 340.0ms. $~$\\ 
You see colour1 and press $<<$X$>>$ in 603.0ms. $~$\\ 
You see colour1 and press $<<$X$>>$ in 289.0ms. $~$\\ 
You see colour1 and press $<<$X$>>$ in 284.0ms. $~$\\ 
You see colour1 and press $<<$X$>>$ in 275.0ms. $~$\\ 
You see colour1 and press $<<$X$>>$ in 299.0ms. $~$\\ 
You see colour2 and press nothing. $~$\\ 
You see colour1 and press $<<$X$>>$ in 265.0ms. $~$\\ 
You see colour1 and press $<<$X$>>$ in 267.0ms. $~$\\ 
You see colour2 and press nothing. $~$\\ 
You see colour1 and press $<<$X$>>$ in 274.0ms. $~$\\ 
You see colour1 and press $<<$X$>>$ in 382.0ms. $~$\\ 
You see colour1 and press $<<$X$>>$ in 272.0ms. $~$\\ 
You see colour2 and press nothing. $~$\\ 
You see colour1 and press $<<$X$>>$ in 258.0ms. $~$\\ 
You see colour1 and press $<<$X$>>$ in 305.0ms. $~$\\ 
You see colour1 and press $<<$X$>>$ in 320.0ms. $~$\\ 
You see colour1 and press $<<$X$>>$ in 261.0ms. $~$\\ 
You see colour1 and press $<<$X$>>$ in 275.0ms. $~$\\ 
You see colour2 and press nothing. $~$\\ 
You see colour1 and press $<<$X$>>$ in 424.0ms. $~$\\ 
You see colour1 and press $<<$X$>>$ in 266.0ms. $~$\\ 
You see colour1 and press $<<$X$>>$ in 273.0ms. $~$\\ 
You see colour1 and press $<<$X$>>$ in 287.0ms. $~$\\ 
You see colour1 and press $<<$X$>>$ in 437.0ms. $~$\\ 
You see colour1 and press $<<$X$>>$ in 293.0ms. $~$\\ 
You see colour1 and press $<<$X$>>$ in 297.0ms. $~$\\ 
You see colour1 and press $<<$X$>>$ in 308.0ms. $~$\\ 
You see colour1 and press $<<$X$>>$ in 313.0ms. $~$\\ 
You see colour1 and press $<<$X$>>$ in 373.0ms. $~$\\ 
You see colour1 and press $<<$X$>>$ in 390.0ms. $~$\\ 
You see colour1 and press $<<$X$>>$ in 304.0ms. $~$\\ 
You see colour1 and press $<<$X$>>$ in 334.0ms. $~$\\ 
You see colour1 and press $<<$X$>>$ in 326.0ms. $~$\\ 
You see colour1 and press $<<$X$>>$ in 382.0ms. $~$\\ 
You see colour1 and press $<<$X$>>$ in 803.0ms. $~$\\ 
You see colour1 and press $<<$X$>>$ in 430.0ms. $~$\\ 
You see colour1 and press $<<$X$>>$ in 324.0ms. $~$\\ 
You see colour

\subsubsection*{Recent probes}
Data source: \cite{enkavi2019large} \\ $~$ \\
Number of experiments: 1 $~$\\ 
Number of participants: 471 $~$\\ 
Number of choices: 34714 $~$\\ 
 $~$\\ 
\textbf{Example prompt:}
 $~$\\ 
You will repeatedly observe sequences of six letters. $~$\\ 
You have to remember these letters before they disappear. $~$\\ 
Afterward, you will be prompted with one letter. You have to answer whether the letter was part of the six previous letters. $~$\\ 
If you think it was, you have to press C. If you think it was not, press Q. $~$\\ 
 $~$\\ 
You are shown the letters ['C', 'I', 'Q', 'F', 'W', 'Z']. You see the letter Y. You press $<<$Q$>>$. $~$\\ 
You are shown the letters ['I', 'Q', 'C', 'D', 'M', 'V']. You see the letter U. You press $<<$Q$>>$. $~$\\ 
You are shown the letters ['I', 'O', 'C', 'X', 'A', 'Q']. You see the letter M. You press $<<$C$>>$. $~$\\ 
You are shown the letters ['Z', 'C', 'W', 'I', 'J', 'O']. You see the letter C. You press $<<$Q$>>$. $~$\\ 
You are shown the letters ['Q', 'M', 'F', 'V', 'P', 'E']. You see the letter W. You press $<<$C$>>$. $~$\\ 
You are shown the letters ['W', 'F', 'U', 'M', 'B', 'Q']. You see the letter V. You press $<<$Q$>>$. $~$\\ 
You are shown the letters ['R', 'U', 'F', 'J', 'W', 'D']. You see the letter W. You press $<<$C$>>$. $~$\\ 
You are shown the letters ['X', 'U', 'R', 'Y', 'H', 'F']. You see the letter X. You press $<<$Q$>>$. $~$\\ 
You are shown the letters ['R', 'Q', 'M', 'X', 'V', 'U']. You see the letter W. You press $<<$C$>>$. $~$\\ 
You are shown the letters ['G', 'Q', 'M', 'N', 'R', 'O']. You see the letter V. You press $<<$Q$>>$. $~$\\ 
You are shown the letters ['T', 'P', 'Q', 'M', 'W', 'G']. You see the letter X. You press $<<$Q$>>$. $~$\\ 
You are shown the letters ['P', 'J', 'Q', 'S', 'D', 'T']. You see the letter J. You press $<<$Q$>>$. $~$\\ 
You are shown the letters ['J', 'R', 'H', 'Q', 'F', 'P']. You see the letter F. You press $<<$C$>>$. $~$\\ 
You are shown the letters ['B', 'V', 'J', 'G', 'R', 'H']. You see the letter R. You press $<<$Q$>>$. $~$\\ 
You are shown the letters ['X', 'J', 'V', 'L', 'B', 'D']. You see the letter B. You press $<<$Q$>>$. $~$\\ 
You are shown the letters ['N', 'M', 'J', 'C', 'V', 'X']. You see the letter C. You press $<<$Q$>>$. $~$\\ 
You are shown the letters ['N', 'T', 'J', 'R', 'M', 'W']. You see the letter C. You press $<<$Q$>>$. $~$\\ 
You are shown the letters ['N', 'T', 'J', 'E', 'I', 'D']. You see the letter N. You press $<<$Q$>>$. $~$\\ 
You are shown the letters ['J', 'T', 'N', 'K', 'C', 'B']. You see the letter T. You press $<<$C$>>$. $~$\\ 
You are shown the letters ['M', 'O', 'N', 'T', 'P', 'J']. You see the letter O. You press $<<$Q$>>$. $~$\\ 
You are shown the letters ['O', 'Q', 'W', 'U', 'M', 'N']. You see the letter Q. You press $<<$C$>>$. $~$\\ 
You are shown the letters ['Q', 'Z', 'Y', 'O', 'I', 'W']. You see the letter I. You press $<<$C$>>$. $~$\\ 
You are shown the letters ['Z', 'A', 'Y', 'F', 'Q', 'G']. You see the letter Q. You press $<<$C$>>$. $~$\\ 
You are shown the letters ['Z', 'M', 'Y', 'P', 'A', 'B']. You see the letter X. You press $<<$Q$>>$. $~$\\ 
You are shown the letters ['L', 'X', 'M', 'Z', 'Y', 'N']. You see the letter B. You press $<<$Q$>>$. $~$\\ 
You are shown the letters ['M', 'J', 'X', 'C', 'L', 'U']. You see the letter P. You press $<<$Q$>>$. $~$\\ 
You are shown the letters ['S', 'J', 'E', 'H', 'X', 'M']. You see the letter M. You press $<<$C$>>$. $~$\\ 
You are shown the letters ['F', 'I', 'E', 'A', 'J', 'S']. You see the letter X. You press $<<$Q$>>$. $~$\\ 
You are shown the letters ['D', 'U', 'I', 'M', 'O', 'T']. You see the letter Q. You press $<<$Q$>>$. $~$\\ 
You are shown the letters ['I', 'D', 'U', 'G', 'Q', 'W']. You see the letter B. You press $<<$Q$>>$. $~$\\ 
You are shown the letters ['U', 'I', 'B', 'L', 'D', 'Z']. You see the letter W. You press $<<$Q$>>$. $~$\\ 
You are shown the letters ['U', 'E', 'B', 'I', 'H', 'R']. You see the letter I. You press $<<$C$>>$. $~$\\ 
You are shown the letters ['S', 'B', 'U', 'E', 'F', 'N']. You see the letter I. You press $<<$Q$>>$. $~$\\ 
You are shown the letters ['Y', 'S', 'Q', 'U', 'B', 'A']. You see the letter Y. You press $<<$C$>>$. $~$\\ 
You are shown the letters ['Q', 'V', 'Y', 'S', 'H', 'O']. You see the letter H. You press $<<$C$>>$. $~$\\ 
You are shown the letters ['V', 'Z', 'P', 'D', 'Q', 'Y']. You see the letter H. You press $<<$C$>>$. $~$\\ 
You are shown the letters ['Z', 'P', 'V', 'W', 'L', 'N']. You see the letter H. You press $<<$Q$>>$. $~$\\ 
You are shown the letters ['U', 'S', 'Z', 'V', 'P', 'M']. You see the letter W. You press $<<$Q$>>$. $~$\\ 
You are shown the letters ['U', 'Y', 'Z', 'S', 'C', 'G']. You see the letter W. You press $<<$Q$>>$.

\subsubsection*{Horizon task}
Data source: \cite{feng2021dynamics} \\ $~$ \\
Number of experiments: 1 $~$\\ 
Number of participants: 26 $~$\\ 
Number of choices: 29120 $~$\\ 
 $~$\\ 
\textbf{Example prompt:}
 $~$\\ 
You are participating in multiple games involving two slot machines, labeled I and H. $~$\\ 
The two slot machines are different across different games. $~$\\ 
Each time you choose a slot machine, you get some points. $~$\\ 
You choose a slot machine by pressing the corresponding key. $~$\\ 
Each slot machine tends to pay out about the same amount of points on average. $~$\\ 
Your goal is to choose the slot machines that will give you the most points across the experiment. $~$\\ 
The first 4 trials in each game are instructed trials where you will be told which slot machine to choose. $~$\\ 
After these instructed trials, you will have the freedom to choose for either 1 or 6 trials. $~$\\ 
 $~$\\ 
Game 1. There are 10 trials in this game. $~$\\ 
You are instructed to press I and get 73 points. $~$\\ 
You are instructed to press H and get 91 points. $~$\\ 
You are instructed to press I and get 68 points. $~$\\ 
You are instructed to press H and get 95 points. $~$\\ 
You press $<<$I$>>$ and get 71 points. $~$\\ 
You press $<<$H$>>$ and get 96 points. $~$\\ 
You press $<<$I$>>$ and get 57 points. $~$\\ 
You press $<<$I$>>$ and get 30 points. $~$\\ 
You press $<<$I$>>$ and get 54 points. $~$\\ 
You press $<<$H$>>$ and get 81 points. $~$\\ 
 $~$\\ 
Game 2. There are 10 trials in this game. $~$\\ 
You are instructed to press I and get 38 points. $~$\\ 
You are instructed to press I and get 1 points. $~$\\ 
You are instructed to press I and get 18 points. $~$\\ 
You are instructed to press H and get 44 points. $~$\\ 
You press $<<$I$>>$ and get 5 points. $~$\\ 
You press $<<$I$>>$ and get 3 points. $~$\\ 
You press $<<$H$>>$ and get 53 points. $~$\\ 
You press $<<$I$>>$ and get 6 points. $~$\\ 
You press $<<$H$>>$ and get 37 points. $~$\\ 
You press $<<$H$>>$ and get 50 points. $~$\\ 
 $~$\\ 
Game 3. There are 5 trials in this game. $~$\\ 
You are instructed to press H and get 24 points. $~$\\ 
You are instructed to press H and get 34 points. $~$\\ 
You are instructed to press I and get 68 points. $~$\\ 
You are instructed to press I and get 53 points. $~$\\ 
You press $<<$I$>>$ and get 57 points. $~$\\ 
 $~$\\ 
Game 4. There are 10 trials in this game. $~$\\ 
You are instructed to press H and get 53 points. $~$\\ 
You are instructed to press I and get 57 points. $~$\\ 
You are instructed to press H and get 45 points. $~$\\ 
You are instructed to press H and get 51 points. $~$\\ 
You press $<<$H$>>$ and get 37 points. $~$\\ 
You press $<<$I$>>$ and get 35 points. $~$\\ 
You press $<<$I$>>$ and get 51 points. $~$\\ 
You press $<<$I$>>$ and get 37 points. $~$\\ 
You press $<<$I$>>$ and get 29 points. $~$\\ 
You press $<<$H$>>$ and get 53 points. $~$\\ 
 $~$\\ 
Game 5. There are 5 trials in this game. $~$\\ 
You are instructed to press H and get 46 points. $~$\\ 
You are instructed to press I and get 9 points. $~$\\ 
You are instructed to press H and get 38 points. $~$\\ 
You are instructed to press H and get 36 points. $~$\\ 
You press $<<$H$>>$ and get 37 points. $~$\\ 
 $~$\\ 
Game 6. There are 5 trials in this game. $~$\\ 
You are instructed to press I and get 35 points. $~$\\ 
You are instructed to press I and get 36 points. $~$\\ 
You are instructed to press I and get 27 points. $~$\\ 
You are instructed to press H and get 49 points. $~$\\ 
You press $<<$H$>>$ and get 41 points. $~$\\ 
 $~$\\ 
Game 7. There are 5 trials in this game. $~$\\ 
You are instructed to press H and get 53 points. $~$\\ 
You are instructed to press H and get 59 points. $~$\\ 
You are instructed to press I and get 80 points. $~$\\ 
You are instructed to press I and get 99 points. $~$\\ 
You press $<<$H$>>$ and get 54 points. $~$\\ 
 $~$\\ 
Game 8. There are 10 trials in this game. $~$\\ 
You are instructed to press H and get 21 points. $~$\\ 
You are instructed to press I and get 43 points. $~$\\ 
You are instructed to press H and get 18 points. $~$\\ 
You are instructed to press H and get 19 points. $~$\\ 
You press $<<$H$>>$ and get 26 points. $~$\\ 
You press $<<$H$>>$ and get 21 points. $~$\\ 
You press $<<$I$>>$ and get 47 points. $~$\\ 
You press $<<$I$>>$ and get 37 points. $~$\\ 
You press $<<$I$>>$ and get 37 points. $~$\\ 
You press $<<$I$>>$ and get 36 points. $~$\\ 
 $~$\\ 
Game 9. There are 5 trials in this game. $~$\\ 
You are instructed to press H and get 47 points. $~$\\ 
You are instructed to press H and get 41 points. $~$\\ 
You are instructed to press I and get 40 points. $~$\\ 
You are instructed to press I and get 35 points. $~$\\ 
You press $<<$I$>>$ and get 30 points. $~$\\ 
 $~$\\ 
Game 10. There are 5 trials in this game. $~$\\ 
You are instructed to press I and get 69 points. $~$\\ 
You are instructed to press I and get 69 points. $~$\\ 
You are instructed to press H and get 65 points. $~$\\ 
You are instructed to press I and get 68 points. $~$\\ 
You press $<<$I$>>$ and get 63 points. $~$\\ 
 $~$\\ 
Game 11. There are 10  

\subsubsection*{Gardening task}
Data source: \cite{flesch2018comparing}  \\ $~$ \\
Number of experiments: 1 $~$\\ 
Number of participants: 320 $~$\\ 
Number of choices: 192000 $~$\\ 
 $~$\\ 
\textbf{Example prompt:}
 $~$\\ 
You are going to plant trees in two different gardens labeled North and South. $~$\\ 
The trees look different from each other regarding their leafiness and branchiness. $~$\\ 
There are 5 levels of leafiness (0, 1, 2, 3, 4) and 5 levels of branchiness (0, 1, 2, 3, 4). $~$\\ 
In each round, you get presented with a tree. $~$\\ 
You can accept to plant the tree by pressing T and reject to plant it by pressing N. $~$\\ 
If you accept to plant the tree and your answer is correct, you will be rewarded with points, otherwise, you will lose some points. $~$\\ 
If you reject to plant the tree, you will not be rewarded (0 points). $~$\\ 
Your task is to learn which type of tree grows best in each garden. $~$\\ 
During the training phase, there will be feedback on every trial about your decisions. $~$\\ 
During the testing phase, there will be no feedback for your decision. $~$\\ 
 $~$\\ 
You get a tree with level 3 of leafiness and level 0 of branchiness in the South garden. You press $<<$T$>>$ and get -50 points. You would have gotten 0 points, had you rejected to plant the tree. $~$\\ 
You get a tree with level 4 of leafiness and level 0 of branchiness in the North garden. You press $<<$T$>>$ and get -50 points. You would have gotten 0 points, had you rejected to plant the tree. $~$\\ 
You get a tree with level 1 of leafiness and level 1 of branchiness in the South garden. You press $<<$T$>>$ and get -25 points. You would have gotten 0 points, had you rejected to plant the tree. $~$\\ 
You get a tree with level 3 of leafiness and level 1 of branchiness in the North garden. You press $<<$T$>>$ and get -25 points. You would have gotten 0 points, had you rejected to plant the tree. $~$\\ 
You get a tree with level 0 of leafiness and level 4 of branchiness in the North garden. You press $<<$T$>>$ and get 50 points. You would have gotten 0 points, had you rejected to plant the tree. $~$\\ 
You get a tree with level 2 of leafiness and level 2 of branchiness in the North garden. You press $<<$N$>>$ and get 0 points. You would have gotten 0 points, had you accepted to plant the tree. $~$\\ 
You get a tree with level 0 of leafiness and level 2 of branchiness in the South garden. You press $<<$T$>>$ and get 0 points. You would have gotten 0 points, had you rejected to plant the tree. $~$\\ 
You get a tree with level 4 of leafiness and level 3 of branchiness in the North garden. You press $<<$N$>>$ and get 0 points. You would have gotten -50 points, had you accepted to plant the tree. $~$\\ 
You get a tree with level 1 of leafiness and level 1 of branchiness in the North garden. You press $<<$T$>>$ and get 25 points. You would have gotten 0 points, had you rejected to plant the tree. $~$\\ 
You get a tree with level 2 of leafiness and level 1 of branchiness in the South garden. You press $<<$N$>>$ and get 0 points. You would have gotten -25 points, had you accepted to plant the tree. $~$\\ 
You get a tree with level 4 of leafiness and level 4 of branchiness in the North garden. You press $<<$N$>>$ and get 0 points. You would have gotten -50 points, had you accepted to plant the tree. $~$\\ 
You get a tree with level 1 of leafiness and level 3 of branchiness in the South garden. You press $<<$N$>>$ and get 0 points. You would have gotten 25 points, had you accepted to plant the tree. $~$\\ 
You get a tree with level 0 of leafiness and level 2 of branchiness in the South garden. You press $<<$N$>>$ and get 0 points. You would have gotten 0 points, had you accepted to plant the tree. $~$\\ 
You get a tree with level 4 of leafiness and level 1 of branchiness in the South garden. You press $<<$T$>>$ and get -25 points. You would have gotten 0 points, had you rejected to plant the tree. $~$\\ 
You get a tree with level 4 of leafiness and level 0 of branchiness in the North garden. You press $<<$N$>>$ and get 0 points. You would have gotten -50 points, had you accepted to plant the tree. $~$\\ 
You get a tree with level 1 of leafiness and level 1 of branchiness in the South garden. You press $<<$T$>>$ and get -25 points. You would have gotten 0 points, had you rejected to plant the tree. $~$\\ 
You get a tree with level 0 of leafiness and level 2 of branchiness in the South garden. You press $<<$N$>>$ and get 0 points. You would have gotten 0 points, had you accepted to plant the tree. $~$\\ 
You get a  

\subsubsection*{Columbia card task}
Data source: \cite{frey2017risk} \\ $~$ \\ 
Number of experiments: 1 $~$\\ 
Number of participants: 1368 $~$\\ 
Number of choices: 613299 $~$\\ 
 $~$\\ 
\textbf{Example prompt:}
 $~$\\ 
You will play a games with 84 rounds. $~$\\ 
In each round, you will be presented with 32 face-down cards. $~$\\ 
Every card is either a gain card or a loss card. $~$\\ 
If you turn over a gain card, the gain amount of that card (between 10 and 600 points) will be added to your current game score. $~$\\ 
If you turn over a loss card, the loss amount of that card (between 25 and 750 points) will be subtracted from your game score. $~$\\ 
In different rounds, between 1 and 28 cards are loss cards. $~$\\ 
Loss and gain amounts also differ between rounds. $~$\\ 
You may keep turning over cards as long as you keep encountering gain cards. $~$\\ 
You may also stop the round at any point and claim your current payout.  $~$\\ 
If you encounter a loss card, the round ends immediately. $~$\\ 
Your gains and losses will be summed up to give you your final score for each round. $~$\\ 
Press E to turn a card over, or C to stop the round and claim your current payout. $~$\\ 
 $~$\\ 
Round 1: $~$\\ 
You will be awarded 150 points for turning over a gain card. $~$\\ 
You will lose 75 points for turning over a loss card. $~$\\ 
There are 20 loss cards in this round. $~$\\ 
You press $<<$E$>>$ and turn over a loss card. Your current score is -75. The round has now ended because you encountered a loss card. $~$\\ 
Your final score for this round is -75. $~$\\ 
 $~$\\ 
Round 2: $~$\\ 
You will be awarded 50 points for turning over a gain card. $~$\\ 
You will lose 100 points for turning over a loss card. $~$\\ 
There are 1 loss cards in this round. $~$\\ 
You press $<<$E$>>$ and turn over a gain card. Your current score is 50. $~$\\ 
You press $<<$E$>>$ and turn over a gain card. Your current score is 100. $~$\\ 
You press $<<$E$>>$ and turn over a gain card. Your current score is 150. $~$\\ 
You press $<<$E$>>$ and turn over a gain card. Your current score is 200. $~$\\ 
You press $<<$E$>>$ and turn over a loss card. Your current score is 100. The round has now ended because you encountered a loss card. $~$\\ 
Your final score for this round is 100. $~$\\ 
 $~$\\ 
Round 3: $~$\\ 
You will be awarded 200 points for turning over a gain card. $~$\\ 
You will lose 100 points for turning over a loss card. $~$\\ 
There are 10 loss cards in this round. $~$\\ 
You press $<<$E$>>$ and turn over a gain card. Your current score is 200. $~$\\ 
You press $<<$E$>>$ and turn over a gain card. Your current score is 400. $~$\\ 
You press $<<$E$>>$ and turn over a gain card. Your current score is 600. $~$\\ 
You press $<<$C$>>$ and claim your payout. $~$\\ 
Your final score for this round is 600. $~$\\ 
 $~$\\ 
Round 4: $~$\\ 
You will be awarded 200 points for turning over a gain card. $~$\\ 
You will lose 50 points for turning over a loss card. $~$\\ 
There are 28 loss cards in this round. $~$\\ 
You press $<<$C$>>$ and claim your payout. $~$\\ 
Your final score for this round is 0. $~$\\ 
 $~$\\ 
Round 5: $~$\\ 
You will be awarded 20 points for turning over a gain card. $~$\\ 
You will lose 750 points for turning over a loss card. $~$\\ 
There are 1 loss cards in this round. $~$\\ 
You press $<<$E$>>$ and turn over a gain card. Your current score is 20. $~$\\ 
You press $<<$E$>>$ and turn over a gain card. Your current score is 40. $~$\\ 
You press $<<$E$>>$ and turn over a gain card. Your current score is 60. $~$\\ 
You press $<<$E$>>$ and turn over a gain card. Your current score is 80. $~$\\ 
You press $<<$E$>>$ and turn over a gain card. Your current score is 100. $~$\\ 
You press $<<$C$>>$ and claim your payout. $~$\\ 
Your final score for this round is 100. $~$\\ 
 $~$\\ 
Round 6: $~$\\ 
You will be awarded 300 points for turning over a gain card. $~$\\ 
You will lose 100 points for turning over a loss card. $~$\\ 
There are 16 loss cards in this round. $~$\\ 
You press $<<$E$>>$ and turn over a loss card. Your current score is -100. The round has now ended because you encountered a loss card. $~$\\ 
Your final score for this round is -100. $~$\\ 
 $~$\\ 
Round 7: $~$\\ 
You will be awarded 10 points for turning over a gain card. $~$\\ 
You will lose 500 points for turning over a loss card. $~$\\ 
There are 3 loss cards in this round. $~$\\ 
You press $<<$E$>>$ and turn over a gain card. Your current score is 10. $~$\\ 
You press $<<$E$>>$ and turn over a gain card. Your current score is 20. $~$\\ 
You press $<<$E$>>$ and turn over a gain card. Your current score is 30. $~$\\ 
You press $<<$E$>>$ and turn over a gain card. Your current score is 40. $~$\\ 
You press $<<$C$>>$ and claim your payout. $~$\\ 
Your final score for this round is 40. $~$\\ 
 $~$\\ 
Round 8: $~$\\ 
You will be awarded 10 points for turning over a gain card. $~$\\ 
You will lose 250 points for turning over a loss 

\subsubsection*{Balloon analog risk task}
Data source: \cite{frey2017risk} \\ $~$ \\
Number of experiments: 1 $~$\\ 
Number of participants: 1331 $~$\\ 
Number of choices: 1496974 $~$\\ 
 $~$\\ 
\textbf{Example prompt:}
 $~$\\ 
Throughout the task, you will be presented with balloons, one at a time. $~$\\ 
In each step, you can choose to pump up the balloon by pressing H and you will accumulate 1 point for each pump. $~$\\ 
At any point, you can stop pumping up the balloon by pressing W and you will collect your accumulated points. $~$\\ 
You will repeat this procedure on multiple different balloons. $~$\\ 
It is your choice to determine how much to pump up the balloon, but be aware that at some point the balloon will explode. $~$\\ 
If the balloon explodes before you collect your accumulated points, then you move on to the next balloon and the points are lost. $~$\\ 
 $~$\\ 
Balloon 1: $~$\\ 
You press $<<$H$>>$ $<<$H$>>$ $<<$H$>>$ $<<$H$>>$ $<<$H$>>$ $<<$H$>>$ $<<$H$>>$ $<<$H$>>$ $<<$H$>>$ $<<$H$>>$ $<<$H$>>$ $<<$H$>>$ $<<$H$>>$. The balloon was inflated too much and explodes. $~$\\ 
 $~$\\ 
Balloon 2: $~$\\ 
You press $<<$H$>>$ $<<$H$>>$ $<<$H$>>$ $<<$H$>>$ $<<$H$>>$ $<<$H$>>$ $<<$H$>>$ $<<$H$>>$ $<<$H$>>$ $<<$H$>>$ $<<$H$>>$ $<<$H$>>$ $<<$H$>>$ $<<$H$>>$ $<<$H$>>$ $<<$H$>>$ $<<$H$>>$ $<<$H$>>$ $<<$H$>>$ $<<$H$>>$ $<<$H$>>$ $<<$H$>>$ $<<$H$>>$ $<<$H$>>$ $<<$H$>>$ $<<$H$>>$ $<<$H$>>$ $<<$H$>>$ $<<$H$>>$ $<<$H$>>$ $<<$H$>>$ $<<$H$>>$ $<<$H$>>$ $<<$H$>>$ $<<$H$>>$ $<<$H$>>$ $<<$H$>>$ $<<$H$>>$ $<<$H$>>$ $<<$H$>>$ $<<$H$>>$ $<<$H$>>$ $<<$H$>>$ $<<$H$>>$ $<<$H$>>$ $<<$H$>>$ $<<$H$>>$ $<<$H$>>$ $<<$H$>>$ $<<$H$>>$ $<<$H$>>$ $<<$H$>>$ $<<$H$>>$ $<<$H$>>$ $<<$H$>>$ $<<$H$>>$ $<<$H$>>$ $<<$H$>>$ $<<$H$>>$ $<<$H$>>$ $<<$W$>>$. You stop inflating the balloon and get 60 points. $~$\\ 
 $~$\\ 
Balloon 3: $~$\\ 
You press $<<$H$>>$ $<<$H$>>$ $<<$H$>>$ $<<$H$>>$ $<<$H$>>$ $<<$H$>>$ $<<$H$>>$ $<<$H$>>$ $<<$H$>>$ $<<$H$>>$ $<<$H$>>$ $<<$H$>>$ $<<$H$>>$ $<<$H$>>$ $<<$H$>>$ $<<$H$>>$ $<<$H$>>$ $<<$H$>>$ $<<$H$>>$ $<<$H$>>$ $<<$H$>>$ $<<$H$>>$. The balloon was inflated too much and explodes. $~$\\ 
 $~$\\ 
Balloon 4: $~$\\ 
You press $<<$H$>>$ $<<$H$>>$ $<<$H$>>$ $<<$H$>>$ $<<$H$>>$ $<<$H$>>$ $<<$H$>>$ $<<$H$>>$ $<<$H$>>$ $<<$H$>>$ $<<$H$>>$ $<<$H$>>$ $<<$H$>>$ $<<$H$>>$ $<<$H$>>$ $<<$H$>>$ $<<$H$>>$ $<<$H$>>$ $<<$H$>>$ $<<$H$>>$ $<<$H$>>$ $<<$H$>>$ $<<$H$>>$ $<<$H$>>$ $<<$H$>>$ $<<$H$>>$ $<<$H$>>$ $<<$H$>>$ $<<$H$>>$ $<<$H$>>$. The balloon was inflated too much and explodes. $~$\\ 
 $~$\\ 
Balloon 5: $~$\\ 
You press $<<$H$>>$ $<<$H$>>$ $<<$H$>>$ $<<$H$>>$ $<<$H$>>$ $<<$H$>>$ $<<$H$>>$ $<<$H$>>$ $<<$H$>>$ $<<$H$>>$ $<<$H$>>$ $<<$H$>>$ $<<$H$>>$ $<<$H$>>$ $<<$H$>>$ $<<$H$>>$ $<<$H$>>$ $<<$H$>>$ $<<$H$>>$ $<<$H$>>$ $<<$H$>>$ $<<$H$>>$ $<<$H$>>$ $<<$H$>>$ $<<$H$>>$ $<<$H$>>$ $<<$H$>>$ $<<$H$>>$ $<<$H$>>$ $<<$H$>>$ $<<$H$>>$ $<<$H$>>$ $<<$H$>>$ $<<$H$>>$ $<<$H$>>$ $<<$H$>>$ $<<$H$>>$. The balloon was inflated too much and explodes. $~$\\ 
 $~$\\ 
Balloon 6: $~$\\ 
You press $<<$H$>>$ $<<$H$>>$ $<<$H$>>$ $<<$H$>>$ $<<$H$>>$ $<<$H$>>$ $<<$H$>>$ $<<$H$>>$ $<<$H$>>$ $<<$H$>>$ $<<$H$>>$ $<<$H$>>$ $<<$H$>>$ $<<$H$>>$ $<<$H$>>$ $<<$H$>>$ $<<$H$>>$ $<<$H$>>$ $<<$W$>>$. You stop inflating the balloon and get 18 points. $~$\\ 
 $~$\\ 
Balloon 7: $~$\\ 
You press $<<$H$>>$ $<<$H$>>$ $<<$H$>>$ $<<$H$>>$ $<<$H$>>$ $<<$H$>>$ $<<$H$>>$ $<<$H$>>$ $<<$H$>>$ $<<$H$>>$ $<<$H$>>$ $<<$H$>>$ $<<$H$>>$ $<<$H$>>$ $<<$H$>>$ $<<$H$>>$ $<<$H$>>$ $<<$H$>>$ $<<$H$>>$ $<<$H$>>$ $<<$H$>>$ $<<$H$>>$ $<<$H$>>$ $<<$H$>>$ $<<$H$>>$ $<<$H$>>$ $<<$H$>>$ $<<$H$>>$. The balloon was inflated too much and explodes. $~$\\ 
 $~$\\ 
Balloon 8: $~$\\ 
You press $<<$H$>>$ $<<$H$>>$ $<<$H$>>$ $<<$H$>>$ $<<$H$>>$ $<<$H$>>$ $<<$H$>>$ $<<$H$>>$ $<<$H$>>$ $<<$H$>>$ $<<$H$>>$ $<<$H$>>$ $<<$H$>>$ $<<$H$>>$ $<<$H$>>$ $<<$H$>>$ $<<$H$>>$ $<<$H$>>$ $<<$H$>>$ $<<$H$>>$ $<<$H$>>$ $<<$H$>>$ $<<$H$>>$ $<<$W$>>$. You stop inflating the balloon and get 23 points. $~$\\ 
 $~$\\ 
Balloon 9: $~$\\ 
You press $<<$H$>>$ $<<$H$>>$ $<<$H$>>$ $<<$H$>>$ $<<$H$>>$ $<<$H$>>$ $<<$H$>>$ $<<$H$>>$ $<<$H$>>$ $<<$H$>>$ $<<$H$>>$ $<<$H$>>$ $<<$H$>>$ $<<$H$>>$ $<<$H$>>$ $<<$H$>>$ $<<$H$>>$ $<<$H$>>$ $<<$H$>>$ $<<$H$>>$ $<<$H$>>$ $<<$H$>>$ $<<$H$>>$ $<<$H$>>$ $<<$H$>>$ $<<$H$>>$ $<<$W$>>$. You stop inflating the balloon and get 26 points. $~$\\ 
 $~$\\ 
Balloon 10: $~$\\ 
You press $<<$H$>>$ $<<$H$>>$ $<<$H$>>$ $<<$H$>>$ $<<$H$>>$ $<<$H$>>$ $<<$H$>>$ $<<$H$>>$ $<<$H$>>$ $<<$H$>>$ $<<$H$>>$ $<<$H$>>$ $<<$H$>>$ $<<$H$>>$ $<<$H$>>$ $<<$H$>>$ $<<$H$>>$ $<<$H$>>$ $<<$H$>>$ $<<$H$>>$ $<<$H$>>$ $<<$H$>>$ $<<$H$>>$ $<<$H$>>$ $<<$H$>>$ $<<$H$>>$ $<<$H$>>$ $<<$H$>>$ $<<$H$>>$ $<<$H$>>$ $<<$H$>>$ $<<$H$>>$ $<<$H$>>$ $<<$H$>>$ $<<$H$>>$ $<<$W$>>$. You stop inflating the balloon and get 35 points. $~$\\ 
 $~$\\ 
Balloon 11: $~$\\ 
You press $<<$H$>>$ $<<$H$>>$ $<<$H$>>$ $<<$H$>>$ $<<$H$>>$ $<<$H$>>$ $<<$H$>>$ $<<$H$>>$ $<<$H$>>$ $<<$H$>>$ $<<$H$>>$ $<<$H$>>$ $<<$H$>>$ $<<$H$>>$ $<<$H$>>$ $<<$H$>>$ $<<$H$>>$ $<<$H$>>$ $<<$H$>>$ $<<$H$>>$ $<<$H$>>$ $<<$H$>>$ $<<$H$>>$ $<<$H$>>$ $<<$H$>>$ $<<$H$>>$ $<<$H$>>$ $<<$H$>>$ $<<$H$>>$ $<<$H$>>$ $<<$H$>>$ $<<$H$>>$ $<<$H$>>$ $<<$H$>>$ $<<$H$>>$ $<<$W$>>$. You stop inflating the balloon and get 35 points. $~$\\ 
 $~$\\ 
Balloon 12: $~$\\ 
You press $<<$H$>>$ $<<$H$>>$ $<<$H$>>$ $<<$H$>>$ $<<$H$>>$ $<<$H$>>$ $<<$H$>>$ $<<$H$>>$ $<<$H$>>$ $<<$H$>>$ $<<$H$>>$ $<<$H$>>$ $<<$H$>>$ $<<$H$>>$ $<<$H$>>$ $<<$H$>>$ $<<$H$>>$ $<<$H$>>$ $<<$H$>>$ $<<$H$>>$ $<<$H$>>$ $<<$H$>>$ $<<$H$>>$ $<<$H$>>$ $<<$H$>>$ $<<$H$>>$ $<<$H$>>$ $<<$H$>>$ $<<$H$>>$ $<<$H$>>$ $<<$H$>>$ $<<$H$>>$ $<<$H$>>$ $<<$H$>>$ $<<$H$>>$ $<<$H$>>$ $<<$H$>>$ $<<$H$>>$ $<<$H$>>$ $<<$H$>>$ $<<$H$>>$ $<<$H$>>$ $<<$H$>>$ $<<$H$>>$ $<<$H$>>$ $<<$H$>>$ $<<$H$>>$ $<<$H$>>$ $<<$H$>>$ $<<$H$>>$ $<<$W$>>$. You stop inflating the balloon and get 50 points. $~$\\ 
 $~$\\ 
Balloon 13: $~$\\ 
You press $<<$H$>>$ $<<$H$>>$ $<<$H$>>$ $<<$H$>>$ $<<$H$>>$ $<<$H$>>$ $<<$H$>>$ $<<$H$>>$ $<<$H$>>$ $<<$H$>>$ $<<$H$>>$ $<<$H$>>$ $<<$H$>>$ $<<$H$>>$ $<<$H$>>$ $<<$H$>>$ $<<$H$>>$ $<<$H$>>$ $<<$H$>>$ $<<$H$>>$ $<<$H$>>$ $<<$H$>>$ $<<$H$>>$ $<<$H$>>$ $<<$H$>>$ $<<$H$>>$ $<<$H$>>$ $<<$H$>>$ $<<$H$>>$ $<<$H$>>$ $<<$H$>>$ $<<$H$>>$ $<<$H$>>$ $<<$H$>>$ $<<$H$>>$ $<<$H$>>$ $<<$H$>>$ $<<$H$>>$ $<<$H$>>$ $<<$H$>>$ $<<$H$>>$ $<<$H$>>$ $<<$H$>>$ $<<$H$>>$ $<<$H$>>$ $<<$H$>>$ $<<$H$>>$ $<<$H$>>$ 

\subsubsection*{Experiential-symbolic task}
Data source: \cite{garcia2023experiential} \\ $~$ \\
Number of experiments: 4 $~$\\ 
Number of participants: 346 $~$\\ 
Number of choices: 70608 $~$\\ 
 $~$\\ 
\textbf{Example prompt:}
 $~$\\ 
This experiment is composed of three parts. $~$\\ 
In each round of the first part, you have to choose between one of two options represented by letters. $~$\\ 
In a given pair, one option is, on average, more advantageous compared to the other. $~$\\ 
You can win or lose the following outcomes: 1.0 points and -1.0 points. $~$\\ 
In the second part, there will be two types of options. $~$\\ 
The first type of option is represented by the letters you already encountered during the previous part. $~$\\ 
Note that the options maintain the same odds of winning or losing a point as in the first part. $~$\\ 
The second type of option is represented by an explicit description of the odds of winning or losing a point. $~$\\ 
In each round of the third part, you will be presented with the options you met in the first and the second part. $~$\\ 
This is the occasion to test your knowledge of each options's outcome. $~$\\ 
You will be asked to indicate (in percentages), what are the odds that a given option makes you win a point. $~$\\ 
You can choose an option by pressing its corresponding key. $~$\\ 
Your goal for the first two parts is to maximize the amount of received points. $~$\\ 
Your goal in the third part is to guess as accurately as possible. $~$\\ 
 $~$\\ 
You can choose between option Q and option L. You press $<<$Q$>>$ and get 1.0 points. You would have gotten 1.0 points had you chosen option L instead.  $~$\\ 
You can choose between option Z and option H. You press $<<$Z$>>$ and get 1.0 points. You would have gotten 1.0 points had you chosen option H instead.  $~$\\ 
You can choose between option D and option X. You press $<<$X$>>$ and get 1.0 points. You would have gotten -1.0 points had you chosen option D instead.  $~$\\ 
You can choose between option U and option C. You press $<<$C$>>$ and get 1.0 points. You would have gotten 1.0 points had you chosen option U instead.  $~$\\ 
You can choose between option U and option C. You press $<<$C$>>$ and get -1.0 points. You would have gotten 1.0 points had you chosen option U instead.  $~$\\ 
You can choose between option D and option X. You press $<<$X$>>$ and get 1.0 points. You would have gotten 1.0 points had you chosen option D instead.  $~$\\ 
You can choose between option Z and option H. You press $<<$Z$>>$ and get 1.0 points. You would have gotten -1.0 points had you chosen option H instead.  $~$\\ 
You can choose between option Z and option H. You press $<<$Z$>>$ and get 1.0 points. You would have gotten 1.0 points had you chosen option H instead.  $~$\\ 
You can choose between option D and option X. You press $<<$X$>>$ and get -1.0 points. You would have gotten 1.0 points had you chosen option D instead.  $~$\\ 
You can choose between option D and option X. You press $<<$D$>>$ and get 1.0 points. You would have gotten 1.0 points had you chosen option X instead.  $~$\\ 
You can choose between option Q and option L. You press $<<$Q$>>$ and get 1.0 points. You would have gotten -1.0 points had you chosen option L instead.  $~$\\ 
You can choose between option D and option X. You press $<<$X$>>$ and get -1.0 points. You would have gotten 1.0 points had you chosen option D instead.  $~$\\ 
You can choose between option D and option X. You press $<<$D$>>$ and get 1.0 points. You would have gotten -1.0 points had you chosen option X instead.  $~$\\ 
You can choose between option Q and option L. You press $<<$Q$>>$ and get -1.0 points. You would have gotten -1.0 points had you chosen option L instead.  $~$\\ 
You can choose between option Q and option L. You press $<<$L$>>$ and get -1.0 points. You would have gotten 1.0 points had you chosen option Q instead.  $~$\\ 
You can choose between option Q and option L. You press $<<$Q$>>$ and get 1.0 points. You would have gotten -1.0 points had you chosen option L instead.  $~$\\ 
You can choose between option Q and option L. You press $<<$Q$>>$ and get -1.0 points. You would have gotten 1.0 points had you chosen option L instead.  $~$\\ 
You can choose between option U and option C. You press $<<$C$>>$ and get -1.0 points. You would have gotten 1.0 points had you chosen option U instead.  $~$\\ 
You can choose between option U and option C. You press $<<$C$>>$ and get -1.0 points. You would have gotten -1.0 points had you chosen option U instead.  $~$\\ 
You can choose between option Q and option L. You press $<<$Q$>>$ and get 1.0 poi 

\subsubsection*{Two-armed bandit}
Data source: \cite{gershman2018deconstructing} \\ $~$ \\ 
Number of experiments: 2 $~$\\ 
Number of participants: 80 $~$\\ 
Number of choices: 16000 $~$\\ 
 $~$\\ 
\textbf{Example prompt:}
 $~$\\ 
In this task, you have to repeatedly choose between two slot machines labeled U and P. $~$\\ 
You can choose a slot machine by pressing its corresponding key. $~$\\ 
When you select one of the machines, you will win or lose points. $~$\\ 
Machine U will not always give you the same points when you select it again, but machine P will always give 0 points when you select it. $~$\\ 
Your goal is to choose the slot machines that will give you the most points. $~$\\ 
You will receive feedback about the outcome after making a choice. $~$\\ 
You will play 20 games in total, each with a different pair of slot machines. $~$\\ 
Each game will consist of 10 trials. $~$\\ 
 $~$\\ 
Game 1: $~$\\ 
You press $<<$U$>>$ and get -1 points. $~$\\ 
You press $<<$U$>>$ and get 0 points. $~$\\ 
You press $<<$U$>>$ and get 2 points. $~$\\ 
You press $<<$U$>>$ and get -1 points. $~$\\ 
You press $<<$U$>>$ and get 1 points. $~$\\ 
You press $<<$U$>>$ and get -1 points. $~$\\ 
You press $<<$U$>>$ and get -1 points. $~$\\ 
You press $<<$U$>>$ and get -1 points. $~$\\ 
You press $<<$U$>>$ and get 1 points. $~$\\ 
You press $<<$U$>>$ and get 1 points. $~$\\ 
 $~$\\ 
Game 2: $~$\\ 
You press $<<$U$>>$ and get -1 points. $~$\\ 
You press $<<$U$>>$ and get -1 points. $~$\\ 
You press $<<$U$>>$ and get -2 points. $~$\\ 
You press $<<$U$>>$ and get 0 points. $~$\\ 
You press $<<$U$>>$ and get -1 points. $~$\\ 
You press $<<$P$>>$ and get 0 points. $~$\\ 
You press $<<$P$>>$ and get 0 points. $~$\\ 
You press $<<$P$>>$ and get 0 points. $~$\\ 
You press $<<$P$>>$ and get 0 points. $~$\\ 
You press $<<$P$>>$ and get 0 points. $~$\\ 
 $~$\\ 
Game 3: $~$\\ 
You press $<<$U$>>$ and get -2 points. $~$\\ 
You press $<<$U$>>$ and get -1 points. $~$\\ 
You press $<<$P$>>$ and get 0 points. $~$\\ 
You press $<<$P$>>$ and get 0 points. $~$\\ 
You press $<<$P$>>$ and get 0 points. $~$\\ 
You press $<<$U$>>$ and get -4 points. $~$\\ 
You press $<<$P$>>$ and get 0 points. $~$\\ 
You press $<<$P$>>$ and get 0 points. $~$\\ 
You press $<<$P$>>$ and get 0 points. $~$\\ 
You press $<<$P$>>$ and get 0 points. $~$\\ 
 $~$\\ 
Game 4: $~$\\ 
You press $<<$U$>>$ and get 2 points. $~$\\ 
You press $<<$U$>>$ and get 0 points. $~$\\ 
You press $<<$U$>>$ and get 2 points. $~$\\ 
You press $<<$U$>>$ and get 1 points. $~$\\ 
You press $<<$U$>>$ and get 2 points. $~$\\ 
You press $<<$U$>>$ and get -1 points. $~$\\ 
You press $<<$U$>>$ and get 0 points. $~$\\ 
You press $<<$U$>>$ and get 2 points. $~$\\ 
You press $<<$U$>>$ and get 1 points. $~$\\ 
You press $<<$U$>>$ and get 0 points. $~$\\ 
 $~$\\ 
Game 5: $~$\\ 
You press $<<$U$>>$ and get 0 points. $~$\\ 
You press $<<$U$>>$ and get 1 points. $~$\\ 
You press $<<$U$>>$ and get 1 points. $~$\\ 
You press $<<$U$>>$ and get 0 points. $~$\\ 
You press $<<$U$>>$ and get 2 points. $~$\\ 
You press $<<$U$>>$ and get 1 points. $~$\\ 
You press $<<$U$>>$ and get 2 points. $~$\\ 
You press $<<$U$>>$ and get 1 points. $~$\\ 
You press $<<$U$>>$ and get 2 points. $~$\\ 
You press $<<$U$>>$ and get 1 points. $~$\\ 
 $~$\\ 
Game 6: $~$\\ 
You press $<<$U$>>$ and get 3 points. $~$\\ 
You press $<<$U$>>$ and get 2 points. $~$\\ 
You press $<<$U$>>$ and get 0 points. $~$\\ 
You press $<<$U$>>$ and get 2 points. $~$\\ 
You press $<<$U$>>$ and get 2 points. $~$\\ 
You press $<<$U$>>$ and get 2 points. $~$\\ 
You press $<<$U$>>$ and get 1 points. $~$\\ 
You press $<<$U$>>$ and get 3 points. $~$\\ 
You press $<<$U$>>$ and get 2 points. $~$\\ 
You press $<<$U$>>$ and get 1 points. $~$\\ 
 $~$\\ 
Game 7: $~$\\ 
You press $<<$U$>>$ and get -1 points. $~$\\ 
You press $<<$U$>>$ and get -3 points. $~$\\ 
You press $<<$P$>>$ and get 0 points. $~$\\ 
You press $<<$P$>>$ and get 0 points. $~$\\ 
You press $<<$P$>>$ and get 0 points. $~$\\ 
You press $<<$P$>>$ and get 0 points. $~$\\ 
You press $<<$P$>>$ and get 0 points. $~$\\ 
You press $<<$P$>>$ and get 0 points. $~$\\ 
You press $<<$P$>>$ and get 0 points. $~$\\ 
You press $<<$P$>>$ and get 0 points. $~$\\ 
 $~$\\ 
Game 8: $~$\\ 
You press $<<$U$>>$ and get 0 points. $~$\\ 
You press $<<$U$>>$ and get -1 points. $~$\\ 
You press $<<$P$>>$ and get 0 points. $~$\\ 
You press $<<$P$>>$ and get 0 points. $~$\\ 
You press $<<$P$>>$ and get 0 points. $~$\\ 
You press $<<$P$>>$ and get 0 points. $~$\\ 
You press $<<$P$>>$ and get 0 points. $~$\\ 
You press $<<$P$>>$ and get 0 points. $~$\\ 
You press $<<$P$>>$ and get 0 points. $~$\\ 
You press $<<$P$>>$ and get 0 points. $~$\\ 
 $~$\\ 
Game 9: $~$\\ 
You press $<<$U$>>$ and get 2 points. $~$\\ 
You press $<<$U$>>$ and get 3 points. $~$\\ 
You press $<<$U$>>$ and get 4 points. $~$\\ 
You press $<<$U$>>$ and get 1 points. $~$\\ 
You press $<<$U$>>$ and get 1 points. $~$\\ 
You press $<<$U$>>$ and get 2 points. $~$\\ 
You press $<<$U$>>$ and get 2 points. $~$\\ 
You press $<<$U$>>$ and get 2 points. $~$\\ 
You press $<<$U$>>$ and get -1 points. $~$\\ 
You press $<<$U$>>$ and get 2 points. $~$\\ 
 $~$\\ 
Game 10: $~$\\ 
You press $<<$U$>>$ and get 1 points. $~$\\ 
You press $<<$U$>>$ and get 0 points. $~$\\ 
You press $<<$U$>>$ and get 0 points. $~$\\ 
You press $<<$U$>>$ and get 1 points. $~$\\ 
You press $<<$U$>>$ and get -1 points. $~$\\ 
You press $<<$U$>>$ and get -2 points. $~$\\ 
You press $<<$U$>>$ and get 1 points. $~$\\ 
You press $<<$U$>>$ and get 2 points. $~$\\ 
You press $<<$U$>>$ and get 1 points. $~$\\ 
You pr 

\subsubsection*{Conditional associative learning}
Data source: \cite{collins2014working} \\ $~$ \\
Number of experiments: 1 $~$\\ 
Number of participants: 74 $~$\\ 
Number of choices: 40539 $~$\\ 
 $~$\\ 
\textbf{Example prompt:}
 $~$\\ 
You are presented with a series of stimuli, each associated with one of three possible responses. $~$\\ 
Your goal is to learn which response is the correct one for each stimulus. $~$\\ 
When a stimulus is presented, you can press one of three keys to respond. $~$\\ 
The three responses available are S, F, and A. $~$\\ 
After your response, you will receive feedback: 1 point for a correct response, or 0 points for an incorrect response. $~$\\ 
The correct response for one stimulus does not inform you about the correct response for another stimulus. $~$\\ 
You will play 13 games, each with a different mapping from stimuli to responses. $~$\\ 
 $~$\\ 
Game 1: $~$\\ 
There are 6 different stimuli. $~$\\ 
You see stimulus 1. You press $<<$S$>>$ and get 0 points. $~$\\ 
You see stimulus 0. You press $<<$F$>>$ and get 0 points. $~$\\ 
You see stimulus 4. You press $<<$A$>>$ and get 1 points. $~$\\ 
You see stimulus 5. You press $<<$S$>>$ and get 0 points. $~$\\ 
You see stimulus 3. You press $<<$F$>>$ and get 1 points. $~$\\ 
You see stimulus 3. You press $<<$F$>>$ and get 1 points. $~$\\ 
You see stimulus 4. You press $<<$A$>>$ and get 1 points. $~$\\ 
You see stimulus 2. You press $<<$S$>>$ and get 0 points. $~$\\ 
You see stimulus 1. You press $<<$S$>>$ and get 0 points. $~$\\ 
You see stimulus 5. You press $<<$S$>>$ and get 0 points. $~$\\ 
You see stimulus 2. You press $<<$S$>>$ and get 0 points. $~$\\ 
You see stimulus 0. You press $<<$A$>>$ and get 0 points. $~$\\ 
You see stimulus 3. You press $<<$F$>>$ and get 1 points. $~$\\ 
You see stimulus 3. You press $<<$F$>>$ and get 1 points. $~$\\ 
You see stimulus 5. You press $<<$A$>>$ and get 1 points. $~$\\ 
You see stimulus 1. You press $<<$A$>>$ and get 0 points. $~$\\ 
You see stimulus 4. You press $<<$A$>>$ and get 1 points. $~$\\ 
You see stimulus 1. You press $<<$S$>>$ and get 0 points. $~$\\ 
You see stimulus 0. You press $<<$S$>>$ and get 1 points. $~$\\ 
You see stimulus 5. You press $<<$S$>>$ and get 0 points. $~$\\ 
You see stimulus 4. You press $<<$A$>>$ and get 1 points. $~$\\ 
You see stimulus 2. You press $<<$S$>>$ and get 0 points. $~$\\ 
You see stimulus 0. You press $<<$S$>>$ and get 1 points. $~$\\ 
You see stimulus 2. You press $<<$S$>>$ and get 0 points. $~$\\ 
You see stimulus 3. You press $<<$F$>>$ and get 1 points. $~$\\ 
You see stimulus 1. You press $<<$A$>>$ and get 0 points. $~$\\ 
You see stimulus 5. You press $<<$A$>>$ and get 1 points. $~$\\ 
You see stimulus 4. You press $<<$A$>>$ and get 1 points. $~$\\ 
You see stimulus 0. You press $<<$S$>>$ and get 1 points. $~$\\ 
You see stimulus 1. You press $<<$S$>>$ and get 0 points. $~$\\ 
You see stimulus 3. You press $<<$F$>>$ and get 1 points. $~$\\ 
You see stimulus 4. You press $<<$A$>>$ and get 1 points. $~$\\ 
You see stimulus 0. You press $<<$S$>>$ and get 1 points. $~$\\ 
You see stimulus 2. You press $<<$S$>>$ and get 0 points. $~$\\ 
You see stimulus 2. You press $<<$S$>>$ and get 0 points. $~$\\ 
You see stimulus 3. You press $<<$F$>>$ and get 1 points. $~$\\ 
You see stimulus 5. You press $<<$S$>>$ and get 0 points. $~$\\ 
You see stimulus 1. You press $<<$A$>>$ and get 0 points. $~$\\ 
You see stimulus 0. You press $<<$F$>>$ and get 0 points. $~$\\ 
You see stimulus 5. You press $<<$F$>>$ and get 0 points. $~$\\ 
You see stimulus 4. You press $<<$A$>>$ and get 1 points. $~$\\ 
You see stimulus 4. You press $<<$A$>>$ and get 1 points. $~$\\ 
You see stimulus 1. You press $<<$A$>>$ and get 0 points. $~$\\ 
You see stimulus 3. You press $<<$F$>>$ and get 1 points. $~$\\ 
You see stimulus 2. You press $<<$F$>>$ and get 1 points. $~$\\ 
You see stimulus 2. You press $<<$F$>>$ and get 1 points. $~$\\ 
You see stimulus 3. You press $<<$F$>>$ and get 1 points. $~$\\ 
You see stimulus 0. You press $<<$F$>>$ and get 0 points. $~$\\ 
You see stimulus 1. You press $<<$F$>>$ and get 1 points. $~$\\ 
You see stimulus 5. You press $<<$F$>>$ and get 0 points. $~$\\ 
You see stimulus 0. You press $<<$F$>>$ and get 0 points. $~$\\ 
You see stimulus 5. You press $<<$F$>>$ and get 0 points. $~$\\ 
You see stimulus 4. You press $<<$A$>>$ and get 1 points. $~$\\ 
You see stimulus 2. You press $<<$A$>>$ and get 0 points. $~$\\ 
You see stimulus 0. You press $<<$F$>>$ and get 0 points. $~$\\ 
You see stimulus 0. You press $<<$F$>>$ and get 0 points. $~$\\ 
You see stimulus 1. You press $<<$A$>>$ and get 0 points. $~$\\ 
You see stimulus 2. You press $<<$A$>>$ and get 0 points. $~$\\ 
You see stimulus 1. You press $<<$A$>>$ and get 0 points. $~$\\ 
You see stimulus 4. You press $<<$A$>>$ and get 1 points. $~$\\ 
You see stimulus 4. You press $<<$A$>>$ and get 1 points. $~$\\ 
You see stimulus 2. You press $<<$S$>>$ and get 0 points. $~$\\ 
You see stimulus 5. You press $<<$S$>>$ and get 0 points. $~$\\ 
You see stimulus 5. You press $<<$F$>>$ and get 0 points. 

\subsubsection*{THINGS odd-one-out}
Data source: \cite{hebart2023things} \\ $~$ \\
Number of experiments: 1 $~$\\ 
Number of participants: 11122 $~$\\ 
Number of choices: 2611240 $~$\\ 
 $~$\\ 
\textbf{Example prompt:}
 $~$\\ 
You will be presented with triplets of objects, which will be assigned to the keys B, J, and K. $~$\\ 
In each trial, please indicate which object you think is the odd one out by pressing the corresponding key. $~$\\ 
In other words, please choose the object that is the least similar to the other two. $~$\\ 
 $~$\\ 
B: prune, J: nail polish, and K: diskette. You press $<<$K$>>$. $~$\\ 
B: ladle, J: water bottle, and K: pug. You press $<<$K$>>$. $~$\\ 
B: punch, J: hair, and K: lollipop. You press $<<$J$>>$. $~$\\ 
B: oar, J: mug, and K: macaroni. You press $<<$B$>>$. $~$\\ 
B: towel, J: hot tub, and K: mallet. You press $<<$K$>>$. $~$\\ 
B: train set, J: hot-water bottle, and K: treasure. You press $<<$J$>>$. $~$\\ 
B: shutter, J: cleaver, and K: toe. You press $<<$K$>>$. $~$\\ 
B: straw, J: sweeper, and K: baton. You press $<<$J$>>$. $~$\\ 
B: footprint, J: beehive, and K: skunk. You press $<<$B$>>$. $~$\\ 
B: cash machine, J: thermostat, and K: mandolin. You press $<<$K$>>$. $~$\\ 
B: orange, J: throne, and K: stir fry. You press $<<$J$>>$. $~$\\ 
B: boy, J: burrito, and K: microscope. You press $<<$J$>>$. $~$\\ 
B: pheasant, J: sponge, and K: orchid. You press $<<$J$>>$. $~$\\ 
B: forklift, J: clipper, and K: hip. You press $<<$J$>>$. $~$\\ 
B: candelabra, J: beard, and K: glue. You press $<<$J$>>$. $~$\\ 
B: raccoon, J: hammer, and K: roulette wheel. You press $<<$B$>>$. $~$\\ 
B: wing, J: beanie, and K: girl. You press $<<$B$>>$. $~$\\ 
B: piggy bank, J: footrest, and K: sandal. You press $<<$B$>>$. $~$\\ 
B: knee, J: cornhusk, and K: tuning fork. You press $<<$B$>>$. $~$\\ 
B: anklet, J: bedpost, and K: ice cube. You press $<<$B$>>$. $~$\\ 
B: mannequin, J: stove, and K: coin. You press $<<$K$>>$. $~$\\ 
B: tortellini, J: cantaloupe, and K: sequin. You press $<<$K$>>$. $~$\\ 
B: coffee filter, J: fingerprint, and K: rose. You press $<<$B$>>$. $~$\\ 
B: porcupine, J: christmas tree, and K: corkscrew. You press $<<$B$>>$. $~$\\ 
B: freezer, J: coat rack, and K: puffin. You press $<<$K$>>$. $~$\\ 
B: maggot, J: mouth, and K: stockings. You press $<<$B$>>$. $~$\\ 
B: soap, J: hot-water bottle, and K: knitting needle. You press $<<$K$>>$. $~$\\ 
B: mosquito net, J: baklava, and K: beanbag. You press $<<$J$>>$. $~$\\ 
B: skewer, J: baklava, and K: propeller. You press $<<$K$>>$. $~$\\ 
B: nail polish, J: goose, and K: pizza. You press $<<$J$>>$. $~$\\ 
B: face mask, J: cinnamon, and K: toilet paper. You press $<<$B$>>$. $~$\\ 
B: bag, J: eel, and K: trampoline. You press $<<$J$>>$. $~$\\ 
B: lightbulb, J: moose, and K: curling iron. You press $<<$J$>>$. $~$\\ 
B: dumbwaiter, J: jigsaw puzzle, and K: lamb. You press $<<$K$>>$. $~$\\ 
B: eyeliner, J: shopping basket, and K: flipper. You press $<<$K$>>$. $~$\\ 
B: bowtie, J: wooden leg, and K: kangaroo. You press $<<$K$>>$. $~$\\ 
B: puffin, J: wire cutters, and K: battery. You press $<<$B$>>$. $~$\\ 
B: toilet, J: rack, and K: star fruit. You press $<<$K$>>$. $~$\\ 
B: stingray, J: cork, and K: fire pit. You press $<<$B$>>$. $~$\\ 
B: bun, J: snow, and K: tinsel. You press $<<$J$>>$. $~$\\ 
B: almond, J: trailer, and K: paper. You press $<<$B$>>$. $~$\\ 
B: anteater, J: chalice, and K: wedge. You press $<<$B$>>$. $~$\\ 
B: scone, J: pie, and K: ant. You press $<<$K$>>$. $~$\\ 
B: heater, J: aircraft carrier, and K: joystick. You press $<<$B$>>$. $~$\\ 
B: comic book, J: playing card, and K: organ. You press $<<$K$>>$. $~$\\ 
B: flag, J: measuring cup, and K: strawberry. You press $<<$B$>>$. $~$\\ 
B: bunkbed, J: tractor, and K: windshield. You press $<<$B$>>$. $~$\\ 
B: aircraft carrier, J: prism, and K: turban. You press $<<$K$>>$. $~$\\ 
B: egg, J: scrambled egg, and K: doorknocker. You press $<<$K$>>$. $~$\\ 
B: microscope, J: stained glass, and K: strainer. You press $<<$J$>>$. $~$\\ 
B: polygraph, J: hairdryer, and K: harness. You press $<<$J$>>$. $~$\\ 
B: chip, J: iguana, and K: hedge. You press $<<$B$>>$. $~$\\ 
B: tick, J: binder, and K: shoelace. You press $<<$B$>>$. $~$\\ 
B: nut, J: yogurt, and K: jug. You press $<<$K$>>$. $~$\\ 
B: hairpin, J: giraffe, and K: fur coat. You press $<<$B$>>$. $~$\\ 
B: lasagna, J: statue, and K: bookshelf. You press $<<$B$>>$. $~$\\ 
B: grill, J: catapult, and K: sonogram. You press $<<$K$>>$. $~$\\ 
B: fondue, J: pill, and K: firewood. You press $<<$K$>>$. $~$\\ 
B: sunroof, J: onion, and K: flan. You press $<<$B$>>$. $~$\\ 
B: wreck, J: bungee, and K: cockroach. You press $<<$K$>>$. 

\subsubsection*{Multi-attribute decision-making}
Data source: \cite{hilbig2014generalized} \\ $~$ \\
Number of experiments: 1 $~$\\ 
Number of participants: 73 $~$\\ 
Number of choices: 7008 $~$\\ 
 $~$\\ 
\textbf{Example prompt:}
 $~$\\ 
You are repeatedly presented with two options, labeled A and R. $~$\\ 
Each option represents a fictitious product and you have to infer which product is superior in terms of quality. $~$\\ 
You select a product by pressing the corresponding key. $~$\\ 
For each decision, you are provided with four expert ratings (with 1 representing a positive and 0 representing a negative rating). $~$\\ 
The four experts differ in their validity. $~$\\ 
The ratings of experts are given in descending order of their validity (having validities of 90\%, 80\%, 70\%, and 60\%). $~$\\ 
 $~$\\ 
Product A ratings: [0 1 1 1]. Product R ratings: [1 0 0 1]. You press $<<$A$>>$. $~$\\ 
Product A ratings: [1 1 1 1]. Product R ratings: [0 0 1 1]. You press $<<$A$>>$. $~$\\ 
Product A ratings: [1 0 0 0]. Product R ratings: [0 0 0 1]. You press $<<$A$>>$. $~$\\ 
Product A ratings: [1 1 1 0]. Product R ratings: [0 0 1 0]. You press $<<$A$>>$. $~$\\ 
Product A ratings: [0 1 1 1]. Product R ratings: [1 1 1 0]. You press $<<$R$>>$. $~$\\ 
Product A ratings: [0 1 0 1]. Product R ratings: [1 1 0 0]. You press $<<$R$>>$. $~$\\ 
Product A ratings: [0 0 1 1]. Product R ratings: [1 0 1 0]. You press $<<$R$>>$. $~$\\ 
Product A ratings: [1 0 0 1]. Product R ratings: [0 1 1 1]. You press $<<$R$>>$. $~$\\ 
Product A ratings: [0 1 1 1]. Product R ratings: [1 0 0 1]. You press $<<$A$>>$. $~$\\ 
Product A ratings: [0 1 1 1]. Product R ratings: [1 1 1 0]. You press $<<$R$>>$. $~$\\ 
Product A ratings: [1 0 1 0]. Product R ratings: [0 0 1 1]. You press $<<$A$>>$. $~$\\ 
Product A ratings: [1 1 0 0]. Product R ratings: [0 1 0 1]. You press $<<$A$>>$. $~$\\ 
Product A ratings: [1 0 0 0]. Product R ratings: [0 1 1 0]. You press $<<$R$>>$. $~$\\ 
Product A ratings: [1 0 0 1]. Product R ratings: [0 1 1 1]. You press $<<$R$>>$. $~$\\ 
Product A ratings: [1 1 1 1]. Product R ratings: [0 0 1 1]. You press $<<$A$>>$. $~$\\ 
Product A ratings: [1 0 0 0]. Product R ratings: [0 0 0 1]. You press $<<$A$>>$. $~$\\ 
Product A ratings: [1 1 0 1]. Product R ratings: [0 0 0 1]. You press $<<$A$>>$. $~$\\ 
Product A ratings: [1 0 0 1]. Product R ratings: [0 1 1 1]. You press $<<$R$>>$. $~$\\ 
Product A ratings: [0 1 0 1]. Product R ratings: [1 1 0 0]. You press $<<$R$>>$. $~$\\ 
Product A ratings: [1 0 1 1]. Product R ratings: [0 0 0 0]. You press $<<$A$>>$. $~$\\ 
Product A ratings: [1 0 0 0]. Product R ratings: [0 0 0 1]. You press $<<$A$>>$. $~$\\ 
Product A ratings: [0 1 1 1]. Product R ratings: [1 1 1 0]. You press $<<$R$>>$. $~$\\ 
Product A ratings: [0 1 1 0]. Product R ratings: [1 0 0 0]. You press $<<$A$>>$. $~$\\ 
Product A ratings: [1 1 1 0]. Product R ratings: [0 1 1 1]. You press $<<$A$>>$. $~$\\ 
Product A ratings: [0 0 0 0]. Product R ratings: [1 1 0 0]. You press $<<$R$>>$. $~$\\ 
Product A ratings: [0 1 1 1]. Product R ratings: [1 0 0 1]. You press $<<$A$>>$. $~$\\ 
Product A ratings: [0 0 0 0]. Product R ratings: [1 1 0 0]. You press $<<$R$>>$. $~$\\ 
Product A ratings: [1 0 0 0]. Product R ratings: [0 1 1 0]. You press $<<$R$>>$. $~$\\ 
Product A ratings: [0 0 0 0]. Product R ratings: [1 1 0 0]. You press $<<$R$>>$. $~$\\ 
Product A ratings: [0 1 1 1]. Product R ratings: [1 0 0 1]. You press $<<$A$>>$. $~$\\ 
Product A ratings: [0 0 0 1]. Product R ratings: [1 0 0 0]. You press $<<$R$>>$. $~$\\ 
Product A ratings: [1 0 0 0]. Product R ratings: [0 1 1 0]. You press $<<$R$>>$. $~$\\ 
Product A ratings: [1 1 0 0]. Product R ratings: [0 0 0 0]. You press $<<$A$>>$. $~$\\ 
Product A ratings: [1 1 0 0]. Product R ratings: [0 1 0 1]. You press $<<$A$>>$. $~$\\ 
Product A ratings: [1 0 0 0]. Product R ratings: [0 0 0 1]. You press $<<$A$>>$. $~$\\ 
Product A ratings: [0 0 0 1]. Product R ratings: [1 0 0 0]. You press $<<$R$>>$. $~$\\ 
Product A ratings: [1 0 1 0]. Product R ratings: [0 0 1 1]. You press $<<$A$>>$. $~$\\ 
Product A ratings: [1 1 0 0]. Product R ratings: [0 0 0 0]. You press $<<$A$>>$. $~$\\ 
Product A ratings: [0 0 1 1]. Product R ratings: [1 0 1 0]. You press $<<$R$>>$. $~$\\ 
Product A ratings: [1 0 0 0]. Product R ratings: [0 1 1 0]. You press $<<$R$>>$. $~$\\ 
Product A ratings: [1 1 1 0]. Product R ratings: [0 0 1 0]. You press $<<$A$>>$. $~$\\ 
Product A ratings: [1 0 0 1]. Product R ratings: [0 1 1 1]. You press $<<$R$>>$. $~$\\ 
Product A ratings: [1 0 1 0]. Product R ratings: [0 0 1 1]. You press $<<$A$>>$. $~$\\ 
Product A ratings: [0 1 1 0]. Product R ratings: [1 0 0 0]. You press $<<$A$>>$. $~$\\ 
Product A ratings: [1 1 0 0]. Product R ratings: [0 1 0 1]. You press $<<$A$>>$. $~$\\ 
Product A ratings: [0 1 1 0]. Product R ratings: [1 0 0 0]. You press $<<$A$>>$. $~$\\ 
Product A ratings: [1 0 0 0 

\subsubsection*{Grammar judgement}
Data source: \cite{jansen2021rational} \\ $~$ \\
Number of experiments: 1 $~$\\ 
Number of participants: 3192 $~$\\ 
Number of choices: 89376 $~$\\ 
 $~$\\ 
\textbf{Example prompt:}
 $~$\\ 
You're about to answer a set of 20 questions about grammar. How many of the 20 questions do you think you will answer correctly? $~$\\ 
You say $<<$4$>>$. $~$\\ 
Compared to other participants in this study, how well do you think you will do? Marking 90\% means you will do better than 90\% of participants, marking 10\% means you will do better than only 10\%, and marking 50\% means that you will perform better than half of the participants. $~$\\ 
You say $<<$50$>>$\%. $~$\\ 
On a scale of 0 to 10, how difficult is recognizing correct grammar for the average participant? $~$\\ 
You say $<<$6$>>$. $~$\\ 
On a scale of 0 to 10, how difficult is recognizing correct grammar for you? $~$\\ 
You say $<<$6$>>$. $~$\\ 
 $~$\\ 
You will now see twenty questions. $~$\\ 
In each question, some part of each sentence is in square brackets. $~$\\ 
Five choices for rephrasing that part follow each sentence; one choice repeats the original, and the other four are different. $~$\\ 
Your task is to use the buttons J, E, V, H, and G to select the grammatically correct choice. $~$\\ 
 $~$\\ 
Q1. The school-age child faces a formidable task when during the first few years of classroom experiences [he or she is expected to master the printed form of language.] $~$\\ 
The choices are: $~$\\ 
V: he or she expects to master the printed form of language. $~$\\ 
E: he or she is expected to master the printed form of language. $~$\\ 
H: he or she faces expectations of mastering the printed form of language. $~$\\ 
J: mastery of the printed form of language is expected of him or her. $~$\\ 
G: mastery of print is expected by his or her teacher. $~$\\ 
You press $<<$E$>>$. $~$\\ 
 $~$\\ 
Q2. He came to the United States as a young [man, he found] a job as a coal miner. $~$\\ 
The choices are: $~$\\ 
H: man, he found $~$\\ 
G: man and found $~$\\ 
E: man and there he was able to find $~$\\ 
V: man and then finding $~$\\ 
J: man and had found $~$\\ 
You press $<<$H$>>$. $~$\\ 
 $~$\\ 
Q3. To a large degree, [poetry, along with all the other arts, is] a form of imitation. $~$\\ 
The choices are: $~$\\ 
E: poetry, along with all the other arts, is $~$\\ 
V: poetry along with all the other arts is $~$\\ 
J: poetry, along with all the other arts, are $~$\\ 
G: poetry, and other arts, is $~$\\ 
H: poetry and art are $~$\\ 
You press $<<$V$>>$. $~$\\ 
 $~$\\ 
Q4. Delegates to the political convention found [difficulty to choose] a candidate from among the few nominated. $~$\\ 
The choices are: $~$\\ 
E: difficulty to choose $~$\\ 
G: it difficult in making the choice of $~$\\ 
H: it difficult to choose $~$\\ 
V: choosing difficult when selecting $~$\\ 
J: making a choice difficult in selecting $~$\\ 
You press $<<$E$>>$. $~$\\ 
 $~$\\ 
Q5. Reading in any language can be viewed as a developmental task much the same as learning to walk, to cross the street independently, to care for one's possessions, or [accepting responsibility for one's own decisions.] $~$\\ 
The choices are: $~$\\ 
J: accepting responsibility for one's own decisions. $~$\\ 
V: accepting one's own decisions responsibly. $~$\\ 
H: to accept responsibility for one's own decisions. $~$\\ 
E: accepting responsibility and making one's own decisions. $~$\\ 
G: to make one's own decisions. $~$\\ 
You press $<<$G$>>$. $~$\\ 
 $~$\\ 
Q6. Sea forests of giant kelp, which fringe only one coastline in the Northern Hemisphere, [is native to shores] throughout the Southern Hemisphere. $~$\\ 
The choices are: $~$\\ 
E: is native to shores $~$\\ 
V: is native to most shores $~$\\ 
G: are native only in shores $~$\\ 
J: are native $~$\\ 
H: are native to shores $~$\\ 
You press $<<$J$>>$. $~$\\ 
 $~$\\ 
Q7. Taking an occasional respite between chapters or assignments is more desirable [than a long, continuous period of study. $~$\\ 
The choices are: $~$\\ 
E: than a long, continuous period of study. $~$\\ 
G: than a period of long, continuous study. $~$\\ 
V: than a long period of continuous study. $~$\\ 
J: than studying for a long, continuous period. $~$\\ 
H: than a study period long and continuous. $~$\\ 
You press $<<$V$>>$. $~$\\ 
 $~$\\ 
Q8. Like so many characters in Russian fiction, [Crime and Punishment exhibits] a behavior so foreign to the American temperament that many readers find the story rather incredible. $~$\\ 
The choices are: $~$\\ 
J: Crime and Punishment exhibits $~$\\ 
H: those in Crime and Punishment exhibit $~$\\ 
G: those in Crime and Punishment exhibits $~$\\ 
V: often exhibiting $~$\\ 
E: characterized by $~$\\ 
You press $<<$G$>>$. $~$\\ 
 $~$\\ 
Q9. Don Quixote provides a cross section of Spanish life, thought, and [portrays the feelings of many Spaniards] at the end of the chivalric age. $~$\\ 
The choices are: $~$\\ 
H: portrays th

\subsubsection*{Two-step task}
Data source: \cite{kool2016does} \\ $~$ \\
Number of experiments: 2 $~$\\ 
Number of participants: 181 $~$\\ 
Number of choices: 52861 $~$\\ 
 $~$\\ 
\textbf{Example prompt:}
 $~$\\ 
Each day you will either be presented with spaceships P and F or with spaceships Z and J. $~$\\ 
These spaceships will take you to two different planets L and Q. $~$\\ 
You can take a spaceship by pressing the corresponding key. $~$\\ 
Each planet has one alien on it and each alien has its own space treasure mine. $~$\\ 
When you arrive at a planet, you will ask the alien for space treasure from its mine. $~$\\ 
When you ask the alien, you will find out whether you got space treasure. $~$\\ 
However, sometimes the alien will not bring up any treasure. $~$\\ 
The quality of each alien's mine will change during the game. $~$\\ 
Before you choose a spaceship, you will be told whether there is a treasure multiplier. $~$\\ 
If there is a treasure multiplier, you will receive 5 times the amount of treasure you will find. $~$\\ 
Your goal is to get as much treasure as possible over the next 200 days. $~$\\ 
 $~$\\ 
There is no treasure multiplier. You are presented with spaceships F and P. You press $<<$P$>>$. You end up on planet L. You find 7 pieces of space treasure. You receive 7 pieces of space treasure. $~$\\ 
There is no treasure multiplier. You are presented with spaceships F and P. You press $<<$F$>>$. You end up on planet Q. You find 0 pieces of space treasure. You receive 0 pieces of space treasure. $~$\\ 
There is no treasure multiplier. You are presented with spaceships J and Z. You press $<<$Z$>>$. You end up on planet L. You find 7 pieces of space treasure. You receive 7 pieces of space treasure. $~$\\ 
There is no treasure multiplier. You are presented with spaceships Z and J. You press $<<$Z$>>$. You end up on planet L. You find 7 pieces of space treasure. You receive 7 pieces of space treasure. $~$\\ 
There is no treasure multiplier. You are presented with spaceships Z and J. You press $<<$Z$>>$. You end up on planet L. You find 8 pieces of space treasure. You receive 8 pieces of space treasure. $~$\\ 
There is a treasure multiplier. You are presented with spaceships F and P. You press $<<$P$>>$. You end up on planet L. You find 6 pieces of space treasure. You receive 30 pieces of space treasure. $~$\\ 
There is a treasure multiplier. You are presented with spaceships F and P. You press $<<$P$>>$. You end up on planet L. You find 8 pieces of space treasure. You receive 40 pieces of space treasure. $~$\\ 
There is no treasure multiplier. You are presented with spaceships J and Z. You press $<<$Z$>>$. You end up on planet L. You find 9 pieces of space treasure. You receive 9 pieces of space treasure. $~$\\ 
There is no treasure multiplier. You are presented with spaceships P and F. You press $<<$P$>>$. You end up on planet L. You find 8 pieces of space treasure. You receive 8 pieces of space treasure. $~$\\ 
There is a treasure multiplier. You are presented with spaceships J and Z. You press $<<$Z$>>$. You end up on planet L. You find 9 pieces of space treasure. You receive 45 pieces of space treasure. $~$\\ 
There is a treasure multiplier. You are presented with spaceships P and F. You press $<<$P$>>$. You end up on planet L. You find 9 pieces of space treasure. You receive 45 pieces of space treasure. $~$\\ 
There is no treasure multiplier. You are presented with spaceships J and Z. You press $<<$Z$>>$. You end up on planet L. You find 8 pieces of space treasure. You receive 8 pieces of space treasure. $~$\\ 
There is a treasure multiplier. You are presented with spaceships J and Z. You press $<<$Z$>>$. You end up on planet L. You find 9 pieces of space treasure. You receive 45 pieces of space treasure. $~$\\ 
There is no treasure multiplier. You are presented with spaceships Z and J. You press $<<$Z$>>$. You end up on planet L. You find 5 pieces of space treasure. You receive 5 pieces of space treasure. $~$\\ 
There is no treasure multiplier. You are presented with spaceships Z and J. You press $<<$Z$>>$. You end up on planet L. You find 6 pieces of space treasure. You receive 6 pieces of space treasure. $~$\\ 
There is a treasure multiplier. You are presented with spaceships P and F. You press $<<$P$>>$. You end up on planet L. You find 8 pieces of space treasure. You receive 40 pieces of space treasure. $~$\\ 
There is no treasure multiplier. You are presented with spaceships J and Z. You press $<<$Z$>>$. You end up on planet L. You find 7 pieces of space treasure.  

\subsubsection*{Two-step task}
Data source:  \cite{kool2017cost} \\ $~$ \\
Number of experiments: 2 $~$\\ 
Number of participants: 367 $~$\\ 
Number of choices: 67041 $~$\\ 
 $~$\\ 
\textbf{Example prompt:}
 $~$\\ 
Each day you will either be presented with spaceships G and S or with spaceships T and N. $~$\\ 
These spaceships will take you to two different planets R and Z. $~$\\ 
You can take a spaceship by pressing the corresponding key. $~$\\ 
Each planet has one alien on it and each alien has its own space treasure mine. $~$\\ 
When you arrive at a planet, you will ask the alien for space treasure from its mine. $~$\\ 
When you ask the alien, you will find out whether you got space treasure. $~$\\ 
However, sometimes the alien's mine will dig up antimatter. $~$\\ 
Antimatter is bad because each piece will destroy a piece of space treasure, reducing the total amount of treasure that you have. $~$\\ 
The quality of each alien's mine will change during the game. $~$\\ 
Your goal is to get as much treasure and as little antimatter as possible over the next 125 days. $~$\\ 
 $~$\\ 
You are presented with spaceships N and T. You press $<<$T$>>$. You end up on planet R. You find 1 pieces of antimatter. $~$\\ 
You are presented with spaceships T and N. You do not respond in time on this day. You do not go to any planet. You find nothing. $~$\\ 
You are presented with spaceships N and T. You press $<<$N$>>$. You end up on planet Z. You find 4 pieces of space treasure. $~$\\ 
You are presented with spaceships T and N. You press $<<$N$>>$. You end up on planet Z. You find 4 pieces of space treasure. $~$\\ 
You are presented with spaceships S and G. You press $<<$S$>>$. You end up on planet Z. You find 5 pieces of space treasure. $~$\\ 
You are presented with spaceships T and N. You press $<<$N$>>$. You end up on planet Z. You find 3 pieces of space treasure. $~$\\ 
You are presented with spaceships N and T. You press $<<$N$>>$. You end up on planet Z. You find 4 pieces of space treasure. $~$\\ 
You are presented with spaceships T and N. You press $<<$N$>>$. You end up on planet Z. You find 2 pieces of space treasure. $~$\\ 
You are presented with spaceships S and G. You press $<<$S$>>$. You end up on planet Z. You find 1 pieces of space treasure. $~$\\ 
You are presented with spaceships T and N. You press $<<$N$>>$. You end up on planet Z. You find 3 pieces of antimatter. $~$\\ 
You are presented with spaceships T and N. You press $<<$T$>>$. You end up on planet R. You find 2 pieces of space treasure. $~$\\ 
You are presented with spaceships G and S. You press $<<$G$>>$. You end up on planet R. You find 4 pieces of space treasure. $~$\\ 
You are presented with spaceships S and G. You press $<<$G$>>$. You end up on planet R. You find 2 pieces of space treasure. $~$\\ 
You are presented with spaceships T and N. You press $<<$T$>>$. You end up on planet R. You find 3 pieces of space treasure. $~$\\ 
You are presented with spaceships T and N. You press $<<$T$>>$. You end up on planet R. You find 5 pieces of space treasure. $~$\\ 
You are presented with spaceships N and T. You press $<<$T$>>$. You end up on planet R. You find 4 pieces of space treasure. $~$\\ 
You are presented with spaceships S and G. You press $<<$G$>>$. You end up on planet R. You find 4 pieces of space treasure. $~$\\ 
You are presented with spaceships S and G. You press $<<$G$>>$. You end up on planet R. You find nothing. $~$\\ 
You are presented with spaceships T and N. You press $<<$T$>>$. You end up on planet R. You find 3 pieces of antimatter. $~$\\ 
You are presented with spaceships T and N. You press $<<$N$>>$. You end up on planet Z. You find 2 pieces of space treasure. $~$\\ 
You are presented with spaceships S and G. You press $<<$S$>>$. You end up on planet Z. You find 2 pieces of space treasure. $~$\\ 
You are presented with spaceships S and G. You press $<<$S$>>$. You end up on planet Z. You find 4 pieces of space treasure. $~$\\ 
You are presented with spaceships S and G. You press $<<$S$>>$. You end up on planet Z. You find 4 pieces of space treasure. $~$\\ 
You are presented with spaceships N and T. You press $<<$N$>>$. You end up on planet Z. You find 4 pieces of space treasure. $~$\\ 
You are presented with spaceships N and T. You press $<<$N$>>$. You end up on planet Z. You find 5 pieces of space treasure. $~$\\ 
You are presented with spaceships S and G. You press $<<$S$>>$. You end up on planet Z. You find 3 pieces of space treasure. $~$\\ 
You are presented with spaceships G and S. You press $<<$S$>>$. You end up on planet Z. You find 4 pieces of space treasure. $~$\\ 
You are presented with spaceships S and G. Yo 

\subsubsection*{Risky choice}
Data source: \cite{krueger2024identifying} \\ $~$ \\
Number of experiments: 1 $~$\\ 
Number of participants: 1755 $~$\\ 
Number of choices: 499728 $~$\\ 
 $~$\\ 
\textbf{Example prompt:}
 $~$\\ 
You will play multiple rounds of a gambling game. $~$\\ 
In each round, you will be presented with 6 different gambles labeled: Q, N, E, S, H, and K. $~$\\ 
You will have to choose one of the gambles and receive a payoff for doing so. $~$\\ 
The payoff you receive depends on both the gamble you choose and also the color of a ball we pull out of a jar with 100 colored balls. $~$\\ 
There are a different number of balls of each color on every round. $~$\\ 
The colors with more balls are more likely to be chosen. $~$\\ 
Before making your choice, you may check how much different gambles are worth for different ball colors. $~$\\ 
Each time you check a gamble will cost you 4 points. $~$\\ 
To choose or check a gamble, first press the corresponding key, followed by typing "stop" (for choosing) or the ball color you would like to check. $~$\\ 
 $~$\\ 
A new round begins. $~$\\ 
There are 50 pink balls, 19 red balls, 15 black balls, and 16 maroon balls. $~$\\ 
You press $<<$Q$>>$ and then type $<<$pink$>>$. The payoff for this combination would be -105 points. $~$\\ 
You press $<<$E$>>$ and then type $<<$pink$>>$. The payoff for this combination would be 47 points. $~$\\ 
You press $<<$H$>>$ and then type $<<$pink$>>$. The payoff for this combination would be 91 points. $~$\\ 
You press $<<$H$>>$ and then type $<<$stop$>>$. A maroon ball is chosen, and you earn -16 points. $~$\\ 
 $~$\\ 
A new round begins. $~$\\ 
There are 2 pink balls, 47 red balls, 14 black balls, and 37 maroon balls. $~$\\ 
You press $<<$Q$>>$ and then type $<<$red$>>$. The payoff for this combination would be -168 points. $~$\\ 
You press $<<$E$>>$ and then type $<<$red$>>$. The payoff for this combination would be 209 points. $~$\\ 
You press $<<$H$>>$ and then type $<<$red$>>$. The payoff for this combination would be 22 points. $~$\\ 
You press $<<$E$>>$ and then type $<<$stop$>>$. A maroon ball is chosen, and you earn 25 points. $~$\\ 
 $~$\\ 
A new round begins. $~$\\ 
There are 13 pink balls, 1 red balls, 47 black balls, and 39 maroon balls. $~$\\ 
You press $<<$Q$>>$ and then type $<<$black$>>$. The payoff for this combination would be -101 points. $~$\\ 
You press $<<$E$>>$ and then type $<<$black$>>$. The payoff for this combination would be -98 points. $~$\\ 
You press $<<$H$>>$ and then type $<<$black$>>$. The payoff for this combination would be 32 points. $~$\\ 
You press $<<$H$>>$ and then type $<<$stop$>>$. A maroon ball is chosen, and you earn -85 points. $~$\\ 
 $~$\\ 
A new round begins. $~$\\ 
There are 7 pink balls, 15 red balls, 76 black balls, and 2 maroon balls. $~$\\ 
You press $<<$N$>>$ and then type $<<$black$>>$. The payoff for this combination would be -108 points. $~$\\ 
You press $<<$Q$>>$ and then type $<<$black$>>$. The payoff for this combination would be -191 points. $~$\\ 
You press $<<$S$>>$ and then type $<<$black$>>$. The payoff for this combination would be -222 points. $~$\\ 
You press $<<$K$>>$ and then type $<<$stop$>>$. A black ball is chosen, and you earn -90 points. $~$\\ 
 $~$\\ 
A new round begins. $~$\\ 
There are 17 pink balls, 54 red balls, 16 black balls, and 13 maroon balls. $~$\\ 
You press $<<$Q$>>$ and then type $<<$red$>>$. The payoff for this combination would be 138 points. $~$\\ 
You press $<<$N$>>$ and then type $<<$red$>>$. The payoff for this combination would be 171 points. $~$\\ 
You press $<<$E$>>$ and then type $<<$red$>>$. The payoff for this combination would be -197 points. $~$\\ 
You press $<<$N$>>$ and then type $<<$stop$>>$. A red ball is chosen, and you earn 171 points. $~$\\ 
 $~$\\ 
A new round begins. $~$\\ 
There are 1 pink balls, 38 red balls, 25 black balls, and 36 maroon balls. $~$\\ 
You press $<<$N$>>$ and then type $<<$red$>>$. The payoff for this combination would be -124 points. $~$\\ 
You press $<<$E$>>$ and then type $<<$red$>>$. The payoff for this combination would be -158 points. $~$\\ 
You press $<<$S$>>$ and then type $<<$red$>>$. The payoff for this combination would be -114 points. $~$\\ 
You press $<<$H$>>$ and then type $<<$stop$>>$. A black ball is chosen, and you earn 338 points. $~$\\ 
 $~$\\ 
A new round begins. $~$\\ 
There are 37 pink balls, 44 red balls, 17 black balls, and 2 maroon balls. $~$\\ 
You press $<<$Q$>>$ and then type $<<$red$>>$. The payoff for this combination would be -198 points. $~$\\ 
You press $<<$S$>>$ and then type $<<$red$>>$. The payoff for this combination would be 39 points. $~$\\ 
You press $<<$H$>>$ and then type $<<$red$>>$. The payoff for this combination would be -63 points. $~$\\ 
You press $<<$E$>>$ and then type $<<$stop$>>$. A red ball is chosen, and you earn 51 points. $~$\\ 
 $~$\\ 
A new round begins. $~$\\ 
There are 12 pink balls, 32 red balls, 45  

\subsubsection*{Tile-revealing task}
Data source: \cite{kumar2023disentangling} \\ $~$ \\
Number of experiments: 1 $~$\\ 
Number of participants: 707 $~$\\ 
Number of choices: 321738 $~$\\ 
 $~$\\ 
\textbf{Example prompt:}
 $~$\\ 
You are playing a game where you are revealing patterns on a binary grid. $~$\\ 
Each turn you will be presented with a 7x7 grid of tiles. $~$\\ 
If you give the location on the grid with an "X" on it, then it will either turn into a "1" or a "0". $~$\\ 
Please give the location as a [row, column] pair where both row and column take values between 0 and 6. $~$\\ 
Do not give the location on the grid where there is already a "1" or a "0". $~$\\ 
Your job is to reveal as many 1s as possible while revealing as little 0s as possible. $~$\\ 
The game ends when you have revealed all 1s on the grid. $~$\\ 
You will have 25 boards to complete, each with a different underlying pattern. $~$\\ 
 $~$\\ 
You are currently solving board 1. $~$\\ 
The board state is now: $~$\\ 
XXXXXXX $~$\\ 
XXXXXXX $~$\\ 
XXXXXXX $~$\\ 
XXX1XXX $~$\\ 
XXXXXXX $~$\\ 
XXXXXXX $~$\\ 
XXXXXXX $~$\\ 
You choose $<<$[2, 2]$>>$ and get a miss! $~$\\ 
 $~$\\ 
The board state is now: $~$\\ 
XXXXXXX $~$\\ 
XXXXXXX $~$\\ 
XX0XXXX $~$\\ 
XXX1XXX $~$\\ 
XXXXXXX $~$\\ 
XXXXXXX $~$\\ 
XXXXXXX $~$\\ 
You choose $<<$[2, 4]$>>$ and get a miss! $~$\\ 
 $~$\\ 
The board state is now: $~$\\ 
XXXXXXX $~$\\ 
XXXXXXX $~$\\ 
XX0X0XX $~$\\ 
XXX1XXX $~$\\ 
XXXXXXX $~$\\ 
XXXXXXX $~$\\ 
XXXXXXX $~$\\ 
You choose $<<$[2, 3]$>>$ and get a hit! $~$\\ 
 $~$\\ 
The board state is now: $~$\\ 
XXXXXXX $~$\\ 
XXXXXXX $~$\\ 
XX010XX $~$\\ 
XXX1XXX $~$\\ 
XXXXXXX $~$\\ 
XXXXXXX $~$\\ 
XXXXXXX $~$\\ 
You choose $<<$[1, 3]$>>$ and get a miss! $~$\\ 
 $~$\\ 
The board state is now: $~$\\ 
XXXXXXX $~$\\ 
XXX0XXX $~$\\ 
XX010XX $~$\\ 
XXX1XXX $~$\\ 
XXXXXXX $~$\\ 
XXXXXXX $~$\\ 
XXXXXXX $~$\\ 
You choose $<<$[3, 4]$>>$ and get a hit! $~$\\ 
 $~$\\ 
The board state is now: $~$\\ 
XXXXXXX $~$\\ 
XXX0XXX $~$\\ 
XX010XX $~$\\ 
XXX11XX $~$\\ 
XXXXXXX $~$\\ 
XXXXXXX $~$\\ 
XXXXXXX $~$\\ 
You choose $<<$[3, 2]$>>$ and get a hit! $~$\\ 
 $~$\\ 
The board state is now: $~$\\ 
XXXXXXX $~$\\ 
XXX0XXX $~$\\ 
XX010XX $~$\\ 
XX111XX $~$\\ 
XXXXXXX $~$\\ 
XXXXXXX $~$\\ 
XXXXXXX $~$\\ 
You choose $<<$[3, 5]$>>$ and get a hit! $~$\\ 
 $~$\\ 
The board state is now: $~$\\ 
XXXXXXX $~$\\ 
XXX0XXX $~$\\ 
XX010XX $~$\\ 
XX1111X $~$\\ 
XXXXXXX $~$\\ 
XXXXXXX $~$\\ 
XXXXXXX $~$\\ 
You choose $<<$[3, 6]$>>$ and get a hit! $~$\\ 
 $~$\\ 
The board state is now: $~$\\ 
XXXXXXX $~$\\ 
XXX0XXX $~$\\ 
XX010XX $~$\\ 
XX11111 $~$\\ 
XXXXXXX $~$\\ 
XXXXXXX $~$\\ 
XXXXXXX $~$\\ 
You choose $<<$[3, 1]$>>$ and get a hit! $~$\\ 
 $~$\\ 
You are currently solving board 2. $~$\\ 
The board state is now: $~$\\ 
XXXXXXX $~$\\ 
X1XXXXX $~$\\ 
XXXXXXX $~$\\ 
XXXXXXX $~$\\ 
XXXXXXX $~$\\ 
XXXXXXX $~$\\ 
XXXXXXX $~$\\ 
You choose $<<$[2, 1]$>>$ and get a miss! $~$\\ 
 $~$\\ 
The board state is now: $~$\\ 
XXXXXXX $~$\\ 
X1XXXXX $~$\\ 
X0XXXXX $~$\\ 
XXXXXXX $~$\\ 
XXXXXXX $~$\\ 
XXXXXXX $~$\\ 
XXXXXXX $~$\\ 
You choose $<<$[0, 1]$>>$ and get a miss! $~$\\ 
 $~$\\ 
The board state is now: $~$\\ 
X0XXXXX $~$\\ 
X1XXXXX $~$\\ 
X0XXXXX $~$\\ 
XXXXXXX $~$\\ 
XXXXXXX $~$\\ 
XXXXXXX $~$\\ 
XXXXXXX $~$\\ 
You choose $<<$[1, 2]$>>$ and get a hit! $~$\\ 
 $~$\\ 
The board state is now: $~$\\ 
X0XXXXX $~$\\ 
X11XXXX $~$\\ 
X0XXXXX $~$\\ 
XXXXXXX $~$\\ 
XXXXXXX $~$\\ 
XXXXXXX $~$\\ 
XXXXXXX $~$\\ 
You choose $<<$[1, 0]$>>$ and get a hit! $~$\\ 
 $~$\\ 
The board state is now: $~$\\ 
X0XXXXX $~$\\ 
111XXXX $~$\\ 
X0XXXXX $~$\\ 
XXXXXXX $~$\\ 
XXXXXXX $~$\\ 
XXXXXXX $~$\\ 
XXXXXXX $~$\\ 
You choose $<<$[1, 3]$>>$ and get a hit! $~$\\ 
 $~$\\ 
The board state is now: $~$\\ 
X0XXXXX $~$\\ 
1111XXX $~$\\ 
X0XXXXX $~$\\ 
XXXXXXX $~$\\ 
XXXXXXX $~$\\ 
XXXXXXX $~$\\ 
XXXXXXX $~$\\ 
You choose $<<$[1, 4]$>>$ and get a hit! $~$\\ 
 $~$\\ 
The board state is now: $~$\\ 
X0XXXXX $~$\\ 
11111XX $~$\\ 
X0XXXXX $~$\\ 
XXXXXXX $~$\\ 
XXXXXXX $~$\\ 
XXXXXXX $~$\\ 
XXXXXXX $~$\\ 
You choose $<<$[1, 5]$>>$ and get a miss! $~$\\ 
 $~$\\ 
The board state is now: $~$\\ 
X0XXXXX $~$\\ 
111110X $~$\\ 
X0XXXXX $~$\\ 
XXXXXXX $~$\\ 
XXXXXXX $~$\\ 
XXXXXXX $~$\\ 
XXXXXXX $~$\\ 
You choose $<<$[0, 0]$>>$ and get a miss! $~$\\ 
 $~$\\ 
The board state is now: $~$\\ 
00XXXXX $~$\\ 
111110X $~$\\ 
X0XXXXX $~$\\ 
XXXXXXX $~$\\ 
XXXXXXX $~$\\ 
XXXXXXX $~$\\ 
XXXXXXX $~$\\ 
You choose $<<$[2, 2]$>>$ and get a miss! $~$\\ 
 $~$\\ 
The board state is now: $~$\\ 
00XXXXX $~$\\ 
111110X $~$\\ 
X00XXXX $~$\\ 
XXXXXXX $~$\\ 
XXXXXXX $~$\\ 
XXXXXXX $~$\\ 
XXXXXXX $~$\\ 
You choose $<<$[2, 3]$>>$ and get a hit! $~$\\ 
 $~$\\ 
The board state is now: $~$\\ 
00XXXXX $~$\\ 
111110X $~$\\ 
X001XXX $~$\\ 
XXXXXXX $~$\\ 
XXXXXXX $~$\\ 
XXXXXXX $~$\\ 
XXXXXXX $~$\\ 
You choose $<<$[3, 3]$>>$ and get a hit! $~$\\ 
 $~$\\ 
The board state is now: $~$\\ 
00XXXXX $~$\\ 
111110X $~$\\ 
X001XXX $~$\\ 
XXX1XXX $~$\\ 
XXXXXXX $~$\\ 
XXXXXXX $~$\\ 
XXXXXXX $~$\\ 
You choose $<<$[0, 3]$>>$ and get a hit! $~$\\ 
 $~$\\ 
The board state is now: $~$\\ 
00X1XXX $~$\\ 
111110X $~$\\ 
X001XXX $~$\\ 
XXX1XXX $~$\\ 
XXXXXXX $~$\\ 
XXXXXXX $~$\\ 
XXXXXXX $~$\\ 
You choose $<<$[4, 3]$>>$ and get a hit! $~$\\ 
 $~$\\ 
You are currently solving board 3. $~$\\ 
The board state is now: $~$\\ 
XXXXXXX $~$\\ 
XXXXXXX $~$\\ 
XXXXXXX $~$\\ 
XXXXXXX $~$\\ 
XXXXXXX $~$\\ 
XXX1XXX $~$\\ 
XXXXXXX $~$\\ 
You choose $<<$[4, 2]$>>$ and get a hit! $~$\\ 
 $~$\\ 
The board state is now: $~$\\ 
XXXXXXX $~$\\ 
XXXXXXX $~$\\ 
XXXXXXX $~$\\ 
XXXXXXX $~$\\ 
XX1XXXX $~$\\ 
XXX1XXX $~$\\ 
XXXXXXX $~$\\ 
You choose $<<$[3, 1]$>>$ and get a hit! $~$\\ 
 $~$\\ 
The board state is now: $~$\\ 
XXXXXXX $~$\\ 
XXXXXXX $~$\\ 
XXXXXXX $~$\\ 
X1XXXXX $~$\\ 
XX1XXXX $~$\\ 
XXX1XXX $~$\\ 
XXXXXXX $~$\\ 
You choose $<<$[2, 0]$>>$ and get a hit! $~$\\ 
 $~$\\ 
The board state is now: $~$\\ 
XXXXXXX $~$\\ 
XXXXXXX $~$\\ 
1XXXXXX $~$\\ 
X1XXXXX $~$\\ 
XX1XXXX $~$\\ 
XXX1XXX $~$\\ 
XXXXXXX $~$\\ 
You choose $<<$[1, 1]$>>$ and get a miss! $~$\\ 
 $~$\\ 
The board state is now: $~$\\ 
XXXXXXX $~$\\ 
X0XXXXX $~$\\ 
1XXXXXX $~$\\ 
X1XXXXX $~$\\ 
XX1XXXX $~$\\ 
XXX1XXX $~$\\ 
XXXXXXX $~$\\ 
You choose $<<$[2, 2]$>>$ and get a miss! $~$\\ 
 $~$\\ 
The board state is now: $~$\\ 
XXXXXXX $~$\\ 
X0XXXXX $~$\\ 
1X0XXXX $~$\\ 
X1XXXXX $~$\\ 
XX1XXXX $~$\\ 
XXX1XXX $~$\\ 
XXXXXXX $~$\\ 
You choose $<<$[3, 3]$>>$ and get a hit! $~$\\ 
 $~$\\ 
The board state is now: $~$\\ 
XXXXXXX $~$\\ 
X0XXXXX 

\subsubsection*{Probabilistic instrumental learning}
Data source:  \cite{lefebvre2017behavioural} \\ $~$ \\
Number of experiments: 2 $~$\\ 
Number of participants: 77 $~$\\ 
Number of choices: 7392 $~$\\ 
 $~$\\ 
\textbf{Example prompt:}
 $~$\\ 
You are going to visit four different casinos (named 1, 2, 3, and 4) 24 times each. $~$\\ 
Each casino owns two slot machines that return either 0 or 0.5 points stochastically with different probabilities. $~$\\ 
You can play one of the machines in order to win points by pressing the corresponding key. $~$\\ 
Your goal is to maximize the sum of received points within all visits. $~$\\ 
 $~$\\ 
You go to casino 3. You can choose between machines B and C. You press $<<$B$>>$ and receive 0.0 points. $~$\\ 
You go to casino 4. You can choose between machines P and T. You press $<<$T$>>$ and receive 0.5 points. $~$\\ 
You go to casino 4. You can choose between machines P and T. You press $<<$T$>>$ and receive 0.5 points. $~$\\ 
You go to casino 3. You can choose between machines B and C. You press $<<$B$>>$ and receive 0.0 points. $~$\\ 
You go to casino 1. You can choose between machines F and I. You press $<<$I$>>$ and receive 0.0 points. $~$\\ 
You go to casino 1. You can choose between machines F and I. You press $<<$F$>>$ and receive 0.0 points. $~$\\ 
You go to casino 2. You can choose between machines L and J. You press $<<$L$>>$ and receive 0.5 points. $~$\\ 
You go to casino 3. You can choose between machines B and C. You press $<<$B$>>$ and receive 0.0 points. $~$\\ 
You go to casino 4. You can choose between machines P and T. You press $<<$T$>>$ and receive 0.0 points. $~$\\ 
You go to casino 3. You can choose between machines B and C. You press $<<$C$>>$ and receive 0.0 points. $~$\\ 
You go to casino 4. You can choose between machines P and T. You press $<<$P$>>$ and receive 0.5 points. $~$\\ 
You go to casino 1. You can choose between machines F and I. You press $<<$F$>>$ and receive 0.5 points. $~$\\ 
You go to casino 2. You can choose between machines L and J. You press $<<$L$>>$ and receive 0.5 points. $~$\\ 
You go to casino 2. You can choose between machines L and J. You press $<<$L$>>$ and receive 0.5 points. $~$\\ 
You go to casino 1. You can choose between machines F and I. You press $<<$F$>>$ and receive 0.0 points. $~$\\ 
You go to casino 2. You can choose between machines L and J. You press $<<$J$>>$ and receive 0.5 points. $~$\\ 
You go to casino 1. You can choose between machines F and I. You press $<<$I$>>$ and receive 0.0 points. $~$\\ 
You go to casino 1. You can choose between machines F and I. You press $<<$F$>>$ and receive 0.0 points. $~$\\ 
You go to casino 2. You can choose between machines L and J. You press $<<$J$>>$ and receive 0.0 points. $~$\\ 
You go to casino 2. You can choose between machines L and J. You press $<<$L$>>$ and receive 0.5 points. $~$\\ 
You go to casino 3. You can choose between machines B and C. You press $<<$B$>>$ and receive 0.5 points. $~$\\ 
You go to casino 3. You can choose between machines B and C. You press $<<$B$>>$ and receive 0.0 points. $~$\\ 
You go to casino 2. You can choose between machines L and J. You press $<<$L$>>$ and receive 0.5 points. $~$\\ 
You go to casino 3. You can choose between machines B and C. You press $<<$C$>>$ and receive 0.5 points. $~$\\ 
You go to casino 2. You can choose between machines L and J. You press $<<$L$>>$ and receive 0.0 points. $~$\\ 
You go to casino 1. You can choose between machines F and I. You press $<<$I$>>$ and receive 0.0 points. $~$\\ 
You go to casino 4. You can choose between machines P and T. You press $<<$P$>>$ and receive 0.0 points. $~$\\ 
You go to casino 4. You can choose between machines P and T. You press $<<$T$>>$ and receive 0.5 points. $~$\\ 
You go to casino 4. You can choose between machines P and T. You press $<<$T$>>$ and receive 0.5 points. $~$\\ 
You go to casino 1. You can choose between machines F and I. You press $<<$I$>>$ and receive 0.5 points. $~$\\ 
You go to casino 4. You can choose between machines P and T. You press $<<$P$>>$ and receive 0.5 points. $~$\\ 
You go to casino 3. You can choose between machines B and C. You press $<<$C$>>$ and receive 0.5 points. $~$\\ 
You go to casino 2. You can choose between machines L and J. You press $<<$L$>>$ and receive 0.5 points. $~$\\ 
You go to casino 3. You can choose between machines B and C. You press $<<$C$>>$ and receive 0.5 points. $~$\\ 
You go to casino 4. You can choose between machines P and T. You press $<<$P$>>$ and receive 0.5 points. $~$\\ 
You go to casino 3. You can choose between machines B and C. You press $<<$C$>>$ and receive 0.5 points. $~$\\ 
You go to casino 4. You can choose between machines P and T. You press $<<$P$>>$ and receive 0.5 point 

\subsubsection*{Medin categorization}
Data source: \cite{levering2020revisiting} \\ $~$ \\
Number of experiments: 2 $~$\\ 
Number of participants: 228 $~$\\ 
Number of choices: 37848 $~$\\ 
 $~$\\ 
\textbf{Example prompt:}
 $~$\\ 
You will observe a series of objects, one at a time. $~$\\ 
The objects differ along three binary dimensions: shape (square vs. triangle), size (1.50 inch vs. 0.75 inch), and shading (black vs. white). $~$\\ 
Each dimension is indicated by the three digits, for example, '121' means a square, 0.75 inch, black object. $~$\\ 
Based on some combination of the three dimensions, each object belongs to one of two categories, W or N. $~$\\ 
You have to assign each object to one of the two categories by pressing the corresponding key. $~$\\ 
If your choice is correct, you get a point, otherwise you lose a point. $~$\\ 
Your goal is to get as many points as possible. $~$\\ 
At some point, you begin a 'test block' in which you will see eight objects. $~$\\ 
Here, you have to assign each object to one of the two categories as before. $~$\\ 
Furthermore, you have to rate how typical the object is for the category you chose, on a scale from 1 to 9. $~$\\ 
1 means 'not at all typical', and 9 means 'most typical'. $~$\\ 
 $~$\\ 
You see the image 112, press $<<$W$>>$ and get 1 points. $~$\\ 
You see the image 121, press $<<$N$>>$ and get 0 points. $~$\\ 
You see the image 212, press $<<$W$>>$ and get 0 points. $~$\\ 
You see the image 211, press $<<$N$>>$ and get 0 points. $~$\\ 
You see the image 221, press $<<$W$>>$ and get 0 points. $~$\\ 
You see the image 122, press $<<$W$>>$ and get 0 points. $~$\\ 
You see the image 122, press $<<$N$>>$ and get 1 points. $~$\\ 
You see the image 221, press $<<$W$>>$ and get 0 points. $~$\\ 
You see the image 121, press $<<$N$>>$ and get 0 points. $~$\\ 
You see the image 212, press $<<$N$>>$ and get 1 points. $~$\\ 
You see the image 211, press $<<$N$>>$ and get 0 points. $~$\\ 
You see the image 112, press $<<$W$>>$ and get 1 points. $~$\\ 
You see the image 112, press $<<$W$>>$ and get 1 points. $~$\\ 
You see the image 212, press $<<$N$>>$ and get 1 points. $~$\\ 
You see the image 221, press $<<$W$>>$ and get 0 points. $~$\\ 
You see the image 211, press $<<$N$>>$ and get 0 points. $~$\\ 
You see the image 121, press $<<$N$>>$ and get 0 points. $~$\\ 
You see the image 122, press $<<$W$>>$ and get 0 points. $~$\\ 
You see the image 212, press $<<$N$>>$ and get 1 points. $~$\\ 
You see the image 221, press $<<$W$>>$ and get 0 points. $~$\\ 
You see the image 211, press $<<$N$>>$ and get 0 points. $~$\\ 
You see the image 121, press $<<$N$>>$ and get 0 points. $~$\\ 
You see the image 112, press $<<$W$>>$ and get 1 points. $~$\\ 
You see the image 122, press $<<$W$>>$ and get 0 points. $~$\\ 
You see the image 122, press $<<$N$>>$ and get 1 points. $~$\\ 
You see the image 112, press $<<$W$>>$ and get 1 points. $~$\\ 
You see the image 121, press $<<$N$>>$ and get 0 points. $~$\\ 
You see the image 221, press $<<$N$>>$ and get 1 points. $~$\\ 
You see the image 212, press $<<$N$>>$ and get 1 points. $~$\\ 
You see the image 211, press $<<$N$>>$ and get 0 points. $~$\\ 
You see the image 112, press $<<$W$>>$ and get 1 points. $~$\\ 
You see the image 212, press $<<$N$>>$ and get 1 points. $~$\\ 
You see the image 122, press $<<$W$>>$ and get 0 points. $~$\\ 
You see the image 211, press $<<$N$>>$ and get 0 points. $~$\\ 
You see the image 221, press $<<$W$>>$ and get 0 points. $~$\\ 
You see the image 121, press $<<$N$>>$ and get 0 points. $~$\\ 
You see the image 122, press $<<$N$>>$ and get 1 points. $~$\\ 
You see the image 212, press $<<$W$>>$ and get 0 points. $~$\\ 
You see the image 211, press $<<$N$>>$ and get 0 points. $~$\\ 
You see the image 221, press $<<$N$>>$ and get 1 points. $~$\\ 
You see the image 112, press $<<$W$>>$ and get 1 points. $~$\\ 
You see the image 121, press $<<$W$>>$ and get 1 points. $~$\\ 
You see the image 121, press $<<$W$>>$ and get 1 points. $~$\\ 
You see the image 211, press $<<$N$>>$ and get 0 points. $~$\\ 
You see the image 221, press $<<$N$>>$ and get 1 points. $~$\\ 
You see the image 212, press $<<$N$>>$ and get 1 points. $~$\\ 
You see the image 122, press $<<$W$>>$ and get 0 points. $~$\\ 
You see the image 112, press $<<$W$>>$ and get 1 points. $~$\\ 
You see the image 212, press $<<$N$>>$ and get 1 points. $~$\\ 
You see the image 112, press $<<$W$>>$ and get 1 points. $~$\\ 
You see the image 121, press $<<$N$>>$ and get 0 points. $~$\\ 
You see the image 122, press $<<$W$>>$ and get 0 points. $~$\\ 
You see the image 221, press $<<$N$>>$ and get 1 points. $~$\\ 
You see the image 211, press $<<$N$>>$ and get 0 points. $~$\\ 
You see the image 112, press $<<$W$>>$ and get 1 points. $~$\\ 
You see the image 122, press $<<$W$>>$ and get 0 points. $~$\\ 
You see the image 121, press $<<$N$>>$ and get 0 points. $~$\\ 
You see the image 221, press $<<$N$>>$ and get 1 points. $~$\\ 
You see the image 212, press $<<$W$>>$ and get 0 points. $~$\\ 
You see the image 211, p 

\subsubsection*{Zoopermarket}
Data source: \cite{Ludwig2023} \\ $~$ \\
Number of experiments: 3 $~$\\ 
Number of participants: 96 $~$\\ 
Number of choices: 34442 $~$\\ 
 $~$\\ 
\textbf{Example prompt:}
 $~$\\ 
You will have to repeatedly feed animals with fruits. $~$\\ 
Each fruit contains two vitamins. $~$\\ 
Every animal has a different preference for the vitamins. $~$\\ 
The vitamin contents and the preferences are both given as vectors with two entries. $~$\\ 
Your points are calculated as the dot product of the vitamin content with the preference of the current animal. $~$\\ 
For example, let us assume that you have to feed the elephant who has a preference [-1  1]. $~$\\ 
Then, if you feed the elephant a fruit with vitamin content [-1  1], this would yield 2 points. $~$\\ 
If you feed it a fruit with vitamins [1 0], this would yield -1 points. $~$\\ 
You have to buy the fruits in a market, in which you can go left or right for two steps. $~$\\ 
You can press I to go left, and V to go right. $~$\\ 
Per round, you always collect two fruits. $~$\\ 
There are eight animals in total and you have to feed one, two, or three of them in each block. $~$\\ 
In each block, there are twelve trials with different animals in random order. $~$\\ 
After these twelve trials, there are three more in which you have to feed new animals. $~$\\ 
The fruits in the market are rearranged after each block, meaning that you have to relearn the positions. $~$\\ 
Your goal is to maximize the points obtained. $~$\\ 
 $~$\\ 
A new block starts. The locations of the fruits in the market got scrambled. $~$\\ 
You have to feed the crocodile. It has the preference [1 1]. $~$\\ 
You press $<<$V$>>$ and find the apple which has the vitamins [-1 -1]. You get -2 points. $~$\\ 
You press $<<$V$>>$ and find the orange which has the vitamins [0 1]. You get 1 points. $~$\\ 
You have to feed the crocodile. It has the preference [1 1]. $~$\\ 
You press $<<$V$>>$ and find the apple which has the vitamins [-1 -1]. You get -2 points. $~$\\ 
You press $<<$I$>>$ and find the blueberry which has the vitamins [-1  1]. You get 0 points. $~$\\ 
You have to feed the crocodile. It has the preference [1 1]. $~$\\ 
You press $<<$I$>>$ and find the strawberry which has the vitamins [1 1]. You get 2 points. $~$\\ 
You press $<<$V$>>$ and find the grapes which has the vitamins [ 1 -1]. You get 0 points. $~$\\ 
You have to feed the crocodile. It has the preference [1 1]. $~$\\ 
You press $<<$I$>>$ and find the strawberry which has the vitamins [1 1]. You get 2 points. $~$\\ 
You press $<<$V$>>$ and find the grapes which has the vitamins [ 1 -1]. You get 0 points. $~$\\ 
You have to feed the kangaroo. It has the preference [-1  0]. $~$\\ 
You press $<<$V$>>$ and find the apple which has the vitamins [-1 -1]. You get 1 points. $~$\\ 
You press $<<$I$>>$ and find the blueberry which has the vitamins [-1  1]. You get 1 points. $~$\\ 
You have to feed the kangaroo. It has the preference [-1  0]. $~$\\ 
You press $<<$V$>>$ and find the apple which has the vitamins [-1 -1]. You get 1 points. $~$\\ 
You press $<<$I$>>$ and find the blueberry which has the vitamins [-1  1]. You get 1 points. $~$\\ 
You have to feed the crocodile. It has the preference [1 1]. $~$\\ 
You press $<<$I$>>$ and find the strawberry which has the vitamins [1 1]. You get 2 points. $~$\\ 
You press $<<$V$>>$ and find the grapes which has the vitamins [ 1 -1]. You get 0 points. $~$\\ 
You have to feed the kangaroo. It has the preference [-1  0]. $~$\\ 
You press $<<$V$>>$ and find the apple which has the vitamins [-1 -1]. You get 1 points. $~$\\ 
You press $<<$I$>>$ and find the blueberry which has the vitamins [-1  1]. You get 1 points. $~$\\ 
You have to feed the kangaroo. It has the preference [-1  0]. $~$\\ 
You press $<<$V$>>$ and find the apple which has the vitamins [-1 -1]. You get 1 points. $~$\\ 
You press $<<$I$>>$ and find the blueberry which has the vitamins [-1  1]. You get 1 points. $~$\\ 
You have to feed the kangaroo. It has the preference [-1  0]. $~$\\ 
You press $<<$V$>>$ and find the apple which has the vitamins [-1 -1]. You get 1 points. $~$\\ 
You press $<<$I$>>$ and find the blueberry which has the vitamins [-1  1]. You get 1 points. $~$\\ 
You have to feed the kangaroo. It has the preference [-1  0]. $~$\\ 
You press $<<$V$>>$ and find the apple which has the vitamins [-1 -1]. You get 1 points. $~$\\ 
You press $<<$I$>>$ and find the blueberry which has the vitamins [-1  1]. You get 1 points. $~$\\ 
You have to feed the crocodile. It has the preference [1 1]. $~$\\ 
You press $<<$V$>>$ and find the apple which has the vitamins [-1 -1]. You get -2 points. $~$\\ 
You press $<<$V$>>$ and find the orange which has the vitamins [0 1]. You get 1 points. $~$\\ 
You 

\subsubsection*{choices13k}
Data source: \cite{peterson2021using} \\ $~$ \\
Number of experiments: 1 $~$\\ 
Number of participants: 13735 $~$\\ 
Number of choices: 1097375 $~$\\ 
 $~$\\ 
\textbf{Example prompt:}
 $~$\\ 
You will encounter a series of gambling problems where you have to select between two options. $~$\\ 
You can select an option by pressing the corresponding key. $~$\\ 
For some problems, you are told the points you received and missed out on after each selection, while for others this information is suppressed. $~$\\ 
In cases where the probabilities are unknown, they sum up to one and remain constant within a problem. $~$\\ 
 $~$\\ 
Option L delivers 10.0 points with 80.0\% chance, or -25.0 points with 20.0\% chance. $~$\\ 
Option B delivers 0.0 points with 20.0\% chance, or 5.0 points with 80.0\% chance. $~$\\ 
You press $<<$B$>>$. You receive 5.0 points by selecting this option. You would have received 10.0 points had you chosen the other option. $~$\\ 
You press $<<$B$>>$. You receive 5.0 points by selecting this option. You would have received -25.0 points had you chosen the other option. $~$\\ 
You press $<<$B$>>$. You receive 5.0 points by selecting this option. You would have received 10.0 points had you chosen the other option. $~$\\ 
You press $<<$B$>>$. You receive 5.0 points by selecting this option. You would have received -25.0 points had you chosen the other option. $~$\\ 
You press $<<$B$>>$. You receive 0.0 points by selecting this option. You would have received -25.0 points had you chosen the other option. $~$\\ 
 $~$\\ 
Option L delivers either 30.0 points with 100.0\% chance, or 30.0 points with 0.0\% chance. $~$\\ 
Option B delivers either 0.0 points with unknown chance, or 42.0 points with unknown chance. $~$\\ 
You press $<<$L$>>$. You receive 30.0 points by selecting this option. You would have received 42.0 points had you chosen the other option. $~$\\ 
You press $<<$L$>>$. You receive 30.0 points by selecting this option. You would have received 0.0 points had you chosen the other option. $~$\\ 
You press $<<$L$>>$. You receive 30.0 points by selecting this option. You would have received 42.0 points had you chosen the other option. $~$\\ 
You press $<<$L$>>$. You receive 30.0 points by selecting this option. You would have received 42.0 points had you chosen the other option. $~$\\ 
You press $<<$B$>>$. You receive 42.0 points by selecting this option. You would have received 30.0 points had you chosen the other option. $~$\\ 
 $~$\\ 
Option L delivers 8.0 points with 100.0\% chance, or 8.0 points with 0.0\% chance. $~$\\ 
Option B delivers 5.0 points with 95.0\% chance, or 54.0 points with 5.0\% chance. $~$\\ 
You press $<<$B$>>$. You receive 5.0 points by selecting this option. You would have received 8.0 points had you chosen the other option. $~$\\ 
You press $<<$B$>>$. You receive 5.0 points by selecting this option. You would have received 8.0 points had you chosen the other option. $~$\\ 
You press $<<$B$>>$. You receive 5.0 points by selecting this option. You would have received 8.0 points had you chosen the other option. $~$\\ 
You press $<<$B$>>$. You receive 5.0 points by selecting this option. You would have received 8.0 points had you chosen the other option. $~$\\ 
You press $<<$B$>>$. You receive 5.0 points by selecting this option. You would have received 8.0 points had you chosen the other option. $~$\\ 
 $~$\\ 
Option L delivers 20.0 points with 100.0\% chance, or 20.0 points with 0.0\% chance. $~$\\ 
Option B delivers 15.0 points with 99.0\% chance, 47.5 points with 0.0078\% chance, 48.5 points with 0.0547\% chance, 49.5 points with 0.1641\% chance, 50.5 points with 0.2734\% chance, 51.5 points with 0.2734\% chance, 52.5 points with 0.1641\% chance, 53.5 points with 0.0547\% chance, or 54.5 points with 0.0078\% chance. $~$\\ 
You press $<<$B$>>$. You receive 15.0 points by selecting this option. You would have received 20.0 points had you chosen the other option. $~$\\ 
You press $<<$B$>>$. You receive 15.0 points by selecting this option. You would have received 20.0 points had you chosen the other option. $~$\\ 
You press $<<$B$>>$. You receive 15.0 points by selecting this option. You would have received 20.0 points had you chosen the other option. $~$\\ 
You press $<<$B$>>$. You receive 15.0 points by selecting this option. You would have received 20.0 points had you chosen the other option. $~$\\ 
You press $<<$B$>>$. You receive 15.0 points by selecting this option. You would have received 20.0 points had you chosen the other option. $~$\\ 
 $~$\\ 
Option L delivers 15.0 points with 5.0\% chance, or 9.0 points with 95.0\% 

\subsubsection*{CPC18}
Data source: \cite{plonsky2018and} \\ $~$ \\
Number of experiments: 1 $~$\\ 
Number of participants: 216 $~$\\ 
Number of choices: 162000 $~$\\ 
 $~$\\ 
\textbf{Example prompt:}
 $~$\\ 
You will encounter a series of gambling problems where you have to select between two options. $~$\\ 
You can select an option by pressing the corresponding key. $~$\\ 
You will encounter each problem 25 times. $~$\\ 
In the first five encounters, you will not receive feedback. $~$\\ 
In the remaining 20 encounters, you will receive feedback about the outcomes of both options. $~$\\ 
In cases where the probabilities are stated to be unknown, they sum up to one and remain constant within a problem. $~$\\ 
 $~$\\ 
Option F delivers 3 points with 80.0\% chance, 94 points with 1.25\% chance, 95 points with 5.0\% chance, 96 points with 7.5\% chance, 97 points with 5.0\% chance, 98 points with 1.25\% chance. $~$\\ 
Option X delivers -19 points with 50.0\% chance, 59 points with 50.0\% chance. $~$\\ 
You press $<<$X$>>$. $~$\\ 
You press $<<$X$>>$. $~$\\ 
You press $<<$X$>>$. $~$\\ 
You press $<<$X$>>$. $~$\\ 
You press $<<$X$>>$. $~$\\ 
You press $<<$X$>>$ and gain 59 points. You would have gained 3 points had you chosen option F. $~$\\ 
You press $<<$X$>>$ and gain 59 points. You would have gained 98 points had you chosen option F. $~$\\ 
You press $<<$X$>>$ and gain 59 points. You would have gained 3 points had you chosen option F. $~$\\ 
You press $<<$X$>>$ and lose 19 points. You would have gained 95 points had you chosen option F. $~$\\ 
You press $<<$X$>>$ and lose 19 points. You would have gained 3 points had you chosen option F. $~$\\ 
You press $<<$X$>>$ and gain 59 points. You would have gained 3 points had you chosen option F. $~$\\ 
You press $<<$X$>>$ and lose 19 points. You would have gained 3 points had you chosen option F. $~$\\ 
You press $<<$X$>>$ and lose 19 points. You would have gained 3 points had you chosen option F. $~$\\ 
You press $<<$F$>>$ and gain 3 points. You would have lost 19 points had you chosen option X. $~$\\ 
You press $<<$X$>>$ and lose 19 points. You would have gained 3 points had you chosen option F. $~$\\ 
You press $<<$X$>>$ and lose 19 points. You would have gained 3 points had you chosen option F. $~$\\ 
You press $<<$X$>>$ and gain 59 points. You would have gained 3 points had you chosen option F. $~$\\ 
You press $<<$X$>>$ and gain 59 points. You would have gained 3 points had you chosen option F. $~$\\ 
You press $<<$X$>>$ and lose 19 points. You would have gained 3 points had you chosen option F. $~$\\ 
You press $<<$X$>>$ and lose 19 points. You would have gained 96 points had you chosen option F. $~$\\ 
You press $<<$X$>>$ and gain 59 points. You would have gained 3 points had you chosen option F. $~$\\ 
You press $<<$X$>>$ and lose 19 points. You would have gained 3 points had you chosen option F. $~$\\ 
You press $<<$X$>>$ and gain 59 points. You would have gained 98 points had you chosen option F. $~$\\ 
You press $<<$X$>>$ and gain 59 points. You would have gained 3 points had you chosen option F. $~$\\ 
You press $<<$X$>>$ and lose 19 points. You would have gained 3 points had you chosen option F. $~$\\ 
 $~$\\ 
Option F delivers -12 points with 95.0\% chance, 47 points with 2.5\% chance, 49 points with 1.25\% chance, 53 points with 0.625\% chance, 61 points with 0.3125\% chance, 77 points with 0.15625\% chance, 109 points with 0.078125\% chance, 173 points with 0.078125\% chance. $~$\\ 
Option X delivers -9 points with 100.0\% chance. $~$\\ 
You press $<<$F$>>$. $~$\\ 
You press $<<$F$>>$. $~$\\ 
You press $<<$F$>>$. $~$\\ 
You press $<<$F$>>$. $~$\\ 
You press $<<$F$>>$. $~$\\ 
You press $<<$F$>>$ and lose 12 points. You would have lost 9 points had you chosen option X. $~$\\ 
You press $<<$F$>>$ and lose 12 points. You would have lost 9 points had you chosen option X. $~$\\ 
You press $<<$F$>>$ and lose 12 points. You would have lost 9 points had you chosen option X. $~$\\ 
You press $<<$F$>>$ and lose 12 points. You would have lost 9 points had you chosen option X. $~$\\ 
You press $<<$F$>>$ and lose 12 points. You would have lost 9 points had you chosen option X. $~$\\ 
You press $<<$F$>>$ and lose 12 points. You would have lost 9 points had you chosen option X. $~$\\ 
You press $<<$F$>>$ and lose 12 points. You would have lost 9 points had you chosen option X. $~$\\ 
You press $<<$F$>>$ and lose 12 points. You would have lost 9 points had you chosen option X. $~$\\ 
You press $<<$F$>>$ and gain 49 points. You would have lost 9 points had you chosen option X. $~$\\ 
You press $<<$F$>>$ and gain 53 points. You would have lost 9 points had you chosen option X. $~$\\ 
You press $<<$F$>>$ and lose 12 points. You would have lost 9 points had you chosen option X. $~$\\ 
You press $<<$F$>>$ and lose 12 points. You would 

\subsubsection*{Episodic long-term memory}
Data source: \cite{popov2023intent} \\ $~$ \\
Number of experiments: 3 $~$\\ 
Number of participants: 132 $~$\\ 
Number of choices: 18649 $~$\\ 
 $~$\\ 
\textbf{Example prompt:}
 $~$\\ 
In this experiment, you will go through three cycles of three tasks. $~$\\ 
Each cycle concerns itself with a list of 30 words for you to study. $~$\\ 
In the first task of each cycle, you will go through the list. $~$\\ 
You need to remember the words indicated by a red border. $~$\\ 
For each word, please make a judgement on whether the object is larger or smaller than a football, and press the key "O" if the object is larger than a football, and the key "M" if it is smaller instead. $~$\\ 
In the second task of each cycle, you will try to solve as many arithmetic equations as you can in one minute. $~$\\ 
In the third task of each cycle, you will recall the words that you memorized in the first task of that cycle. $~$\\ 
 $~$\\ 
List 1, task 1:  $~$\\ 
You see the word "church", surrounded by a blue border. You press $<<$O$>>$. The word disappears but the blue border stays for another 3 seconds before disappearing. $~$\\ 
You see the word "robot", surrounded by a blue border. You press $<<$O$>>$. The word disappears but the blue border stays for another 3 seconds before disappearing. $~$\\ 
You see the word "jewelry", surrounded by a red border. You press $<<$M$>>$. The word disappears but the red border stays for another 3 seconds before disappearing. $~$\\ 
You see the word "skull", surrounded by a blue border. You press $<<$O$>>$. The word disappears but the blue border stays for another 3 seconds before disappearing. $~$\\ 
You see the word "apple", surrounded by a blue border. You press $<<$M$>>$. The word disappears but the blue border stays for another 3 seconds before disappearing. $~$\\ 
You see the word "garden", surrounded by a blue border. You press $<<$O$>>$. The word disappears but the blue border stays for another 3 seconds before disappearing. $~$\\ 
You see the word "pipe", surrounded by a red border. You press $<<$M$>>$. The word disappears but the red border stays for another 3 seconds before disappearing. $~$\\ 
You see the word "needle", surrounded by a red border. You press $<<$M$>>$. The word disappears but the red border stays for another 3 seconds before disappearing. $~$\\ 
You see the word "circus", surrounded by a red border. You press $<<$O$>>$. The word disappears but the red border stays for another 3 seconds before disappearing. $~$\\ 
You see the word "towel", surrounded by a blue border. You press $<<$M$>>$. The word disappears but the blue border stays for another 3 seconds before disappearing. $~$\\ 
You see the word "rabbit", surrounded by a blue border. You press $<<$M$>>$. The word disappears but the blue border stays for another 3 seconds before disappearing. $~$\\ 
You see the word "diamond", surrounded by a red border. You press $<<$M$>>$. The word disappears but the red border stays for another 3 seconds before disappearing. $~$\\ 
You see the word "cocktail", surrounded by a red border. You press $<<$M$>>$. The word disappears but the red border stays for another 3 seconds before disappearing. $~$\\ 
You see the word "satellite", surrounded by a red border. You press $<<$O$>>$. The word disappears but the red border stays for another 3 seconds before disappearing. $~$\\ 
You see the word "sweater", surrounded by a blue border. You press $<<$O$>>$. The word disappears but the blue border stays for another 3 seconds before disappearing. $~$\\ 
You see the word "planet", surrounded by a red border. You press $<<$O$>>$. The word disappears but the red border stays for another 3 seconds before disappearing. $~$\\ 
You see the word "pizza", surrounded by a blue border. You press $<<$O$>>$. The word disappears but the blue border stays for another 3 seconds before disappearing. $~$\\ 
You see the word "forest", surrounded by a blue border. You press $<<$O$>>$. The word disappears but the blue border stays for another 3 seconds before disappearing. $~$\\ 
You see the word "carpet", surrounded by a blue border. You press $<<$O$>>$. The word disappears but the blue border stays for another 3 seconds before disappearing. $~$\\ 
You see the word "highway", surrounded by a red border. You press $<<$O$>>$. The word disappears but the red border stays for another 3 seconds before disappearing. $~$\\ 
You see the word "jeep", surrounded by a blue border. You press $<<$O$>>$. The word disappears but the blue border stays for another 3 seconds before disappearing. $~$\\ 
You see t 

\subsubsection*{Intertemporal choice}
Data source: \cite{ruggeri2022globalizability} \\ $~$ \\
Number of experiments: 1 $~$\\ 
Number of participants: 11937 $~$\\ 
Number of choices: 142236 $~$\\ 
 $~$\\ 
\textbf{Example prompt:}
 $~$\\ 
In the following you will be presented with multiple choices between two options G and C. $~$\\ 
Please name which option you would prefer by pressing the corresponding key. $~$\\ 
 $~$\\ 
You have the choice between receiving 500\$ immediately (press G) or receiving 550\$ in one year (press C). You press $<<$G$>>$. $~$\\ 
You have the choice between receiving 500\$ immediately (press G) or receiving 600\$ in one year (press C). You press $<<$C$>>$. $~$\\ 
You have the choice between paying 500\$ immediately (press G) or paying 550\$ in one year (press C). You press $<<$G$>>$. $~$\\ 
You have the choice between paying 500\$ immediately (press G) or paying 510\$ in one year (press C). You press $<<$C$>>$. $~$\\ 
You have the choice between receiving 5000\$ immediately (press G) or receiving 5500\$ in one year (press C). You press $<<$G$>>$. $~$\\ 
You have the choice between receiving 5000\$ immediately (press G) or receiving 6000\$ in one year (press C). You press $<<$C$>>$. $~$\\ 
You have the choice between receiving 500\$ in one year (press G) or receiving 600\$ in two years (press C). You press $<<$C$>>$. $~$\\ 
You have the choice between receiving 500\$ immediately (press G) or receiving 700\$ in two years (press C). You press $<<$C$>>$. $~$\\ 
You have the choice between wait 500\$ immediately (press G) or wait 600\$ in one year (press C). You press $<<$C$>>$. $~$\\ 
You have the choice between receive 500\$ immediately (press G) or receive 600\$ in one year (press C). You press $<<$C$>>$. 

\subsubsection*{Horizon task}
Data source: \cite{sadeghiyeh2020temporal} \\ $~$ \\
Number of experiments: 1 $~$\\ 
Number of participants: 78 $~$\\ 
Number of choices: 25336 $~$\\ 
 $~$\\ 
\textbf{Example prompt:}
 $~$\\ 
You are participating in multiple games involving two slot machines, labeled J and R. $~$\\ 
The two slot machines are different across different games. $~$\\ 
Each time you choose a slot machine, you get some points. $~$\\ 
You choose a slot machine by pressing the corresponding key. $~$\\ 
Each slot machine tends to pay out about the same amount of points on average. $~$\\ 
Your goal is to choose the slot machines that will give you the most points across the experiment. $~$\\ 
The first 4 trials in each game are instructed trials where you will be told which slot machine to choose. $~$\\ 
After these instructed trials, you will have the freedom to choose for either 1 or 6 trials. $~$\\ 
 $~$\\ 
Game 1. There are 10 trials in this game. $~$\\ 
You are instructed to press J and get 57 points. $~$\\ 
You are instructed to press R and get 29 points. $~$\\ 
You are instructed to press J and get 66 points. $~$\\ 
You are instructed to press R and get 38 points. $~$\\ 
You press $<<$R$>>$ and get 45 points. $~$\\ 
You press $<<$R$>>$ and get 38 points. $~$\\ 
You press $<<$R$>>$ and get 49 points. $~$\\ 
You press $<<$J$>>$ and get 59 points. $~$\\ 
You press $<<$R$>>$ and get 28 points. $~$\\ 
You press $<<$R$>>$ and get 51 points. $~$\\ 
 $~$\\ 
Game 2. There are 10 trials in this game. $~$\\ 
You are instructed to press J and get 76 points. $~$\\ 
You are instructed to press J and get 89 points. $~$\\ 
You are instructed to press R and get 61 points. $~$\\ 
You are instructed to press J and get 74 points. $~$\\ 
You press $<<$R$>>$ and get 59 points. $~$\\ 
You press $<<$J$>>$ and get 72 points. $~$\\ 
You press $<<$J$>>$ and get 61 points. $~$\\ 
You press $<<$R$>>$ and get 70 points. $~$\\ 
You press $<<$R$>>$ and get 73 points. $~$\\ 
You press $<<$J$>>$ and get 65 points. $~$\\ 
 $~$\\ 
Game 3. There are 10 trials in this game. $~$\\ 
You are instructed to press R and get 60 points. $~$\\ 
You are instructed to press R and get 43 points. $~$\\ 
You are instructed to press J and get 54 points. $~$\\ 
You are instructed to press J and get 65 points. $~$\\ 
You press $<<$R$>>$ and get 56 points. $~$\\ 
You press $<<$J$>>$ and get 38 points. $~$\\ 
You press $<<$R$>>$ and get 56 points. $~$\\ 
You press $<<$J$>>$ and get 61 points. $~$\\ 
You press $<<$R$>>$ and get 59 points. $~$\\ 
You press $<<$J$>>$ and get 56 points. $~$\\ 
 $~$\\ 
Game 4. There are 5 trials in this game. $~$\\ 
You are instructed to press J and get 36 points. $~$\\ 
You are instructed to press J and get 43 points. $~$\\ 
You are instructed to press R and get 57 points. $~$\\ 
You are instructed to press R and get 49 points. $~$\\ 
You press $<<$R$>>$ and get 63 points. $~$\\ 
 $~$\\ 
Game 5. There are 5 trials in this game. $~$\\ 
You are instructed to press R and get 44 points. $~$\\ 
You are instructed to press J and get 45 points. $~$\\ 
You are instructed to press R and get 29 points. $~$\\ 
You are instructed to press J and get 38 points. $~$\\ 
You press $<<$R$>>$ and get 43 points. $~$\\ 
 $~$\\ 
Game 6. There are 5 trials in this game. $~$\\ 
You are instructed to press R and get 74 points. $~$\\ 
You are instructed to press R and get 70 points. $~$\\ 
You are instructed to press R and get 61 points. $~$\\ 
You are instructed to press J and get 77 points. $~$\\ 
You press $<<$J$>>$ and get 74 points. $~$\\ 
 $~$\\ 
Game 7. There are 5 trials in this game. $~$\\ 
You are instructed to press J and get 36 points. $~$\\ 
You are instructed to press R and get 50 points. $~$\\ 
You are instructed to press J and get 49 points. $~$\\ 
You are instructed to press J and get 34 points. $~$\\ 
You press $<<$R$>>$ and get 48 points. $~$\\ 
 $~$\\ 
Game 8. There are 10 trials in this game. $~$\\ 
You are instructed to press J and get 54 points. $~$\\ 
You are instructed to press R and get 64 points. $~$\\ 
You are instructed to press R and get 63 points. $~$\\ 
You are instructed to press R and get 52 points. $~$\\ 
You press $<<$J$>>$ and get 63 points. $~$\\ 
You press $<<$J$>>$ and get 65 points. $~$\\ 
You press $<<$R$>>$ and get 70 points. $~$\\ 
You press $<<$J$>>$ and get 69 points. $~$\\ 
You press $<<$R$>>$ and get 64 points. $~$\\ 
You press $<<$J$>>$ and get 64 points. $~$\\ 
 $~$\\ 
Game 9. There are 5 trials in this game. $~$\\ 
You are instructed to press J and get 57 points. $~$\\ 
You are instructed to press R and get 50 points. $~$\\ 
You are instructed to press R and get 57 points. $~$\\ 
You are instructed to press J and get 72 points. $~$\\ 
You press $<<$R$>>$ and get 47 points. $~$\\ 
 $~$\\ 
Game 10. There are 10 trials in this game. $~$\\ 
You are instructed to press R and get 21 points. $~$\\ 
You are instructed to press J and get 26 points. $~$\\ 
You are instructed to press J and get 52 points. $~$\\ 
You are instructed to press J and get 27 points. $~$\\ 
You press $<<$R$>>$ and get 21 points. $~$\\ 
You press $<<$J$>>$ a 

\subsubsection*{Structured bandit}
Data source: \cite{schulz2020finding} \\ $~$ \\
Number of experiments: 5 $~$\\ 
Number of participants: 534 $~$\\ 
Number of choices: 160200 $~$\\ 
 $~$\\ 
\textbf{Example prompt:}
 $~$\\ 
You will be playing a game for 30 rounds. $~$\\ 
Each round contains 10 trials. $~$\\ 
In each trial, you have to select one option that will generate a reward between 0 and 50 points. $~$\\ 
You can choose between options 1, 2, 3, 4, 5, 6, 7 and 8 by pressing the corresponding key. $~$\\ 
After each round the options reset and each option can produce different rewards in the following round. $~$\\ 
Your goal is to maximize your reward. $~$\\ 
 $~$\\ 
You are playing round 1: $~$\\ 
You press $<<$1$>>$ and get 13.927462234 points. $~$\\ 
You press $<<$2$>>$ and get 36.7688570508 points. $~$\\ 
You press $<<$1$>>$ and get 14.2022179045 points. $~$\\ 
You press $<<$1$>>$ and get 14.255711791 points. $~$\\ 
You press $<<$1$>>$ and get 14.0630012349 points. $~$\\ 
You press $<<$1$>>$ and get 13.7662251776 points. $~$\\ 
You press $<<$1$>>$ and get 14.0950976416 points. $~$\\ 
You press $<<$1$>>$ and get 13.9059322374 points. $~$\\ 
You press $<<$1$>>$ and get 13.7876455405 points. $~$\\ 
You press $<<$1$>>$ and get 14.0791620504 points. $~$\\ 
 $~$\\ 
You are playing round 2: $~$\\ 
You press $<<$1$>>$ and get 7.9614376644 points. $~$\\ 
You press $<<$1$>>$ and get 8.0581019194 points. $~$\\ 
You press $<<$1$>>$ and get 7.8981838872 points. $~$\\ 
You press $<<$1$>>$ and get 7.6801851393 points. $~$\\ 
You press $<<$1$>>$ and get 7.8750440099 points. $~$\\ 
You press $<<$1$>>$ and get 7.8730616431 points. $~$\\ 
You press $<<$1$>>$ and get 7.9118344028 points. $~$\\ 
You press $<<$1$>>$ and get 8.134691905 points. $~$\\ 
You press $<<$1$>>$ and get 7.3146307967 points. $~$\\ 
You press $<<$1$>>$ and get 7.5832954876 points. $~$\\ 
 $~$\\ 
You are playing round 3: $~$\\ 
You press $<<$1$>>$ and get 45.5086514616 points. $~$\\ 
You press $<<$1$>>$ and get 45.4708060494 points. $~$\\ 
You press $<<$1$>>$ and get 45.9992623342 points. $~$\\ 
You press $<<$1$>>$ and get 45.2562607277 points. $~$\\ 
You press $<<$1$>>$ and get 45.4858045741 points. $~$\\ 
You press $<<$1$>>$ and get 45.5714527483 points. $~$\\ 
You press $<<$1$>>$ and get 45.6341629546 points. $~$\\ 
You press $<<$1$>>$ and get 45.0394158823 points. $~$\\ 
You press $<<$1$>>$ and get 45.4305272183 points. $~$\\ 
You press $<<$1$>>$ and get 45.5876540303 points. $~$\\ 
 $~$\\ 
You are playing round 4: $~$\\ 
You press $<<$1$>>$ and get 11.3846751956 points. $~$\\ 
You press $<<$1$>>$ and get 11.5363886375 points. $~$\\ 
You press $<<$1$>>$ and get 11.7231131775 points. $~$\\ 
You press $<<$1$>>$ and get 11.6522950289 points. $~$\\ 
You press $<<$1$>>$ and get 11.526071202 points. $~$\\ 
You press $<<$1$>>$ and get 11.8137491734 points. $~$\\ 
You press $<<$1$>>$ and get 11.3114379632 points. $~$\\ 
You press $<<$1$>>$ and get 11.9434174388 points. $~$\\ 
You press $<<$1$>>$ and get 11.2375196571 points. $~$\\ 
You press $<<$1$>>$ and get 11.7253908566 points. $~$\\ 
 $~$\\ 
You are playing round 5: $~$\\ 
You press $<<$1$>>$ and get 6.3035553498 points. $~$\\ 
You press $<<$1$>>$ and get 6.1091934289 points. $~$\\ 
You press $<<$1$>>$ and get 5.8593245302 points. $~$\\ 
You press $<<$1$>>$ and get 6.7472106803 points. $~$\\ 
You press $<<$1$>>$ and get 5.8960908181 points. $~$\\ 
You press $<<$1$>>$ and get 6.4056392625 points. $~$\\ 
You press $<<$1$>>$ and get 6.2178725578 points. $~$\\ 
You press $<<$1$>>$ and get 6.0009585684 points. $~$\\ 
You press $<<$1$>>$ and get 6.5478514159 points. $~$\\ 
You press $<<$1$>>$ and get 6.1982281156 points. $~$\\ 
 $~$\\ 
You are playing round 6: $~$\\ 
You press $<<$1$>>$ and get 7.6893330342 points. $~$\\ 
You press $<<$1$>>$ and get 7.3522351643 points. $~$\\ 
You press $<<$1$>>$ and get 7.5121866021 points. $~$\\ 
You press $<<$1$>>$ and get 7.7384186564 points. $~$\\ 
You press $<<$1$>>$ and get 6.8107475324 points. $~$\\ 
You press $<<$4$>>$ and get 24.8575884026 points. $~$\\ 
You press $<<$2$>>$ and get 13.8675683188 points. $~$\\ 
You press $<<$1$>>$ and get 7.8646759927 points. $~$\\ 
You press $<<$4$>>$ and get 24.9148267139 points. $~$\\ 
You press $<<$2$>>$ and get 13.7239940234 points. $~$\\ 
 $~$\\ 
You are playing round 7: $~$\\ 
You press $<<$1$>>$ and get 4.8637779497 points. $~$\\ 
You press $<<$4$>>$ and get 21.4211685718 points. $~$\\ 
You press $<<$2$>>$ and get 9.9283331067 points. $~$\\ 
You press $<<$1$>>$ and get 4.3283739798 points. $~$\\ 
You press $<<$1$>>$ and get 4.2486815414 points. $~$\\ 
You press $<<$2$>>$ and get 9.7358045234 points. $~$\\ 
You press $<<$3$>>$ and get 15.5522984698 points. $~$\\ 
You press $<<$4$>>$ and get 21.1157043351 points. $~$\\ 
You press $<<$4$>>$ and get 21.2293801786 points. $~$\\ 
You press $<<$4$>>$ and get 21.2346665196 points. $~$\\ 
 $~$\\ 
You are playing round 8: $~$\\ 
You press $<<$5$>>$ and get 21.2600225368 points. $~$\\ 
You press $<<$6$>>$ and get 14.9281922416 points. $~$\\ 
You press $<<$6$>>$ and get 15.1378822594 points. $~$\\ 
You press $<<$6$>>$ and get 15.0143768307 points. $~$\\ 
You press $<<$8$>>$ and get 2.1704472833 points. $~$\\ 
You press $<<$7$>>$ and get 8.4601420045 points. $~$\\ 
You press $<<$5$>>$ and get 

\subsubsection*{Horizon task}
Data source: \cite{somerville2017charting} \\ $~$ \\
Number of experiments: 1 $~$\\ 
Number of participants: 78 $~$\\ 
Number of choices: 43680 $~$\\ 
 $~$\\ 
\textbf{Example prompt:}
 $~$\\ 
You are participating in multiple games involving two slot machines, labeled F and N. $~$\\ 
The two slot machines are different across different games. $~$\\ 
Each time you choose a slot machine, you get some points. $~$\\ 
You choose a slot machine by pressing the corresponding key. $~$\\ 
Each slot machine tends to pay out about the same amount of points on average. $~$\\ 
Your goal is to choose the slot machines that will give you the most points across the experiment. $~$\\ 
The first 4 trials in each game are instructed trials where you will be told which slot machine to choose. $~$\\ 
After these instructed trials, you will have the freedom to choose for either 1 or 6 trials. $~$\\ 
 $~$\\ 
Game 1. There are 5 trials in this game. $~$\\ 
You are instructed to press F and get 63 points. $~$\\ 
You are instructed to press N and get 60 points. $~$\\ 
You are instructed to press N and get 64 points. $~$\\ 
You are instructed to press N and get 74 points. $~$\\ 
You press $<<$N$>>$ and get 65 points. $~$\\ 
 $~$\\ 
Game 2. There are 10 trials in this game. $~$\\ 
You are instructed to press N and get 36 points. $~$\\ 
You are instructed to press N and get 41 points. $~$\\ 
You are instructed to press F and get 40 points. $~$\\ 
You are instructed to press F and get 33 points. $~$\\ 
You press $<<$N$>>$ and get 44 points. $~$\\ 
You press $<<$N$>>$ and get 34 points. $~$\\ 
You press $<<$F$>>$ and get 48 points. $~$\\ 
You press $<<$F$>>$ and get 50 points. $~$\\ 
You press $<<$N$>>$ and get 38 points. $~$\\ 
You press $<<$F$>>$ and get 35 points. $~$\\ 
 $~$\\ 
Game 3. There are 5 trials in this game. $~$\\ 
You are instructed to press F and get 18 points. $~$\\ 
You are instructed to press N and get 51 points. $~$\\ 
You are instructed to press N and get 41 points. $~$\\ 
You are instructed to press F and get 23 points. $~$\\ 
You press $<<$N$>>$ and get 42 points. $~$\\ 
 $~$\\ 
Game 4. There are 10 trials in this game. $~$\\ 
You are instructed to press F and get 65 points. $~$\\ 
You are instructed to press N and get 55 points. $~$\\ 
You are instructed to press N and get 68 points. $~$\\ 
You are instructed to press N and get 61 points. $~$\\ 
You press $<<$F$>>$ and get 54 points. $~$\\ 
You press $<<$N$>>$ and get 74 points. $~$\\ 
You press $<<$N$>>$ and get 71 points. $~$\\ 
You press $<<$N$>>$ and get 42 points. $~$\\ 
You press $<<$F$>>$ and get 63 points. $~$\\ 
You press $<<$N$>>$ and get 53 points. $~$\\ 
 $~$\\ 
Game 5. There are 10 trials in this game. $~$\\ 
You are instructed to press N and get 18 points. $~$\\ 
You are instructed to press F and get 53 points. $~$\\ 
You are instructed to press F and get 50 points. $~$\\ 
You are instructed to press N and get 19 points. $~$\\ 
You press $<<$F$>>$ and get 46 points. $~$\\ 
You press $<<$F$>>$ and get 44 points. $~$\\ 
You press $<<$N$>>$ and get 10 points. $~$\\ 
You press $<<$F$>>$ and get 42 points. $~$\\ 
You press $<<$F$>>$ and get 39 points. $~$\\ 
You press $<<$F$>>$ and get 44 points. $~$\\ 
 $~$\\ 
Game 6. There are 10 trials in this game. $~$\\ 
You are instructed to press F and get 13 points. $~$\\ 
You are instructed to press N and get 51 points. $~$\\ 
You are instructed to press F and get 1 points. $~$\\ 
You are instructed to press N and get 36 points. $~$\\ 
You press $<<$N$>>$ and get 38 points. $~$\\ 
You press $<<$N$>>$ and get 36 points. $~$\\ 
You press $<<$N$>>$ and get 52 points. $~$\\ 
You press $<<$N$>>$ and get 36 points. $~$\\ 
You press $<<$N$>>$ and get 27 points. $~$\\ 
You press $<<$F$>>$ and get 12 points. $~$\\ 
 $~$\\ 
Game 7. There are 5 trials in this game. $~$\\ 
You are instructed to press F and get 65 points. $~$\\ 
You are instructed to press N and get 50 points. $~$\\ 
You are instructed to press F and get 74 points. $~$\\ 
You are instructed to press F and get 55 points. $~$\\ 
You press $<<$N$>>$ and get 72 points. $~$\\ 
 $~$\\ 
Game 8. There are 10 trials in this game. $~$\\ 
You are instructed to press N and get 29 points. $~$\\ 
You are instructed to press F and get 42 points. $~$\\ 
You are instructed to press F and get 40 points. $~$\\ 
You are instructed to press F and get 42 points. $~$\\ 
You press $<<$F$>>$ and get 53 points. $~$\\ 
You press $<<$F$>>$ and get 47 points. $~$\\ 
You press $<<$N$>>$ and get 40 points. $~$\\ 
You press $<<$N$>>$ and get 41 points. $~$\\ 
You press $<<$F$>>$ and get 38 points. $~$\\ 
You press $<<$F$>>$ and get 51 points. $~$\\ 
 $~$\\ 
Game 9. There are 5 trials in this game. $~$\\ 
You are instructed to press N and get 60 points. $~$\\ 
You are instructed to press N and get 66 points. $~$\\ 
You are instructed to press F and get 61 points. $~$\\ 
You are instructed to press F and get 65 points. $~$\\ 
You press $<<$N$>>$ and get 54 points. $~$\\ 
 $~$\\ 
Game 10. There are 10 trials in this game. $~$\\ 
You are instructed to press F and get 44 points. $~$\\ 
You are instructed to pr 

\subsubsection*{Weather prediction task}
Data source: \cite{speekenbrink2008learning} \\ $~$ \\
Number of experiments: 1 $~$\\ 
Number of participants: 23 $~$\\ 
Number of choices: 4600 $~$\\ 
 $~$\\ 
\textbf{Example prompt:}
 $~$\\ 
You will be playing a game in which you pretend to be a weather forecaster. $~$\\ 
In each trial, you will see between one and three tarot cards. $~$\\ 
Your task is to decide if the combination of cards presented predicts rainy weather (by pressing E) or fine weather (by pressing J). $~$\\ 
 $~$\\ 
You are seeing the following: card 2, card 4. You press $<<$E$>>$. You are correct, the weather is indeed rainy. $~$\\ 
You are seeing the following: card 1, card 2. You press $<<$J$>>$. You are correct, the weather is indeed fine. $~$\\ 
You are seeing the following: card 3, card 4. You press $<<$J$>>$. You are wrong, the weather is rainy. $~$\\ 
You are seeing the following: card 2, card 3. You press $<<$J$>>$. You are correct, the weather is indeed fine. $~$\\ 
You are seeing the following: card 1. You press $<<$E$>>$. You are wrong, the weather is fine. $~$\\ 
You are seeing the following: card 1, card 2, card 3. You press $<<$E$>>$. You are wrong, the weather is fine. $~$\\ 
You are seeing the following: card 1, card 2, card 4. You press $<<$J$>>$. You are wrong, the weather is rainy. $~$\\ 
You are seeing the following: card 1, card 2, card 4. You press $<<$E$>>$. You are correct, the weather is indeed rainy. $~$\\ 
You are seeing the following: card 3, card 4. You press $<<$E$>>$. You are correct, the weather is indeed rainy. $~$\\ 
You are seeing the following: card 3, card 4. You press $<<$E$>>$. You are correct, the weather is indeed rainy. $~$\\ 
You are seeing the following: card 1. You press $<<$J$>>$. You are correct, the weather is indeed fine. $~$\\ 
You are seeing the following: card 1, card 4. You press $<<$J$>>$. You are correct, the weather is indeed fine. $~$\\ 
You are seeing the following: card 4. You press $<<$E$>>$. You are wrong, the weather is fine. $~$\\ 
You are seeing the following: card 3, card 4. You press $<<$E$>>$. You are correct, the weather is indeed rainy. $~$\\ 
You are seeing the following: card 1, card 2. You press $<<$J$>>$. You are correct, the weather is indeed fine. $~$\\ 
You are seeing the following: card 3, card 4. You press $<<$E$>>$. You are correct, the weather is indeed rainy. $~$\\ 
You are seeing the following: card 2. You press $<<$J$>>$. You are correct, the weather is indeed fine. $~$\\ 
You are seeing the following: card 1, card 4. You press $<<$J$>>$. You are wrong, the weather is rainy. $~$\\ 
You are seeing the following: card 1, card 4. You press $<<$J$>>$. You are correct, the weather is indeed fine. $~$\\ 
You are seeing the following: card 1. You press $<<$J$>>$. You are correct, the weather is indeed fine. $~$\\ 
You are seeing the following: card 1, card 3, card 4. You press $<<$J$>>$. You are correct, the weather is indeed fine. $~$\\ 
You are seeing the following: card 2. You press $<<$J$>>$. You are correct, the weather is indeed fine. $~$\\ 
You are seeing the following: card 1, card 2. You press $<<$J$>>$. You are correct, the weather is indeed fine. $~$\\ 
You are seeing the following: card 2, card 3, card 4. You press $<<$E$>>$. You are correct, the weather is indeed rainy. $~$\\ 
You are seeing the following: card 3. You press $<<$J$>>$. You are correct, the weather is indeed fine. $~$\\ 
You are seeing the following: card 1, card 2, card 3. You press $<<$J$>>$. You are wrong, the weather is rainy. $~$\\ 
You are seeing the following: card 3, card 4. You press $<<$E$>>$. You are correct, the weather is indeed rainy. $~$\\ 
You are seeing the following: card 1, card 3. You press $<<$J$>>$. You are correct, the weather is indeed fine. $~$\\ 
You are seeing the following: card 1, card 4. You press $<<$J$>>$. You are wrong, the weather is rainy. $~$\\ 
You are seeing the following: card 2, card 3, card 4. You press $<<$E$>>$. You are wrong, the weather is fine. $~$\\ 
You are seeing the following: card 2, card 4. You press $<<$E$>>$. You are correct, the weather is indeed rainy. $~$\\ 
You are seeing the following: card 4. You press $<<$J$>>$. You are wrong, the weather is rainy. $~$\\ 
You are seeing the following: card 1, card 2. You press $<<$J$>>$. You are correct, the weather is indeed fine. $~$\\ 
You are seeing the following: card 1, card 3, card 4. You press $<<$E$>>$. You are correct, the weather is indeed rainy. $~$\\ 
You are seeing the following: card 4. You press $<<$J$>>$. You are wrong, the weather is rainy. $~$\\ 
You are seeing the following: card 1, card 2. You press $<<$J$>>$. You are correct, the weather is indeed fine. $~$\\ 
You are seeing the follow 

\subsubsection*{Iowa gambling task}
Data source: \cite{steingroever2015data} \\ $~$ \\
Number of experiments: 9 $~$\\ 
Number of participants: 511 $~$\\ 
Number of choices: 55435 $~$\\ 
 $~$\\ 
\textbf{Example prompt:}
 $~$\\ 
You see in front of you four decks of cards labeled H, V, J, and D. $~$\\ 
You get a loan of 2000\$ of play money. $~$\\ 
You have to select one card at a time, from any of the four decks, for 100 trials. $~$\\ 
You select a card from a deck by pressing the corresponding key. $~$\\ 
After turning a card, you win some money, the amount varies with the deck. $~$\\ 
You sometimes also have to pay a penalty, which also varies with the deck. $~$\\ 
Your goal is to maximize profit on the loan of the play money. $~$\\ 
 $~$\\ 
You press $<<$H$>>$. You win 100.0\$ and lose 200.0\$. $~$\\ 
You press $<<$H$>>$. You win 100.0\$ and lose 150.0\$. $~$\\ 
You press $<<$V$>>$. You win 100.0\$ and lose 0.0\$. $~$\\ 
You press $<<$D$>>$. You win 50.0\$ and lose 250.0\$. $~$\\ 
You press $<<$J$>>$. You win 50.0\$ and lose 0.0\$. $~$\\ 
You press $<<$V$>>$. You win 100.0\$ and lose 0.0\$. $~$\\ 
You press $<<$H$>>$. You win 100.0\$ and lose 0.0\$. $~$\\ 
You press $<<$V$>>$. You win 100.0\$ and lose 0.0\$. $~$\\ 
You press $<<$D$>>$. You win 50.0\$ and lose 0.0\$. $~$\\ 
You press $<<$V$>>$. You win 100.0\$ and lose 0.0\$. $~$\\ 
You press $<<$J$>>$. You win 50.0\$ and lose 0.0\$. $~$\\ 
You press $<<$J$>>$. You win 50.0\$ and lose 0.0\$. $~$\\ 
You press $<<$H$>>$. You win 100.0\$ and lose 300.0\$. $~$\\ 
You press $<<$V$>>$. You win 100.0\$ and lose 0.0\$. $~$\\ 
You press $<<$V$>>$. You win 100.0\$ and lose 1250.0\$. $~$\\ 
You press $<<$D$>>$. You win 50.0\$ and lose 0.0\$. $~$\\ 
You press $<<$J$>>$. You win 50.0\$ and lose 50.0\$. $~$\\ 
You press $<<$H$>>$. You win 100.0\$ and lose 0.0\$. $~$\\ 
You press $<<$V$>>$. You win 100.0\$ and lose 0.0\$. $~$\\ 
You press $<<$D$>>$. You win 50.0\$ and lose 0.0\$. $~$\\ 
You press $<<$J$>>$. You win 50.0\$ and lose 0.0\$. $~$\\ 
You press $<<$D$>>$. You win 50.0\$ and lose 0.0\$. $~$\\ 
You press $<<$V$>>$. You win 100.0\$ and lose 0.0\$. $~$\\ 
You press $<<$H$>>$. You win 100.0\$ and lose 0.0\$. $~$\\ 
You press $<<$D$>>$. You win 50.0\$ and lose 0.0\$. $~$\\ 
You press $<<$H$>>$. You win 100.0\$ and lose 0.0\$. $~$\\ 
You press $<<$J$>>$. You win 50.0\$ and lose 50.0\$. $~$\\ 
You press $<<$D$>>$. You win 50.0\$ and lose 0.0\$. $~$\\ 
You press $<<$V$>>$. You win 100.0\$ and lose 0.0\$. $~$\\ 
You press $<<$H$>>$. You win 100.0\$ and lose 0.0\$. $~$\\ 
You press $<<$V$>>$. You win 100.0\$ and lose 0.0\$. $~$\\ 
You press $<<$D$>>$. You win 50.0\$ and lose 0.0\$. $~$\\ 
You press $<<$H$>>$. You win 100.0\$ and lose 250.0\$. $~$\\ 
You press $<<$J$>>$. You win 50.0\$ and lose 50.0\$. $~$\\ 
You press $<<$D$>>$. You win 50.0\$ and lose 0.0\$. $~$\\ 
You press $<<$V$>>$. You win 100.0\$ and lose 0.0\$. $~$\\ 
You press $<<$H$>>$. You win 100.0\$ and lose 350.0\$. $~$\\ 
You press $<<$V$>>$. You win 100.0\$ and lose 0.0\$. $~$\\ 
You press $<<$D$>>$. You win 50.0\$ and lose 0.0\$. $~$\\ 
You press $<<$J$>>$. You win 50.0\$ and lose 0.0\$. $~$\\ 
You press $<<$H$>>$. You win 100.0\$ and lose 300.0\$. $~$\\ 
You press $<<$V$>>$. You win 100.0\$ and lose 0.0\$. $~$\\ 
You press $<<$D$>>$. You win 50.0\$ and lose 0.0\$. $~$\\ 
You press $<<$V$>>$. You win 100.0\$ and lose 1250.0\$. $~$\\ 
You press $<<$H$>>$. You win 100.0\$ and lose 0.0\$. $~$\\ 
You press $<<$V$>>$. You win 100.0\$ and lose 0.0\$. $~$\\ 
You press $<<$D$>>$. You win 50.0\$ and lose 0.0\$. $~$\\ 
You press $<<$J$>>$. You win 50.0\$ and lose 50.0\$. $~$\\ 
You press $<<$H$>>$. You win 100.0\$ and lose 0.0\$. $~$\\ 
You press $<<$V$>>$. You win 100.0\$ and lose 0.0\$. $~$\\ 
You press $<<$D$>>$. You win 50.0\$ and lose 0.0\$. $~$\\ 
You press $<<$J$>>$. You win 50.0\$ and lose 50.0\$. $~$\\ 
You press $<<$H$>>$. You win 100.0\$ and lose 0.0\$. $~$\\ 
You press $<<$V$>>$. You win 100.0\$ and lose 0.0\$. $~$\\ 
You press $<<$D$>>$. You win 50.0\$ and lose 250.0\$. $~$\\ 
You press $<<$J$>>$. You win 50.0\$ and lose 0.0\$. $~$\\ 
You press $<<$D$>>$. You win 50.0\$ and lose 0.0\$. $~$\\ 
You press $<<$V$>>$. You win 100.0\$ and lose 0.0\$. $~$\\ 
You press $<<$H$>>$. You win 100.0\$ and lose 150.0\$. $~$\\ 
You press $<<$H$>>$. You win 100.0\$ and lose 200.0\$. $~$\\ 
You press $<<$H$>>$. You win 100.0\$ and lose 350.0\$. $~$\\ 
You press $<<$V$>>$. You win 100.0\$ and lose 0.0\$. $~$\\ 
You press $<<$V$>>$. You win 100.0\$ and lose 0.0\$. $~$\\ 
You press $<<$V$>>$. You win 100.0\$ and lose 0.0\$. $~$\\ 
You press $<<$V$>>$. You win 100.0\$ and lose 0.0\$. $~$\\ 
You press $<<$V$>>$. You win 100.0\$ and lose 0.0\$. $~$\\ 
You press $<<$V$>>$. You win 100.0\$ and lose 0.0\$. $~$\\ 
You press $<<$D$>>$. You win 50.0\$ and lose 0.0\$. $~$\\ 
You press $<<$V$>>$. You win 100.0\$ and lose 0.0\$. $~$\\ 
You press $<<$V$>>$. You win 100.0\$ and lose 1250.0\$. $~$\\ 
You press $<<$J$>>$. You win 50.0\$ and lose 0.0\$. $~$\\ 
You press $<<$J$>>$. You win 50.0\$ and lose 50.0\$. $~$\\ 
You press $<<$D$>>$. You win 50.0\$ and lose 0.0\$. $~$\\ 
You press $<<$V$>>$. You win 100.0\$ and lose 0.0\$. $~$\\ 
You press $<<$V$>>$. You win 100.0\$ and lose 0.0\$. $~$\\ 
You press $<<$V$>>$. You win 100.0\$ and lose 0.0\$. $~$\\ 
You press $<<$D$>>$. You win 50.0\$ and lose 0.0\$. $~$\\ 
Yo 

\subsubsection*{Virtual subway network}
Data source: \cite{tomov2020discovery} \\ $~$ \\
Number of experiments: 4 $~$\\ 
Number of participants: 789 $~$\\ 
Number of choices: 227923 $~$\\ 
 $~$\\ 
\textbf{Example prompt:}
 $~$\\ 
Imagine that you are a tourist and you have to navigate the subway network in an unfamiliar town. $~$\\ 
In each round, you will have to navigate from a starting station to a goal station. $~$\\ 
Please try to plan your trip as quickly as possible. $~$\\ 
During your trip, you will see the name of the current station and its neighboring stations in all four directions. $~$\\ 
If there is no neighboring station in a particular direction, there will be a circle instead of a station name. $~$\\ 
You can go north by pressing G, west by pressing B, south by pressing V, and east by pressing C. $~$\\ 
When you reach the goal station, press Z to end the round and start the next round. $~$\\ 
 $~$\\ 
The new starting station is 1 and the goal station is 3. $~$\\ 
Your station: 1. Neighboring stations: circle on the north, 2 on the east, circle on the south, and 9 on the west. You press $<<$B$>>$. $~$\\ 
Your station: 9. Neighboring stations: circle on the north, 1 on the east, 8 on the south, and circle on the west. You press $<<$V$>>$. $~$\\ 
Your station: 8. Neighboring stations: 9 on the north, circle on the east, 7 on the south, and circle on the west. You press $<<$V$>>$. $~$\\ 
Your station: 7. Neighboring stations: 8 on the north, 6 on the east, circle on the south, and circle on the west. You press $<<$C$>>$. $~$\\ 
Your station: 6. Neighboring stations: circle on the north, 5 on the east, circle on the south, and 7 on the west. You press $<<$C$>>$. $~$\\ 
Your station: 5. Neighboring stations: 4 on the north, circle on the east, circle on the south, and 6 on the west. You press $<<$G$>>$. $~$\\ 
Your station: 4. Neighboring stations: 3 on the north, circle on the east, 5 on the south, and circle on the west. You press $<<$G$>>$. $~$\\ 
Your station: 3. Neighboring stations: 2 on the north, circle on the east, 4 on the south, and circle on the west. You press $<<$Z$>>$. $~$\\ 
You are successful. $~$\\ 
 $~$\\ 
The new starting station is 1 and the goal station is 3. $~$\\ 
Your station: 1. Neighboring stations: circle on the north, 2 on the east, circle on the south, and 9 on the west. You press $<<$B$>>$. $~$\\ 
Your station: 9. Neighboring stations: circle on the north, 1 on the east, 8 on the south, and circle on the west. You press $<<$V$>>$. $~$\\ 
Your station: 8. Neighboring stations: 9 on the north, circle on the east, 7 on the south, and circle on the west. You press $<<$V$>>$. $~$\\ 
Your station: 7. Neighboring stations: 8 on the north, 6 on the east, circle on the south, and circle on the west. You press $<<$C$>>$. $~$\\ 
Your station: 6. Neighboring stations: circle on the north, 5 on the east, circle on the south, and 7 on the west. You press $<<$C$>>$. $~$\\ 
Your station: 5. Neighboring stations: 4 on the north, circle on the east, circle on the south, and 6 on the west. You press $<<$G$>>$. $~$\\ 
Your station: 4. Neighboring stations: 3 on the north, circle on the east, 5 on the south, and circle on the west. You press $<<$G$>>$. $~$\\ 
Your station: 3. Neighboring stations: 2 on the north, circle on the east, 4 on the south, and circle on the west. You press $<<$Z$>>$. $~$\\ 
You are successful. $~$\\ 
 $~$\\ 
The new starting station is 4 and the goal station is 6. $~$\\ 
Your station: 4. Neighboring stations: 3 on the north, circle on the east, 5 on the south, and circle on the west. You press $<<$V$>>$. $~$\\ 
Your station: 5. Neighboring stations: 4 on the north, circle on the east, circle on the south, and 6 on the west. You press $<<$B$>>$. $~$\\ 
Your station: 6. Neighboring stations: circle on the north, 5 on the east, circle on the south, and 7 on the west. You press $<<$Z$>>$. $~$\\ 
You are successful. $~$\\ 
 $~$\\ 
The new starting station is 1 and the goal station is 7. $~$\\ 
Your station: 1. Neighboring stations: circle on the north, 2 on the east, circle on the south, and 9 on the west. You press $<<$B$>>$. $~$\\ 
Your station: 9. Neighboring stations: circle on the north, 1 on the east, 8 on the south, and circle on the west. You press $<<$V$>>$. $~$\\ 
Your station: 8. Neighboring stations: 9 on the north, circle on the east, 7 on the south, and circle on the west. You press $<<$V$>>$. $~$\\ 
Your station: 7. Neighboring stations: 8 on the north, 6 on the east, circle on the south, and circle on the west. You press $<<$Z$>>$. $~$\\ 
You are successful. $~$\\ 
 $~$\\ 
The new starting station is 9 and the goal station is 7. $~$\\ 
Your station: 9. Neighboring stations: circle  

\subsubsection*{Multi-task reinforcement learning}
Data source: \cite{tomov2021multi} \\ $~$ \\
Number of experiments: 2 $~$\\ 
Number of participants: 380 $~$\\ 
Number of choices: 76760 $~$\\ 
 $~$\\ 
\textbf{Example prompt:}
 $~$\\ 
You will explore a castle, walking from room to room. $~$\\ 
In each room, you will find different amounts of resources: wood, stone, and iron. $~$\\ 
In each room, there are three doors that lead to different rooms. $~$\\ 
The doors are labeled I, P, and G. $~$\\ 
You have to choose the right doors to find the most valuable resources. $~$\\ 
You choose a door by pressing the corresponding key. $~$\\ 
At the beginning of each round, you will be shown how valuable the resources are. $~$\\ 
These values are given as market prices for wood, stone, and iron. $~$\\ 
Multiplying the prices with the amounts of resources and adding them up yields a reward. $~$\\ 
You want to maximize the cumulative reward. $~$\\ 
After every round, you will start in room 0 again and see the new market prices. $~$\\ 
 $~$\\ 
The current market prices are 1 for wood, -1 for stone, and 0 for iron. $~$\\ 
You are in room 0. You press $<<$P$>>$ and you find 0 wood, 0 stone, and 0 iron. You get 0 points. $~$\\ 
You are in room 2. You press $<<$P$>>$ and you find 100 wood, 100 stone, and 0 iron. You get 0 points. $~$\\ 
The current market prices are -1 for wood, 1 for stone, and 0 for iron. $~$\\ 
You are in room 0. You press $<<$I$>>$ and you find 0 wood, 0 stone, and 0 iron. You get 0 points. $~$\\ 
You are in room 1. You press $<<$G$>>$ and you find 70 wood, 70 stone, and 70 iron. You get 0 points. $~$\\ 
The current market prices are -1 for wood, 1 for stone, and 0 for iron. $~$\\ 
You are in room 0. You press $<<$P$>>$ and you find 0 wood, 0 stone, and 0 iron. You get 0 points. $~$\\ 
You are in room 2. You press $<<$G$>>$ and you find 0 wood, 90 stone, and 0 iron. You get 90 points. $~$\\ 
The current market prices are -2 for wood, 1 for stone, and 0 for iron. $~$\\ 
You are in room 0. You press $<<$P$>>$ and you find 0 wood, 0 stone, and 0 iron. You get 0 points. $~$\\ 
You are in room 2. You press $<<$G$>>$ and you find 0 wood, 90 stone, and 0 iron. You get 90 points. $~$\\ 
The current market prices are 1 for wood, -2 for stone, and 0 for iron. $~$\\ 
You are in room 0. You press $<<$I$>>$ and you find 0 wood, 0 stone, and 0 iron. You get 0 points. $~$\\ 
You are in room 1. You press $<<$G$>>$ and you find 70 wood, 70 stone, and 70 iron. You get -70 points. $~$\\ 
The current market prices are -1 for wood, 1 for stone, and 0 for iron. $~$\\ 
You are in room 0. You press $<<$P$>>$ and you find 0 wood, 0 stone, and 0 iron. You get 0 points. $~$\\ 
You are in room 2. You press $<<$G$>>$ and you find 0 wood, 90 stone, and 0 iron. You get 90 points. $~$\\ 
The current market prices are 1 for wood, -2 for stone, and 0 for iron. $~$\\ 
You are in room 0. You press $<<$P$>>$ and you find 0 wood, 0 stone, and 0 iron. You get 0 points. $~$\\ 
You are in room 2. You press $<<$G$>>$ and you find 0 wood, 90 stone, and 0 iron. You get -180 points. $~$\\ 
The current market prices are 1 for wood, -1 for stone, and 0 for iron. $~$\\ 
You are in room 0. You press $<<$G$>>$ and you find 0 wood, 0 stone, and 0 iron. You get 0 points. $~$\\ 
You are in room 3. You press $<<$P$>>$ and you find 0 wood, 100 stone, and 60 iron. You get -100 points. $~$\\ 
The current market prices are 1 for wood, -2 for stone, and 0 for iron. $~$\\ 
You are in room 0. You press $<<$I$>>$ and you find 0 wood, 0 stone, and 0 iron. You get 0 points. $~$\\ 
You are in room 1. You press $<<$G$>>$ and you find 70 wood, 70 stone, and 70 iron. You get -70 points. $~$\\ 
The current market prices are -1 for wood, 1 for stone, and 0 for iron. $~$\\ 
You are in room 0. You press $<<$P$>>$ and you find 0 wood, 0 stone, and 0 iron. You get 0 points. $~$\\ 
You are in room 2. You press $<<$G$>>$ and you find 0 wood, 90 stone, and 0 iron. You get 90 points. $~$\\ 
The current market prices are 1 for wood, -2 for stone, and 0 for iron. $~$\\ 
You are in room 0. You press $<<$I$>>$ and you find 0 wood, 0 stone, and 0 iron. You get 0 points. $~$\\ 
You are in room 1. You press $<<$G$>>$ and you find 70 wood, 70 stone, and 70 iron. You get -70 points. $~$\\ 
The current market prices are -2 for wood, 1 for stone, and 0 for iron. $~$\\ 
You are in room 0. You press $<<$P$>>$ and you find 0 wood, 0 stone, and 0 iron. You get 0 points. $~$\\ 
You are in room 2. You press $<<$G$>>$ and you find 0 wood, 90 stone, and 0 iron. You get 90 points. $~$\\ 
The current market prices are 1 for wood, -2 for stone, and 0 for iron. $~$\\ 
You are in room 0. You press $<<$I$>>$ and you find 0 wood, 0 stone, and 0 iron. You get 0 points. $~$\\ 
You are in room  

\subsubsection*{Horizon task}
Data source: \cite{waltz2020differential} \\ $~$ \\
Number of experiments: 1 $~$\\ 
Number of participants: 36 $~$\\ 
Number of choices: 15290 $~$\\ 
 $~$\\ 
\textbf{Example prompt:}
 $~$\\ 
You are participating in multiple games involving two slot machines, labeled M and U. $~$\\ 
The two slot machines are different across different games. $~$\\ 
Each time you choose a slot machine, you get some points. $~$\\ 
You choose a slot machine by pressing the corresponding key. $~$\\ 
Each slot machine tends to pay out about the same amount of points on average. $~$\\ 
Your goal is to choose the slot machines that will give you the most points across the experiment. $~$\\ 
The first 4 trials in each game are instructed trials where you will be told which slot machine to choose. $~$\\ 
After these instructed trials, you will have the freedom to choose for either 1 or 6 trials. $~$\\ 
 $~$\\ 
Game 1. There are 10 trials in this game. $~$\\ 
You are instructed to press U and get 45 points. $~$\\ 
You are instructed to press M and get 25 points. $~$\\ 
You are instructed to press U and get 38 points. $~$\\ 
You are instructed to press M and get 12 points. $~$\\ 
You press $<<$U$>>$ and get 38 points. $~$\\ 
You press $<<$U$>>$ and get 42 points. $~$\\ 
You press $<<$U$>>$ and get 44 points. $~$\\ 
You press $<<$U$>>$ and get 35 points. $~$\\ 
You press $<<$U$>>$ and get 42 points. $~$\\ 
You press $<<$U$>>$ and get 45 points. $~$\\ 
 $~$\\ 
Game 2. There are 10 trials in this game. $~$\\ 
You are instructed to press U and get 71 points. $~$\\ 
You are instructed to press M and get 35 points. $~$\\ 
You are instructed to press U and get 71 points. $~$\\ 
You are instructed to press M and get 26 points. $~$\\ 
You press $<<$U$>>$ and get 61 points. $~$\\ 
You press $<<$U$>>$ and get 55 points. $~$\\ 
You press $<<$U$>>$ and get 61 points. $~$\\ 
You press $<<$U$>>$ and get 61 points. $~$\\ 
You press $<<$U$>>$ and get 67 points. $~$\\ 
You press $<<$U$>>$ and get 69 points. $~$\\ 
 $~$\\ 
Game 3. There are 5 trials in this game. $~$\\ 
You are instructed to press U and get 50 points. $~$\\ 
You are instructed to press M and get 43 points. $~$\\ 
You are instructed to press M and get 46 points. $~$\\ 
You are instructed to press U and get 45 points. $~$\\ 
You press $<<$U$>>$ and get 61 points. $~$\\ 
 $~$\\ 
Game 4. There are 10 trials in this game. $~$\\ 
You are instructed to press U and get 52 points. $~$\\ 
You are instructed to press M and get 51 points. $~$\\ 
You are instructed to press U and get 57 points. $~$\\ 
You are instructed to press M and get 54 points. $~$\\ 
You press $<<$U$>>$ and get 67 points. $~$\\ 
You press $<<$U$>>$ and get 58 points. $~$\\ 
You press $<<$U$>>$ and get 68 points. $~$\\ 
You press $<<$U$>>$ and get 57 points. $~$\\ 
You press $<<$U$>>$ and get 68 points. $~$\\ 
You press $<<$U$>>$ and get 65 points. $~$\\ 
 $~$\\ 
Game 5. There are 10 trials in this game. $~$\\ 
You are instructed to press U and get 32 points. $~$\\ 
You are instructed to press M and get 32 points. $~$\\ 
You are instructed to press M and get 38 points. $~$\\ 
You are instructed to press U and get 33 points. $~$\\ 
You press $<<$M$>>$ and get 37 points. $~$\\ 
You press $<<$M$>>$ and get 48 points. $~$\\ 
You press $<<$M$>>$ and get 38 points. $~$\\ 
You press $<<$M$>>$ and get 49 points. $~$\\ 
You press $<<$M$>>$ and get 36 points. $~$\\ 
You press $<<$M$>>$ and get 48 points. $~$\\ 
 $~$\\ 
Game 6. There are 5 trials in this game. $~$\\ 
You are instructed to press U and get 74 points. $~$\\ 
You are instructed to press M and get 51 points. $~$\\ 
You are instructed to press M and get 72 points. $~$\\ 
You are instructed to press M and get 76 points. $~$\\ 
You press $<<$U$>>$ and get 57 points. $~$\\ 
 $~$\\ 
Game 7. There are 10 trials in this game. $~$\\ 
You are instructed to press U and get 40 points. $~$\\ 
You are instructed to press U and get 34 points. $~$\\ 
You are instructed to press M and get 31 points. $~$\\ 
You are instructed to press M and get 37 points. $~$\\ 
You press $<<$U$>>$ and get 41 points. $~$\\ 
You press $<<$U$>>$ and get 43 points. $~$\\ 
You press $<<$U$>>$ and get 47 points. $~$\\ 
You press $<<$U$>>$ and get 41 points. $~$\\ 
You press $<<$U$>>$ and get 44 points. $~$\\ 
You press $<<$U$>>$ and get 45 points. $~$\\ 
 $~$\\ 
Game 8. There are 5 trials in this game. $~$\\ 
You are instructed to press U and get 56 points. $~$\\ 
You are instructed to press M and get 59 points. $~$\\ 
You are instructed to press M and get 48 points. $~$\\ 
You are instructed to press M and get 48 points. $~$\\ 
You press $<<$U$>>$ and get 71 points. $~$\\ 
 $~$\\ 
Game 9. There are 5 trials in this game. $~$\\ 
You are instructed to press U and get 46 points. $~$\\ 
You are instructed to press M and get 46 points. $~$\\ 
You are instructed to press U and get 47 points. $~$\\ 
You are instructed to press U and get 56 points. $~$\\ 
You press $<<$U$>>$ and get 47 points. $~$\\ 
 $~$\\ 
Game 10. There are 10 trials in this game. $~$\\ 
You are instructed to press M and get 49 points. $~$\\ 
You are instructed to p 

\subsubsection*{Horizon task}
Data source: \cite{wilson2014humans} and unpublished data from the authors \\ $~$ \\
Number of experiments: 5 $~$\\ 
Number of participants: 221 $~$\\ 
Number of choices: 138875 $~$\\ 
 $~$\\ 
\textbf{Example prompt:}
 $~$\\ 
You are participating in multiple games involving two slot machines, labeled C and A. $~$\\ 
The two slot machines are different across different games. $~$\\ 
Each time you choose a slot machine, you get some points. $~$\\ 
You choose a slot machine by pressing the corresponding key. $~$\\ 
Each slot machine tends to pay out about the same amount of points on average. $~$\\ 
Your goal is to choose the slot machines that will give you the most points across the experiment. $~$\\ 
The first 4 trials in each game are instructed trials where you will be told which slot machine to choose. $~$\\ 
After these instructed trials, you will have the freedom to choose for either 1 or 6 trials. $~$\\ 
 $~$\\ 
Game 1. There are 5 trials in this game. $~$\\ 
You are instructed to press A and get 66 points. $~$\\ 
You are instructed to press A and get 80 points. $~$\\ 
You are instructed to press C and get 29 points. $~$\\ 
You are instructed to press A and get 75 points. $~$\\ 
You press $<<$A$>>$ and get 81 points. $~$\\ 
 $~$\\ 
Game 2. There are 10 trials in this game. $~$\\ 
You are instructed to press A and get 69 points. $~$\\ 
You are instructed to press A and get 50 points. $~$\\ 
You are instructed to press C and get 51 points. $~$\\ 
You are instructed to press A and get 64 points. $~$\\ 
You press $<<$C$>>$ and get 42 points. $~$\\ 
You press $<<$A$>>$ and get 54 points. $~$\\ 
You press $<<$A$>>$ and get 64 points. $~$\\ 
You press $<<$A$>>$ and get 64 points. $~$\\ 
You press $<<$A$>>$ and get 57 points. $~$\\ 
You press $<<$C$>>$ and get 55 points. $~$\\ 
 $~$\\ 
Game 3. There are 10 trials in this game. $~$\\ 
You are instructed to press A and get 31 points. $~$\\ 
You are instructed to press C and get 43 points. $~$\\ 
You are instructed to press A and get 26 points. $~$\\ 
You are instructed to press C and get 36 points. $~$\\ 
You press $<<$C$>>$ and get 26 points. $~$\\ 
You press $<<$C$>>$ and get 41 points. $~$\\ 
You press $<<$C$>>$ and get 44 points. $~$\\ 
You press $<<$C$>>$ and get 44 points. $~$\\ 
You press $<<$C$>>$ and get 43 points. $~$\\ 
You press $<<$C$>>$ and get 53 points. $~$\\ 
 $~$\\ 
Game 4. There are 10 trials in this game. $~$\\ 
You are instructed to press C and get 65 points. $~$\\ 
You are instructed to press A and get 77 points. $~$\\ 
You are instructed to press A and get 52 points. $~$\\ 
You are instructed to press C and get 73 points. $~$\\ 
You press $<<$C$>>$ and get 61 points. $~$\\ 
You press $<<$C$>>$ and get 81 points. $~$\\ 
You press $<<$C$>>$ and get 70 points. $~$\\ 
You press $<<$C$>>$ and get 67 points. $~$\\ 
You press $<<$A$>>$ and get 62 points. $~$\\ 
You press $<<$C$>>$ and get 68 points. $~$\\ 
 $~$\\ 
Game 5. There are 10 trials in this game. $~$\\ 
You are instructed to press A and get 70 points. $~$\\ 
You are instructed to press C and get 19 points. $~$\\ 
You are instructed to press A and get 43 points. $~$\\ 
You are instructed to press C and get 41 points. $~$\\ 
You press $<<$A$>>$ and get 53 points. $~$\\ 
You press $<<$C$>>$ and get 19 points. $~$\\ 
You press $<<$A$>>$ and get 61 points. $~$\\ 
You press $<<$A$>>$ and get 68 points. $~$\\ 
You press $<<$A$>>$ and get 62 points. $~$\\ 
You press $<<$C$>>$ and get 46 points. $~$\\ 
 $~$\\ 
Game 6. There are 10 trials in this game. $~$\\ 
You are instructed to press C and get 63 points. $~$\\ 
You are instructed to press A and get 44 points. $~$\\ 
You are instructed to press C and get 49 points. $~$\\ 
You are instructed to press C and get 47 points. $~$\\ 
You press $<<$A$>>$ and get 52 points. $~$\\ 
You press $<<$A$>>$ and get 52 points. $~$\\ 
You press $<<$C$>>$ and get 55 points. $~$\\ 
You press $<<$A$>>$ and get 51 points. $~$\\ 
You press $<<$A$>>$ and get 34 points. $~$\\ 
You press $<<$C$>>$ and get 56 points. $~$\\ 
 $~$\\ 
Game 7. There are 5 trials in this game. $~$\\ 
You are instructed to press C and get 61 points. $~$\\ 
You are instructed to press A and get 44 points. $~$\\ 
You are instructed to press A and get 41 points. $~$\\ 
You are instructed to press A and get 47 points. $~$\\ 
You press $<<$A$>>$ and get 29 points. $~$\\ 
 $~$\\ 
Game 8. There are 5 trials in this game. $~$\\ 
You are instructed to press C and get 51 points. $~$\\ 
You are instructed to press A and get 76 points. $~$\\ 
You are instructed to press C and get 54 points. $~$\\ 
You are instructed to press A and get 84 points. $~$\\ 
You press $<<$C$>>$ and get 58 points. $~$\\ 
 $~$\\ 
Game 9. There are 10 trials in this game. $~$\\ 
You are instructed to press C and get 54 points. $~$\\ 
You are instructed to press A and get 14 points. $~$\\ 
You are instructed to press A and get 15 points. $~$\\ 
You are instructed to press C and get 46 points. $~$\\ 
You press $<<$A$>>$ and get 20 points. $~$\\ 
You press $<<$C$>>$ and get 44 points. $~$\\ 
You press $<<$A$>>$ and get 16 points. $~$\\ 
You press $<<$C$>>$ and get 46 points. $~$\\ 
You press  

\subsubsection*{Aversive learning}
Data source: \cite{wise2019computational} \\ $~$ \\
Number of experiments: 1 $~$\\ 
Number of participants: 57 $~$\\ 
Number of choices: 18240 $~$\\ 
 $~$\\ 
\textbf{Example prompt:}
 $~$\\ 
You are going to predict the probability of electric shocks associated with two visual stimuli. $~$\\ 
First, you will have to indicate the probability that a stimulus predicts a shock at the current moment in time on a rating bar (between 0 and 100 percent). $~$\\ 
After that, the outcome for each stimulus will be presented visually. $~$\\ 
An upcoming shock will be indicated by a square over the stimulus, while a no-shock outcome will be indicated by a circle. $~$\\ 
Finally, shocks will be delivered after you have learned about the outcome visually. $~$\\ 
If both stimuli indicate a shock, they will be presented one after the other in random order. $~$\\ 
The shock probability fluctuates over time such that one stimulus has a stable probability while the other varies. $~$\\ 
 $~$\\ 
Stimulus J and K are shown on the screen. You predict that the shock probability for stimulus J is $<<$50.0$>>$\% and the shock probability for stimulus K is $<<$50.0$>>$\%. After that, a circle is shown over stimulus J, and a square is shown over stimulus K. Finally, a shock is delivered for stimulus K. $~$\\ 
Stimulus J and K are shown on the screen. You predict that the shock probability for stimulus J is $<<$50.0$>>$\% and the shock probability for stimulus K is $<<$59.33$>>$\%. After that, a square is shown over stimulus J, and a circle is shown over stimulus K. Finally, a shock is delivered for stimulus J. $~$\\ 
Stimulus J and K are shown on the screen. You predict that the shock probability for stimulus J is $<<$55.72$>>$\% and the shock probability for stimulus K is $<<$59.33$>>$\%. After that, a circle is shown over stimulus J, and a circle is shown over stimulus K. Finally, no shocks are delivered. $~$\\ 
Stimulus J and K are shown on the screen. You predict that the shock probability for stimulus J is $<<$55.72$>>$\% and the shock probability for stimulus K is $<<$59.33$>>$\%. After that, a circle is shown over stimulus J, and a circle is shown over stimulus K. Finally, no shocks are delivered. $~$\\ 
Stimulus J and K are shown on the screen. You predict that the shock probability for stimulus J is $<<$55.72$>>$\% and the shock probability for stimulus K is $<<$59.33$>>$\%. After that, a circle is shown over stimulus J, and a circle is shown over stimulus K. Finally, no shocks are delivered. $~$\\ 
Stimulus J and K are shown on the screen. You predict that the shock probability for stimulus J is $<<$55.72$>>$\% and the shock probability for stimulus K is $<<$59.33$>>$\%. After that, a circle is shown over stimulus J, and a square is shown over stimulus K. Finally, a shock is delivered for stimulus K. $~$\\ 
Stimulus J and K are shown on the screen. You predict that the shock probability for stimulus J is $<<$55.72$>>$\% and the shock probability for stimulus K is $<<$62.42$>>$\%. After that, a circle is shown over stimulus J, and a circle is shown over stimulus K. Finally, no shocks are delivered. $~$\\ 
Stimulus J and K are shown on the screen. You predict that the shock probability for stimulus J is $<<$55.72$>>$\% and the shock probability for stimulus K is $<<$62.42$>>$\%. After that, a circle is shown over stimulus J, and a circle is shown over stimulus K. Finally, no shocks are delivered. $~$\\ 
Stimulus J and K are shown on the screen. You predict that the shock probability for stimulus J is $<<$55.72$>>$\% and the shock probability for stimulus K is $<<$62.42$>>$\%. After that, a square is shown over stimulus J, and a square is shown over stimulus K. Finally, shocks are delivered for both stimulus J and stimulus K. $~$\\ 
Stimulus J and K are shown on the screen. You predict that the shock probability for stimulus J is $<<$58.2$>>$\% and the shock probability for stimulus K is $<<$65.0$>>$\%. After that, a circle is shown over stimulus J, and a square is shown over stimulus K. Finally, a shock is delivered for stimulus K. $~$\\ 
Stimulus J and K are shown on the screen. You predict that the shock probability for stimulus J is $<<$58.2$>>$\% and the shock probability for stimulus K is $<<$67.48$>>$\%. After that, a circle is shown over stimulus J, and a circle is shown over stimulus K. Finally, no shocks are delivered. $~$\\ 
Stimulus J and K are shown on the screen. You predict that the shock probability for stimulus J is $<<$58.2$>>$\% and the shock probability  

\subsubsection*{Spatially correlated multi-armed bandit}
Data source: \cite{wu2018generalization}  \\ $~$ \\
Number of experiments: 1 $~$\\ 
Number of participants: 78 $~$\\ 
Number of choices: 9360 $~$\\ 
 $~$\\ 
\textbf{Example prompt:}
 $~$\\ 
You will be presented with a series of 16 different environments to explore. $~$\\ 
In each trial, you can select an option between numbers 1 and 30 by pressing the corresponding key. $~$\\ 
By selecting any of these options, you will earn points associated with each unique option. $~$\\ 
Imagine these options 1 through 30 as lying next to each other in an ordered line; options closer to each other tend to have similar rewards as rewards tend to cluster together. $~$\\ 
For each environment, you will be able to make either 5 or 10 choices. $~$\\ 
When you made all your choices in a given environment, you will start making choices in the next unexplored environment. $~$\\ 
The rewards underlying the different options are different in each environment so you will learn them anew for each environment. $~$\\ 
Each environment starts with the value of a single option revealed. $~$\\ 
When you choose the number corresponding to a different option, you will be told the value of that option and receive those points. $~$\\ 
Previously revealed options, including the starting option, can also be reselected, although there may be small changes in the point value. $~$\\ 
It is your task to gain as many points as possible across all 16 environments. $~$\\ 
 $~$\\ 
Environment number 1: $~$\\ 
The value of option 4 is 76. You have 5 choices to make in this environment. $~$\\ 
You press $<<$5$>>$ and receive 61 points. $~$\\ 
You press $<<$3$>>$ and receive 64 points. $~$\\ 
You press $<<$6$>>$ and receive 42 points. $~$\\ 
You press $<<$2$>>$ and receive 49 points. $~$\\ 
You press $<<$11$>>$ and receive 22 points. $~$\\ 
 $~$\\ 
Environment number 2: $~$\\ 
The value of option 2 is 65. You have 10 choices to make in this environment. $~$\\ 
You press $<<$3$>>$ and receive 68 points. $~$\\ 
You press $<<$1$>>$ and receive 40 points. $~$\\ 
You press $<<$9$>>$ and receive 10 points. $~$\\ 
You press $<<$17$>>$ and receive 51 points. $~$\\ 
You press $<<$18$>>$ and receive 35 points. $~$\\ 
You press $<<$16$>>$ and receive 54 points. $~$\\ 
You press $<<$14$>>$ and receive 48 points. $~$\\ 
You press $<<$23$>>$ and receive 64 points. $~$\\ 
You press $<<$22$>>$ and receive 53 points. $~$\\ 
You press $<<$23$>>$ and receive 65 points. $~$\\ 
 $~$\\ 
Environment number 3: $~$\\ 
The value of option 22 is 37. You have 5 choices to make in this environment. $~$\\ 
You press $<<$13$>>$ and receive 30 points. $~$\\ 
You press $<<$7$>>$ and receive 67 points. $~$\\ 
You press $<<$8$>>$ and receive 77 points. $~$\\ 
You press $<<$6$>>$ and receive 32 points. $~$\\ 
You press $<<$9$>>$ and receive 57 points. $~$\\ 
 $~$\\ 
Environment number 4: $~$\\ 
The value of option 16 is 34. You have 10 choices to make in this environment. $~$\\ 
You press $<<$5$>>$ and receive 77 points. $~$\\ 
You press $<<$6$>>$ and receive 53 points. $~$\\ 
You press $<<$4$>>$ and receive 79 points. $~$\\ 
You press $<<$3$>>$ and receive 41 points. $~$\\ 
You press $<<$22$>>$ and receive 25 points. $~$\\ 
You press $<<$29$>>$ and receive 41 points. $~$\\ 
You press $<<$26$>>$ and receive 46 points. $~$\\ 
You press $<<$12$>>$ and receive 64 points. $~$\\ 
You press $<<$11$>>$ and receive 41 points. $~$\\ 
You press $<<$13$>>$ and receive 55 points. $~$\\ 
 $~$\\ 
Environment number 5: $~$\\ 
The value of option 19 is 26. You have 5 choices to make in this environment. $~$\\ 
You press $<<$8$>>$ and receive 42 points. $~$\\ 
You press $<<$3$>>$ and receive 44 points. $~$\\ 
You press $<<$24$>>$ and receive 37 points. $~$\\ 
You press $<<$29$>>$ and receive 35 points. $~$\\ 
You press $<<$15$>>$ and receive 48 points. $~$\\ 
 $~$\\ 
Environment number 6: $~$\\ 
The value of option 16 is 44. You have 10 choices to make in this environment. $~$\\ 
You press $<<$9$>>$ and receive 32 points. $~$\\ 
You press $<<$4$>>$ and receive 43 points. $~$\\ 
You press $<<$21$>>$ and receive 26 points. $~$\\ 
You press $<<$26$>>$ and receive 5 points. $~$\\ 
You press $<<$1$>>$ and receive 13 points. $~$\\ 
You press $<<$6$>>$ and receive 9 points. $~$\\ 
You press $<<$12$>>$ and receive 27 points. $~$\\ 
You press $<<$18$>>$ and receive 36 points. $~$\\ 
You press $<<$23$>>$ and receive 23 points. $~$\\ 
You press $<<$17$>>$ and receive 21 points. $~$\\ 
 $~$\\ 
Environment number 7: $~$\\ 
The value of option 2 is 36. You have 5 choices to make in this environment. $~$\\ 
You press $<<$9$>>$ and receive 49 points. $~$\\ 
You press $<<$16$>>$ and receive 72 points. $~$\\ 
You press $<<$17$>>$ and receive 59 points. $~$\\ 
You press $<<$15$>>$ and receive 49 points. $~$\\ 
You press $<<$17$>>$ and receive 58 points. $~$\\ 
 $~$\\ 
Environment number 8: $~$\\ 
The value of option 9 is 66. You have 10 choices to make in this environment. $~$\\ 
You press $<<$10$>>$ and receive 66 points. $~$\\ 
You press $<<$8$>>$ and receive 37 points. $~$\\ 
You press $<<$11$>>$ and receive 38 points. $~$\\ 
You pr

\subsubsection*{Serial reaction time task}
Data source: \cite{wu2023chunking} \\ $~$ \\
Number of experiments: 2 $~$\\ 
Number of participants: 238 $~$\\ 
Number of choices: 238000 $~$\\ 
 $~$\\ 
\textbf{Example prompt:}
 $~$\\ 
Press the instructed key. $~$\\ 
 $~$\\ 
Act as fast and accurately as possible. $~$\\ 
 $~$\\ 
The instruction is to press D, you press $<<$D$>>$ in 1375 ms. That is correct. $~$\\ 
The instruction is to press K, you press $<<$K$>>$ in 1960 ms. That is correct. $~$\\ 
The instruction is to press J, you press $<<$J$>>$ in 1043 ms. That is correct. $~$\\ 
The instruction is to press F, you press $<<$F$>>$ in 718 ms. That is correct. $~$\\ 
The instruction is to press D, you press $<<$D$>>$ in 568 ms. That is correct. $~$\\ 
The instruction is to press K, you press $<<$K$>>$ in 527 ms. That is correct. $~$\\ 
The instruction is to press J, you press $<<$J$>>$ in 587 ms. That is correct. $~$\\ 
The instruction is to press K, you press $<<$K$>>$ in 565 ms. That is correct. $~$\\ 
The instruction is to press J, you press $<<$J$>>$ in 541 ms. That is correct. $~$\\ 
The instruction is to press F, you press $<<$F$>>$ in 631 ms. That is correct. $~$\\ 
The instruction is to press D, you press $<<$D$>>$ in 456 ms. That is correct. $~$\\ 
The instruction is to press K, you press $<<$K$>>$ in 480 ms. That is correct. $~$\\ 
The instruction is to press J, you press $<<$J$>>$ in 627 ms. That is correct. $~$\\ 
The instruction is to press J, you press $<<$J$>>$ in 1816 ms. That is correct. $~$\\ 
The instruction is to press F, you press $<<$F$>>$ in 731 ms. That is correct. $~$\\ 
The instruction is to press D, you press $<<$D$>>$ in 563 ms. That is correct. $~$\\ 
The instruction is to press K, you press $<<$K$>>$ in 401 ms. That is correct. $~$\\ 
The instruction is to press J, you press $<<$J$>>$ in 472 ms. That is correct. $~$\\ 
The instruction is to press D, you press $<<$D$>>$ in 414 ms. That is correct. $~$\\ 
The instruction is to press K, you press $<<$K$>>$ in 412 ms. That is correct. $~$\\ 
The instruction is to press J, you press $<<$J$>>$ in 439 ms. That is correct. $~$\\ 
The instruction is to press F, you press $<<$F$>>$ in 612 ms. That is correct. $~$\\ 
The instruction is to press K, you press $<<$K$>>$ in 468 ms. That is correct. $~$\\ 
The instruction is to press J, you press $<<$J$>>$ in 497 ms. That is correct. $~$\\ 
The instruction is to press F, you press $<<$F$>>$ in 397 ms. That is correct. $~$\\ 
The instruction is to press D, you press $<<$D$>>$ in 318 ms. That is correct. $~$\\ 
The instruction is to press K, you press $<<$K$>>$ in 374 ms. That is correct. $~$\\ 
The instruction is to press J, you press $<<$J$>>$ in 327 ms. That is correct. $~$\\ 
The instruction is to press F, you press $<<$F$>>$ in 604 ms. That is correct. $~$\\ 
The instruction is to press D, you press $<<$D$>>$ in 300 ms. That is correct. $~$\\ 
The instruction is to press K, you press $<<$K$>>$ in 555 ms. That is correct. $~$\\ 
The instruction is to press J, you press $<<$J$>>$ in 314 ms. That is correct. $~$\\ 
The instruction is to press K, you press $<<$K$>>$ in 441 ms. That is correct. $~$\\ 
The instruction is to press J, you press $<<$J$>>$ in 474 ms. That is correct. $~$\\ 
The instruction is to press F, you press $<<$D$>>$ in 394 ms. That is incorrect. $~$\\ 
The instruction is to press K, you press $<<$K$>>$ in 586 ms. That is correct. $~$\\ 
The instruction is to press K, you press $<<$K$>>$ in 528 ms. That is correct. $~$\\ 
The instruction is to press J, you press $<<$J$>>$ in 466 ms. That is correct. $~$\\ 
The instruction is to press F, you press $<<$F$>>$ in 434 ms. That is correct. $~$\\ 
The instruction is to press D, you press $<<$D$>>$ in 355 ms. That is correct. $~$\\ 
The instruction is to press J, you press $<<$K$>>$ in 390 ms. That is incorrect. $~$\\ 
The instruction is to press F, you press $<<$F$>>$ in 769 ms. That is correct. $~$\\ 
The instruction is to press D, you press $<<$D$>>$ in 416 ms. That is correct. $~$\\ 
The instruction is to press K, you press $<<$K$>>$ in 395 ms. That is correct. $~$\\ 
The instruction is to press K, you press $<<$K$>>$ in 601 ms. That is correct. $~$\\ 
The instruction is to press J, you press $<<$J$>>$ in 475 ms. That is correct. $~$\\ 
The instruction is to press J, you press $<<$J$>>$ in 549 ms. That is correct. $~$\\ 
The instruction is to press F, you press $<<$F$>>$ in 438 ms. That is correct. $~$\\ 
The instruction is to press D, you press $<<$D$>>$ in 327 ms. That is correct. $~$\\ 
The instruction is to press K, you press $<<$K$>>$ in 438 ms. That is correct. $~$\\ 
The instruction is to press J, you press $<<$J$>>$ in 342 ms. That is correct. $~$\\ 
The instruction is to press F, you press $<<$F$>>$ in 456 ms. That is correct. $~$\\ 
The instruction is to press D, you press $<<$D$>>$ in 380 ms. That is correct. $~$\\ 
The instruction is to press K, you press $<<$K> 

\subsubsection*{Decisions from description}
Data source: \cite{wulff2018meta} \\ $~$ \\ 
Number of experiments: 1 $~$\\ 
Number of participants: 1981 $~$\\ 
Number of choices: 28153 $~$\\ 
 $~$\\ 
\textbf{Example prompt:}
 $~$\\ 
You will choose from two monetary lotteries by pressing W or H. $~$\\ 
The lotteries offer different points with different probabilities. $~$\\ 
Your choice will trigger a random draw from the chosen lottery that will be added to your bonus. $~$\\ 
Your goal is to maximize your bonus. $~$\\ 
You will be presented with multiple choice problems consisting of different lotteries varying in outcomes and probabilities. $~$\\ 
 $~$\\ 
Lottery W offers 4.0 points with 80.0\% probability or 0.0 points with 20.0\% probability. $~$\\ 
Lottery H offers 3.0 points with 100.0\% probability. $~$\\ 
You press $<<$H$>>$. $~$\\ 
 $~$\\ 
Lottery W offers 4.0 points with 20.0\% probability or 0.0 points with 80.0\% probability. $~$\\ 
Lottery H offers 3.0 points with 25.0\% probability, or 0.0 points with 75.0\% probability. $~$\\ 
You press $<<$H$>>$. $~$\\ 
 $~$\\ 
Lottery W offers -3.0 points with 100.0\% probability. $~$\\ 
Lottery H offers -32.0 points with 10.0\% probability, or 0.0 points with 90.0\% probability. $~$\\ 
You press $<<$W$>>$. 

\subsubsection*{Decisions from experience}
Data source: \cite{wulff2018meta} \\ $~$ \\
Number of experiments: 79 $~$\\ 
Number of participants: 3942 $~$\\ 
Number of choices: 1015249 $~$\\ 
 $~$\\ 
\textbf{Example prompt:}
 $~$\\ 
You can sample from two monetary lotteries by pressing K or D. $~$\\ 
The lotteries offer different points with different probabilities. $~$\\ 
Initially, you will not know the outcomes and probabilities of the lotteries, but you can learn about them through sampling. $~$\\ 
Whenever you sample, a random draw from the selected lottery will be generated, which does not affect your bonus. $~$\\ 
You can sample from the lotteries in whatever order and for as long as you like. $~$\\ 
Whenever you feel ready, you can stop sampling by pressing X and then choose one lottery for real by pressing the corresponding key. $~$\\ 
This choice will then trigger a random draw from the chosen lottery that will be added to your bonus. $~$\\ 
Your goal is to maximize your bonus. $~$\\ 
You will be presented with multiple choice problems consisting of different lotteries varying in outcomes and probabilities. $~$\\ 
 $~$\\ 
You encounter a new choice problem: $~$\\ 
You press $<<$K$>>$ and observe 4.0 points. $~$\\ 
You press $<<$K$>>$ and observe 4.0 points. $~$\\ 
You press $<<$K$>>$ and observe 4.0 points. $~$\\ 
You press $<<$K$>>$ and observe 4.0 points. $~$\\ 
You press $<<$K$>>$ and observe 0.0 points. $~$\\ 
You press $<<$K$>>$ and observe 4.0 points. $~$\\ 
You press $<<$K$>>$ and observe 4.0 points. $~$\\ 
You press $<<$K$>>$ and observe 4.0 points. $~$\\ 
You press $<<$K$>>$ and observe 4.0 points. $~$\\ 
You press $<<$K$>>$ and observe 4.0 points. $~$\\ 
You press $<<$K$>>$ and observe 0.0 points. $~$\\ 
You press $<<$K$>>$ and observe 4.0 points. $~$\\ 
You press $<<$D$>>$ and observe 3.0 points. $~$\\ 
You press $<<$D$>>$ and observe 3.0 points. $~$\\ 
You press $<<$D$>>$ and observe 3.0 points. $~$\\ 
You press $<<$D$>>$ and observe 3.0 points. $~$\\ 
You press $<<$D$>>$ and observe 3.0 points. $~$\\ 
You press $<<$D$>>$ and observe 3.0 points. $~$\\ 
You press $<<$D$>>$ and observe 3.0 points. $~$\\ 
You press $<<$D$>>$ and observe 3.0 points. $~$\\ 
You press $<<$D$>>$ and observe 3.0 points. $~$\\ 
You press $<<$D$>>$ and observe 3.0 points. $~$\\ 
You press $<<$D$>>$ and observe 3.0 points. $~$\\ 
You press $<<$D$>>$ and observe 3.0 points. $~$\\ 
You press $<<$D$>>$ and observe 3.0 points. $~$\\ 
You press $<<$D$>>$ and observe 3.0 points. $~$\\ 
You press $<<$D$>>$ and observe 3.0 points. $~$\\ 
You press $<<$D$>>$ and observe 3.0 points. $~$\\ 
You press $<<$K$>>$ and observe 4.0 points. $~$\\ 
You press $<<$K$>>$ and observe 4.0 points. $~$\\ 
You press $<<$K$>>$ and observe 4.0 points. $~$\\ 
You press $<<$K$>>$ and observe 4.0 points. $~$\\ 
You press $<<$K$>>$ and observe 0.0 points. $~$\\ 
You press $<<$K$>>$ and observe 4.0 points. $~$\\ 
You press $<<$X$>>$ to stop sampling and then press $<<$K$>>$. $~$\\ 
 $~$\\ 
You encounter a new choice problem: $~$\\ 
You press $<<$K$>>$ and observe 0.0 points. $~$\\ 
You press $<<$K$>>$ and observe 0.0 points. $~$\\ 
You press $<<$K$>>$ and observe 0.0 points. $~$\\ 
You press $<<$D$>>$ and observe 0.0 points. $~$\\ 
You press $<<$D$>>$ and observe 0.0 points. $~$\\ 
You press $<<$D$>>$ and observe 0.0 points. $~$\\ 
You press $<<$D$>>$ and observe 0.0 points. $~$\\ 
You press $<<$D$>>$ and observe 3.0 points. $~$\\ 
You press $<<$K$>>$ and observe 0.0 points. $~$\\ 
You press $<<$K$>>$ and observe 0.0 points. $~$\\ 
You press $<<$K$>>$ and observe 4.0 points. $~$\\ 
You press $<<$K$>>$ and observe 0.0 points. $~$\\ 
You press $<<$K$>>$ and observe 0.0 points. $~$\\ 
You press $<<$K$>>$ and observe 0.0 points. $~$\\ 
You press $<<$K$>>$ and observe 0.0 points. $~$\\ 
You press $<<$K$>>$ and observe 4.0 points. $~$\\ 
You press $<<$K$>>$ and observe 0.0 points. $~$\\ 
You press $<<$K$>>$ and observe 4.0 points. $~$\\ 
You press $<<$K$>>$ and observe 0.0 points. $~$\\ 
You press $<<$K$>>$ and observe 0.0 points. $~$\\ 
You press $<<$K$>>$ and observe 0.0 points. $~$\\ 
You press $<<$K$>>$ and observe 0.0 points. $~$\\ 
You press $<<$K$>>$ and observe 0.0 points. $~$\\ 
You press $<<$K$>>$ and observe 0.0 points. $~$\\ 
You press $<<$K$>>$ and observe 0.0 points. $~$\\ 
You press $<<$K$>>$ and observe 0.0 points. $~$\\ 
You press $<<$K$>>$ and observe 0.0 points. $~$\\ 
You press $<<$K$>>$ and observe 0.0 points. $~$\\ 
You press $<<$K$>>$ and observe 0.0 points. $~$\\ 
You press $<<$D$>>$ and observe 0.0 points. $~$\\ 
You press $<<$D$>>$ and observe 0.0 points. $~$\\ 
You press $<<$D$>>$ and observe 0.0 points. $~$\\ 
You press $<<$D$>>$ and observe 3.0 points. $~$\\ 
You press $<<$D$>>$ and observe 0.0 points. $~$\\ 
You press $<<$D$>>$ and observe 0.0 points. $~$\\ 
You press $<<$D$>>$ and observe 0.0 points. $~$\\ 
You press $<<$D$>>$ and observe 0.0 points. $~$\\ 
You press $<<$D$>>$ and observe 0.0 points. $~$\\ 
You press $<<$D$>>$ and observe 0.0 points. $~$\\ 
You press $<<$D$>>$ and observe 3.0 points. $~$\\ 
You press $<<$D$>>$ and observe 0.0 points. $~$\\ 
You press $<<$D$>>$ and observe 0.0 points. $~$\\ 
You press $<<$D$>>$ and observe 0.0 points. $~$\\ 
You press $<<$D$>>$ and observe 0.0 points.

\subsubsection*{Changing bandit}
Data source: \cite{Xiong2023-ie} \\ $~$ \\ 
Number of experiments: 1 $~$\\ 
Number of participants: 30 $~$\\ 
Number of choices: 141000 $~$\\ 
 $~$\\ 
\textbf{Example prompt:}
 $~$\\ 
You are participating in multiple games involving two slot machines, labeled M and V. $~$\\ 
The two slot machines are different in different games. $~$\\ 
Each time you choose a slot machine, you get points (choosing the same slot machine will not always give you the same points). $~$\\ 
You select a slot machine by pressing the corresponding key. $~$\\ 
The expected points change randomly, abruptly, and independently with a hazard rate (which you will be told). $~$\\ 
When the points change, the new expected point value assigned to that slot machine is sampled from a uniform distribution (from 1 to 99 points). $~$\\ 
For example, if the hazard rate is 0.1, the expected points of the machines change with 10\%. $~$\\ 
Your goal is to choose the slot machine that will give you the most points. $~$\\ 
 $~$\\ 
Game 1. The hazard rate is 0.1. There are 100 trials in this game. $~$\\ 
You press $<<$M$>>$ and get 65 points. $~$\\ 
You press $<<$V$>>$ and get 58 points. $~$\\ 
You press $<<$M$>>$ and get 65 points. $~$\\ 
You press $<<$M$>>$ and get 65 points. $~$\\ 
You press $<<$V$>>$ and get 5 points. $~$\\ 
You press $<<$M$>>$ and get 65 points. $~$\\ 
You press $<<$M$>>$ and get 65 points. $~$\\ 
You press $<<$M$>>$ and get 65 points. $~$\\ 
You press $<<$V$>>$ and get 60 points. $~$\\ 
You press $<<$M$>>$ and get 65 points. $~$\\ 
You press $<<$M$>>$ and get 65 points. $~$\\ 
You press $<<$V$>>$ and get 39 points. $~$\\ 
You press $<<$M$>>$ and get 65 points. $~$\\ 
You press $<<$M$>>$ and get 65 points. $~$\\ 
You press $<<$M$>>$ and get 65 points. $~$\\ 
You press $<<$M$>>$ and get 65 points. $~$\\ 
You press $<<$V$>>$ and get 87 points. $~$\\ 
You press $<<$V$>>$ and get 82 points. $~$\\ 
You press $<<$V$>>$ and get 82 points. $~$\\ 
You press $<<$V$>>$ and get 56 points. $~$\\ 
You press $<<$M$>>$ and get 65 points. $~$\\ 
You press $<<$M$>>$ and get 65 points. $~$\\ 
You press $<<$M$>>$ and get 65 points. $~$\\ 
You press $<<$M$>>$ and get 65 points. $~$\\ 
You press $<<$V$>>$ and get 28 points. $~$\\ 
You press $<<$M$>>$ and get 65 points. $~$\\ 
You press $<<$M$>>$ and get 65 points. $~$\\ 
You press $<<$M$>>$ and get 65 points. $~$\\ 
You press $<<$M$>>$ and get 65 points. $~$\\ 
You press $<<$M$>>$ and get 65 points. $~$\\ 
You press $<<$M$>>$ and get 65 points. $~$\\ 
You press $<<$M$>>$ and get 65 points. $~$\\ 
You press $<<$M$>>$ and get 65 points. $~$\\ 
You press $<<$M$>>$ and get 65 points. $~$\\ 
You press $<<$M$>>$ and get 65 points. $~$\\ 
You press $<<$M$>>$ and get 65 points. $~$\\ 
You press $<<$M$>>$ and get 65 points. $~$\\ 
You press $<<$M$>>$ and get 65 points. $~$\\ 
You press $<<$M$>>$ and get 10 points. $~$\\ 
You press $<<$V$>>$ and get 28 points. $~$\\ 
You press $<<$V$>>$ and get 28 points. $~$\\ 
You press $<<$V$>>$ and get 28 points. $~$\\ 
You press $<<$V$>>$ and get 28 points. $~$\\ 
You press $<<$V$>>$ and get 28 points. $~$\\ 
You press $<<$V$>>$ and get 28 points. $~$\\ 
You press $<<$V$>>$ and get 28 points. $~$\\ 
You press $<<$M$>>$ and get 13 points. $~$\\ 
You press $<<$V$>>$ and get 28 points. $~$\\ 
You press $<<$V$>>$ and get 28 points. $~$\\ 
You press $<<$V$>>$ and get 28 points. $~$\\ 
You press $<<$V$>>$ and get 28 points. $~$\\ 
You press $<<$V$>>$ and get 28 points. $~$\\ 
You press $<<$M$>>$ and get 13 points. $~$\\ 
You press $<<$V$>>$ and get 29 points. $~$\\ 
You press $<<$M$>>$ and get 88 points. $~$\\ 
You press $<<$M$>>$ and get 88 points. $~$\\ 
You press $<<$M$>>$ and get 88 points. $~$\\ 
You press $<<$M$>>$ and get 88 points. $~$\\ 
You press $<<$M$>>$ and get 88 points. $~$\\ 
You press $<<$M$>>$ and get 88 points. $~$\\ 
You press $<<$V$>>$ and get 47 points. $~$\\ 
You press $<<$M$>>$ and get 88 points. $~$\\ 
You press $<<$M$>>$ and get 53 points. $~$\\ 
You press $<<$V$>>$ and get 18 points. $~$\\ 
You press $<<$M$>>$ and get 3 points. $~$\\ 
You press $<<$V$>>$ and get 18 points. $~$\\ 
You press $<<$V$>>$ and get 18 points. $~$\\ 
You press $<<$V$>>$ and get 18 points. $~$\\ 
You press $<<$V$>>$ and get 18 points. $~$\\ 
You press $<<$V$>>$ and get 18 points. $~$\\ 
You press $<<$V$>>$ and get 18 points. $~$\\ 
You press $<<$V$>>$ and get 18 points. $~$\\ 
You press $<<$M$>>$ and get 49 points. $~$\\ 
You press $<<$M$>>$ and get 49 points. $~$\\ 
You press $<<$M$>>$ and get 49 points. $~$\\ 
You press $<<$M$>>$ and get 49 points. $~$\\ 
You press $<<$M$>>$ and get 49 points. $~$\\ 
You press $<<$M$>>$ and get 49 points. $~$\\ 
You press $<<$M$>>$ and get 49 points. $~$\\ 
You press $<<$M$>>$ and get 49 points. $~$\\ 
You press $<<$V$>>$ and get 71 points. $~$\\ 
You press $<<$V$>>$ and get 71 points. $~$\\ 
You press $<<$M$>>$ and get 68 points. $~$\\ 
You press $<<$V$>>$ and get 59 points. $~$\\ 
You press $<<$M$>>$ and get 68 points. $~$\\ 
You press $<<$M$>>$ and get 68 points. $~$\\ 
You press $<<$M$>>$ and get 68 points. $~$\\ 
You press $<<$M$>>$ and get 68 points. $~$\\ 
You press $<<$M$>>$ and get 68 points. $~$\\ 
You press $<<$M$>>$ and get 68 points. $~$\\ 
You press $<<$M$>>$ and get 68 points. $~$\\ 
You press $<<$M$>>$ and get 68 points. $~$\\ 
You press $<<$M$>>$ and get 68 points. $~$\\ 
You press $<<$M$>>$ and g 

\subsubsection*{Probabilistic reasoning}
Data source: \cite{zhu2020bayesian} \\ $~$ \\
Number of experiments: 2 $~$\\ 
Number of participants: 128 $~$\\ 
Number of choices: 19740 $~$\\ 
 $~$\\ 
\textbf{Example prompt:}
 $~$\\ 
You have to estimate the probability of a series of weather-related queries. $~$\\ 
Please respond by typing your estimated probabilities. $~$\\ 
 $~$\\ 
What is the probability that the weather will be icy and not frosty on a random day in England? You estimate $<<$20$>>$\%. $~$\\ 
What is the probability that the weather will be not normal or not typical on a random day in England? You estimate $<<$50$>>$\%. $~$\\ 
What is the probability that the weather will be not frosty on a random day in England? You estimate $<<$65$>>$\%. $~$\\ 
If the weather in England is not frosty on a random day, what is the probability that the weather will also be not icy on the same day? You estimate $<<$40$>>$\%. $~$\\ 
If the weather in England is icy on a random day, what is the probability that the weather will also be not frosty on the same day? You estimate $<<$20$>>$\%. $~$\\ 
If the weather in England is typical on a random day, what is the probability that the weather will also be normal on the same day? You estimate $<<$50$>>$\%. $~$\\ 
What is the probability that the weather will be icy and frosty on a random day in England? You estimate $<<$35$>>$\%. $~$\\ 
What is the probability that the weather will be normal on a random day in England? You estimate $<<$50$>>$\%. $~$\\ 
What is the probability that the weather will be icy or frosty on a random day in England? You estimate $<<$35$>>$\%. $~$\\ 
What is the probability that the weather will be not icy or not frosty on a random day in England? You estimate $<<$60$>>$\%. $~$\\ 
If the weather in England is not typical on a random day, what is the probability that the weather will also be not normal on the same day? You estimate $<<$40$>>$\%. $~$\\ 
What is the probability that the weather will be normal and typical on a random day in England? You estimate $<<$30$>>$\%. $~$\\ 
What is the probability that the weather will be not normal on a random day in England? You estimate $<<$20$>>$\%. $~$\\ 
If the weather in England is not frosty on a random day, what is the probability that the weather will also be icy on the same day? You estimate $<<$20$>>$\%. $~$\\ 
If the weather in England is not normal on a random day, what is the probability that the weather will also be typical on the same day? You estimate $<<$40$>>$\%. $~$\\ 
If the weather in England is icy on a random day, what is the probability that the weather will also be frosty on the same day? You estimate $<<$60$>>$\%. $~$\\ 
If the weather in England is not icy on a random day, what is the probability that the weather will also be not frosty on the same day? You estimate $<<$60$>>$\%. $~$\\ 
What is the probability that the weather will be typical on a random day in England? You estimate $<<$70$>>$\%. $~$\\ 
What is the probability that the weather will be typical or not normal on a random day in England? You estimate $<<$50$>>$\%. $~$\\ 
What is the probability that the weather will be typical and not normal on a random day in England? You estimate $<<$30$>>$\%. $~$\\ 
What is the probability that the weather will be not normal and not typical on a random day in England? You estimate $<<$40$>>$\%. $~$\\ 
If the weather in England is frosty on a random day, what is the probability that the weather will also be not icy on the same day? You estimate $<<$60$>>$\%. $~$\\ 
What is the probability that the weather will be frosty or not icy on a random day in England? You estimate $<<$65$>>$\%. $~$\\ 
If the weather in England is not typical on a random day, what is the probability that the weather will also be normal on the same day? You estimate $<<$34$>>$\%. $~$\\ 
If the weather in England is normal on a random day, what is the probability that the weather will also be not typical on the same day? You estimate $<<$45$>>$\%. $~$\\ 
If the weather in England is normal on a random day, what is the probability that the weather will also be typical on the same day? You estimate $<<$10$>>$\%. $~$\\ 
What is the probability that the weather will be icy or not frosty on a random day in England? You estimate $<<$40$>>$\%. $~$\\ 
What is the probability that the weather will be frosty on a random day in England? You estimate $<<$40$>>$\%. $~$\\ 
If the weather in England is frosty on a random day, what is the probability that the weather will also be icy on the same day? You estimate $<<$40$>>$\%. $~$\\ 
If the weather in England is not icy on a random day, what is the probability that the wea 

\subsubsection*{Two-step task}
Data source: \cite{Zorowitz_Niv_2023} \\ $~$ \\
Number of experiments: 1 $~$\\ 
Number of participants: 139 $~$\\ 
Number of choices: 55878 $~$\\ 
 $~$\\ 
\textbf{Example prompt:}
 $~$\\ 
You are participating in a space treasure game. $~$\\ 
In this game, you will be visiting two alien planets in search of treasure. $~$\\ 
Each planet has two aliens on it. $~$\\ 
The blue aliens live on the blue planet. $~$\\ 
The red aliens live on the red planet. $~$\\ 
When you visit a planet, you can choose an alien to trade with by pressing the corresponding button. $~$\\ 
When you trade with an alien, it will either give you treasure or junk. $~$\\ 
Your goal is to figure out, and trade with, the aliens that are most likely to give you treasure. $~$\\ 
To visit a planet, you will choose one rocket ship from two by pressing the corresponding button. $~$\\ 
They have different designations. $~$\\ 
Each rocket ship has a planet it will fly to most of the time. $~$\\ 
But sometimes they will take you to the other planet! $~$\\ 
Remember the following hints: $~$\\ 
1. How likely an alien is to give you treasure will change over time, but this change will be slow. $~$\\ 
2. Whether you get treasure depends only on the alien you choose to trade with. $~$\\ 
3. If there is an alien you want to trade with, remember to pick the rocket ship that is most likely to take you to that alien’s planet. $~$\\ 
 $~$\\ 
You are presented with two spaceships called S and C. You press $<<$S$>>$. You end up on the blue planet. You see a blue alien named D and a blue alien named R. You press $<<$R$>>$. You find junk. $~$\\ 
You are presented with two spaceships called S and C. You press $<<$S$>>$. You end up on the blue planet. You see a blue alien named D and a blue alien named R. You press $<<$D$>>$. You find treasure. $~$\\ 
You are presented with two spaceships called S and C. You press $<<$S$>>$. You end up on the blue planet. You see a blue alien named D and a blue alien named R. You press $<<$D$>>$. You find junk. $~$\\ 
You are presented with two spaceships called S and C. You press $<<$C$>>$. You end up on the red planet. You see a red alien named G and a red alien named V. You press $<<$V$>>$. You find junk. $~$\\ 
You are presented with two spaceships called S and C. You press $<<$S$>>$. You end up on the red planet. You see a red alien named G and a red alien named V. You press $<<$G$>>$. You find treasure. $~$\\ 
You are presented with two spaceships called S and C. You press $<<$S$>>$. You end up on the blue planet. You see a blue alien named D and a blue alien named R. You press $<<$D$>>$. You find junk. $~$\\ 
You are presented with two spaceships called S and C. You press $<<$S$>>$. You end up on the blue planet. You see a blue alien named D and a blue alien named R. You press $<<$R$>>$. You find junk. $~$\\ 
You are presented with two spaceships called S and C. You press $<<$C$>>$. You end up on the red planet. You see a red alien named G and a red alien named V. You press $<<$G$>>$. You find junk. $~$\\ 
You are presented with two spaceships called S and C. You press $<<$C$>>$. You end up on the blue planet. You see a blue alien named D and a blue alien named R. You press $<<$D$>>$. You find junk. $~$\\ 
You are presented with two spaceships called S and C. You press $<<$S$>>$. You end up on the blue planet. You see a blue alien named D and a blue alien named R. You press $<<$R$>>$. You find treasure. $~$\\ 
You are presented with two spaceships called S and C. You press $<<$S$>>$. You end up on the red planet. You see a red alien named G and a red alien named V. You press $<<$V$>>$. You find treasure. $~$\\ 
You are presented with two spaceships called S and C. You press $<<$S$>>$. You end up on the red planet. You see a red alien named G and a red alien named V. You press $<<$V$>>$. You find treasure. $~$\\ 
You are presented with two spaceships called S and C. You press $<<$S$>>$. You end up on the blue planet. You see a blue alien named D and a blue alien named R. You press $<<$D$>>$. You find treasure. $~$\\ 
You are presented with two spaceships called S and C. You press $<<$S$>>$. You end up on the blue planet. You see a blue alien named D and a blue alien named R. You press $<<$R$>>$. You find junk. $~$\\ 
You are presented with two spaceships called S and C. You press $<<$S$>>$. You end up on the red planet. You see a red alien named G and a red alien named V. You press $<<$V$>>$. You find junk. $~$\\ 
You are presented with two spaceships called S and C. You press $<<$S$>>$. You end up on the blue planet. You see a blue alien named D and a blue alien 

\subsection*{Evaluation data}

\subsubsection*{Two-step task (modified cover story)}

Data source: \cite{feher2020humans} \\ $~$ \\
Number of experiments: 1 $~$\\ 
Number of participants: 24 $~$\\ 
Number of choices: 9702 $~$\\ 
 $~$\\ 
Example prompt: $~$\\ 
 $~$\\ 
You are playing the role of a musician living in a fantasy land. $~$\\ 
You play the flute for gold coins to an audience of genies, who live inside magic lamps on Pink Mountain and Blue Mountain. $~$\\ 
Pink Mountain has genies H and J, and Blue Mountain has genies A and E. $~$\\ 
Each genie lives in a lamp with the corresponding letter on it. $~$\\ 
When you arrive on a mountain, you can pick up a lamp and rub it. $~$\\ 
If the genie is in the mood for music, he will come out of his lamp, listen to a song, and give you a gold coin. $~$\\ 
Each genie’s interest in music changes with time. $~$\\ 
To go to the mountains, you chose one of two magic carpets, which you purchase from a magician, who enchants them to fly. $~$\\ 
Magic carpet K generally flies to Pink Mountain, and magic carpet O generally flies to Blue Mountain. $~$\\ 
However, on rare occasions a strong wind blowing from that mountain makes flying there too dangerous because the wind might blow you off the carpet. $~$\\ 
In this case, the carpet is forced to land instead on the other mountain. $~$\\ 
You can take a magic carpet or pick up a lamp and rub it by pressing the corresponding key. $~$\\ 
Your goal is to get as many coins as possible over the next 201 days. $~$\\ 
 $~$\\ 
You are presented with magic carpets K and O. You press $<<$K$>>$. You end up on Pink Mountain. You see lamp H and lamp J. You rub lamp $<<$H$>>$. You receive 0 coins. $~$\\ 
You are presented with magic carpets O and K. You press $<<$K$>>$. You end up on Pink Mountain. You see lamp H and lamp J. You rub lamp $<<$J$>>$. You receive 1 coins. $~$\\ 
You are presented with magic carpets K and O. You press $<<$K$>>$. You end up on Pink Mountain. You see lamp H and lamp J. You rub lamp $<<$J$>>$. You receive 1 coins. $~$\\ 
You are presented with magic carpets O and K. You press $<<$K$>>$. You end up on Pink Mountain. You see lamp H and lamp J. You rub lamp $<<$J$>>$. You receive 1 coins. $~$\\ 
You are presented with magic carpets K and O. You press $<<$K$>>$. You end up on Pink Mountain. You see lamp H and lamp J. You rub lamp $<<$J$>>$. You receive 0 coins. $~$\\ 
You are presented with magic carpets O and K. You press $<<$O$>>$. You end up on Blue Mountain. You see lamp A and lamp E. You rub lamp $<<$A$>>$. You receive 1 coins. $~$\\ 
You are presented with magic carpets O and K. You press $<<$O$>>$. You end up on Blue Mountain. You see lamp A and lamp E. You rub lamp $<<$A$>>$. You receive 0 coins. $~$\\ 
You are presented with magic carpets K and O. You press $<<$O$>>$. You end up on Blue Mountain. You see lamp A and lamp E. You rub lamp $<<$E$>>$. You receive 1 coins. $~$\\ 
You are presented with magic carpets K and O. You press $<<$O$>>$. You end up on Blue Mountain. You see lamp A and lamp E. You rub lamp $<<$E$>>$. You receive 0 coins. $~$\\ 
You are presented with magic carpets O and K. You press $<<$K$>>$. You end up on Pink Mountain. You see lamp H and lamp J. You rub lamp $<<$J$>>$. You receive 0 coins. $~$\\ 
You are presented with magic carpets O and K. You press $<<$K$>>$. You end up on Pink Mountain. You see lamp H and lamp J. You rub lamp $<<$H$>>$. You receive 0 coins. $~$\\ 
You are presented with magic carpets O and K. You press $<<$O$>>$. You end up on Blue Mountain. You see lamp A and lamp E. You rub lamp $<<$A$>>$. You receive 1 coins. $~$\\ 
You are presented with magic carpets K and O. You press $<<$O$>>$. You end up on Pink Mountain. You see lamp H and lamp J. You rub lamp $<<$J$>>$. You receive 1 coins. $~$\\ 
You are presented with magic carpets K and O. You press $<<$O$>>$. You end up on Blue Mountain. You see lamp A and lamp E. You rub lamp $<<$A$>>$. You receive 0 coins. $~$\\ 
You are presented with magic carpets O and K. You press $<<$K$>>$. You end up on Blue Mountain. You see lamp A and lamp E. You rub lamp $<<$A$>>$. You receive 1 coins. $~$\\ 
You are presented with magic carpets K and O. You press $<<$O$>>$. You end up on Pink Mountain. You see lamp H and lamp J. You rub lamp $<<$J$>>$. You receive 0 coins. $~$\\ 
You are presented with magic carpets K and O. You press $<<$K$>>$. You end up on Pink Mountain. You see lamp H and lamp J. You rub lamp $<<$J$>>$. You receive 0 coins. $~$\\ 
You are presented with magic carpets O and K. You press $<<$O$>>$. You end up on Blue Mountain. You see lamp A and lamp E. You rub lamp $<<$A$>>$. You receive 0 coins. $~$\\ 
You are presented with magic carpets O and K. You pr

\subsubsection*{Maggie's farm (modified problem structure)}

Data source: \cite{dubois2022value} \\ $~$ \\
Number of experiments: 1 $~$\\ 
Number of participants: 658 $~$\\ 
Number of choices: 921200 $~$\\ 
 $~$\\ 
Example prompt: $~$\\ 
 $~$\\ 
 You are participating in multiple games involving three apple trees, labeled S, F, and N. $~$\\ 
The three apple trees are different across different games. $~$\\ 
Each time you choose an apple tree, you get an apple of a given size. $~$\\ 
You choose an apple tree by pressing the corresponding key. $~$\\ 
Each apple tree tends to provide apples of about the same size on average. $~$\\ 
Your goal is to choose the apple trees that will give you the largest apples across the experiment. $~$\\ 
The first few trials in each game are instructed trials where you will be told which apple tree to choose. $~$\\ 
After these instructed trials, you will have the freedom to choose for either 1 or 6 trials. $~$\\ 
 $~$\\ 
Game 1. There are 8 trials in this game. $~$\\ 
You are instructed to press F and get an apple with size 3.0 centimeters. $~$\\ 
You are instructed to press N and get an apple with size 2.0 centimeters. $~$\\ 
You press $<<$S$>>$ and get an apple with size 4.0 centimeters. $~$\\ 
You press $<<$S$>>$ and get an apple with size 4.0 centimeters. $~$\\ 
You press $<<$S$>>$ and get an apple with size 4.0 centimeters. $~$\\ 
You press $<<$F$>>$ and get an apple with size 5.0 centimeters. $~$\\ 
You press $<<$N$>>$ and get an apple with size 5.0 centimeters. $~$\\ 
You press $<<$N$>>$ and get an apple with size 4.0 centimeters. $~$\\ 
 $~$\\ 
Game 2. There are 10 trials in this game. $~$\\ 
You are instructed to press F and get an apple with size 9.0 centimeters. $~$\\ 
You are instructed to press N and get an apple with size 2.0 centimeters. $~$\\ 
You are instructed to press F and get an apple with size 10.0 centimeters. $~$\\ 
You are instructed to press F and get an apple with size 10.0 centimeters. $~$\\ 
You press $<<$F$>>$ and get an apple with size 10.0 centimeters. $~$\\ 
You press $<<$F$>>$ and get an apple with size 8.0 centimeters. $~$\\ 
You press $<<$F$>>$ and get an apple with size 9.0 centimeters. $~$\\ 
You press $<<$F$>>$ and get an apple with size 7.0 centimeters. $~$\\ 
You press $<<$F$>>$ and get an apple with size 9.0 centimeters. $~$\\ 
You press $<<$F$>>$ and get an apple with size 10.0 centimeters. $~$\\ 
 $~$\\ 
Game 3. There are 8 trials in this game. $~$\\ 
You are instructed to press S and get an apple with size 2.0 centimeters. $~$\\ 
You are instructed to press F and get an apple with size 6.0 centimeters. $~$\\ 
You press $<<$F$>>$ and get an apple with size 5.0 centimeters. $~$\\ 
You press $<<$F$>>$ and get an apple with size 6.0 centimeters. $~$\\ 
You press $<<$F$>>$ and get an apple with size 6.0 centimeters. $~$\\ 
You press $<<$F$>>$ and get an apple with size 5.0 centimeters. $~$\\ 
You press $<<$F$>>$ and get an apple with size 6.0 centimeters. $~$\\ 
You press $<<$F$>>$ and get an apple with size 7.0 centimeters. $~$\\ 
 $~$\\ 
Game 4. There are 11 trials in this game. $~$\\ 
You are instructed to press F and get an apple with size 5.0 centimeters. $~$\\ 
You are instructed to press S and get an apple with size 6.0 centimeters. $~$\\ 
You are instructed to press F and get an apple with size 4.0 centimeters. $~$\\ 
You are instructed to press F and get an apple with size 4.0 centimeters. $~$\\ 
You are instructed to press N and get an apple with size 2.0 centimeters. $~$\\ 
You press $<<$S$>>$ and get an apple with size 6.0 centimeters. $~$\\ 
You press $<<$S$>>$ and get an apple with size 5.0 centimeters. $~$\\ 
You press $<<$S$>>$ and get an apple with size 3.0 centimeters. $~$\\ 
You press $<<$S$>>$ and get an apple with size 4.0 centimeters. $~$\\ 
You press $<<$N$>>$ and get an apple with size 2.0 centimeters. $~$\\ 
You press $<<$S$>>$ and get an apple with size 4.0 centimeters. $~$\\ 
 $~$\\ 
Game 5. There are 5 trials in this game. $~$\\ 
You are instructed to press S and get an apple with size 6.0 centimeters. $~$\\ 
You are instructed to press F and get an apple with size 7.0 centimeters. $~$\\ 
You are instructed to press S and get an apple with size 6.0 centimeters. $~$\\ 
You are instructed to press S and get an apple with size 5.0 centimeters. $~$\\ 
You press $<<$S$>>$ and get an apple with size 6.0 centimeters. $~$\\ 
 $~$\\ 
Game 6. There are 11 trials in this game. $~$\\ 
You are instructed to press S and get an apple with size 6.0 centimeters. $~$\\ 
You are instructed to press S and get an apple with size 4.0 centimeters. $~$\\ 
You are instructed to press S and get an apple with size 5.0 centimeters. $~$\\ 
You are instructed to press F and get an apple with size 3.0 centimeters. $~$\\ 
You are instructed to press N and get an apple with size 4.0 centimeters. $~$\\ 
You press $<<$S$>>$ and get an apple with size 6.0 centime

\subsubsection*{Logical reasoning (entirely novel domain)}

Data source: \cite{jansen2021rational} \\ $~$ \\
Number of experiments: 1 $~$\\ 
Number of participants: 3543 $~$\\ 
Number of choices: 99204 $~$\\ 
 $~$\\ 
Example prompt: $~$\\ 
 $~$\\ 
 You're about to answer a set of 20 questions about logical reasoning. How many of the 20 questions do you think you will answer correctly? $~$\\ 
You say $<<$12$>>$. $~$\\ 
Compared to other participants in this study, how well do you think you will do? Marking 90\% means you will do better than 90\% of participants, marking 10\% means you will do better than only 10\%, and marking 50\% means that you will perform better than half of the participants. $~$\\ 
You say $<<$70$>>$\%. $~$\\ 
On a scale of 0 to 10, how difficult is solving logical reasoning problems for the average participant? $~$\\ 
You say $<<$6$>>$. $~$\\ 
On a scale of 0 to 10, how difficult is solving logical reasoning problems for you? $~$\\ 
You say $<<$4$>>$. $~$\\ 
 $~$\\ 
You will be presented with brief passages or statements and will be required to evaluate their reasoning or determine what inferences you can logically draw from the passage. $~$\\ 
Your task is to use the buttons D, Z, F, O, and X to select the best answer choice, even though more than one choice may present a possible answer. $~$\\ 
 $~$\\ 
Q1. Life imitates art. Which of the following, if true, most strongly supports the previous statement? $~$\\ 
The choices are: $~$\\ 
Z: When Warren Beatty filmed Reds, he tried to suggest not only the chaos of the Russian Revolution but also its relationship to the present. $~$\\ 
X: The number of professional ballet companies has increased over the last five years, but the number of dance majors has decreased. $~$\\ 
D: On Tuesday, the business section of the newspaper had predicted the drop in interest rates that occurred on Friday. $~$\\ 
O: Truman Capote wrote In Cold Blood as a result of a series of brutal slayings by two crazed killers. $~$\\ 
F: Soon after the advent of color television, white shirts became less popular as dressy attire for men, and pastel-colored shirts began to sell well. $~$\\ 
You press $<<$F$>>$. $~$\\ 
 $~$\\ 
Q2. On average, federal workers receive salaries 35.5 percent higher than private-sector salaries. For instance, federal workers in California average \$19,206 a year, 25 percent higher than the average pay in the private sector, which is \$15,365. This information would best support which of the following opinions? $~$\\ 
The choices are: $~$\\ 
Z: Private-sector salaries in California are above average. $~$\\ 
X: The private sector is being paid fairly. $~$\\ 
O: Federal jobs are more secure than private-sector jobs. $~$\\ 
D: Public-sector work is more difficult than private-sector work. $~$\\ 
F: Federal pay is out of line. $~$\\ 
You press $<<$Z$>>$. $~$\\ 
 $~$\\ 
Q3. No high jumper entered the track meet unless he or she was a track club member. No track club member both entered the meet and was a high jumper. Which of the following conclusions can be correctly drawn from the two previous sentences? $~$\\ 
The choices are: $~$\\ 
F: No one but high jumpers entered the meet. $~$\\ 
D: Only track club members entered the meet. $~$\\ 
X: No track club members entered the meet. $~$\\ 
Z: No high jumper entered the meet. $~$\\ 
O: Some track club members entered the meet. $~$\\ 
You press $<<$Z$>>$. $~$\\ 
 $~$\\ 
Q4. About 33\% of American men between 25 and 50 are overweight. Research has shown that in most cases men between 25 and 50 who are overweight are more subject to heart disease than men who are not overweight. Which of the following is the most logical conclusion to this argument? $~$\\ 
The choices are: $~$\\ 
D: Therefore, 33\% of the American men between 25 and 50 should lose weight. $~$\\ 
O: Therefore, if 33\% of the American men between 25 and 50 were to lose weight, they would reduce their risk of heart disease. $~$\\ 
X: Therefore, if the men between 25 and 50 who are overweight were to lose weight, they would reduce their risk of heart disease by 33\%. $~$\\ 
Z: Therefore, if 33\% of American men were to lose weight, they would reduce their risk of heart disease. $~$\\ 
F: Therefore, if the overweight men between 25 and 50 were to lose weight, their risk of heart disease would be reduced. $~$\\ 
You press $<<$F$>>$. $~$\\ 
 $~$\\ 
Q5. All computer geniuses are also brilliant mathematicians. Therefore, some computer geniuses don't require calculators for simple multiplication facts. Which of the following is the least necessary assumption for the previous conclusion to be logically correct? $~$\\ 
The choices are: $~$\\ 
F: Some brilliant mathematicians don't require calcul

\subsubsection*{Two-step task (neural alignment)}

Data source: \cite{feher2023rethinking} \\ $~$ \\
Number of experiments: 1 $~$\\ 
Number of participants: 94 $~$\\ 
Number of choices: 28153 $~$\\ 
 $~$\\ 
Example prompt: $~$\\ 
 $~$\\ 
 You are playing multiple rounds of a game. $~$\\ 
Your goal is to collect as many gold coins as possible as you visit different states. $~$\\ 
If you are in state C, you have the choice between options B and H. $~$\\ 
If you are in state S, you have the choice between options G and X. $~$\\ 
Picking one of these options may result in a gold coin. $~$\\ 
How likely an option leads to a gold coin slowly changes during the game. $~$\\ 
Picking option O generally leads to state C, and picking option N generally leads to state S. $~$\\ 
However, on rare occasions you will end up in the other state. $~$\\ 
You can select an option by pressing the corresponding key. $~$\\ 
 $~$\\ 
You are presented with options O and N. You press $<<$N$>>$. You end up in state S. You are presented with option G and option X. You press $<<$X$>>$. You receive 1 coins. $~$\\ 
You are presented with options O and N. You press $<<$N$>>$. You end up in state S. You are presented with option G and option X. You press $<<$X$>>$. You receive 1 coins. $~$\\ 
You are presented with options N and O. You press $<<$N$>>$. You end up in state S. You are presented with option G and option X. You press $<<$X$>>$. You receive 1 coins. $~$\\ 
You are presented with options N and O. You press $<<$O$>>$. You end up in state C. You are presented with option B and option H. You press $<<$B$>>$. You receive 1 coins. $~$\\ 
You are presented with options N and O. You press $<<$O$>>$. You end up in state S. You are presented with option G and option X. You press $<<$X$>>$. You receive 1 coins. $~$\\ 
You are presented with options N and O. You press $<<$O$>>$. You end up in state C. You are presented with option B and option H. You press $<<$B$>>$. You receive 1 coins. $~$\\ 
You are presented with options O and N. You press $<<$N$>>$. You end up in state S. You are presented with option G and option X. You press $<<$X$>>$. You receive 0 coins. $~$\\ 
You are presented with options O and N. You press $<<$N$>>$. You end up in state C. You are presented with option B and option H. You press $<<$H$>>$. You receive 1 coins. $~$\\ 
You are presented with options N and O. You press $<<$O$>>$. You end up in state C. You are presented with option B and option H. You press $<<$H$>>$. You receive 1 coins. $~$\\ 
You are presented with options N and O. You press $<<$O$>>$. You end up in state S. You are presented with option G and option X. You press $<<$G$>>$. You receive 0 coins. $~$\\ 
You are presented with options O and N. You press $<<$N$>>$. You end up in state S. You are presented with option G and option X. You press $<<$X$>>$. You receive 1 coins. $~$\\ 
You are presented with options N and O. You press $<<$O$>>$. You end up in state C. You are presented with option B and option H. You press $<<$H$>>$. You receive 1 coins. $~$\\ 
You are presented with options O and N. You press $<<$N$>>$. You end up in state S. You are presented with option G and option X. You press $<<$X$>>$. You receive 0 coins. $~$\\ 
You are presented with options O and N. You press $<<$N$>>$. You end up in state S. You are presented with option G and option X. You press $<<$G$>>$. You receive 0 coins. $~$\\ 
You are presented with options N and O. You press $<<$O$>>$. You end up in state C. You are presented with option B and option H. You press $<<$B$>>$. You receive 1 coins. $~$\\ 
You are presented with options O and N. You press $<<$N$>>$. You end up in state S. You are presented with option G and option X. You press $<<$X$>>$. You receive 1 coins. $~$\\ 
You are presented with options N and O. You press $<<$O$>>$. You end up in state S. You are presented with option G and option X. You press $<<$X$>>$. You receive 1 coins. $~$\\ 
You are presented with options O and N. You press $<<$N$>>$. You end up in state S. You are presented with option G and option X. You press $<<$X$>>$. You receive 1 coins. $~$\\ 
You are presented with options O and N. You press $<<$N$>>$. You end up in state S. You are presented with option G and option X. You press $<<$X$>>$. You receive 0 coins. $~$\\ 
You are presented with options O and N. You press $<<$N$>>$. You end up in state C. You are presented with option B and option H. You press $<<$H$>>$. You receive 1 coins. $~$\\ 
You are presented with options O and N. You press $<<$N$>>$. You end up in state C. You are presented with option B and option H. You press $<<$H$>>$. You receive 1 coins. $~$\\ 
You are presented with options N and O.

\subsubsection*{Sentence reading (neural alignment)}

Data source: \cite{tuckute2024driving} \\ $~$ \\
Number of experiments: 1 $~$\\ 
Number of participants: 5 $~$\\ 
Number of choices: 0 $~$\\ 
 $~$\\ 
Example prompt: $~$\\ 
 $~$\\ 
We were sitting on the couch.





\newpage
\bibliography{sn-bibliography}

\end{document}